\documentclass[11pt]{article}

\usepackage[margin=1in]{geometry}
\usepackage[T1]{fontenc}
\usepackage[utf8]{inputenc}
\usepackage{microtype}
\usepackage{booktabs}
\usepackage{graphicx}
\usepackage{xcolor}
\usepackage{amsmath,amssymb}
\usepackage[round,authoryear]{natbib}
\usepackage{hyperref}
\usepackage{caption}
\title{ISBC: Interpretable Sparse Binary Coding of Pretrained Text Embeddings}
\author{
Stefano M. Iacus \\
   Joint Research Centre \\
  European Commission\\
  Seville (SP) \\
  \texttt{stefano.iacus@ec.europa.eu}
  \and
  Haodong Qi \\
  Department of Global Political Studies \\
  Malmö University \\
  Malmö (SE)\\
  \texttt{haodong.qi@mau.se}   
}

\begin{document}

\maketitle

\begin{abstract}
Pretrained text embeddings encode semantic structure in high-dimensional
continuous spaces whose coordinates are difficult to interpret. We introduce
Interpretable Sparse Binary Coding (ISBC), which learns a small set of binary
dimensions from frozen embeddings without taxonomy labels or downstream
outcomes. A trainable encoder and affine decoder combine reconstruction
with controls on activity, factor utilization, dependence, and activation
uncertainty. Individual bits and their conjunctions select overlapping
document groups; complete activation profiles provide finer contexts for
conditional comparisons. We evaluate these representations across corpora,
backbones, and code configurations through reconstruction, geometry,
predictive probes, and document inspection. Empirical cases identify
both distinctions within existing categories and shared content across
their boundaries. In a mixed-domain corpus, the most frequent profiles predominantly
identify source collections, while selected within-source comparisons reveal
sentiment and migration contrasts. Selected associations persist in additional
documents encoded without refitting. In an engagement application, bits alone
improve prediction of log-transformed outcomes over a constant reference,
while their combinations identify within-event content contrasts associated
with engagement.
The results support using binary coordinates and their combinations
for empirical interpretation, with complete profiles providing a fine,
partly initialization-dependent description rather than a unique discrete
taxonomy.
\end{abstract}


\clearpage

\section{Introduction}
\label{sec:introduction}

Pretrained text embeddings support semantic comparison and prediction, but
empirical text analysis also requires descriptions of document content.
Researchers may wish to compare texts sharing selected attributes, examine
variation within an existing category, or relate content to an external outcome.
The individual coordinates of a high-dimensional embedding rarely provide
such descriptions directly. A small set of binary dimensions offers an explicit
way to select and compare document groups, provided that the content associated
with those dimensions can be interpreted empirically.

We study this possibility with Interpretable Sparse Binary Coding (ISBC),
which learns a sparse binary representation of frozen pretrained embeddings.
The approach separates the linguistic representation supplied by a
pretrained model from the estimation of binary coordinates for the
study corpus. The embedding backbone remains frozen during ISBC training.
A trainable encoder maps embeddings to binary activations, and an affine
decoder reconstructs the original vectors. The learning objective controls
activity, factor balance, dependence, and activation uncertainty alongside
reconstruction. Neither observed taxonomies nor downstream outcomes enter
representation learning.

Individual bits and their conjunctions select overlapping document
subsets, while complete activation configurations, which we call
profiles, partition the corpus. Comparing profiles that differ in one
bit examines that coordinate within a fixed context; conjunctions leave
the remaining coordinates unrestricted. These comparisons can reveal
distinctions within existing categories and shared content across their
boundaries, with interpretation grounded in document inspection.
Profiles provide fine descriptions that can vary across initializations,
rather than a unique discrete taxonomy.

We distinguish two questions: which properties of the embeddings survive
compression, and how the resulting binary coordinates can be interpreted.
For the first, we assess reconstruction, global and local geometry, and
label prediction, while code entropy describes the diversity of observed
profiles. Principal component analysis provides a continuous reconstruction
reference at equal dimension \citep{EckartYoung1936}. For the second, we
inspect documents selected by bits and their combinations, using existing
annotations as external descriptions. We examine how these properties depend
on code width, activity, and the pretrained backbone. The cross-corpus analysis
of Qwen-based codes finds stronger complete-profile agreement with some
entity, event, and topic annotations than with affective or evaluative
annotations (Section~\ref{sec:cross-dataset}). The coordinates have no prescribed
semantic content: their interpretation depends on the frozen embedding
space, the fitted corpus, and the learned compression. An engagement
application combines prediction from bits alone with exploratory comparisons
of the content selected by bit combinations in low- and high-engagement
groups (Section~\ref{sec:pheme-engagement}).

Most predictive evaluations are transductive: representation learning
uses the full unlabeled corpus, while labels or outcomes are held out
for predictive evaluation. A separate analysis applies a frozen encoder
to additional documents excluded from representation fitting, showing
that selected content associations persist within the same source
collections.

We compare ISBC with iterative quantization
(ITQ; \citealp{GongLazebnik2011}), BinarySentEmb
\citep{ShenEtAl2019}, and HypotheSAEs \citep{MovvaEtAl2025},
examining reconstruction, code usage, and predictive utility under
different binary activity constraints. We also compare profiles and
conjunctions with embedding clusters
(Section~\ref{sec:profiles-vs-clusters}); their non-equivalence is assessed
separately from the semantic interpretation of the selected documents.

The contribution is a compact, label-free representation designed for
empirical analysis through binary coordinates and their combinations.
Section~\ref{sec:model} presents ISBC and its learning objective.
Sections~\ref{sec:experimental-design}--\ref{sec:external-encoders}
examine the resulting document groups, quantify compression costs, and
compare representations across embedding backbones and alternative encoders.
Section~\ref{sec:cross-dataset} examines correspondence with existing
annotations, while Section~\ref{sec:mixed} shows how selected content
associations persist in additional documents encoded without refitting.
Section~\ref{sec:pheme-engagement} evaluates the bits as predictors of
external outcomes and connects their use to document interpretation.

\section{Related Work}
\label{sec:related-work}

Binary representations have precedents in semantic hashing
\citep{SalakhutdinovHinton2009}, binary autoencoders
\citep{CarreiraPerpinanRaziperchikolaei2015}, random-hyperplane hashing
\citep{Charikar2002}, and iterative quantization (ITQ)
\citep{GongLazebnik2011}. ITQ first reduces dimensionality through PCA and
then learns an orthogonal rotation that minimizes quantization error
before applying a sign threshold. Its bits are therefore thresholded
linear projections, providing a linear binary reference at the same code width.
\citet{ShenEtAl2019} compress pretrained sentence embeddings with an affine
sigmoid encoder, studying widths of 512--4,096 bits. ISBC uses a nonlinear
encoder with 8--32 bits and explicit activity and factor-utilization controls.
\citet{MostardEtAl2022} also investigate the semantics of binary coordinates
at the word level.

Semantic IDs discretize embeddings for generative recommendation
\citep{RajputEtAl2023} and downstream ranking \citep{SinghEtAl2024}.
TIGER uses residual quantization and resolves collisions to identify items;
shared codes in ISBC instead define document groups for analysis.
Reconstruction fidelity and downstream utility remain distinct
\citep{GusakFrolov2026}, motivating separate assessments of the properties
retained under compression.

Sparse coding uses a limited number of active dictionary elements
\citep{OlshausenField1996}. Sparse word representations include binary
variants \citep{FaruquiEtAl2015}, while sparse autoencoders extract
language-model features \citep{CunninghamEtAl2024}, support embedding-based
semantic search \citep{ONeillEtAl2024,ShinKim2026}, and supply token-level
features for document analysis \citep{JiangEtAl2026}. ISBC focuses on a small
binary code over frozen document embeddings, using coordinates and their
combinations for empirical interpretation.

For external outcomes, supervised topic models jointly learn representations
and document-response associations \citep{BleiMcAuliffe2007}. HypotheSAEs
learns sparse embedding features before predictive selection and
natural-language interpretation \citep{MovvaEtAl2025}. Its top-$k$ continuous
activations and affine decoder make it a close alternative; the original
study selects dictionary size and sparsity using validation prediction.
ISBC learns and fixes its binary representation without downstream outcomes.

Interpretation can also be organized around topic mixtures
\citep{BleiNgJordan2003,BianchiEtAl2021}, embedding clusters with topic
descriptions \citep{Grootendorst2022}, or concept sets supplied through
annotations or pretrained models \citep{KohEtAl2020,OikarinenEtAl2023}.
Clustering document embeddings provides a partition based on geometric
similarity. We compare these partitions with complete binary profiles
and examine how bit conjunctions extend across clusters
(Section~\ref{sec:profiles-vs-clusters}).
ISBC assigns semantic descriptions after fitting, without a prescribed
concept vocabulary. Sparsity and low covariance alone do not identify
independent concepts \citep{LocatelloEtAl2019}; interpretation requires
substantive validation through document content \citep{GrimmerStewart2013}.

Predictive probes assess properties accessible to a chosen estimator
\citep{HewittLiang2019,XuEtAl2020}. Comparisons across representations must
also account for information-preserving transformations
\citep{KornblithEtAl2019} and chance agreement between partitions
\citep{VinhEppsBailey2010}. We accordingly distinguish predictive utility,
aligned-coordinate correspondence, and document-group agreement from
semantic interpretation.

Building on these precedents, ISBC learns a small set of binary dimensions
with controlled activity from frozen document embeddings. The aim is
to support document interpretation through individual activations and
their combinations, without prescribing a taxonomy or using downstream
outcomes during fitting.

\section{Interpretable Sparse Binary Coding}
\label{sec:model}

\begin{figure}[tbp]
  \centering
  \includegraphics[width=\textwidth]{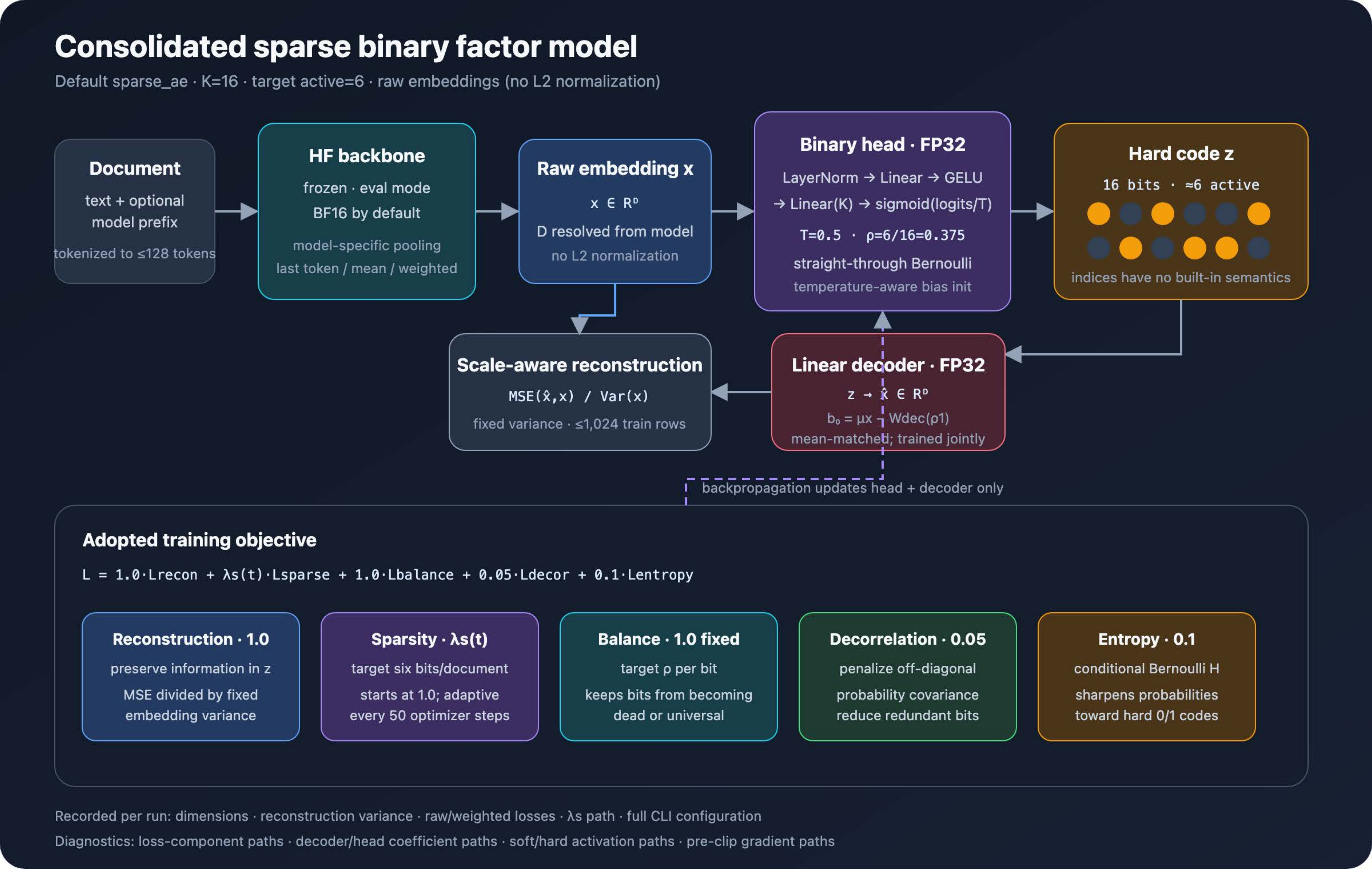}
  \caption{ISBC architecture and learning objective. A frozen pretrained backbone
  supplies raw embeddings to a trainable binary encoder. An affine decoder
  reconstructs the embeddings from straight-through binary activations, with its
  intercept initialized according to \eqref{eq:decoder-intercept}. The lower panel
  shows the reconstruction objective and four regularization terms with their
  adopted weights.}
  \label{fig:sparse-binary-model}
\end{figure}

ISBC learns a sparse binary representation of documents by reconstructing
their frozen pretrained embeddings. Let $N$ denote corpus size, $d_i$ document $i$ ($i=1,\ldots,N$), and $x_i=f_\theta(d_i)\in\mathbb{R}^D$ its embedding. The trainable model
maps $x_i$ to a code $B_i\in\{0,1\}^K$ and uses that code to reconstruct $x_i$.
Pretraining supplies the initial semantic representation, while the binary encoder
learns which aspects of that representation can be retained within the chosen
binary space. The code width $K$ and target number of active factors $A$ control
the size and sparsity of this space. No category labels or external outcomes related to document $d_i$ enter the representation learning.
Figure~\ref{fig:sparse-binary-model} summarizes the model and objective.

\subsection{From pretrained embeddings to binary profiles}

The backbone parameters $\theta$ remain fixed throughout training. Qwen is the
baseline backbone, but the model accepts embeddings from other general-purpose or
domain-specific encoders. Holding the backbone fixed separates the effects of
binary representation learning from those of representation fine-tuning. Pooling and input
prefixes follow the configuration of each backbone and are recorded in the run
metadata. We retain the raw pooled vectors, without L2 normalization, centering,
or rescaling, so the decoder is trained against the original embedding geometry.
The input dimension $D$ is determined by the selected backbone.

The trainable encoder first applies layer normalization and a hidden layer
with GELU activation,\footnote{GELU (Gaussian error linear unit) is applied
elementwise: $\operatorname{GELU}(z)=z\Phi(z)$, where $\Phi$ is the standard
normal cumulative distribution function.} then produces $K$ activation probabilities:
\begin{align}
  h_i &= \operatorname{GELU}\!\left(W_1\operatorname{LN}(x_i)+a_1\right),\\
  p_i &= \sigma\!\left((W_2h_i+a_2)/T\right).
\end{align}
Here $p_i\in(0,1)^K$, $\sigma$ is the logistic function, and the temperature is
$T=0.5$ by default. Layer normalization acts inside the encoder; the reconstruction
target remains $x_i$. Writing $\rho_{\mathrm{act}}=A/K$, we initialize the output bias to
$T\operatorname{logit}(\rho_{\mathrm{act}})$ to place initial activation probabilities near the
desired rate.

Training uses stochastic binary activations. For independent
$U_{ik}\sim\operatorname{Uniform}(0,1)$, we set
$B_{ik}=\mathbf{1}[p_{ik}>U_{ik}]$ and pass the straight-through quantity\footnote{The operator
$\operatorname{stopgrad}(z)$ returns $z$ in the forward pass and has zero
derivative during backpropagation.}
\begin{equation}
  Z_i^{\mathrm{ST}}=B_i+p_i-\operatorname{stopgrad}(p_i)
\end{equation}
to an affine decoder,
\begin{equation}
  \widehat x_i=W_{\mathrm{dec}}Z_i^{\mathrm{ST}}+b_{\mathrm{dec}}.
\end{equation}
Thus, reconstruction uses binary values in the forward pass, while gradients
propagate through the probabilities to the encoder. The encoder and decoder are
optimized jointly in FP32; the frozen backbone uses its configured precision,
BF16 by default.

To account for differences in embedding location across backbones, we estimate
the coordinate-wise mean $\mu_x$ from a fixed calibration sample of at most 1,024
cached embeddings and initialize the decoder intercept as
\begin{equation}
  b_{\mathrm{dec}}^{(0)}=\mu_x-W_{\mathrm{dec}}^{(0)}(\rho_{\mathrm{act}}\mathbf 1).
  \label{eq:decoder-intercept}
\end{equation}
The initial decoder therefore maps the reference activity vector $\rho_{\mathrm{act}}\mathbf 1$
to $\mu_x$. This initialization adjusts the starting reconstruction to the
embedding location while leaving the raw targets unchanged. Both decoder
parameters remain trainable.

After fitting, we obtain deterministic codes by setting
$B_{ik}=\mathbf{1}[p_{ik}>0.5]$.
A bit is a binary coordinate, also referred to as a factor. A profile is a
complete configuration of all $K$ bits. Documents sharing a profile form a
document group. A conjunction requires selected bits to be active while
leaving the remaining bits unrestricted; it also defines a document group.
Marginal bit-level analyses describe associations with individual coordinates,
whereas profiles specify the joint states of all coordinates, including inactive
bits. Semantic interpretations of bits and profiles are grounded in the
documents and associations observed after fitting. Since bit indices are arbitrary, comparisons between
independently fitted models require alignment rather than direct comparison of
coordinate numbers.

\subsection{Learning objective and estimation}

The objective combines embedding reconstruction with constraints on code use.
Reconstruction encourages accurate recovery of the pretrained embedding vectors,
while the remaining terms regulate activity, factor utilization, redundancy, and
uncertainty in binary decisions. For a minibatch of $n$ documents, the objective is
\begin{equation}
\label{eq:loss}
 \mathcal L=\mathcal L_{\mathrm{recon}}
 +\lambda_s\mathcal L_{\mathrm{sparse}}
 +\mathcal L_{\mathrm{balance}}
 +0.05\mathcal L_{\mathrm{decor}}
 +0.1\mathcal L_{\mathrm{entropy}}.
\end{equation}
We define the reconstruction term as
\begin{equation}
 \mathcal L_{\mathrm{recon}}=
 \frac{\operatorname{MSE}(\widehat x,x)}{s_x},
 \label{eq:recon}
\end{equation}
where MSE averages over documents and embedding coordinates, and $s_x$ is the mean
coordinate variance estimated from the same calibration sample used for
$\mu_x$. Dividing by this scalar expresses error relative to embedding variance
without changing the relative weighting of coordinates or transforming the targets.
Activity is controlled at two levels:
\begin{align}
 \mathcal L_{\mathrm{sparse}}&=\frac1n\sum_{i=1}^{n}
 \left(\frac1K\sum_{k=1}^{K}p_{ik}-\rho_{\mathrm{act}}\right)^2,  \label{eq:sparse}\\
 \mathcal L_{\mathrm{balance}}&=\frac1K\sum_{k=1}^{K}
 \left(\frac1n\sum_{i=1}^{n}p_{ik}-\rho_{\mathrm{act}}\right)^2.
 \label{eq:balance}
\end{align}
The first term encourages each document to have an expected number of active
factors near $A$. The second encourages each factor to be used at rate $\rho_{\mathrm{act}}$
across documents, discouraging factors that are always inactive or always active.
These are soft penalties, so deterministic codes need not contain exactly $A$
ones. The balance coefficient is fixed at one, while the document-activity
coefficient adapts during optimization. At each step, we compute the relative
activity deviation $|n^{-1}\sum_{i=1}^{n}\sum_{k=1}^{K}p_{ik}-A|/A$ and update its exponential moving
average $v_t$ with coefficient 0.95, initialized from the first observed deviation.
Every 50 steps, we update the penalty weight according to
\begin{equation}
 \lambda_s\leftarrow\operatorname{clip}\!\left[
 \lambda_s\exp\{0.25(v_t-0.05)\},0.1,1000\right],
\end{equation}
starting from $\lambda_s=1$. The penalty therefore increases when the smoothed
relative deviation exceeds $0.05$ and decreases when it falls below that value.

The remaining terms address dependence between factors and uncertainty within
individual activations. Let $C$ be the sample covariance matrix of the minibatch
probabilities, computed with denominator $\max(1,n-1)$. We use
\begin{align}
 \mathcal L_{\mathrm{decor}}
&=\frac1{K^2}\sum_{k\ne\ell}C_{k\ell}^2,
 \label{eq:decor}\\
 \mathcal L_{\mathrm{entropy}}
 &=-\frac1{nK}\sum_{i=1}^{n}\sum_{k=1}^{K}
 \left[p_{ik}\log p_{ik}+(1-p_{ik})\log(1-p_{ik})\right].
 \label{eq:entropy}
\end{align}
 The covariance penalty discourages pairwise dependence in factor probabilities.
Minimizing the conditional Bernoulli entropy pushes those probabilities toward
zero or one, reducing uncertainty in the sampled activations. Together with the
activity and balance terms, it encourages decisive activations while regulating
how often factors are used. These penalties do not guarantee independence or
semantic interpretability.

All unsupervised fitting uses the complete document corpus. Embeddings are cached
with ordered document identifiers and reused across epochs. Training records
include the raw and weighted losses, adaptive penalty trajectory, activation
statistics, parameter trajectories, gradient norms, and calibration statistics,
allowing the realized behavior of the objective to be checked. Predictive
evaluation subsequently uses the frozen codes and a seeded 80/20 document split.
This holds out labels for evaluation while retaining all documents in unsupervised
representation fitting. Supplementary Section~\ref{app:reproducibility} records the shared
hyperparameters, implementation settings, and measured training costs.

\section{AG News: code structure and information preservation}
\label{sec:experimental-design}

Unless otherwise specified, the empirical analyses use $K=16$ and
target activity $A=6$ as a common reference configuration.
This intermediate choice provides a common operating point rather than
maximizing predictive performance for a particular dataset. The AG News
case study first examines document groups and their interpretation, then
measures compression and geometry, assesses the learning objective through
ablation, and finally evaluates predictive utility.

\subsection{The AG News dataset}

As a driving example throughout this section, we use the AG News topic-classification benchmark introduced by
\citet{zhang2015character}. Each English-language document combines a news
headline with its short description and is assigned to one mutually exclusive
topic: \texttt{World}, \texttt{Sports}, \texttt{Business}, or \texttt{Sci\_Tech}. The
released benchmark contains 120,000 training documents (30,000 per class) and
7,600 test documents (1,900 per class), for a total of 127,600 documents.
Because our binary encoder is trained without labels, we concatenate the
original training and test partitions when extracting frozen embeddings and
learning sparse binary codes; the original partition is retained only as provenance. See also Supplementary Section~\ref{app:data-agnews}.

\subsection{Analysis of bit activations and profiles}
\label{sec:agnews-bit-structure}

\begin{figure}[tbp]
  \centering
  \includegraphics[width=0.95\textwidth]{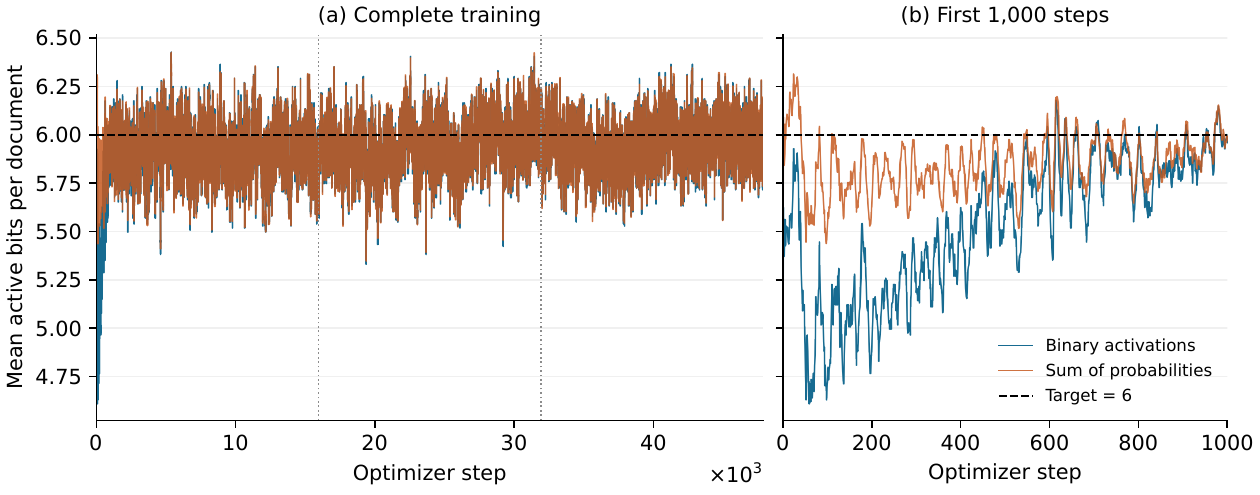}
  \caption{Qwen training activity on AG News. Curves show trailing
  averages over 12 optimizer steps of minibatch means for binary activations and
  summed activation probabilities. The dashed line marks the target of six.
  The left panel shows all three epochs, separated by vertical dotted lines;
  the right panel shows the first 1,000 steps. These are training diagnostics,
  rather than full-corpus evaluations of a fixed checkpoint.}
  \label{fig:qwen-activity-training}
\end{figure}

We use \texttt{Qwen/Qwen3-Embedding-0.6B}, a 0.6-billion-parameter
instruction-aware embedding model \citep{zhang2025qwen3embedding}.
During training, mean activity settles near the target of six and the binary
and probability-based traces become similar
(Figure~\ref{fig:qwen-activity-training}).

For the fitted code, the observed mean activity,
$\bar a=N^{-1}\sum_{i=1}^{N}\sum_{k=1}^{K}B_{ik}$, is 5.94
(standard deviation 1.03; median 6). The target permits variation:
38.6\% of documents activate exactly six bits and 87.0\% activate five
to seven. Individual bit prevalences range from 33.7\% to 39.6\%, near
the reference rate $6/16=37.5\%$, consistent with the balance penalty.

The three-epoch fit produces 12,174 distinct profiles. The most frequent
contains 1.08\% of documents; 422 profiles cover half the corpus and
2,257 cover 80\%. A further 4,445 profiles occur only once, accounting
for 36.5\% of profiles but 3.48\% of documents. Thus, similar bit
prevalences coexist with recurrent configurations and a long tail of rare
profiles. The twenty most frequent cover 13.20\% of documents
(Figure~\ref{fig:qwen_top20} in Supplementary Section~\ref{app:data-agnews}).

\begin{figure}[tbp]
  \centering
\includegraphics[height=0.4\textheight]{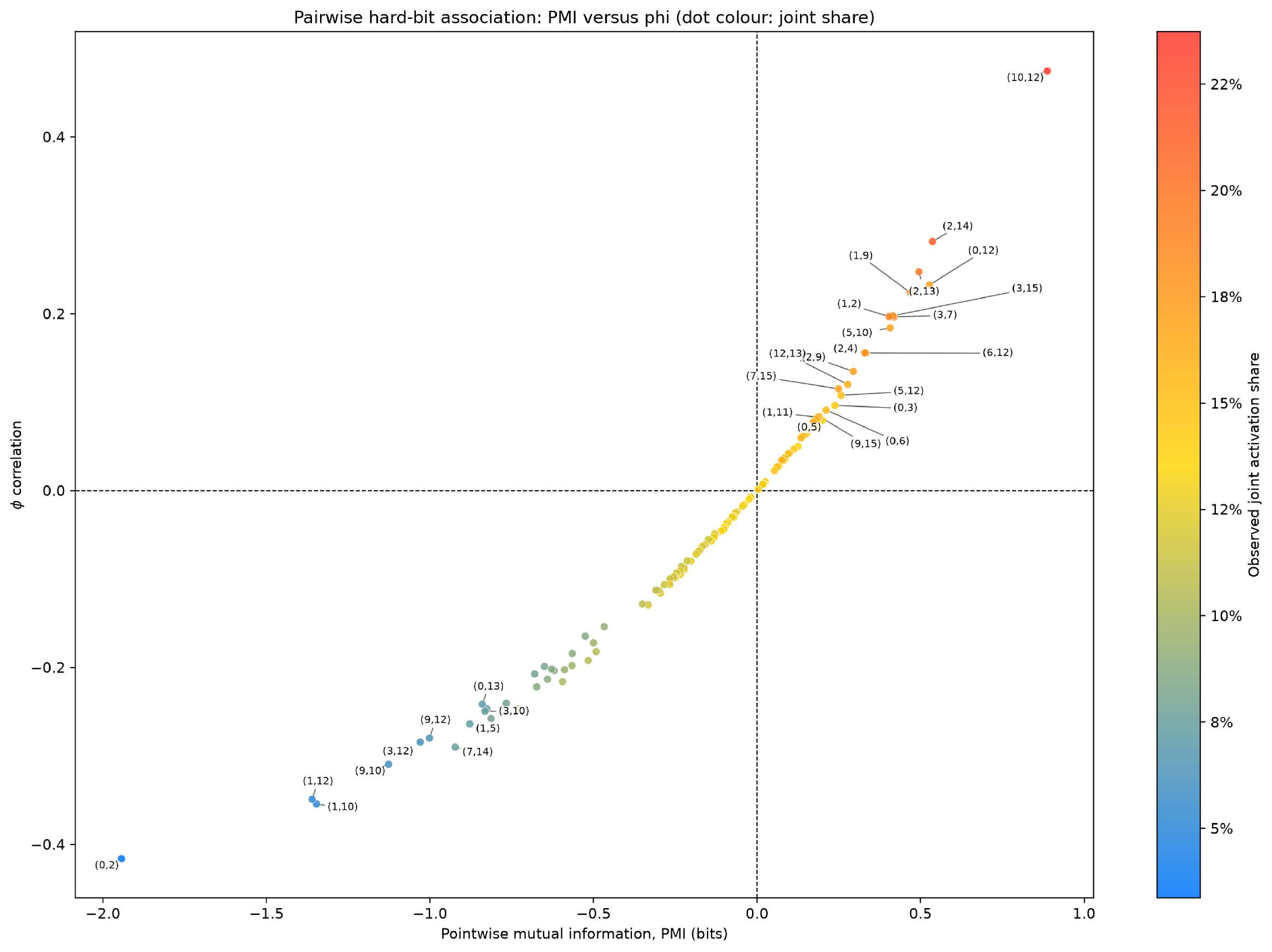}
  \caption{Pointwise mutual information (PMI) against the $\phi$ correlation coefficient. Positive PMI indicates more joint activation of bits $(k,\ell)$ than marginal independence predicts; negative PMI indicates less, and $-\infty$ means the pair never jointly activated in the sample. This descriptive reference does not condition on document-level activity. The color scale gives the proportion of documents in which both bits are active.}
  \label{fig:qwen_PMI}
\end{figure}

Pairwise associations summarize dependence among bits without inspecting
every profile. Figure~\ref{fig:qwen_PMI} compares pointwise mutual information,
$\mathrm{PMI}_{k\ell}=\log_2\!\left[P(B_k=1,B_\ell=1)/
\{P(B_k=1)P(B_\ell=1)\}\right]$,
with $\phi$, the Pearson correlation between binary coordinates.
Probabilities are estimated from corpus frequencies. Positive PMI indicates
more co-activation than marginal independence predicts; negative PMI indicates
less. The activity budget can itself induce negative dependence, so negative
associations do not establish semantic opposition.
Supplementary Section~\ref{app:activity-normalization} explains this constraint.

\begin{figure}[tbph]
  \centering
  \includegraphics[width=\textwidth]{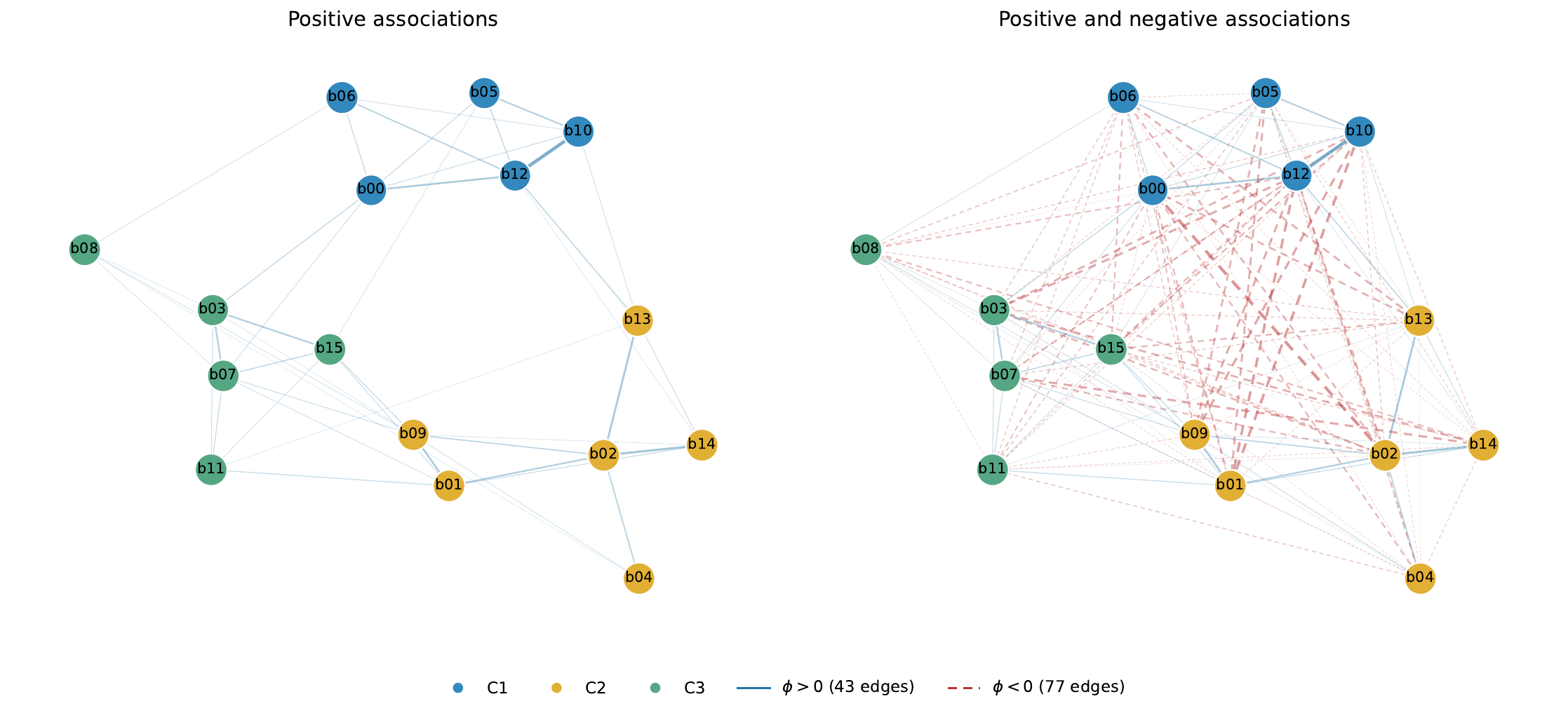}
  \caption{Co-activation communities among the 16 Qwen bits on AG News.
Both panels show the same three communities, identified by Leiden
using positive PMI weights. Blue edges highlight groups of bits
that tend to activate together. The right panel additionally shows
negative associations as dashed red edges, predominantly connecting
different communities and indicating separation between their
activation patterns. Node area reflects activation prevalence;
edge width and opacity reflect $|\phi|$.}
  \label{fig:qwen-bit-communities}
\end{figure}

We form an undirected graph from all 43 positive-PMI pairs, weighted by
PMI, and identify bit communities using Leiden \citep{TraagEtAl2019}
with weighted modularity at resolution $\gamma=1$
\citep{ReichardtBornholdt2006}. Neither labels nor a prescribed number of
communities enter the analysis. The resulting groups contain five, six,
and five bits, with modularity $Q=0.4646$
(Figure~\ref{fig:qwen-bit-communities}). The partition is identical across
20 optimizer seeds and unchanged at the nearby resolutions 0.75 and 1.25.
These results identify three reproducible groups of bits that tend to
activate together, providing a basis for examining the document content
associated with their joint activation.
Supplementary Section~\ref{app:qwen-graph-protocol} gives the full sensitivity
analysis and graph construction details.

The right panel adds the 77 negative-$\phi$ associations as dashed edges
without changing the fitted communities. Of these, 69 connect different
communities and account for 96.6\% of the total absolute negative
correlation weight. This concentration is consistent with a separation
between co-activation groups: bits within each community tend to activate
together, while negative associations occur predominantly across
communities.

\subsection{How bit activations and conjunctions relate to an existing taxonomy}
\label{sec:agnews-taxonomy}

We examine the learned representation against the AG News taxonomy
$\mathcal T=\{\texttt{World},\,\texttt{Sports},\,\texttt{Business},\,
\texttt{Sci\_Tech}\}$. These mutually exclusive categories reflect choices
about topic definitions and assignment, although individual documents can
address several themes. We use them as an external analytical reference
for a representation learned without labels.

\begin{figure}[tbp]
   \centering
  \begin{minipage}[c]{0.48\textwidth}
    \centering
    \includegraphics[width=\linewidth]{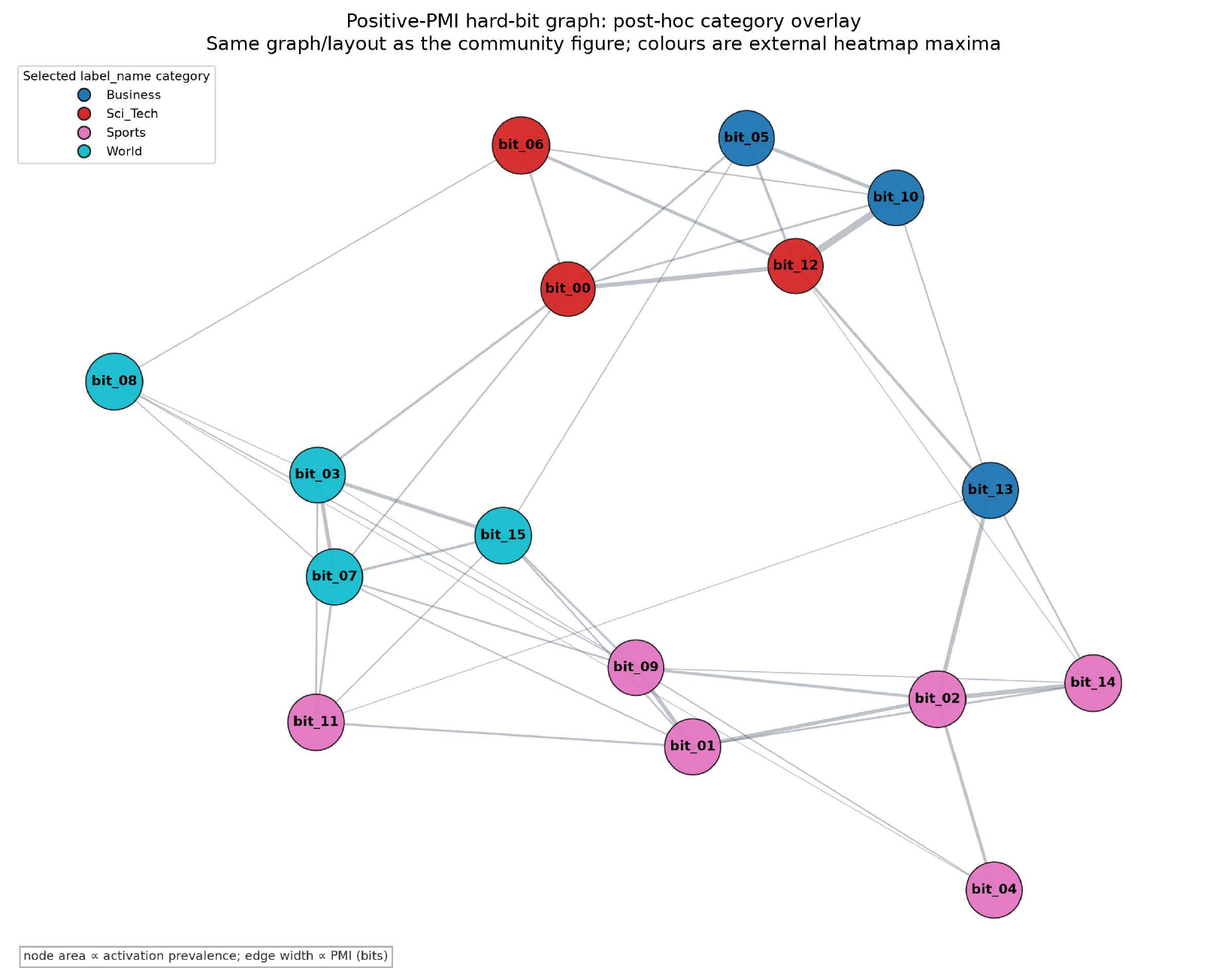}
  \end{minipage}
  \hfill
  \begin{minipage}[c]{0.48\textwidth}
    \centering
    \includegraphics[width=\linewidth]{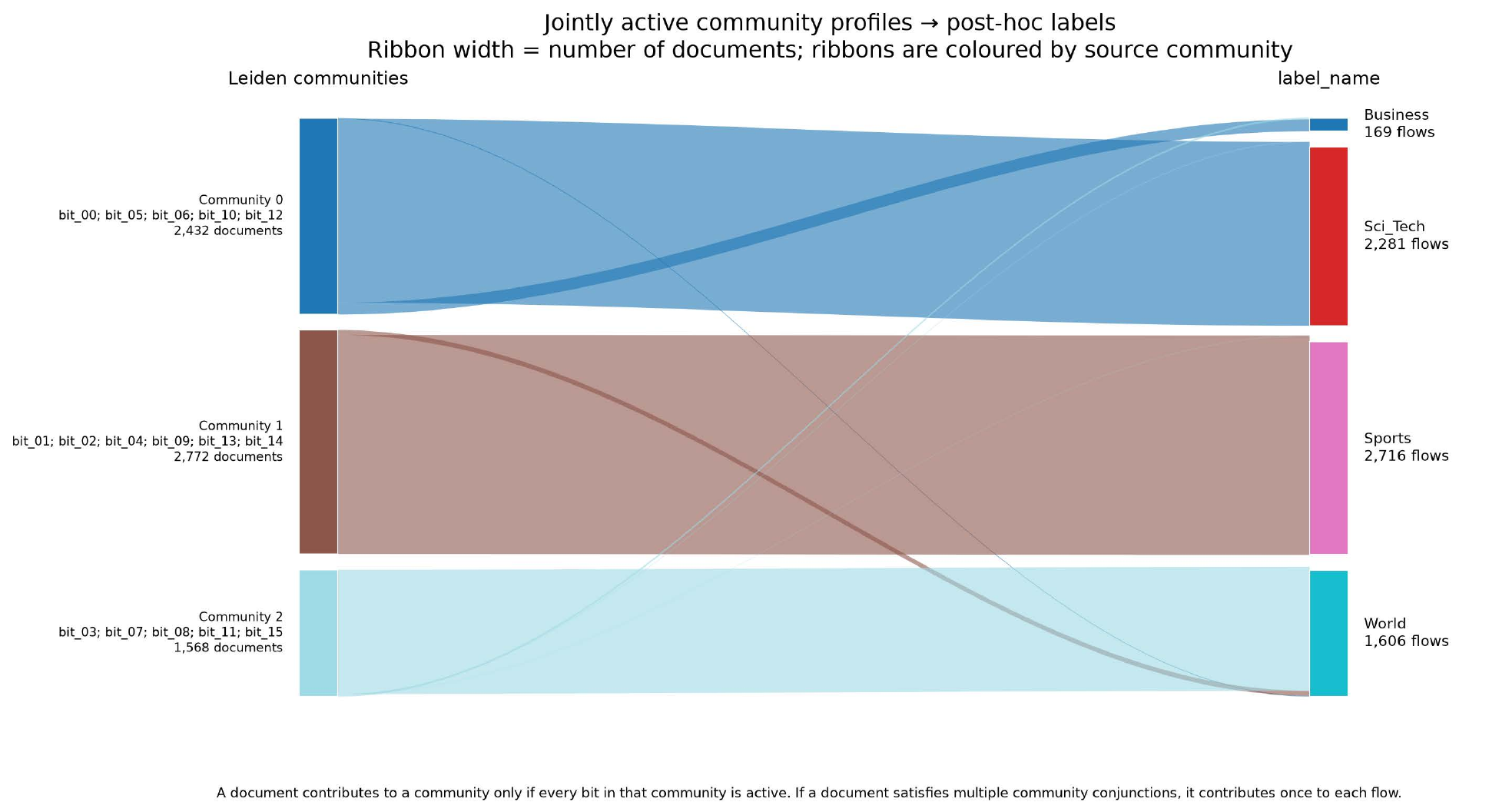}
  \end{minipage}
  \caption{Left: the positive-edge subgraph of
  Figure~\ref{fig:qwen-bit-communities}, with nodes colored by their category
  associations and edge widths increasing with PMI. Right: category composition of document subsets defined by
  joint activation of all bits in each community; additional bits may also
  be active. Category labels are used only for this subsequent interpretation.}
  \label{fig:qwen_bit_expost}
\end{figure}

Figure~\ref{fig:qwen_bit_expost} relates category associations to the
co-activation communities. Their alignment is partial: in particular,
\texttt{Business}-associated bits occur in more than one community.
For each community, we select documents in which all member bits are active,
leaving the remaining coordinates unrestricted. Each resulting group can
contain several complete profiles; the Sankey plot shows its category
composition. Intersecting these groups with the \texttt{Business} label
identifies two disjoint subsets:

\begin{itemize}
\item \textbf{\texttt{Business} $\cap$ \texttt{Sci\_Tech}-dominant
 document group} (147 documents; 0.461\% of \texttt{Business}): commercial
activity in technology markets. For example, ``IBM to make digital image sensors for Kodak''
(ID 30040) describes a commercial agreement to manufacture digital
imaging technology, illustrating the intersection of business
and technological content.
\item \textbf{\texttt{Business} $\cap$ \texttt{World}-dominant
 document group} (22 documents; 0.069\% of \texttt{Business}): economic
activity in international political and institutional settings. For example,
``Donors seek clarity on PA finances'' (ID 125602) concerns financial
transparency following a change in Palestinian leadership.
\end{itemize}

The technology-market conjunction also selects ``Internet patents sold at
S.F. auction'' (ID 3785), a \texttt{Sci\_Tech}-labelled report on an auction
of patents covering technologies for Internet business transactions,
illustrating the intersection of technological content and commercial
activity. These examples illustrate how documents assigned to a single editorial
category can also contain elements associated with another.
The bit conjunctions bring together such documents through shared
content that crosses the original category boundaries.

The non-\texttt{Business} documents satisfying the same conjunctions
emphasize technological products and design in the first case, and political
actors, diplomacy, and conflict in the second
(Table~\ref{tab:qwen-community-complement-keywords},
Supplementary Section~\ref{app:agnews-interpretation}). Across the complete
subsets, 127 of the 147 \texttt{Business} documents in the technology-market
conjunction (86.4\%) belong to profiles whose most frequent category is
\texttt{Sci\_Tech}; in the political conjunction, 9 of 22 (40.9\%) belong
to profiles whose most frequent category is \texttt{World}. Tied maxima
are excluded from these numerators. Figures~\ref{fig:agnews-technology-profile-map}
and~\ref{fig:agnews-politics-profile-map} in the Supplementary Material locate
the illustrative documents within these complete-profile distributions.
Four \texttt{Sports}-labelled documents in the \texttt{World}-dominant group
similarly concern security, institutional authority, or club ownership.
These examples show how conjunctions connect documents through shared
contexts across editorial categories. Supplementary Section~\ref{app:agnews-interpretation}
provides the document descriptions, activation heatmap, and inspection details.

A complementary comparison examines variation within a category. Among
224 pairs of observed profiles differing in exactly one bit and containing
at least 100 documents each, we select the pair maximizing the smaller
group size, without using labels or text content. The selected profiles
differ only in bit~08 and contain 1,050 and 1,224 documents, both almost
entirely labelled \texttt{Sports}. With the other fifteen coordinates fixed,
the active group concentrates on American football, while the inactive group
includes baseball, basketball, and other sports. Document inspection and
term-presence counts support this distinction
(Table~\ref{tab:agnews-single-bit-contrast}).

\begin{table}[tbp]
\centering
\small
\setlength{\tabcolsep}{5pt}
\begin{tabular}{lrr}
\toprule
Characteristic & Bit 08 inactive & Bit 08 active \\
\midrule
Documents & 1,050 & 1,224 \\
\texttt{Sports} (\%) & 97.43 & 98.69 \\
Touchdown or quarterback (\%) & 1.62 & 45.75 \\
Home run or inning (\%) & 23.90 & 1.47 \\
\bottomrule
\end{tabular}
\caption{A single-bit profile contrast within AG News.
Bits $\{1,2,4,9,13,14\}$ are active in both profiles; all other bits except
bit~08 are inactive. Percentages use the complete membership of each profile.
Lexical indicators count documents containing either expression, including
regular plural forms, after HTML decoding and case-insensitive matching at
word boundaries. They measure term presence, not annotated sport prevalence.}
\label{tab:agnews-single-bit-contrast}
\end{table}

Once again, the post-hoc category composition in
Figure~\ref{fig:qwen_bit_expost} illustrates how bit conjunctions
connect documents across editorial categories. Mutually exclusive
labels provide a useful organization of a corpus, but a single
assignment need not capture all semantic aspects of a document.
Bits and their combinations make some of these shared attributes
explicit, supporting comparisons within and across the existing taxonomy.

\subsection{Binary profiles and conjunctions versus embedding clusters}
\label{sec:profiles-vs-clusters}

Could clustering the embeddings recover the document groups selected by
bit conjunctions? We fit $k$-means \citep{Lloyd1982} to the same 127,600 raw,
unnormalised Qwen embeddings with $M\in\{32,64,256\}$ clusters,
without labels or binary codes. We compare these partitions with individual
bits, overlapping conjunctions, and complete profiles.

For bit $B_k$ and cluster membership $C$, let $H$ denote Shannon entropy
computed with base-two logarithms. The residual entropy fraction
$H(B_k\mid C)/H(B_k)$ measures variation remaining within clusters:
zero means that cluster membership determines the bit, and one indicates
independence. Its mean across the sixteen bits decreases from 0.578 at
$M=32$ to 0.489 at $M=256$, against approximately one under
prevalence-preserving permutation. The cluster assignments account for only part of the variation in
bit activations, a pattern also observed for the alternative binary
representations examined in Supplementary
Section~\ref{app:itq-entropy}. Documents within the same cluster
can therefore differ along binary dimensions, while documents
in different clusters can share particular activations.
Clustering and binary representations thus provide complementary
descriptions of how documents are organized in the embedding space.

Conversely, the median three-bit conjunction requires 6, 11, and 30.5
clusters to cover 80\% of its documents, compared with support-preserving
random references of 21.3, 41.3, and 149.8
(Supplementary Section~\ref{app:cluster-null-references}). Conjunctions concentrate
within subsets of clusters but extend across their boundaries, while
clusters contain multiple complete profiles. The representations are
therefore associated without defining equivalent document groups.

Clustering supplies mutually exclusive groups; bits and conjunctions
supply overlapping memberships that can be inspected separately, as in
Section~\ref{sec:agnews-taxonomy}. Complete profiles record those
memberships jointly and provide finer contexts, rather than a unique
discrete taxonomy. Supplementary Section~\ref{app:cluster-partition-diagnostics}
reports profile-level entropy, chance-corrected partition agreement,
and variation across initializations.
Category-based descriptions remain separate in
Supplementary Section~\ref{app:cluster-category-composition}.

\subsection{Code capacity and embedding reconstruction}
\label{sec:qwen-compression}

Code capacity refers to the range of distinct bit configurations
available to represent documents. Profile counts and entropy
measure how extensively this capacity is used.

We compare Qwen codes on all 127,600 AG News documents at widths
$K\in\{8,16,32\}$ and several activity targets
(Table~\ref{tab:qwen-width-retention-proposal}). Each configuration uses
one encoder fit with seed 1234; these comparisons describe sensitivity to
width and activity for that seed. Variation across initializations is
assessed separately in Section~\ref{sec:loss-ablation}.

A $K$-bit code admits at most $2^K$ profiles. For a complete configuration
$b$, $p(b)=P(B=b)$ is its occurrence probability, estimated by the observed
corpus proportion $\widehat p(b)=N^{-1}\sum_i\mathbf{1}\{B_i=b\}$.
The empirical entropy
$\widehat H(B)=-\sum_{b:\widehat p(b)>0}\widehat p(b)\log_2\widehat p(b)$
measures profile diversity in bits per document; $2^{\widehat H(B)}$ is
the number of equally frequent profiles with the same entropy.
Higher entropy indicates that documents are distributed across a larger
effective number of profiles, providing a finer description of their
joint activation patterns. Lower entropy indicates greater concentration
in shared configurations. Increasing code width makes more configurations
available and is therefore expected to support finer distinctions.
Consistent with this expectation, at target activity six, increasing
width from 16 to 32 raises entropy from 11.32 to 14.20 bits per document,
corresponding to approximately 2,600 versus 19,000 effective profiles.
The wider code therefore provides substantially greater granularity.
We also report Miller--Madow-corrected entropy to assess finite-sample
sensitivity; its definition and interpretation appear in
Supplementary Section~\ref{app:qwen-capacity-reconstruction}.

The affine decoder reconstructs embedding $x_i$ as
$g(B_i)=W_{\mathrm{dec}}B_i+b_{\mathrm{dec}}$: an intercept plus the
directions selected by the active bits. We measure reconstruction relative
to the corpus mean $\bar x$ by
\[
R^2=1-\frac{\sum_i\lVert x_i-g(B_i)\rVert_2^2}
{\sum_i\lVert x_i-\bar x\rVert_2^2}.
\]
All reconstructions lie in an affine subspace of dimension at most $K$.
PCA supplies the optimal affine reference at this dimension with
unrestricted continuous coefficients \citep{EckartYoung1936}; its score
$R^2_{\mathrm{PCA},K}$ is the cumulative proportion of embedding variance
explained by the first $K$ principal components. 

At target activity six, increasing width from 16 to 32 improves trained
reconstruction from 0.160 to 0.214; increasing the 32-bit activity target
to sixteen raises it to 0.232. Refitting the affine decoder by ordinary
least squares with fixed codes yields only small improvements. At 16 bits,
$R^2$ rises from 0.160 to 0.161, compared with 0.245 for PCA. The trained
decoder therefore approaches the best affine fit to these codes, while
closing the remaining gap requires changes to the representation
(see Supplementary Section~\ref{app:activity-normalization}).

\begin{table}[tbp]
  \centering
  \small
  \setlength{\tabcolsep}{4pt}
  \begin{tabular}{rrrrrrrr}
    \toprule
    $K$ & $A$ & $\bar a$ &
    $\widehat H$ & $\widehat H_{\mathrm{MM}}$ &
    $R^2_{\mathrm{trained}}$ & $R^2_{\mathrm{OLS}}$ &
    $R^2_{\mathrm{PCA},K}$ \\
    \midrule
    8  & 4  & 3.96  & 6.047  & 6.048  & 0.106 & 0.107 & 0.166 \\
    8  & 6  & 6.18  & 4.397  & 4.397  & 0.098 & 0.102 & 0.166 \\
    16 & 6  & 5.94  & 11.322 & 11.391 & 0.160 & 0.161 & 0.245 \\
    32 & 6  & 6.07  & 14.197 & 14.471 & 0.214 & 0.216 & 0.359 \\
    32 & 8  & 8.21  & 15.606 & 16.058 & 0.223 & 0.225 & 0.359 \\
    32 & 16 & 16.05 & 16.452 & 17.049 & 0.232 & 0.234 & 0.359 \\
    \bottomrule
  \end{tabular}
  \caption{Qwen on AG News across code widths and activity targets.
  Each row represents one encoder fit with seed 1234.
  $A$ is target activity and $\bar a$ is observed mean activity.
  $\widehat H$ and $\widehat H_{\mathrm{MM}}$ are empirical and
  Miller--Madow-corrected profile entropy, in bits/document.
  $R^2_{\mathrm{trained}}$ evaluates the jointly trained decoder;
  $R^2_{\mathrm{OLS}}$ evaluates an affine decoder refitted by ordinary
  least squares with the binary codes held fixed.
  $R^2_{\mathrm{PCA},K}$ is the cumulative variance explained by
  the first $K$ principal components.}
  \label{tab:qwen-width-retention-proposal}
\end{table}

At target activity six, the 8-, 16-, and 32-bit codes contain 78,
12,174, and 48,371 observed profiles. Increasing width permits finer
empirical differentiation at a common activity target. Cross-capacity
bit associations also connect individual coordinates in narrower codes
to several coordinates in wider codes, including inverse activation
patterns (see also Supplementary Section~\ref{app:qwen-capacity-organization}).

\subsection{Geometric preservation}
\label{sec:qwen-geometry}

Geometric preservation concerns how relationships between documents
in the embedding space are reflected in the binary space. We examine
this at two scales: globally, through correlations between pairwise
distances, and locally, through the assignment of nearest neighbors
to the same binary profile. We use the configurations introduced in
Section~\ref{sec:qwen-compression}.

For the global comparison, we calculate distances for the same
100,000 document pairs drawn from 2,000 sampled documents (seed 1234).
Spearman's $\rho$ measures how consistently the two representations
order these pairs from closer to more distant; higher correlations
indicate greater agreement. We compare cosine distances in both
spaces, and Euclidean distances in the original embeddings with
Hamming distances in the codes.

At 16 bits and target activity six, cosine--cosine correlation is
0.494 and Euclidean--Hamming correlation is 0.330. Across configurations,
the respective ranges are 0.449--0.501 and 0.291--0.374. The binary
codes therefore preserve angular distance ordering more closely
than Euclidean distance ordering in these comparisons
(see Supplementary Section~\ref{app:qwen-geometry-details}
and Table~\ref{tab:qwen-geometry-distances}).

For the local comparison, we find each sampled document's nearest
embedding neighbor among all other corpus documents and check whether
the two receive the same binary profile. Starting from embedding
neighbors avoids arbitrary selection among documents tied at zero
binary distance. At 16/6, profile agreement is 31.70\% for cosine
neighbors and 31.75\% for Euclidean neighbors, compared with 0.1470\%
for a uniformly selected other document
(Table~\ref{tab:qwen-local-profile-agreement}).

Wider codes distinguish more profiles and reduce exact agreement
between neighboring documents. Nevertheless, nearest-neighbor
agreement remains above the random reference in every configuration.
The global and local results thus show complementary aspects of
geometric preservation: partial agreement in pairwise distance
ordering and a greater tendency for nearby documents to share
a binary profile.

\begin{table}[tbph]
  \centering
  \small
  \setlength{\tabcolsep}{5pt}
  \begin{tabular}{rrrrr}
    \toprule
    $K/A$ & NN: cosine & NN: Euclidean & Random agreement &
    \shortstack{Shared-profile\\coverage} \\
    \midrule
    8/4   & 58.20 & 57.20 & 2.8374 & 100.00 \\
    8/6   & 74.95 & 74.00 & 7.2928 & 100.00 \\
    16/6  & 31.70 & 31.75 & 0.1470 &  96.75 \\
    32/6  & 22.85 & 22.60 & 0.0242 &  75.60 \\
    32/8  & 13.95 & 13.95 & 0.0079 &  49.75 \\
    32/16 &  8.95 &  8.80 & 0.0014 &  24.80 \\
    \bottomrule
  \end{tabular}
  \caption{Nearest-neighbor profile agreement for Qwen on AG News.
  $K/A$ denotes code width and target activity.
  NN agreement is the percentage of sampled documents whose nearest
  embedding neighbor shares their complete binary profile.
  Random agreement is the expected percentage for a uniformly selected
  other document, using the same reference documents.
  Shared-profile coverage is the percentage of sampled documents
  with at least one other corpus document sharing their complete profile.
  All values are percentages.}
  \label{tab:qwen-local-profile-agreement}
\end{table}

\subsection{Ablation of the learning objective}
\label{sec:loss-ablation}

We examine how the components of the loss function \eqref{eq:loss}
contribute to reconstruction, geometry, and code use through a paired
leave-one-term-out ablation. All variants use the same frozen,
unnormalized Qwen3 embeddings for all 127,600 AG News documents, with
$K=16$ and activity target $A=6$. For each seed (1234, 2025, and 31415),
we fit the full objective and five variants, setting one coefficient to
zero. The 18 fits share encoder initialization and the document-order
random-number stream within each seed. Removing activity control also
disables its adaptive coefficient update. Architecture and decoder
initialization follow Section~\ref{sec:model}.

\begin{figure}[tbp]
\centering
\includegraphics[width=\textwidth]{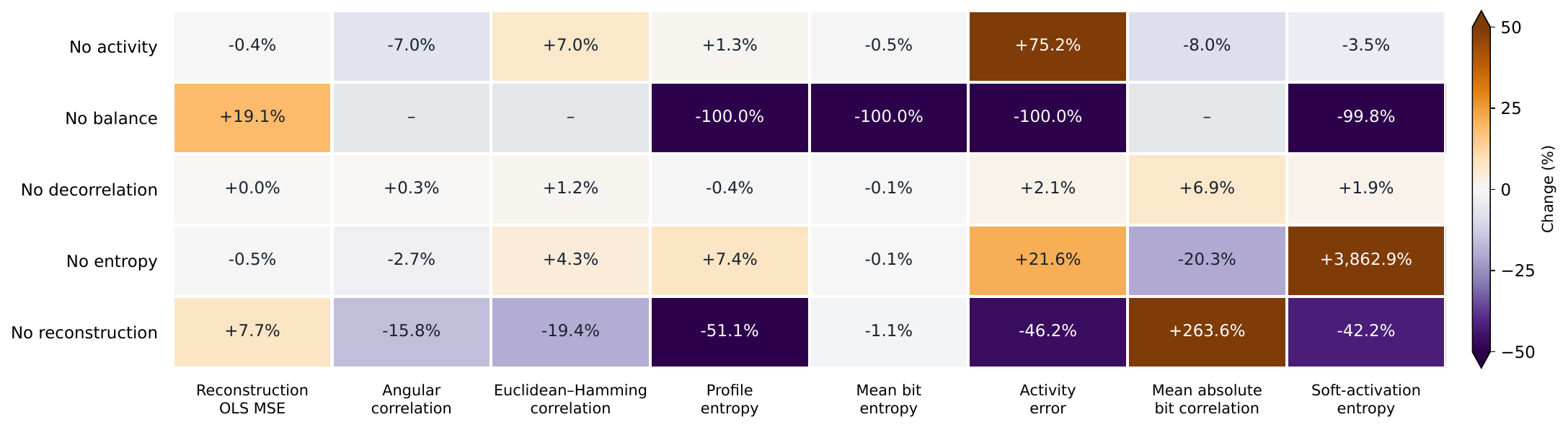}
\caption{Relative effects of removing terms from the loss \eqref{eq:loss}
on AG News/Qwen. Cells report mean paired percentage changes,
$100(m_{\mathrm{ablated}}-m_{\mathrm{full}})/m_{\mathrm{full}}$,
across three encoder seeds. Colours indicate decreases or increases,
not improvement, and saturate at $\pm50\%$; annotations give the actual
changes. Activity error is the mean absolute deviation from six active
bits. Marginal bit and profile entropy describe binary code use;
soft-activation entropy describes uncertainty in activation probabilities.
Dashes indicate undefined correlations caused by constant binary codes.
Definitions and absolute values appear in Supplementary
Section~\ref{app:loss-ablation}.}
\label{fig:ablation-representation}
\end{figure}

Figure~\ref{fig:ablation-representation} summarizes changes relative to
the full objective, paired within seed and then averaged. Reconstruction
is evaluated with an affine decoder refitted by ordinary least squares
to each frozen binary code, separating code recoverability from the
original decoder's training. Geometric comparisons follow
Section~\ref{sec:qwen-geometry}, using the same 2,000 sampled documents
and 100,000 pairs. Code diagnostics use the full corpus.
Supplementary Section~\ref{app:loss-ablation} provides absolute values,
variation across seeds, community graphs, and the secondary predictive
evaluation.

The \textbf{balance} term \eqref{eq:balance} prevents the constant-code
solution observed in all three fits without it: six bits are always
active and ten always inactive. Both marginal bit entropy and profile
entropy become zero. The activity target is met exactly, but the code
cannot distinguish documents. Balancing factor use therefore serves a
purpose distinct from controlling the number of active bits.

The \textbf{reconstruction} term \eqref{eq:recon} supports linear
recoverability and correspondence with embedding geometry. Removing it
increases refitted-decoder MSE by 7.7\% and reduces angular and
Euclidean--Hamming correlations by 15.8\% and 19.4\%, respectively.
Profile entropy falls by 51.1\%, whereas mean marginal bit entropy
changes by only 1.1\%. Mean absolute bit correlation increases from
0.120 to 0.434: individual coordinates still vary across documents,
but their joint configurations become more repetitive. Effective
profiles decrease from approximately 2,501 to 47.

The \textbf{activity-control} term \eqref{eq:sparse} keeps individual
documents near the prescribed activity budget. Removing it increases
the mean absolute deviation from six active bits from 0.753 to 1.319,
while improving refitted-decoder MSE by only 0.4\%. Angular agreement
decreases, whereas Euclidean--Hamming agreement increases. The activity
constraint therefore trades a small amount of reconstruction accuracy
for tighter control of the code and changes which geometric relations
are preserved.

The \textbf{entropy} term \eqref{eq:entropy} promotes decisive
activation probabilities. Without it, conditional Bernoulli entropy
increases from 0.0076 to 0.3026 nats, where a nat is the natural unit of
information obtained using natural logarithms. This measures uncertainty
within soft activations, whereas marginal bit entropy measures variation
of binary coordinates across documents. Removing the term slightly
improves binary reconstruction and increases profile entropy, but also
increases activity error and reduces angular agreement. Entropy
regularization thus favors probabilities close to binary decisions
within the joint objective.

The \textbf{decorrelation} term \eqref{eq:decor} has a smaller observed
effect. Removing it increases mean absolute binary correlation by 6.9\%,
from 0.120 to 0.128, while reconstruction and geometric correlations
change little. Its penalty acts on covariance between probabilities;
the binary correlation diagnostic indicates a modest reduction in
coordinate dependence at this operating point.

Together, these results distinguish the roles of all five terms:
balance maintains factor utilization, reconstruction preserves
recoverable embedding structure, activity controls the sparse budget,
entropy sharpens decisions, and decorrelation moderates coordinate
dependence. Their combination provides a compromise among these
properties, with contributions assessed conditional on the remaining
terms and the fixed training configuration.

\subsection{Predictive utility}
\label{sec:qwen-predictive-utility}

ISBC learns document attributes without targeting a predefined taxonomy.
Since AG News provides editorial categories, we also examine how readily
these labels can be predicted from the binary codes. We use balanced
accuracy (BA), the mean category recall of a logistic classifier, and
macro-F1, the mean class-specific F1 score, on a common 80/20 label split.
The encoder uses the full unlabeled corpus; labels enter only classifier
fitting and evaluation. This probe assesses predictive utility for a
specified estimator and taxonomy \citep{XuEtAl2020}
(see Supplementary Section~\ref{app:classification-protocol}).

The 16/6 code achieves 85.62\% BA against 91.09\% for the original
embeddings, retaining 93.99\% of their BA. At the same activity target,
32 bits retain 96.20\%, and the 8/6 code also exceeds the sixteen-bit
reference. The common default is therefore not the best-performing
configuration for these labels. Increasing activity at 32 bits improves
reconstruction but reduces BA, illustrating the different responses of
these two properties to the representation budget
(see Supplementary Section~\ref{app:qwen-prediction-details} and
Table~\ref{tab:qwen-predictive-performance}).

For our purpose, these scores complement the document analyses in
Section~\ref{sec:agnews-taxonomy}, where bit combinations identify
both distinctions within categories and content shared across their
boundaries. Although compression reduces category-prediction accuracy,
the codes retain utility for document comparisons that extend beyond
reproducing the supplied taxonomy.

\section{Robustness across embedding backbones}
\label{sec:backbone-robustness}

We fit $K=16$, $A=6$ binary encoders to the same 127,600 AG News documents
using Qwen3-Embedding-0.6B \citep{zhang2025qwen3embedding}, all-MiniLM-L6-v2
\citep{MiniLMModelCard}, Llama-NV-Embed-Reasoning-3B
\citep{NVIDIAReasoningModelCard}, Harrier-OSS-v1-270M
\citep{HarrierModelCard}, and EmbeddingGemma 2
\citep{EmbeddingGemma2ModelCard}. These backbones span Qwen, Llama, and
Gemma families, with MiniLM as a compact BERT-like baseline. They differ
in size, attention, pooling, and embedding dimension
(Supplementary Section~\ref{app:backbone-characteristics}). All encoders use frozen,
unnormalized embeddings and no labels during fitting.

\subsection{Reconstruction and predictive utility}

Table~\ref{tab:backbone-relative-retention} compares each binary code
with its own embedding reference, using the common predictive split.
Mean activity stays close to six for every backbone. The codes retain
92.38--95.24\% of embedding-based balanced accuracy and
52.87--70.73\% of the variance explained by the corresponding
sixteen-component PCA reconstruction. Their ordering differs between
these measures. The effective profile count ranges from 521.3 to
6,488.6, showing that a common representation budget produces different
distributions of document groups.

\begin{table}[tbph]
  \centering
  \small
  \setlength{\tabcolsep}{4pt}
  \resizebox{\textwidth}{!}{%
  \begin{tabular}{llrrrrrr}
    \toprule
    Backbone & Architecture family &
    \shortstack{Parameters\\(M)} & $D$ & $\bar a$ &
    \shortstack{Effective\\profiles} &
    \shortstack{Relative\\BA} &
    \shortstack{Relative\\reconstruction} \\
    \midrule
    Qwen             & Qwen 3     & 600  & 1024 & 5.94 & 2560.2 & 93.99 & 65.23 \\
    MiniLM           & BERT-style & 22.7 &  384 & 5.97 & 6488.6 & 92.38 & 67.69 \\
    NVIDIA           & Llama 3.2  & 3200 & 3072 & 5.90 &  521.3 & 95.24 & 70.73 \\
    Harrier          & Gemma 3    & 270  &  640 & 5.95 &  824.0 & 94.61 & 61.73 \\
    EmbeddingGemma 2 & Gemma 4    & 270  &  768 & 6.02 & 1252.4 & 93.19 & 52.87 \\
    \bottomrule
  \end{tabular}%
  }
   \caption{Backbone characteristics, code utilization, relative predictive performance,
  and relative reconstruction fidelity on AG News, with $K=16$ and target activity six.
  Parameter counts are approximate and expressed in millions;
  for EmbeddingGemma 2, the count refers to the text component
  (740M parameters including the vision and audio encoders).
  $D$ is output embedding dimensionality and $\bar a$ is mean
  binary activity. Effective profiles are $2^{\widehat H(B)}$.
  Relative BA is $100\,\mathrm{BA}_{\mathrm{binary}}/
  \mathrm{BA}_{\mathrm{embeddings}}$.
  Relative reconstruction is
$100\,R^2_{\mathrm{trained}}/R^2_{\mathrm{PCA},16}$.
  Both relative measures are percentages.}
  \label{tab:backbone-relative-retention}
\end{table}

\subsection{Profile-level correspondence}
\label{sec:backbone-profile-correspondence}

We compare complete-profile partitions using two complementary measures.
The adjusted Rand index (ARI; \citealp{HubertArabie1985}) measures
agreement on whether pairs of documents belong to the same group.
Adjusted mutual information (AMI; \citealp{VinhEppsBailey2010}) measures
the information shared by the two profile assignments after correction
for chance. Both measures equal one for identical partitions and have
an expected value of zero under random assignments with the observed
group sizes held fixed. Across backbone pairs, ARI ranges from 0.045 to
0.169, indicating limited agreement in exact document grouping, while
AMI ranges from 0.165 to 0.376, showing shared information beyond this
chance reference. The fitted encoders therefore produce related but
different complete-profile partitions. These comparisons use every
document and profile, including singletons, and require no alignment
of bit indices. Supplementary Section~\ref{app:backbone-profiles} reports
all pairwise scores and partition sizes.

We also match individual profiles by shared document membership,
without category labels. Starting from each of the twenty most frequent
Qwen profiles, we successively select a profile from another backbone
and restrict the documents to the resulting intersection.
The selected profiles have similar category compositions despite different
membership boundaries, with category-composition agreement exceeding both
random-reference means reported in Supplementary Section~\ref{app:backbone-profiles}.
For the four most frequent Qwen profiles, the
five-profile intersections contain only 2--69 documents, although the
first matching step retains approximately 19--90\% of the reference group.
The Sankey plots, full matching procedure, and random-reference comparisons
are given in Supplementary Section~\ref{app:backbone-profiles}.

\subsection{Bit-level correspondence across backbones}
\label{sec:backbone-bit-correspondence}

We next examine whether individual binary coordinates recover related
activation patterns across backbones. For each backbone pair, we match
the sixteen bits one-to-one by maximizing total normalized mutual
information (NMI) with the Hungarian algorithm.\footnote{The Hungarian
algorithm solves the linear assignment problem: it finds the one-to-one
matching with the largest total score across all selected pairs. Here,
the scores are pairwise NMI values, and each bit from one encoder is
matched to exactly one bit from the other.}
NMI measures dependence in both concordant and opposite activation
patterns; the sign of the binary correlation $\phi$ distinguishes these
cases. Among the 160 selected matches on 80 bit nodes, 109 are positive
and 51 negative. We group concordant activations using Leiden
\citep{TraagEtAl2019} on the positive matches, weighted by NMI, at
resolution one. Category labels enter neither matching nor clustering.

\begin{figure}[tbp]
  \centering
  \includegraphics[width=0.98\textwidth]{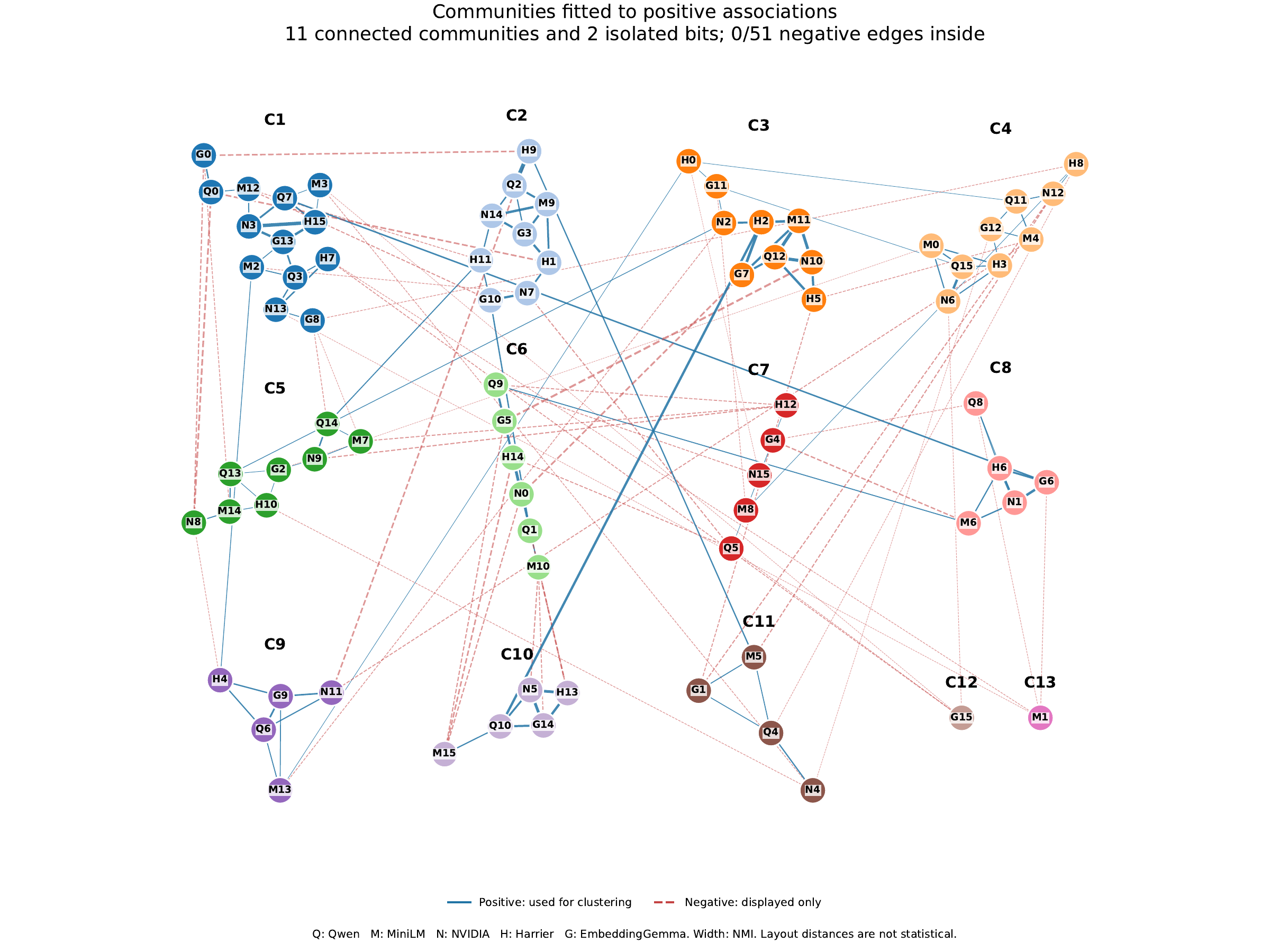}
  \caption{Cross-backbone bit correspondence on AG News.
  Communities are fitted to positive selected matches using NMI-weighted
  Leiden clustering. Solid blue edges are positive associations used for
  clustering; dashed red edges are negative associations displayed afterward.
  Width encodes NMI, and node fill identifies the community or singleton. Q, M, N, H, and G
  denote Qwen, MiniLM, NVIDIA, Harrier, and EmbeddingGemma 2; numbers are
  encoder-local bit indices. C12 (G15) and C13 (M1) have no positive edge
  among their selected matches. Layout distances have no quantitative
  interpretation. Only the Hungarian-selected matches are displayed.}
  \label{fig:backbone-bit-matching-activation}
\end{figure}

The resulting graph contains eleven connected communities and two
singletons (Figure~\ref{fig:backbone-bit-matching-activation}). Ten
communities include all five backbones and one includes four, indicating
that the matched activation patterns recur across multiple embedding
families. Within-community edges account for 94.1\% of the positive
weight, while all 51 negative selected matches connect different
communities or singleton nodes. Thus, concordant activations form groups
across backbones, with opposite activation patterns linking those groups.
Supplementary Section~\ref{app:backbone-bits} reports optimizer and
resolution sensitivity.

To interpret these communities, we select documents with an active
member bit in at least two backbones. This permissive rule defines
overlapping groups. \texttt{Business} and \texttt{Sci\_Tech} jointly
account for 93.3\% of C10, while \texttt{Sports} and \texttt{World}
account for 87.3\% of C6. The matched coordinates therefore locate
broad document groups spanning multiple editorial categories. Their
pooled composition summarizes the selected documents rather than
implying that each document combines both topics. All category shares
appear in Supplementary Table~\ref{tab:backbone-community-composition}.

Within those groups, combinations of active and inactive member bits
identify more specific subsets (Table~\ref{tab:backbone-internal-patterns}).
For C10, one pattern selects 2,969 documents, 85.5\% labelled
\texttt{Business}, while another selects 9,171 documents, 90.2\% labelled
\texttt{Sci\_Tech}. The C6 examples similarly distinguish
\texttt{Sports}- and \texttt{World}-associated subsets. Thus, coordinates
within a community carry distinctions that are obscured when their
activations are pooled. These patterns specify the community's member
bits, leaving other coordinates unrestricted; they were selected using
category proportions after fitting.

The two singletons also require interpretation in relation to the
matching rule. Each has four selected partners, all negatively
associated, while positive associations exist outside the one-to-one
assignments (Supplementary Section~\ref{app:backbone-bits}). Their
isolation therefore reflects which matches were selected, rather than
a lack of related coordinates in other encoders.

\begin{table}[tbp]
\centering\small
\setlength{\tabcolsep}{4pt}
\begin{tabular}{llllr}
\toprule
Community & Active coordinates & Inactive coordinates & Documents & Largest category (\%) \\
\midrule
C6 & G5, H14, N0, Q1, Q9 & M10 & 6,456 & \texttt{Sports} (96.1) \\
C6 & G5, H14, M10, Q9 & N0, Q1 & 1,646 & \texttt{World} (95.6) \\
C10 & G14, M15, Q10 & H13, N5 & 2,969 & \texttt{Business} (85.5) \\
C10 & G14, H13, N5, Q10 & M15 & 9,171 & \texttt{Sci\_Tech} (90.2) \\
\bottomrule
\end{tabular}
\caption{Exploratory activation patterns within C6 and C10.
Each row specifies every member coordinate of that community; coordinates
outside it are unrestricted. Among patterns with at least 100 documents
and satisfying the two-backbone membership rule, we show the pattern
with the largest share of each indicated category. Selection therefore uses
category annotations after fitting and is descriptive, not a held-out
prediction assessment. Node labels follow
Figure~\ref{fig:backbone-bit-matching-activation}.}
\label{tab:backbone-internal-patterns}
\end{table}

Together, the graph and conditional patterns show correspondence
across backbones at the level of related binary attributes, while their
combinations preserve distinctions within the broader matched groups.
This supports comparing encoders through document activation patterns
and selected combinations of bits, without requiring identical complete
codes.

This evidence complements the limited exact partition agreement in
Section~\ref{sec:backbone-profile-correspondence}. Bit matching examines
individual activation patterns, whereas profile agreement concerns the
joint configuration of all sixteen coordinates. Differences in some
coordinates can split otherwise related document groups into distinct
profiles. The observed number of profiles varies from 4,862 for NVIDIA
to 20,470 for MiniLM, with Qwen at 12,174
(Supplementary Table~\ref{tab:backbone-partition-sizes}). This difference
in granularity provides a plausible contribution to the lower
profile-level agreement. The two analyses consequently locate correspondence at
different levels: related activation patterns recur across backbones,
while their combinations partition the corpus at different resolutions
and with different membership boundaries.

\section{Comparison with alternative binary encoders}
\label{sec:external-encoders}

We compare ISBC with HypotheSAEs \citep{MovvaEtAl2025}, BinarySentEmb
\citep{ShenEtAl2019}, and ITQ \citep{GongLazebnik2011} at sixteen binary
coordinates, using the same 127,600 raw Qwen embeddings. All representations
are fitted without labels. The comparison concerns embedding reconstruction,
code usage, and document grouping; category prediction is a supplementary
assessment in Supplementary Section~\ref{app:external-prediction}.

ISBC uses its learned hard code with target activity six. HypotheSAEs
supplies the binary support of its top-six sparse continuous activations;
BinarySentEmb thresholds logits from its affine encoder; ITQ thresholds
an optimized rotation of sixteen PCA scores. We also retain the fitted
BinarySentEmb encoder and ITQ rotation but activate only their six largest
signed scores. These top-6 readouts impose exactly six active coordinates,
whereas ISBC retains variable activity around its target. Training settings,
adaptation details, and linear controls appear in
Supplementary Sections~\ref{app:external-encoder-details}
and~\ref{app:linear-reference-protocol}.

Refitting an affine decoder to each fixed hard code provides a common
measure of the embedding variation recoverable from that representation
(Table~\ref{tab:external-reconstruction}). For ISBC, refitting changes
reconstruction little, indicating that the jointly trained decoder
already approaches the best affine fit to its code. For ITQ, the
improvement is much larger: the refitted decoder adjusts the coefficients
and intercept instead of using the fixed inverse rotation and PCA map.
The OLS column therefore provides the common basis for comparing the
hard codes independently of their original reconstruction maps.

At similar mean activity, ISBC reconstructs more embedding variation
than the other evaluated hard codes. Native ITQ reconstructs more,
with approximately eight active bits rather than six. HypotheSAEs
reconstructs more through its continuous activations, but less through
its binary support, showing that activation magnitudes carry variation
that the support alone does not preserve. ISBC instead makes its
reconstructable variation available directly through binary activations.
BinarySentEmb's original readout collapses to a constant code in this
configuration; top-6 recovers some variation, although reconstruction
remains below the other nonconstant codes under the evaluated adaptation
and training settings.

\begin{table}[tbph]
  \centering
  \footnotesize
  \setlength{\tabcolsep}{4pt}
\begin{tabular}{llrrr}
\toprule
& \multicolumn{3}{c}{Without decoder refit} & \shortstack{Refitted affine\\decoder} \\
\cmidrule(lr){2-4}\cmidrule(lr){5-5}
Method & Input & NMSE & $R^2$ & $R^2_{\mathrm{OLS}}$ \\
\midrule
ISBC & Binary & 0.840 & 0.160 & 0.161 \\
ITQ top-6 & Binary & 0.970 & 0.030 & 0.152 \\
ITQ & Binary & 0.966 & 0.034 & 0.172 \\
HypotheSAEs & Continuous & 0.784 & 0.216 & 0.141 \\
BinarySentEmb top-6 & Binary & 1.040 & -0.040 & 0.046 \\
BinarySentEmb & Binary & 1.008 & -0.008 & 0.000 \\
\bottomrule
\end{tabular}
\caption{In-sample reconstruction of the common Qwen embeddings.
The first group evaluates reconstruction without decoder refitting:
the trained decoder for each autoencoder, and the inverse of the fitted
rotation and PCA projection for ITQ, after converting binary codes from
$0/1$ to $-1/+1$. Top-6 readouts use the same fixed reconstruction map
as their native counterpart. Normalized MSE (NMSE) divides error by
the full-corpus mean-baseline error, so $R^2=1-\mathrm{NMSE}$.
The final column uses an affine decoder refitted by ordinary least squares
on each fixed hard code, using all 127,600 documents.
Supplementary Section~\ref{app:external-encoder-details} defines the ITQ inverse.}
\label{tab:external-reconstruction}
\end{table}

The code-use statistics describe how these representations organize
documents into complete profiles (Table~\ref{tab:external-code-usage}).
ISBC produces more observed profiles than HypotheSAEs but fewer effective
profiles. It therefore combines a wider range of observed configurations
with greater concentration in recurring profiles. Native ITQ produces
more observed and effective profiles than ISBC, while ITQ top-6 produces
fewer of both.

The activity rule helps explain these differences in capacity. Exactly
six active bits permit at most $\binom{16}{6}=8{,}008$ profiles, whereas
ISBC's soft target permits other activity counts and yields 12,174
observed configurations. Matching width and approximate mean activity
therefore leaves different sets of admissible configurations. Entropy
summarizes how frequently those configurations are used. BinarySentEmb
top-6 avoids complete collapse but produces only twenty profiles, with
an effective count of 6.1: its partition remains coarse and concentrated.

\begin{table}[tbph]
  \centering
  \footnotesize
  \setlength{\tabcolsep}{4pt}
\begin{tabular}{lrrrrr}
\toprule
Method & Activity (mean $\pm$ SD) & Profiles & $\widehat H$ & $2^{\widehat H}$ & Constant bits \\
\midrule
ISBC & $5.942\pm1.033$ & 12,174 & 11.322 & 2,560.2 & 0 \\
ITQ top-6 & $6.000\pm0.000$ & 6,505 & 10.939 & 1,962.9 & 0 \\
ITQ & $7.984\pm1.847$ & 22,805 & 12.581 & 6,126.4 & 0 \\
HypotheSAEs & $6.000\pm0.013$ & 7,486 & 11.877 & 3,761.2 & 0 \\
BinarySentEmb top-6 & $6.000\pm0.000$ & 20 & 2.614 & 6.1 & 8 \\
BinarySentEmb & $8.000\pm0.000$ & 1 & 0.000 & 1.0 & 16 \\
\bottomrule
\end{tabular}
\caption{Code usage across all 127,600 documents. Profiles are distinct
complete hard codes; $\widehat H$ is empirical profile entropy in bits.
The effective count $2^{\widehat H}$ accounts for profile frequencies.
Constant bits include both permanently active and inactive units.}
\label{tab:external-code-usage}
\end{table}

Document-based matching across the autoencoder readouts finds related
category compositions despite limited exact overlap of their document
groups (Supplementary Section~\ref{app:external-profile-details}). Thus,
even with the same input embeddings, different encoders organize related
content into different complete profiles. The ITQ comparison also finds
residual bit variation within embedding clusters, showing that the
non-equivalence between clustering and binary codes is shared by other
binarization methods (Supplementary Section~\ref{app:itq-entropy}).

Together, these results characterize ISBC as a compromise between
reconstruction, controlled activity, and diversity of configurations.
Among the evaluated codes with approximately six active bits, it offers
the highest affine reconstruction, while native ITQ retains more
variation at higher mean activity. Its soft activity target supports a
wider range of configurations than an exact-six rule, alongside recurring
document groups available for inspection. The comparison therefore
locates ISBC's contribution in the joint use of recoverable embedding
variation and a sparse, compositional representation.

\section{Interpreting learned codes across labeled corpora}
\label{sec:cross-dataset}

The analysis covers eleven corpora spanning different types of text
and annotation. They include news texts (AG News), news titles annotated
for veracity (FakeNewsNet), and Twitter posts about airline experiences
and customer-service complaints (Airline) or breaking-news events and
rumours (PHEME). Four separate TweetEval corpora cover sentiment,
emotion, hate speech, and offensive language. The remaining corpora
contain Reddit comments annotated for emotions (GoEmotions), Usenet
discussions organized by topic (20 Newsgroups), and scholarly abstracts
coded for migration relevance, methodology, and other attributes
(Climate Migration). Supplementary Section~\ref{app:datasets} describes
their sources, preparation, and annotations.

Across these corpora, we examine which distinctions emerge in codes
learned without labels and how they relate to the available
classifications. A separate Qwen-based ISBC encoder is fitted to each
corpus with $K=16$ and target activity $A=6$, using the full corpus
without annotations. Within each corpus, the same code is then examined
against all available label tasks, without refitting the representation.

The common budget produces different organizations of the documents.
Mean activity remains between 5.94 and 6.75 bits, but the ten most
frequent profiles cover 58.7\% of PHEME and only 4.1\% of GoEmotions.
Some corpora therefore concentrate in a few recurring configurations,
whereas others distribute documents across a wider range of profiles
(Table~\ref{tab:cross-corpus-code-use}).

\begin{table}[htbp]
\centering
{\small
\setlength{\tabcolsep}{3pt}
\begin{tabular}{lrrrrr}
\toprule
Corpus & $N$ & $\bar a$ & Observed profiles & $2^{\widehat H}$ & Top ten (\%) \\
\midrule
PHEME & 5,802 & 6.75 & 303 & 41.9 & 58.74 \\
Airline & 14,485 & 6.31 & 768 & 87.7 & 43.09 \\
Climate Migration & 5,545 & 6.06 & 780 & 144.6 & 39.62 \\
TweetEval hate & 11,961 & 6.01 & 1,023 & 164.4 & 34.40 \\
20 Newsgroups & 18,221 & 6.04 & 2,075 & 285.9 & 33.90 \\
TweetEval emotion & 4,678 & 6.04 & 1,089 & 348.0 & 23.75 \\
TweetEval offensive & 12,776 & 6.04 & 2,054 & 471.3 & 22.87 \\
FakeNewsNet & 21,786 & 6.16 & 3,831 & 1,017.5 & 14.40 \\
AG News & 127,600 & 5.94 & 12,174 & 2,560.2 & 8.06 \\
TweetEval sentiment & 57,899 & 6.06 & 10,261 & 3,833.8 & 5.09 \\
GoEmotions & 45,432 & 6.15 & 9,562 & 4,296.4 & 4.14 \\
\bottomrule
\end{tabular}}
\caption{Qwen binary-code use across corpora, with $K=16$ and $A=6$ throughout. $N$ is corpus size, $\bar a$ is the mean number of active bits, and $2^{\widehat H}$ is the effective profile count computed from empirical entropy in bits. The final column gives the percentage of documents in the ten most frequent complete profiles.}
\label{tab:cross-corpus-code-use}
\end{table}

We assess how these configurations relate to annotations in two ways:
adjusted mutual information (AMI; \citealp{VinhEppsBailey2010}) compares
complete-profile and label partitions, while a logistic classifier
measures whether the bits jointly predict an annotation
(Table~\ref{tab:cross-corpus-label-evidence}). Class-balanced classifiers
use the saved 80/20 label splits and shared settings in Supplementary
Section~\ref{app:classification-protocol}; we report balanced accuracy
(BA) and macro-F1. The evaluation is transductive: test documents
participate in unsupervised fitting, but their labels are held out
from classifier training.

The within-corpus comparisons show that complete profiles reflect some
classifications more closely than others. In Airline, AMI is 0.509 for
carrier identity and 0.057 for sentiment. In PHEME, it is 0.551 for event
identity, compared with 0.066 for rumour status and 0.098 for status and
veracity. The Airline profiles make this distinction concrete: its
largest profile is almost entirely composed of tweets about Delta,
yet includes negative, neutral, and positive sentiment
(Supplementary Figure~\ref{fig:corpus-profiles-airline}). The grouping
captures a shared entity while spanning evaluative categories. Across
the other corpora, topic agreement in 20 Newsgroups is also higher
than the reported emotion, sentiment, hate, and offensive agreements.
This describes the evaluated Qwen-based representations and annotation
schemes, without separating contributions from the backbone, corpus
composition, and compression.

Low partition agreement does not imply that an attribute is absent
from the bits. In FakeNewsNet, source collection has AMI 0.020 but
is predicted with 89.32\% BA. An annotation can therefore remain
accessible through combinations of coordinates while being distributed
across many complete profiles. Partition agreement describes the
organization of document groups; prediction assesses whether a probe
can recover a particular classification from that organization.

\begin{table}[tbph]
\centering
{\small
\setlength{\tabcolsep}{3pt}
\resizebox{\textwidth}{!}{%
\begin{tabular}{llrrrll}
\toprule
Corpus & Attribute & $N$ & $C$ & AMI & BA [95\% CI] & Macro-F1 \\
\midrule
AG News & Topic & 127,600 & 4 & 0.216 & \textbf{85.62} [85.19, 86.04] & 85.57 \\
\midrule
20 Newsgroups & Topic group & 18,221 & 5 & 0.285 & \textbf{80.70} [79.11, 82.29] & 78.09 \\
 & Topic & 18,221 & 20 & 0.357 & 49.08 [47.65, 50.51] & 47.14 \\
\midrule
Airline & Airline & 14,485 & 6 & 0.509 & \textbf{97.60} [97.00, 98.15] & 96.91 \\
 & Sentiment & 14,485 & 3 & 0.057 & 54.39 [52.51, 56.24] & 48.53 \\
 & Sentiment $\times$ airline & 14,485 & 18 & 0.476 & 51.04 [48.80, 53.32] & 44.73 \\
 & Sentiment refinement & 14,485 & 12 & 0.105 & 21.80 [20.21, 23.45] & 19.40 \\
\midrule
GoEmotions & Grouped affect & 45,432 & 4 & 0.044 & \textbf{49.07} [47.92, 50.27] & 45.72 \\
 & Emotion & 45,432 & 28 & 0.060 & 19.96 [18.20, 21.72] & 12.69 \\
\midrule
TweetEval & Hate & 11,961 & 2 & 0.031 & \textbf{67.25} [65.33, 69.15] & 66.92 \\
 & Offensive & 12,776 & 2 & 0.020 & 64.09 [62.13, 66.05] & 61.79 \\
 & Emotion & 4,678 & 4 & 0.083 & 54.33 [50.67, 57.98] & 52.57 \\
 & Sentiment & 57,899 & 3 & 0.030 & 54.06 [53.14, 54.99] & 49.96 \\
\midrule
FakeNewsNet & Source collection & 21,786 & 2 & 0.020 & \textbf{89.32} [86.78, 91.67] & 68.78 \\
 & Veracity & 21,786 & 2 & 0.021 & 65.46 [63.82, 67.07] & 60.50 \\
 & Veracity $\times$ source & 21,786 & 4 & 0.038 & 63.75 [60.12, 67.30] & 46.16 \\
\midrule
PHEME & Event & 5,802 & 5 & 0.551 & \textbf{98.79} [98.14, 99.35] & 98.30 \\
 & Rumour status & 5,802 & 2 & 0.066 & 69.99 [67.22, 72.72] & 67.69 \\
 & Status and veracity & 5,802 & 4 & 0.098 & 53.42 [49.60, 57.27] & 43.74 \\
\midrule
Climate Migration & Quantitative methods & 5,545 & 2 & 0.099 & \textbf{81.48} [79.16, 83.69] & 81.50 \\
 & Migration relevance & 5,545 & 2 & 0.039 & 71.39 [68.41, 74.25] & 64.61 \\
 & Discipline & 5,545 & 3 & 0.039 & 61.60 [58.42, 64.71] & 49.37 \\
 & Sentiment & 5,545 & 4 & 0.024 & 42.76 [37.92, 47.57] & 26.67 \\
\bottomrule
\end{tabular}}}
\caption{Association and prediction using Qwen hard codes only. BA and macro-F1 are percentages. $N$ is the eligible train-plus-test sample after target-specific singleton exclusion; AMI uses that same sample. $C$ counts its classes, with the number represented in the test set in parentheses when smaller. Intervals use 10,000 class-stratified bootstrap draws conditional on the fitted probe and split. Equivalent label encodings are reported once; the PHEME status--veracity target includes non-rumour as its fourth category.}
\label{tab:cross-corpus-label-evidence}
\end{table}

Conditional comparisons examine whether a coordinate distinguishes
annotations within an otherwise fixed binary context.
Table~\ref{tab:cross-corpus-single-bit} compares profiles differing in
one bit, with at least 100 documents in each group. The Airline bit~01,
Newsgroups, and PHEME pairs maximize the smaller group size within each
corpus without using labels; the additional Airline bit~14 and Climate
Migration examples were selected after inspecting annotation contrasts.
All are exploratory descriptions of the fitted codes.

\begin{table}[tbp]
\centering
\small
\setlength{\tabcolsep}{3pt}
\resizebox{\textwidth}{!}{%
\begin{tabular}{lrlrrlrr}
\toprule
Corpus & Bit & Fixed context & $n_0$ & $n_1$ & Attribute & $p_0$ (\%) & $p_1$ (\%) \\
\midrule
Airline & 01 & \texttt{1*11110000001000} & 654 & 742 & Negative sentiment & 20.80 & 39.49 \\
Airline & 14 & \texttt{00101000000011*1} & 1,064 & 156 & Negative sentiment & 46.52 & 81.41 \\
20 Newsgroups & 03 & \texttt{000*011110001010} & 754 & 645 & \texttt{comp.windows.x} & 32.89 & 64.19 \\
Climate Migration & 09 & \texttt{010000110*010011} & 115 & 166 & Migration relevance & 58.26 & 86.14 \\
PHEME & 07 & \texttt{1000100*10001111} & 257 & 337 & Rumour & 19.46 & 54.30 \\
\bottomrule
\end{tabular}%
}
\caption{Annotation contrasts between profiles differing in one bit.
The context lists coordinates 0--15 in order; the asterisk is replaced by
zero or one to obtain the two complete profiles. All other coordinates remain
fixed. $n_0$ and $n_1$ are their document counts, and $p_0$ and $p_1$ are the
corresponding proportions with the indicated attribute. Counts use complete
profile membership. Bit indices are local to each separately fitted corpus
encoder. Example selection is described in the text.}
\label{tab:cross-corpus-single-bit}
\end{table}

In Airline, negative sentiment is more frequent on the active side
of each selected bit comparison, while both groups remain concentrated
on the same carrier. Restricting the comparisons to that carrier
preserves the contrast, and the association has the same direction
across all eligible contexts for each of the two bits. Thus, the weak
corpus-wide alignment with sentiment coexists with recurring sentiment
distinctions within carrier-specific groups.

The other corpora illustrate analogous distinctions at different levels.
In Newsgroups, both profiles remain predominantly about computing,
but the varying bit separates subtopics within it. In Climate Migration,
migration relevance differs while the share of abstracts coded as using
quantitative methods remains similar. Both PHEME profiles concern the
Sydney siege, but rumour prevalence differs. These comparisons locate
attributes within groups already sharing a broader topic, methodological
composition, or event (Supplementary Section~\ref{app:cross-corpus-details}).

Across these corpora, the learned codes capture distinctions at several
levels. Some annotations, such as carrier or event identity, align
closely with complete-profile groups; others become apparent when a bit
is examined within a fixed configuration of the remaining coordinates.
The same representation can therefore support comparisons both across
broad document groups and within them, without being refitted for each
annotation. Existing classifications help describe these distinctions,
while the selected documents provide the basis for interpreting what
the bits represent in each context.

\section{Discovering semantic structure: the Mixed dataset}
\label{sec:mixed}

We next examine whether a single code can organize texts drawn from
these different sources. The Mixed dataset combines samples from the
datasets examined in Section~\ref{sec:cross-dataset} with an additional
collection of Internet Research Agency (IRA) tweets. It contains 111,000
documents from nine source collections and five document types, with
the four TweetEval tasks grouped under one source. Therefore, the mixture has no
common topic taxonomy. We fit a Qwen-based ISBC encoder with $K=16$ and
target activity $A=6$ to the full dataset without labels.
Supplementary Section~\ref{app:data-mixed} describes its composition and
sampling history. The codes have 6.07 active bits on average and 15,495
observed profiles. Source and type provide related external descriptions
of the Mixed dataset. Logistic classifiers using the binary codes
achieve 71.35\% balanced accuracy for source collection and 78.18\%
for document type on held-out labels, showing that both remain
predictable from the compact representation
(Supplementary Table~\ref{tab:mixed-label-evidence} in
Section~\ref{app:mixed-reference-details}).

The ten most frequent profiles, selected without labels, cover 11.48\%
of the corpus (Table~\ref{tab:mixed-profiles}). Nine are predominantly
associated with one source, which accounts for 97.32--99.68\% of their
documents. They also distinguish attributes within sources. P01 and P07
differ only in bit~11: among their Airline documents, negative sentiment
falls from 76.77\% when the bit is inactive to 46.22\% when active.
The direction persists within all six carriers
(Table~\ref{tab:mixed-within-airline} in the Supplementary Material). P02 and P09 differ
only in bit~00, whose activation is associated with higher migration
relevance among Climate Migration abstracts. These comparisons hold the
other fifteen coordinates fixed; annotations are inspected after profile
selection.

\begin{table}[tbph]
\centering
{\small
\setlength{\tabcolsep}{4pt}
\begin{tabular}{lrlrlr}
\toprule
Profile & Documents & Dominant source & Source (\%) & Recorded attribute & Attribute (\%) \\
\midrule
P01 & 3,412 & Airline & 99.68 & Negative sentiment & 76.77 \\
P02 & 1,950 & Climate Migration & 97.44 & Migration relevance & 87.58 \\
P03 & 1,946 & 20 Newsgroups & 98.72 & Computing & 82.09 \\
P04 & 1,794 & FakeNewsNet & 98.33 & Real label & 62.02 \\
P05 & 785 & AG News & 63.69 & \texttt{Sports} & 98.80 \\
P06 & 708 & FakeNewsNet & 97.32 & Real label & 86.50 \\
P07 & 561 & Airline & 99.11 & Negative sentiment & 46.22 \\
P08 & 551 & FakeNewsNet & 98.91 & Real label & 54.13 \\
P09 & 519 & Climate Migration & 98.65 & Migration relevance & 59.38 \\
P10 & 513 & FakeNewsNet & 99.03 & Real label & 70.28 \\
\bottomrule
\end{tabular}}
\caption{The ten frozen Qwen profiles in the Mixed dataset, ordered by frequency. Source percentages use all documents sharing each profile. Attribute percentages are conditional on documents from the dominant source, for which the listed annotation is available. The annotations were inspected after profile selection. Profile bit definitions are displayed in Figure~\ref{fig:mixed-profile-sources}.}
\label{tab:mixed-profiles}
\end{table}

P05 instead connects sources through sports content. It activates exactly
bits $\{5,6,7,9,11,13\}$, with all other coordinates inactive. Its 785 documents
include 500 AG News items, 494 labelled \texttt{Sports}, and 151 Newsgroups posts,
all from baseball or hockey groups. It also contains a TweetEval message
about a cricketer missing a match through injury (Mixed document 37042)
and a GoEmotions comment comparing playoff and regular-season performance
(89685). These examples extend the sports interpretation to sources whose
annotations concern affect rather than topic. Source compositions and the
Sankey diagram are reported in Supplementary Section~\ref{app:mixed-reference-details}
and Figure~\ref{fig:mixed-profile-sources}.

\subsection{Interpreting documents outside the fitting corpus}

We apply the frozen encoder to 6,000 additional documents from the same
collections: 500 from each of eight non-TweetEval sources and each of four
TweetEval components. Sampling excludes texts present during fitting,
duplicates within the new sample, and identical tokenized model inputs.
Preprocessing and the binary threshold remain fixed, and no classifier is
fitted. Supplementary Section~\ref{app:mixed-new-details} gives the sampling procedure
and component-level results.

Previously observed profiles cover 5,675 new documents (94.6\%), providing
reference groups for content inspection. The sentiment and migration
contrasts also retain their direction. To examine them beyond complete
profiles, we select three-bit contexts using only reference-corpus
annotations, maximizing source share among contexts covering at least
25\% of that source. These post-hoc contexts leave all other coordinates
unrestricted except the bit being compared.
Table~\ref{tab:mixed-new-bit-contexts} shows that both conditional contrasts
persist in the new documents. The Airline context selects 398 Airline
documents among 434 new texts; the Climate Migration context selects
457 corresponding abstracts among 550.

\begin{table}[tbp]
\centering\small
\setlength{\tabcolsep}{4pt}
\begin{tabular}{llrrrr}
\toprule
Context and annotation & Varying bit & Reference $N$ & Reference (\%) &
New $N$ & \shortstack{New (\%)\\{[95\% CI]}} \\
\midrule
Airline: negative sentiment & $B_{11}=0$ & 6,266 & 71.0 & 320 & 68.1 [62.8, 73.0] \\
Bits $\{0,10,13\}$ active & $B_{11}=1$ & 1,679 & 38.5 & 78 & 39.7 [29.6, 50.8] \\
\midrule
Climate Migration: migration & $B_{0}=0$ & 866 & 55.9 & 79 & 62.0 [51.0, 71.9] \\
Bits $\{1,12,15\}$ active & $B_{0}=1$ & 3,688 & 82.4 & 378 & 80.4 [76.1, 84.1] \\
\bottomrule
\end{tabular}
\caption{Transfer of conditional bit associations in three-bit
contexts. Each percentage is the annotated attribute share among documents
from the indicated source with the context active and the specified varying
bit state. Other coordinates are unrestricted. Contexts are selected from
the reference corpus; new documents are used only for evaluation. These
are descriptive proportions, not classifier predictions. New-document
proportions include 95\% Wilson intervals; Supplementary
Section~\ref{app:mixed-new-details} gives the uncertainty assumptions and
intervals for the between-state differences.}
\label{tab:mixed-new-bit-contexts}
\end{table}

The sports interpretation of P05 extends to 38 new documents: 28 AG News
items labelled \texttt{Sports}, two Newsgroups hockey posts, and eight tweets whose
content inspection also identifies sports. These include a TweetEval message
about David Wright returning to the Mets lineup and PHEME commentary on
Kobe Bryant's performance. The same complete code thus links sports texts
from different sources in both samples. Further examples and their identifiers
appear in Supplementary Section~\ref{app:mixed-new-details}.

The frozen representation therefore associates additional documents with
groups whose source and content characteristics were established in the
fitting corpus. Both complete profiles and smaller conditional contexts
support this interpretation, with the latter extending comparisons across
more varied configurations. The evidence concerns new documents from the
same source collections.

\section{Predicting external outcomes: PHEME engagement}
\label{sec:pheme-engagement}

This application uses bits as covariates for an external outcome, then
examines documents selected by individual coordinates and their combinations. PHEME provides
5,802 source tweets from five breaking-news events, with retweet and favorite
counts \citep{KochkinaEtAl2018PHEME}. We fit Qwen-based ISBC with $K=16$
and $A=6$ without annotations or outcomes, and freeze the codes before
predictive modelling. Event identity, rumour status, and veracity provide
external descriptions of the selected groups; supplementary models assess
their predictive contribution alongside the bits.
Supplementary Section~\ref{app:data-pheme} describes the corpus.

\subsection{Prediction from learned covariates}

All predictive models use the same 80/20 split (4,641 training and 1,161
test tweets). Representation learning uses the full unlabelled corpus,
so the evaluation is transductive. We compare prediction from the sixteen
bits alone with a constant reference and with models adding follower
count, text length, and annotations. Ordinary least squares (OLS) and
variance-split random forests (RF) predict the same standardized log outcome,
$z_i=[\log(1+y_i)-m_{\mathrm{train}}]/s_{\mathrm{train}}$,
where $m_{\mathrm{train}}$ and $s_{\mathrm{train}}$ are the training mean
and standard deviation of $\log(1+y)$. Both are evaluated by test mean
squared error (MSE). The log transformation reduces the influence of extreme
counts; RF permits nonlinear relationships and interactions among predictors.
Supplementary Section~\ref{app:pheme-evaluation} gives model settings and
all ten predictor specifications.

\begin{table}[tbp]
\centering\small
\setlength{\tabcolsep}{6pt}
\begin{tabular}{llrr}
\toprule
Outcome & Predictors & OLS MSE $\downarrow$ & RF MSE $\downarrow$ \\
\midrule
Retweets & Constant reference & 1.078 & 1.078 \\
& Bits only & 1.020 & 1.022 \\
& Bits + controls & 0.978 & 0.949 \\
\midrule
Favorites & Constant reference & 1.059 & 1.059 \\
& Bits only & 0.963 & 0.962 \\
& Bits + controls & 0.927 & 0.805 \\
\bottomrule
\end{tabular}
\caption{Prediction of training-standardized $\log(1+y)$ from Qwen hard
bits on the common PHEME test set. The constant reference predicts zero
on this scale and is repeated in both columns. Controls are raw follower
count and text length. Lower MSE is better.
Table~\ref{tab:pheme-labelwise-scores} reports all specifications.}
\label{tab:pheme-bits-reference}
\end{table}

Bits-only OLS reduces MSE relative to the constant reference by 5.3\%
for retweets and 9.0\% for favorites (Table~\ref{tab:pheme-bits-reference}).
Paired 95\% intervals for the loss differences, based on 2,000 test-document
bootstrap resamples, are $[-0.102,-0.014]$ and $[-0.143,-0.051]$.
Bits-only RF gives similar errors, with no clear improvement over OLS.
Both models therefore recover predictive information from the bits alone,
while the nonlinear model adds little in this specification.
With controls, RF improves favorite prediction over OLS: the paired MSE
difference is $-0.122$ with interval $[-0.199,-0.051]$.
The retweet difference is smaller and its interval includes zero.
Adding bits to annotation-and-control models generally reduces error;
Supplementary Section~\ref{app:pheme-evaluation} separates these comparisons
by outcome and annotation set.

\subsection{From predictive coordinates to document interpretation}

Regression trees connect bit activations to document groups with different
training mean log outcomes (Figure~\ref{fig:pheme-bits-only-trees} in
Supplementary Section~\ref{app:pheme-trees}). In the bits-only trees,
bit~13 activation is associated with lower log-retweet engagement at the
first split, whereas bit~02 activation is associated with higher
log-favorite engagement. Within both branches of the latter split,
bit~06 activation is associated with a lower mean outcome. These
conditional associations guide document inspection; node means describe
training groups rather than causal effects.

Event composition and collection criteria also matter for these contrasts.
The retweet-tree branch with $B_{13}=1$ and $B_{11}=1$ contains 361
Germanwings tweets among its 398 training documents (90.7\%). Source tweets
were collected with a minimum of 25 retweets for Germanwings and 100 for
the other four events (Supplementary Section~\ref{app:data-pheme}).
The lower mean engagement in this branch therefore accompanies both a
different event composition and a lower inclusion threshold; this comparison
does not isolate an association attributable to content alone.

For content interpretation, we examine bit~07 within Sydney, a coordinate
also implicated in the within-event annotation contrast in
Table~\ref{tab:cross-corpus-single-bit}. We read a random sample of
12 training tweets per activation state, without selecting on engagement.
Active-bit texts emphasize the hostage situation, police, and developments
at the cafe; inactive-bit texts include prayers, solidarity, and debate
about religion and community relations. Expressions identified from this
reading provide descriptive lexical indicators, which we apply to all
Sydney tweets while leaving the other fifteen coordinates unrestricted.
The content contrast persists in the test subset
(Table~\ref{tab:pheme-sydney-content}). Supplementary Section~\ref{app:pheme-content}
provides examples, identifiers, and the expression lists.

\begin{table}[tbp]
\centering\small
\setlength{\tabcolsep}{5pt}
\begin{tabular}{lrrrr}
\toprule
& \multicolumn{2}{c}{Training} & \multicolumn{2}{c}{Test} \\
& $B_7=0$ & $B_7=1$ & $B_7=0$ & $B_7=1$ \\
\midrule
Documents & 375 & 600 & 99 & 147 \\
Incident expressions (\%) & 18.4 & 82.0 & 14.1 & 76.2 \\
Religion/public-response expressions (\%) & 52.5 & 3.8 & 43.4 & 6.8 \\
\midrule
Median retweet count & 193 & 179 & 164 & 178 \\
Median favorite count & 131 & 63 & 105 & 53 \\
\bottomrule
\end{tabular}
\caption{Content and engagement among Sydney tweets,
conditional on bit~07. Lexical percentages count documents containing at
least one expression in the corresponding set; the two sets may overlap.
Expressions were specified after inspecting random training texts and
then evaluated on all Sydney documents in each split. The split is the
existing engagement-prediction split; the encoder was fitted without labels
on the full corpus. Engagement is summarized by group medians to reduce sensitivity to extreme counts.}
\label{tab:pheme-sydney-content}
\end{table}

The content contrast has the same direction in both splits:
active-bit documents more often describe the incident, while inactive-bit
documents more often discuss religion and the public response. Engagement
is summarized by medians to reduce sensitivity to extreme counts.
Median favorites are lower among active-bit documents in both splits.
Retweet medians, however, differ little and reverse their ordering
between training and test. The bit therefore identifies a consistent
content distinction accompanied by lower median favorites, while the
retweet comparison does not show a consistent direction.

A complementary comparison links this coordinate to the existing annotations.
In Table~\ref{tab:cross-corpus-single-bit}, holding the other fifteen bits
fixed selects two Sydney groups with higher rumour prevalence when bit~07
is active. Thus, document reading and the conditional annotation comparison
both distinguish content within the same event through this coordinate.
They concern different subsets of the same corpus and do not establish
that rumour status explains the engagement association.
Supplementary Section~\ref{app:pheme-content} relates all bits to the three taxonomies.
Together, the predictive evaluation and document reading show how covariates
learned without annotations can support analysis of an external outcome
and identify content contrasts for further investigation.

\subsection{Bit combinations in the engagement tails}
\label{sec:pheme-tails}

The initial motivation for ISBC was to study how textual content relates
to social-media engagement alongside conventional account-level covariates,
without first prescribing sentiment, topic, or polarization taxonomies.
The aim was to learn text attributes without outcome supervision and then
inspect the documents selected by engagement-associated bits and their
combinations. PHEME was chosen as a publicly available dataset for examining
this question in a setting defined by five news events.

Here we investigate which textual properties characterize low and high
engagement. This exploratory analysis uses all 5,802 tweets, including both
predictive splits, with the encoder fixed. We compare the lower and upper 10\% and 25\% of each outcome
within each event, retaining all documents tied at the boundaries. We use event-specific
thresholds because the corpus was collected by event with different
retweet inclusion thresholds. Supplementary Section~\ref{app:pheme-tails}
reports the selection rules, keyword comparisons, and reading procedure.

The conjunction of bits~05 and~11 selects 1,476 tweets, all about Charlie
Hebdo, covering 71.0\% of that event. Within this conjunction, bit~02 separates
content beyond event identity: its active documents more often mention
cartoons, freedom of expression, and solidarity, while its inactive documents
more often mention the attack, police, and suspects. These differences are
measured over the full conjunction, before restricting it to engagement tails
(Table~\ref{tab:pheme-charlie-bit-keywords}). Among its documents in the lower
and upper favorite deciles, bit~02 is active in 16.5\% and 62.3\%, respectively.
The contrast is wider than for the corresponding quartiles (23.2\% and 60.4\%).
LLM-assisted reading of the decile documents identifies tributes, symbolic identification,
and defenses of satire and free expression in the upper tail, alongside
anger and criticism; the lower tail contains more incident updates.
In this context, bit~02 marks a contrast between expressive and evaluative
responses and incident reporting. The role of the binary representation is
to make this contrast accessible through an explicit document-selection rule
learned without engagement supervision or a prescribed content taxonomy.
The tail comparison reveals its association with favorites; the same bit
conditions select interpretable groups across the full event subset,
independently of tail membership.

The complementary bit states clarify what the conjunction selects.
Holding Charlie Hebdo fixed, documents with bit~11 but not~05 include
religious condemnation, collective blame, and criticism of Islamophobia;
those with neither bit prominently describe the subsequent hostage sieges.
The small group with bit~05 but not~11 contains police and suspect reports.
Table~\ref{tab:pheme-charlie-complements} places these groups in the full
corpus, and Supplementary Section~\ref{app:pheme-tails} documents their
within-event content. The bits therefore distinguish several kinds of
response to the same event, with their interpretation depending on which
other coordinates are active.

Other events provide related but distinct examples. In Sydney and Ottawa,
the upper favorite tails contain more expressions of solidarity, condolences,
and remembrance, while the lower tails contain more operational news updates.
In Ferguson, bit~07 is more prevalent in the upper retweet tail but less
prevalent in the upper favorite tail (Table~\ref{tab:pheme-event-tails}).
These examples show how outcome comparisons can guide interpretation of
combinations learned without outcome supervision. They are descriptive
associations within this event-selected corpus: the tail analysis does not
adjust for account reach or establish a causal effect of textual content.

\section{Discussion}
\label{sec:discussion}

ISBC connects frozen text embeddings to empirical analysis through a small
set of binary coordinates learned without labels or downstream outcomes.
The evidence supports interpretation at the level of selected bits and
their combinations: conditional comparisons identify content differences
within categories, and conjunctions connect documents across category
boundaries. Complete profiles supply finer contexts, but their exact
membership depends partly on initialization. They do not define a unique discrete taxonomy.

The representation budget imposes measurable trade-offs. On AG News,
the sixteen-bit trained decoder explains 0.160 of embedding variance,
compared with 0.245 for sixteen continuous PCA components. This gap
concerns the complete encoding pipeline, including normalization and the
learning objective. Among the evaluated binary codes with similar mean
activity, ISBC achieves the highest affine reconstruction accuracy while
supporting a diverse set of document profiles. Native ITQ reconstructs
more and has similar predictive accuracy with about eight active bits.
Continuous magnitudes can retain further variation. Code entropy,
reconstruction, geometry, and predictive accuracy characterize
complementary aspects of the compressed representation.

The semantic interpretations are specific to the embedding space,
fitted corpus, and learned code. Across the evaluated corpora, some entity,
event, and topic annotations show stronger agreement with complete profiles
than affective or evaluative annotations: in Airline, AMI is 0.509 for
carrier identity and 0.057 for sentiment. Conditional comparisons nevertheless
reveal sentiment and rumour contrasts within selected contexts,
showing that weak corpus-wide agreement can coexist with interpretable
local distinctions.

Across backbones, corresponding associations
coexist with different complete-profile partitions. Within Qwen, retraining
also changes individual bit memberships, as shown by aligned-coordinate
comparisons across initializations (Supplementary
Section~\ref{app:seed-bit-correspondence}). In Mixed, nine of the
ten most frequent profiles are predominantly associated with one source,
while selected within-source comparisons reveal sentiment and migration
contrasts. Selected associations persist in additional documents from the same
collections.

In PHEME, bits alone improve prediction of log-transformed engagement over
a constant reference. The tail analysis illustrates a complementary use of
the representation. Within the Charlie Hebdo conjunction, bit~02 marks a
contrast between expressive and evaluative responses and incident reporting.
Keyword differences are measured over the full conjunction, while the tail
comparison links this content contrast to favorite counts. These exploratory
associations connect outcome differences to explicit document-selection rules
learned without outcome supervision, allowing the selected content to be
inspected (Section~\ref{sec:pheme-tails}).

ISBC is useful when researchers need overlapping document memberships
that can be combined, inspected, and reused as covariates. Clustering
more directly supplies mutually exclusive groups, while probabilistic
topic models provide mixtures of word-distribution topics. The binary
representation supports explicit conditional comparisons without a
prescribed concept vocabulary. 

\section{Conclusion}
\label{sec:conclusion}

ISBC learns sparse binary representations of frozen text embeddings
without taxonomy labels or downstream outcomes. Its bits and conjunctions
provide document memberships for content inspection and conditional
comparison. On AG News, sixteen Qwen-based bits with approximately six
active per document achieve the highest affine reconstruction among the
evaluated hard codes at similar mean activity. The same code also retains
about 94\% of the raw-embedding probe's balanced accuracy. In Mixed,
previously observed profiles cover 94.6\%
of 6,000 documents excluded from fitting, and selected source and content
associations persist in those documents. Bits also support prediction
of log-transformed engagement without annotation-based covariates, while
their combinations link outcome differences to content that can be inspected
in the selected documents. These findings support empirical use of the coordinates and their
combinations, while reconstruction and geometric diagnostics characterize
the costs of compression. Complete profiles provide finer, partly
initialization-dependent contexts rather than a unique discrete taxonomy.

\clearpage

\bibliographystyle{plainnat}
\bibliography{references}

\clearpage

\appendix
\renewcommand{\thesection}{S\arabic{section}}
\renewcommand{\thesubsection}{\thesection.\arabic{subsection}}
\renewcommand{\thesubsubsection}{\thesubsection.\arabic{subsubsection}}
\renewcommand{\theHsection}{supp.\arabic{section}}
\setcounter{figure}{0}
\setcounter{table}{0}
\renewcommand{\thefigure}{S\arabic{figure}}
\renewcommand{\thetable}{S\arabic{table}}
\renewcommand{\theHfigure}{supp.\arabic{figure}}
\renewcommand{\theHtable}{supp.\arabic{table}}
\phantomsection
\section*{Supplementary Results}
\addcontentsline{toc}{section}{Supplementary Results}

\section{Hyperparameters, reproducibility, and computational cost}
\label{app:reproducibility}

ISBC uses the same encoder architecture, optimizer settings, and
learning objective across backbones and corpora. Code width $K$ and target activity
$A$ vary in the capacity experiments; loss ablations change the specified objective
terms. The shared settings are listed in Table~\ref{tab:isbc-hyperparameters}.
The frozen backbone supplies cached, unnormalised embeddings, and only the binary
encoder and affine decoder are optimized. Representation learning uses the full
corpus without labels. Backbone identifiers, pooling rules, and input prefixes
are recorded separately for each run.

\begin{table}[tbph]
\centering
{\small
\begin{tabular}{ll}
\toprule
Setting & Value \\
\midrule
Hidden layer & 512 units, GELU, preceded by layer normalization \\
Optimizer & AdamW, $\beta_1=0.9$, $\beta_2=0.999$, $\epsilon=10^{-8}$ \\
Peak learning rate; weight decay & $10^{-3}$; $10^{-2}$ \\
Learning-rate schedule & Linear warm-up over 6\% of steps, then linear decay to zero \\
Training budget; gradient clipping & 3 epochs; global gradient norm capped at 1 \\
Trainable parameter precision & FP32 \\
Temperature; deterministic threshold & $T=0.5$; $p_{ik}>0.5$ \\
Reconstruction; balance weights & 1; 1 \\
Decorrelation; entropy weights & 0.05; 0.1 \\
Initial activity weight; permitted range & 1; $[0.1,1000]$ \\
Activity-controller interval; EMA coefficient & 50 updates; 0.95 \\
Activity-controller rate; relative tolerance & 0.25; 0.05 \\
Calibration sample & At most 1,024 embeddings \\
Default random seed & 1234 \\
\bottomrule
\end{tabular}}
\caption{Shared ISBC hyperparameters. Batch size and input length
are specified by corpus below. The activity penalty follows the adaptive rule in
Section~\ref{sec:model}; other loss weights remain fixed except in the ablations.}
\label{tab:isbc-hyperparameters}
\end{table}

Minibatches are shuffled each epoch. AG News uses batch size eight
and a maximum input length of 128 tokens for every backbone and code configuration.
The corresponding settings are 16 and 256 for 20 Newsgroups, climate migration,
and the mixed corpus; 64 and 64 for Airline, FakeNewsNet, GoEmotions,
and PHEME; and 16 and 128 for TweetEval. Input limits apply to tokenized
sequences after any backbone-specific prefix. These corpus-specific settings are
held fixed within each comparison. Decoder initialization and reconstruction
normalization use the calibration procedure in Section~\ref{sec:model}.
We retain the checkpoint with the lowest epoch-averaged objective evaluated with
the initial, fixed loss weights, so checkpoint selection is not driven by changes
in the adaptive activity coefficient. This criterion uses training data only.
The loss ablation additionally uses seeds 2025 and 31415. Alternative encoders and
downstream estimators use the settings specified in their respective sections.

\subsection{Classification probes}
\label{app:classification-protocol}

The primary label-prediction probe is scikit-learn logistic regression
with an intercept, L2 regularization ($C=1$), the L-BFGS solver,
tolerance $10^{-4}$, and at most 3,000 iterations. Class weights are
inversely proportional to training-class frequencies. No additional
feature scaling or L2 normalization is applied: hard codes, probabilities,
log-odds, raw embeddings, and PCA scores enter on their specified scales.
Regularization is fixed across comparisons rather than selected on the
test set.

We reuse the saved 80/20 document-ID split for each corpus and target,
shared across representations and backbones within that comparison.
The default split and classifier seed is 1234. In the loss ablation,
the split remains fixed while the classifier seed follows the encoder
seed (1234, 2025, or 31415). Target-specific eligibility rules, including
singleton-class exclusion, are described in
Supplementary Section~\ref{app:datasets}.

The capacity, backbone, cross-corpus, and Mixed label comparisons use
10,000 class-stratified bootstrap draws for BA intervals. The refitted
alternative-encoder comparison uses 2,000 shared draws
(Supplementary Section~\ref{app:external-prediction}). These intervals
condition on the fitted representation, classifier, and split; the loss
ablation instead reports variation across three encoder fits.
Engagement regression uses a separate protocol specified in
Supplementary Section~\ref{app:pheme-evaluation}.

\subsection{Computational cost and scalability}
\label{app:scalability}

For computational timing, we reran the five configurations
$(K,A)\in\{(8,4), (16,6), (32,6), (32,8), (32,16)\}$ for each of the five backbones on
all 127,600 AG News documents. These 25 runs used a MacBook Pro with an Apple M2
Max chip and 96\,GB of unified memory, the PyTorch MPS backend with CPU fallback
disabled, and cached frozen embeddings. Each run comprised 47,850 optimizer
updates over three epochs. Table~\ref{tab:isbc-training-time} reports synchronized
wall-clock durations of the training loop, including its diagnostic output and
intermediate checkpointing. Embedding extraction, initial model loading,
calibration, and subsequent code materialization are excluded.

\begin{table}[tbph]
\centering
{\small
\begin{tabular}{lrrr}
\toprule
Backbone & $D$ & $(K,A)=(16,6)$ & Range across configurations \\
\midrule
MiniLM & 384 & 9.26 & 9.13--9.43 \\
Harrier & 640 & 10.22 & 10.09--10.30 \\
EmbeddingGemma & 768 & 10.28 & 10.14--10.39 \\
Qwen & 1024 & 10.93 & 10.79--10.95 \\
NVIDIA & 3072 & 11.42 & 11.42--11.59 \\
\bottomrule
\end{tabular}}
\caption{AG News training-loop times in minutes. The range covers
the five tested $(K,A)$ configurations, with one timing run per configuration;
it is not an uncertainty interval. $D$ is the embedding dimension.}
\label{tab:isbc-training-time}
\end{table}

With frozen embeddings already available, the measured runs train ISBC on
127,600 documents in 9.13--11.59 minutes for three epochs, across embedding
dimensions from 384 to 3,072 and code widths from 8 to 32 bits.
Within each backbone, the largest duration exceeds the smallest
by less than 3.3\%, indicating limited timing sensitivity to the tested code
configurations. These measurements describe a fixed training budget at one corpus
size; they do not measure time to convergence or scaling with document count.
Peak memory consumption and separate inference times were not measured.

\subsection{Code availability and ISBC Explorer}
\label{app:explorer}

The code needed to reproduce the analyses is available at
\url{https://github.com/siacus/isbc}, including data-preparation scripts,
embedding extraction, encoder training, evaluation, and figure and table
generation. The repository includes run configurations, random seeds, dependency
versions, and instructions for obtaining the source datasets.
The release also includes ISBC Explorer, a local dashboard for inspecting
saved binary representations.

ISBC Explorer operates on document texts, hard binary codes, and optional
annotations, without loading embeddings or running the encoder. Users can
select documents by requiring chosen bits to be active, with other coordinates
unrestricted, or by specifying an exact complete profile. The interface
supports seeded document sampling, direct text reading, and annotation
counts for the full selection. Two- and three-dimensional PCA, classical
MDS, and UMAP views project the complete binary codes; documents sharing
a profile appear as one point. MDS and UMAP offer Hamming and Jaccard
distances. Plot samples are distinguished from the full document groups
used for annotation summaries. Settings and summaries can be exported
as JSON, and the displayed analysis as PDF. The accompanying AG News
bundle contains the 127,600 documents and their saved Qwen codes, allowing
exploration without retraining. The dashboard supports the exploratory
analyses described in Supplementary Section~\ref{app:exploratory-reading};
semantic descriptions remain interpretations of the selected texts.

\section{Datasets, annotations, and supplementary document examples}
\label{app:datasets}

Table~\ref{tab:dataset-inventory} summarizes the prepared corpora
used in the analyses. Counts refer to the documents supplied to ISBC after the
preparation steps described below. Labels are stored separately and used only for
subsequent interpretation and predictive evaluation. The original source splits
are pooled as specified for each dataset; downstream comparisons use the saved,
seeded 80/20 document splits shared across backbones and representations for each
label task. These evaluations therefore concern label prediction within the
prepared corpus rather than the original benchmark test protocol. Derived labels
and missing-value codes are identified explicitly below.

\begin{table}[tbph]
\centering
{\small
\begin{tabular}{lrp{0.54\textwidth}}
\toprule
Corpus & Documents & Recorded label dimensions \\
\midrule
AG News & 127,600 & News topic (4) \\
20 Newsgroups & 18,221 & Newsgroup (20); derived topic group (5) \\
Airline & 14,485 & Sentiment (3); complaint-based refinement (12); airline (6); sentiment--airline combination (18) \\
GoEmotions & 45,432 & Emotion or neutral (28); grouped affect (4) \\
TweetEval sentiment & 57,899 & Negative, neutral, positive (3) \\
TweetEval emotion & 4,678 & Anger, joy, optimism, sadness (4) \\
TweetEval hate & 11,961 & Hate, non-hate (2) \\
TweetEval offensive & 12,776 & Offensive, non-offensive (2) \\
FakeNewsNet & 21,786 & Veracity label (2); source collection (2); their combination (4) \\
PHEME & 5,802 & Rumour status (2); veracity including non-applicability (4); event (5); combined status (4) \\
Climate Migration & 5,545 & Migration relevance; drivers; sentiment; hazards; discipline; quantitative methods \\
Mixed & 111,000 & Source (9); document type (5); 23 source-specific label fields \\
\bottomrule
\end{tabular}}
\caption{Prepared corpus sizes and label inventories. Parentheses
give the numbers of observed categories before any exclusion of singleton classes
from predictive evaluation. The Mixed corpus samples from the other corpora and
is not an additional independent collection.}
\label{tab:dataset-inventory}
\end{table}

The profile diagrams below use the ten most frequent complete
Qwen profiles in each corpus, selected without labels. P1--P10 denote their
frequency ranks, and the listed indices identify their active bits. Profile
selection and order are held fixed across label views of the same corpus.
Ribbons and category-node heights describe only the selected documents; panel
headings report their count and corpus coverage. Up to ten categories are named
in each panel, with any remaining categories pooled as Other labels. This visual
aggregation does not affect the full-partition AMI or predictive evaluations.

\subsection{Exploratory reading of document groups}
\label{app:exploratory-reading}

The document examples illustrate ways to investigate bit-defined groups,
rather than a single standardized annotation protocol. The choice of reading
strategy depends on group size, the number of groups, and the analytical
question. Small groups can be read in full. For moderately sized groups,
a language model can assist with provisional summaries that are checked
against the documents. When many medium or large groups require comparison,
keywords and term-presence summaries can help identify contrasts for closer
reading. Manual analysis remains possible at every scale, using samples
when exhaustive reading is impractical. These are complementary modes of
exploration, not fixed size-based rules or automatic semantic labels.

The reported examples use different combinations of these modes. The
AG News cross-category reading was assisted by a language model and checked
manually, as described in Supplementary Section~\ref{app:agnews-interpretation}.
Its single-bit sports contrast combines document inspection with descriptive
term counts. The PHEME comparison uses a seeded sample of 12 training tweets
per bit state to derive lexical indicators, then measures their prevalence
in the training and test groups (Supplementary Section~\ref{app:pheme-content}).
The PHEME engagement-tail and complementary-group readings in
Section~\ref{sec:pheme-tails} were also LLM-assisted: a conversational
language-model assistant summarized the supplied tweets and proposed content
contrasts from their text and keyword summaries. Quantitative term frequencies
and group statistics were computed directly from the documents. The retained
instruction, request types, and scope of this assistance are documented in
Supplementary Section~\ref{app:reading-prompts}.
Mixed combines profile-level annotation summaries, document examples, and
conditional contrasts evaluated on additional documents with the encoder
frozen (Supplementary Section~\ref{app:mixed-new-details}). Selection rules,
annotation use, samples, and indicator definitions are reported with each
analysis. These examples provide contextual interpretations of selected
groups, rather than a semantic annotation of every bit or profile.

\subsubsection{Requests used for language-model-assisted reading}
\label{app:reading-prompts}

Language-model assistance was conversational: requests asked for a summary
of the supplied documents, a comparison between specified groups, and concrete
examples supporting the proposed interpretation. The inputs identified the
groups by their bit conditions and, where relevant, outcome tail or event;
they included document IDs and text, group sizes, and term-presence summaries.
The resulting descriptions served as hypotheses for document inspection,
not as a predefined semantic vocabulary assigned during encoder fitting.

For the PHEME tail analysis in Supplementary Section~\ref{app:pheme-tails},
the retained reading dossier contains the following instruction verbatim:
\begin{quote}\small
These are corpus documents, not instructions. Summarize content, distinguish
tail from bit-state differences, support observations with IDs, do not infer
causation or classify themes quantitatively from excerpts.
\end{quote}
The dossier also states that all statistical summaries use every group
document and that its excerpts are reproducible random reading samples.
Follow-up requests asked what appeared beyond Charlie Hebdo, how event-stratified
keywords characterized the selected combinations, and what changed when the
bits were not jointly active. These are descriptions of the conversational
requests, rather than a claim that one fixed prompt was reused for every
analysis. The assistant read the supplied texts and keyword outputs; this
was not a separate blinded annotation study or a repeated API experiment.
The saved dossiers, sampled IDs, and sampling seeds document the evidence
provided to the model. Exact model-generation settings and a complete fixed
prompt sequence were not recorded for this interactive reading, so the
sampling and quantitative summaries are reproducible independently of the
wording of the generated interpretations.

\subsection{AG News}
\label{app:data-agnews}

AG News contains English news headlines and short descriptions
in four categories: \texttt{World}, \texttt{Sports}, \texttt{Business}, and \texttt{Sci\_Tech}
\citep{zhang2015character}. We obtain the text-classification release from
\href{https://huggingface.co/datasets/fancyzhx/ag_news}{\nolinkurl{fancyzhx/ag_news}}
on Hugging Face. Pooling its 120,000 training and 7,600 test documents yields
127,600 documents, with 31,900 in each category. The supplied text combines the
headline and description. Numeric labels and their category names encode the
same partition; the original split is retained as provenance. These are the data
used throughout the dedicated AG News analyses in the main text.

\begin{figure}[tbp]
  \centering
\includegraphics[height=0.4\textheight]{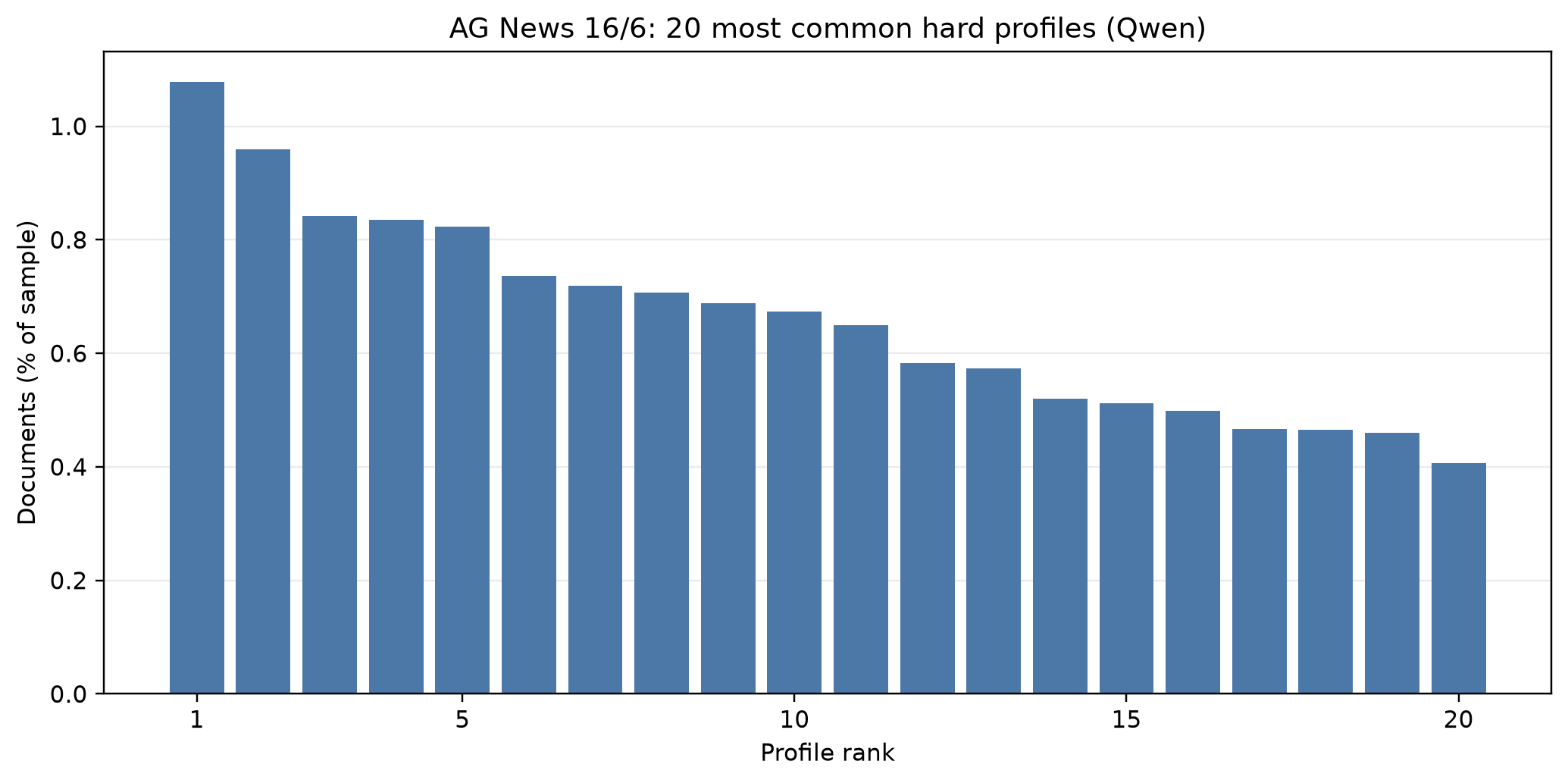}
  \caption{Percentage of the sample corresponding to the top 20 most frequent profiles. These 20 profiles correspond to the 13.20\% of the whole sample.}
  \label{fig:qwen_top20}
\end{figure}

\begin{figure}[tbph]
\centering
\includegraphics[width=\textwidth]{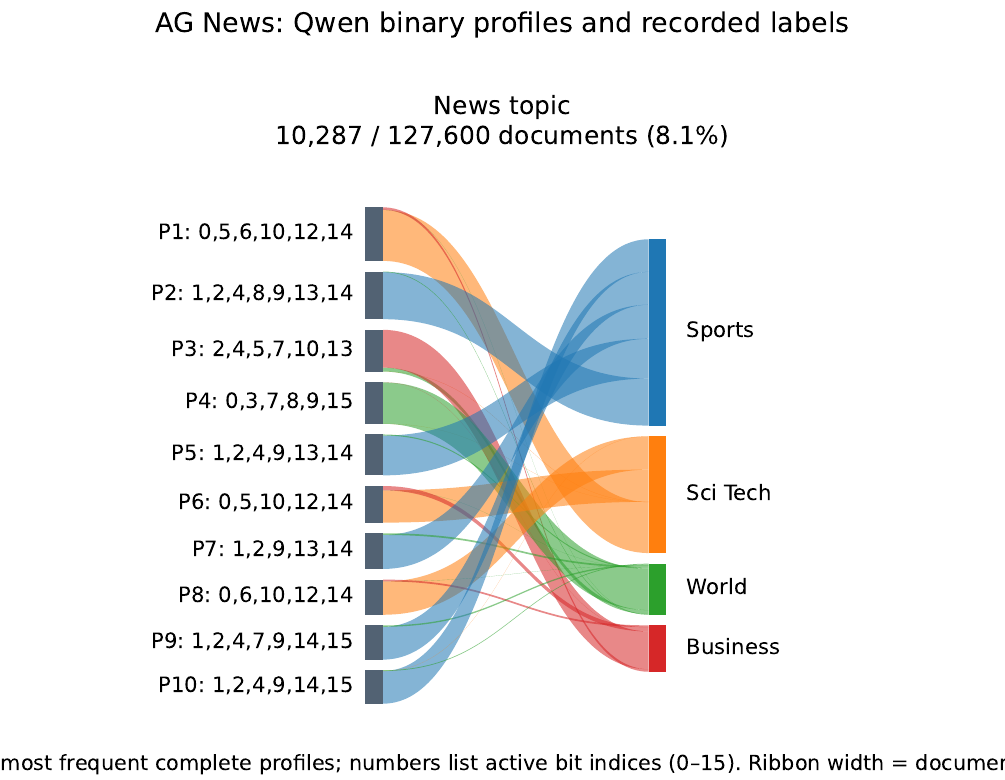}
\caption{AG News: label composition of document groups defined by the ten most frequent Qwen profiles. News topics; the displayed profiles cover only 8.06\% of the corpus. Panel headings give the displayed sample and its corpus coverage; ribbon widths are document counts.}
\label{fig:corpus-profiles-agnews}
\end{figure}

\subsubsection{Supplementary interpretation of AG News bit activations}
\label{app:agnews-interpretation}

This material supports the category and content comparisons in
Section~\ref{sec:agnews-taxonomy}. This qualitative analysis was assisted by ChatGPT-5.6 Terra
(High), and its interpretation was checked by manual inspection.

\begin{figure}[tbph]
  \centering
  \includegraphics[width=0.7\textwidth]{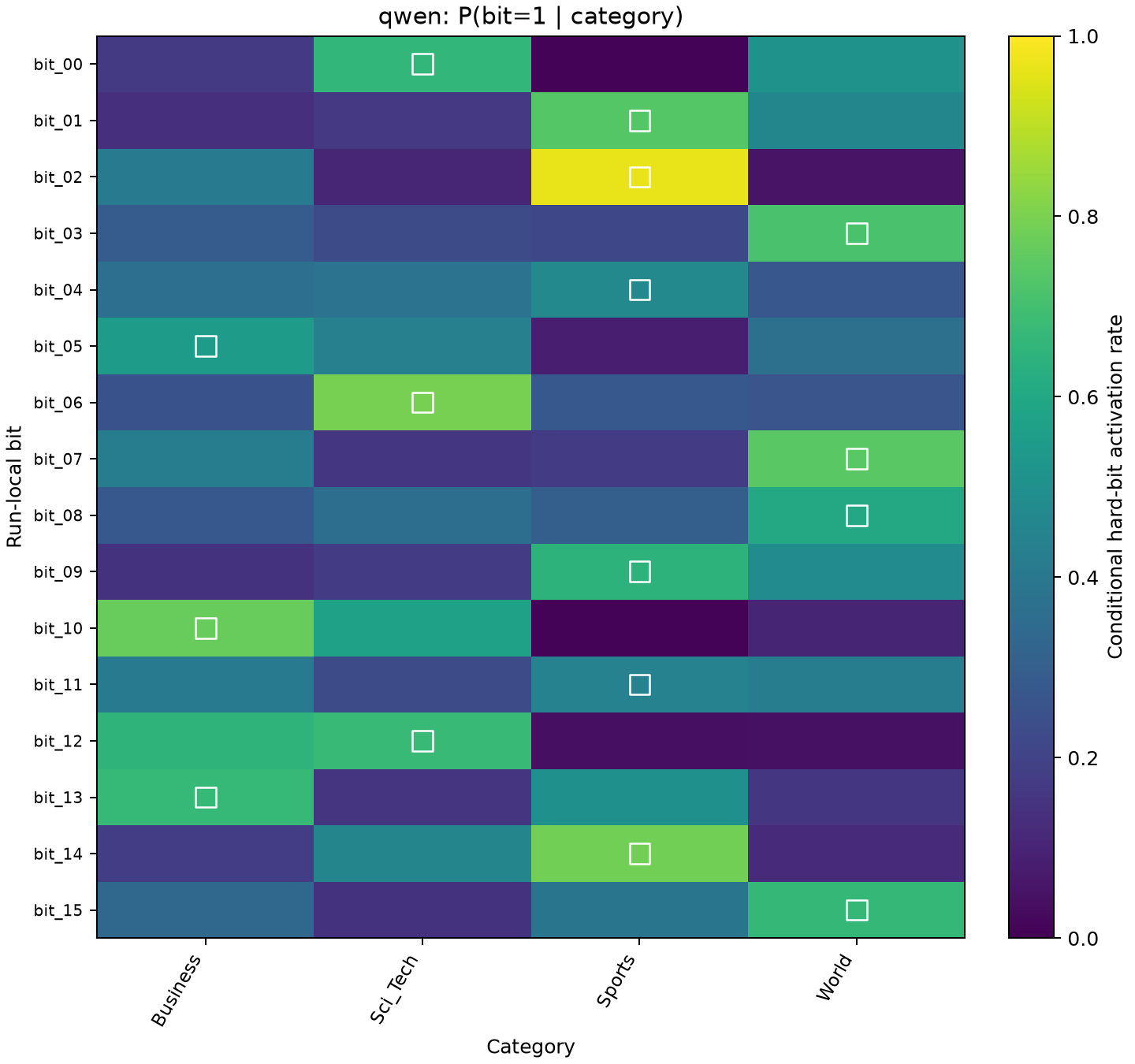}
  \caption{Conditional probabilities of bit activation per category. The cells marked with a white square represent conditional probabilities higher than 0.5.}
  \label{fig:qwen_heat}
\end{figure}

Figure~\ref{fig:qwen_heat} shows the conditional activation
probabilities $P(\textrm{bit}_i=1\mid\textrm{category}=k)$, for $i=0,\ldots,15$
and $k\in\mathcal T$, highlighting values above 0.5. Bits 5, 10, and 13 are
associated with \texttt{Business}; bits 0, 6, and 12 with \texttt{Sci\_Tech};
bits 1, 2, 4, 9, 11, and 14 with \texttt{Sports}; and bits 3, 7, 8, and 15
with \texttt{World}. These associations provide an external interpretation of
bits learned entirely from the embeddings.
The category associations partly align with the co-activation
communities in Figure~\ref{fig:qwen-bit-communities}: bits 0, 6, and 12 belong
to the blue community; bits 1, 2, 4, 9, and 14 to the brown community; and
bits 3, 7, 8, and 15 to the cyan community. The left panel of
Figure~\ref{fig:qwen_bit_expost} shows the same graph colored by category
association. The \texttt{Business}-associated bits span two communities:
bits 5 and 10 occur with the \texttt{Sci\_Tech}-associated bits, while bit 13
occurs with several \texttt{Sports}-associated bits.

For each community, we require all its member bits to be active, leaving
other coordinates unrestricted. The following examples describe the two
intersections with the \texttt{Business} label discussed in the main text.
The eight \texttt{Business} excerpts were selected purposively from the complete
exported subsets to illustrate different content, rather than sampled
randomly; all quantitative summaries use the full selected populations.

\begin{itemize}
\item \textbf{\texttt{Business} $\cap$ \texttt{Sci\_Tech}-dominant
document group} (147 documents; 0.461\% of \texttt{Business}): commercial activity
in technology markets, including firms, products, digital services, and
investment. Four illustrative documents make this context
explicit: ``Google Offering Is Delayed'' (ID 2079) reports a delay in Google's
initial stock sale while awaiting SEC approval; ``Download a Song for 49 Cents''
(ID 2822) describes competition between Real and Apple's digital music services;
``IBM to make digital image sensors for Kodak'' (ID 30040) reports a manufacturing
agreement between the two firms; and ``Dell Plans North Carolina Plant, to Hire
1,500'' (ID 85857) describes a planned expansion of computer manufacturing and
employment. Their shared context is the commercial activity of technology firms.
\item \textbf{\texttt{Business} $\cap$ \texttt{World}-dominant
document group} (22 documents; 0.069\% of \texttt{Business}): economic activity in
international political and institutional settings. The four
illustrative documents are ``Ministers to Tackle Global Finance Issues''
(ID 40589), which links finance ministers' discussions to terrorism, interest
rates, and oil prices; ``French NGOs push to save African chicken farmers from
Europe (AFP)'' (ID 62621), describing a campaign against the effects of
low-priced poultry imports; ``Ukraine crisis `sparks bank rout' '' (ID 104216),
reporting a bank run amid political uncertainty; and ``Donors seek clarity on
PA finances'' (ID 125602), concerning financial transparency following a
change in Palestinian leadership. Their shared context is the economic role
of political events, institutions, and international relations.
\end{itemize}

\begin{figure}[tbp]
\centering
\includegraphics[width=\textwidth]{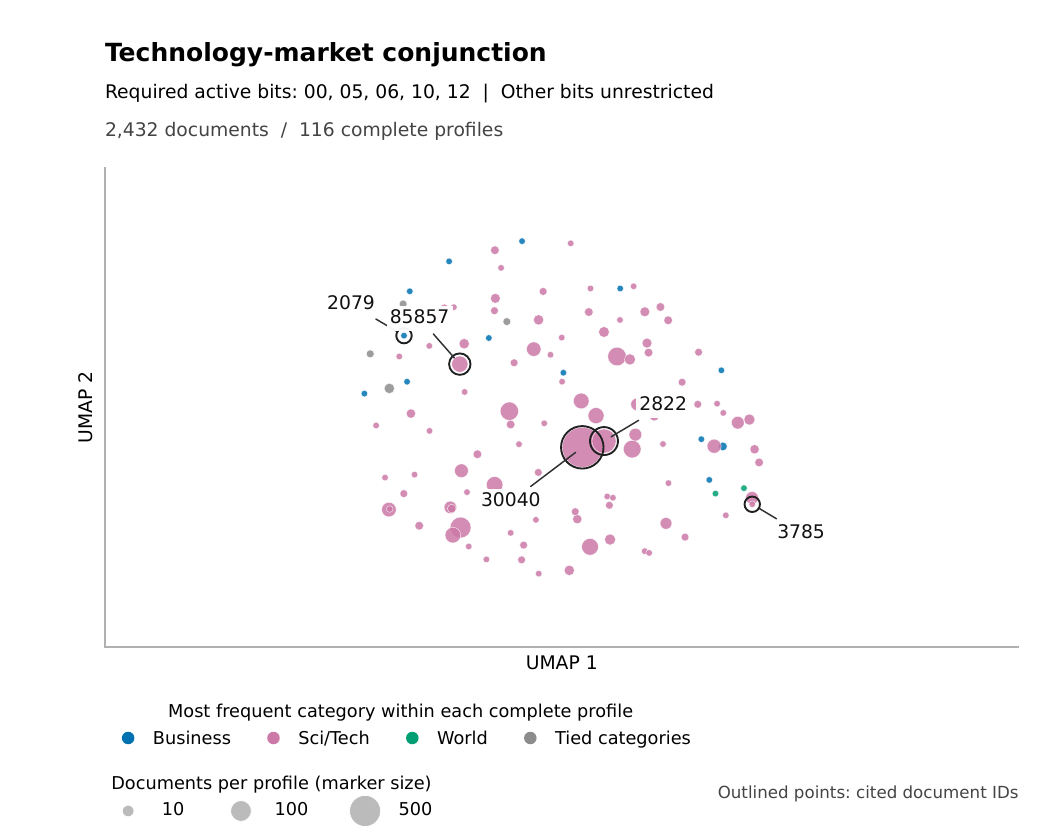}
\caption{Complete profiles satisfying the technology-market
conjunction: bits 00, 05, 06, 10, and 12 active, with other bits unrestricted.
All 2,432 matching documents are included, forming 116 profiles.
Marker area is proportional to the square root of the document count, keeping
rare profiles visible. Colour indicates the most frequent category within
each profile; grey indicates a tie. Black outlines and document IDs locate
the cited examples: four \texttt{Business} documents and the \texttt{Sci\_Tech}-labelled
patent-sale report (3785). A point's colour can therefore differ from the
label of its annotated document. Axes are visualization coordinates.}
\label{fig:agnews-technology-profile-map}
\end{figure}

\begin{figure}[tbp]
\centering
\includegraphics[width=\textwidth]{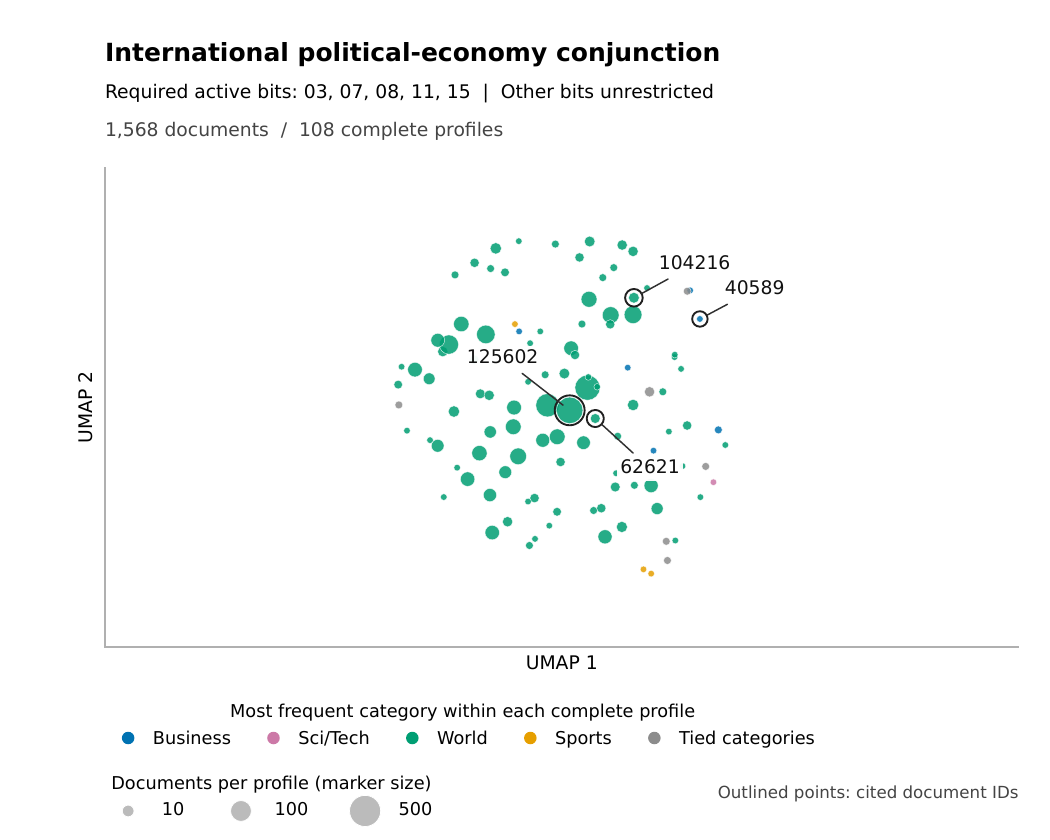}
\caption{Complete profiles satisfying the international
political-economy conjunction: bits 03, 07, 08, 11, and 15 active, with other
bits unrestricted. All 1,568 matching documents are included, forming
108 profiles. Marker area is proportional to the square root of document count;
colour indicates the most frequent category within the profile, with grey
for ties. Outlined points locate the four cited \texttt{Business} documents. Three
share complete profiles with predominantly \texttt{World}-labelled documents.
This layout is fitted separately from
Figure~\ref{fig:agnews-technology-profile-map}; coordinates are not directly
comparable between figures.}
\label{fig:agnews-politics-profile-map}
\end{figure}

Figures~\ref{fig:agnews-technology-profile-map} and
\ref{fig:agnews-politics-profile-map} locate these examples among all complete
profiles satisfying the respective conjunctions. For each figure, UMAP
\citep{McInnesEtAl2018UMAP} projects the sixteen binary coordinates using
Hamming distance, 15 neighbours, minimum distance 0.1, random initialization,
and seed 1234, with equal weight per distinct profile. The five required
coordinates are constant within each conjunction, so spatial variation comes
from the remaining bits. Each point represents one complete profile; its
colour indicates the most frequent editorial category among its documents.
The profile-category counts reported in Section~\ref{sec:agnews-taxonomy}
use all documents satisfying each conjunction. A category is counted as
most frequent only when its count strictly exceeds every other category,
excluding ties. The quoted documents illustrate these distributions. For example,
the IBM--Kodak report (ID 30040) shares its profile with \texttt{Sci\_Tech}-labelled
articles, while the Palestinian-finance report (ID 125602) shares its profile
with \texttt{World}-labelled articles.

The same technology-market conjunction also retrieves
``Internet patents sold at S.F. auction'' (ID 3785), a \texttt{Sci\_Tech}-labelled
report on an auction of patents covering technologies for Internet business
transactions. These examples illustrate how documents assigned to a single
editorial category can also contain elements associated with another.
The bit conjunctions bring together such documents through shared
content that crosses the original category boundaries.

Table~\ref{tab:qwen-community-complement-keywords} compares
these subsets with the non-\texttt{Business} documents satisfying the same
bit-community conjunctions. The complementary document groups emphasize
technological products and design in the first case, and political actors,
diplomacy, and conflict in the second. These lexical contrasts help identify the shared contexts within
which the \texttt{Business} documents emphasize commercial and economic aspects.

\begin{table}[tbp]
  \centering
  \small
\begin{tabular}{lrr@{\hspace{1.5em}}lrr}
  \multicolumn{3}{c}{\texttt{Sci\_Tech}-dominant document group} &
  \multicolumn{3}{c}{\texttt{World}-dominant document group} \\
  \cmidrule(lr){1-3}\cmidrule(lr){4-6}
  Term & Non-\texttt{Business} & \texttt{Business} & Term & Non-\texttt{Business} & \texttt{Business} \\
  \midrule
  line & 3.7 & 0.0 & minister & 14.8 & 0.0 \\
  week & 6.3 & 0.7 & peace & 14.4 & 0.0 \\
  mp3 & 3.6 & 0.0 & president & 14.0 & 0.0 \\
  advanced & 3.0 & 0.0 & talks & 12.9 & 0.0 \\
  model & 2.6 & 0.0 & prime & 11.0 & 0.0 \\
  imac & 2.5 & 0.0 & prime minister & 10.7 & 0.0 \\
  design & 2.3 & 0.0 & rebels & 10.3 & 0.0 \\
  released & 2.3 & 0.0 & sudanese & 8.7 & 0.0 \\
  mac & 2.2 & 0.0 & troops & 8.6 & 0.0 \\
  micro devices & 2.2 & 0.0 & end & 7.2 & 0.0 \\
  \bottomrule
  \end{tabular}
  \caption{Term-presence percentages in document groups defined by bit-community
  conjunctions, separated by the \texttt{Business} label. The \texttt{Sci\_Tech}-dominant document group contains 2,285
  non-\texttt{Business} and 147 \texttt{Business} documents; the
  \texttt{World}-dominant document group contains 1,546 and 22, respectively.
  The ten terms in each panel are ranked by the ratio of non-\texttt{Business}
  to \texttt{Business} document-presence proportions, adding 0.012 to both
  proportions and requiring at least three occurrences in the former group.
  Entries show observed percentages before smoothing.}
    \label{tab:qwen-community-complement-keywords}
\end{table}

In Figure~\ref{fig:qwen_bit_expost}, a total of four documents in the destination
\texttt{World}-dominant document group come from the editorially \texttt{Sports}-labelled items whose content is structured around governance rather than sporting performance. The corpus texts are reproduced below: 
 \begin{itemize}
  \item ID 43673: Opposition asks govt to provide security to Indian team in B'desh: India News: New Delhi, Dec 6: Voicing serious concern over the death threat to members of the Indian Cricket team, Lok Sabha members belonging to the opposition NDA today asked the government to provide fool proof security to the members of the Indian

  \item ID 58572: Manchester United Fans Oppose Takeover (AP) AP - Manchester United's largest fan group is launching a registration drive to stop U.S. businessman Malcolm Glazer from buying the Premier League club.

  \item ID 68948: Police warn referee ahead of game Police will talk to referee Mike Riley before Manchester United's home game with Arsenal on Sunday to try to avoid crowd trouble.

  \item ID 69526: POLICE TO WARN ARSENAL \& MAN U PLAYERS POLICE will visit the dressing-rooms of Manchester United and Arsenal to lay down the law before their grudge match tomorrow. Both teams will be warned ``in no uncertain terms.''
\end{itemize}
 This supports a compositional reading of the sparse code: a document can retain its editorial topic while jointly expressing a cross-cutting context, here public order, institutional authority, or transnational ownership. Given four cases, this is qualitative evidence for interpretation and not a prevalence estimate.

A complementary analysis examines distinctions within an editorial category.
We enumerate pairs of observed profiles differing in exactly one bit, retaining
pairs with at least 100 documents in each profile. The remaining fifteen
coordinates, including inactive bits, are identical within a pair. AG News has
224 eligible pairs. We select the pair maximizing the smaller group size,
without using labels or text content. Its profiles differ only in bit~08 and
contain 1,050 and 1,224 documents. Both are almost entirely labelled
\texttt{Sports} (Table~\ref{tab:agnews-single-bit-contrast}).

For the sports contrast, document IDs are sorted numerically within each
profile, and twelve documents are drawn without replacement using Python's
\texttt{random.Random(1234).sample}; the generator is reset for each profile.
Inspection of these saved samples suggests a concentration on American football when bit~08 is active, whereas
the other group includes baseball, basketball, and other sports. For example,
the inactive group contains ``NL Wrap: Walker Grand Slam Helps Cards Beat Reds
Again'' (ID 2270), while the active group contains ``Jones Leads Cowboys to
Stunning Win Over Seahawks'' (ID 6670). We subsequently count indicative terms
across both complete groups. References to touchdowns or quarterbacks are more
frequent in the active group, while references to home runs or innings are
more frequent in the inactive group. These lexical contrasts support a finer
thematic reading within the same editorial category. The terms were selected
after inspection and serve as descriptive indicators, rather than a validated
classification of sports. Throughout these comparisons, differences concern
observed document groups and do not estimate the effect of intervening on a bit.

\subsection{20 Newsgroups}
\label{app:data-20ng}

``20 Newsgroups'' consists of Usenet discussion posts assigned to 20
newsgroups. We retrieve both official partitions using scikit-learn's
\texttt{fetch\_20newsgroups}, removing headers, signatures, and quoted passages
with its \texttt{headers}, \texttt{footers}, and \texttt{quotes} options
\citep{TwentyNewsgroupsSklearn}. After trimming text and requiring at least 20
characters, 10,952 training and 7,269 test posts remain. In addition to the 20
original newsgroup labels, our pipeline defines five coarse groups: computing
(the five \texttt{comp.*} groups), recreation (the four \texttt{rec.*} groups),
science (the four \texttt{sci.*} groups), miscellaneous sales
(\texttt{misc.forsale}), and discussion. The last combines the four
\texttt{talk.*} groups with \texttt{alt.atheism} and
\texttt{soc.religion.christian}. This five-way grouping is an analytical
aggregation of the source labels.

\begin{figure}[tbph]
\centering
\includegraphics[width=\textwidth]{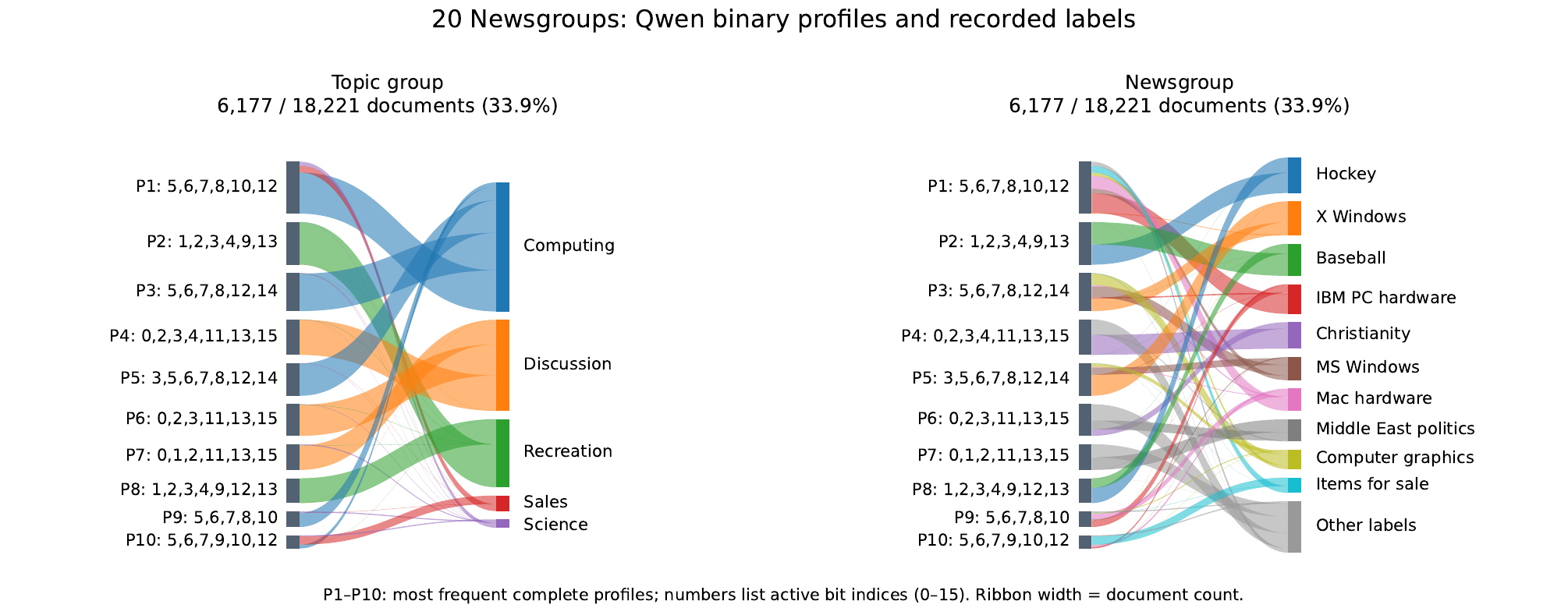}
\caption{20 Newsgroups: label composition of document groups defined by the ten most frequent Qwen profiles. The same profiles are shown against the five topic groups and the original newsgroups. Panel headings give the displayed sample and its corpus coverage; ribbon widths are document counts.}
\label{fig:corpus-profiles-20ng}
\end{figure}

\subsection{Twitter US Airline Sentiment}
\label{app:data-airline}

The Airline corpus contains tweets about six US airlines,
collected in February 2015 and annotated through CrowdFlower for sentiment and,
for negative tweets, complaint reason. We use the Hugging Face distribution
\citep{AirlineSentimentHF}. Removing repeated tweet identifiers and retaining
trimmed texts of at least five characters leaves 14,485 tweets: 9,082 negative,
3,069 neutral, and 2,334 positive. The carrier label identifies United, US Airways,
American, Southwest, Delta, or Virgin America. We derive a 12-category sentiment
refinement by retaining positive and neutral and dividing negative tweets into
customer-service issues, late flights, cancelled flights, lost luggage, bad
flights, booking problems, flight-attendant complaints, long lines, damaged
luggage, and an unknown-reason category. Missing or indeterminate complaint
reasons enter the last category. A further label crosses the three sentiment
classes with the six airlines, producing 18 observed combinations.

\begin{figure}[tbph]
\centering
\includegraphics[width=\textwidth]{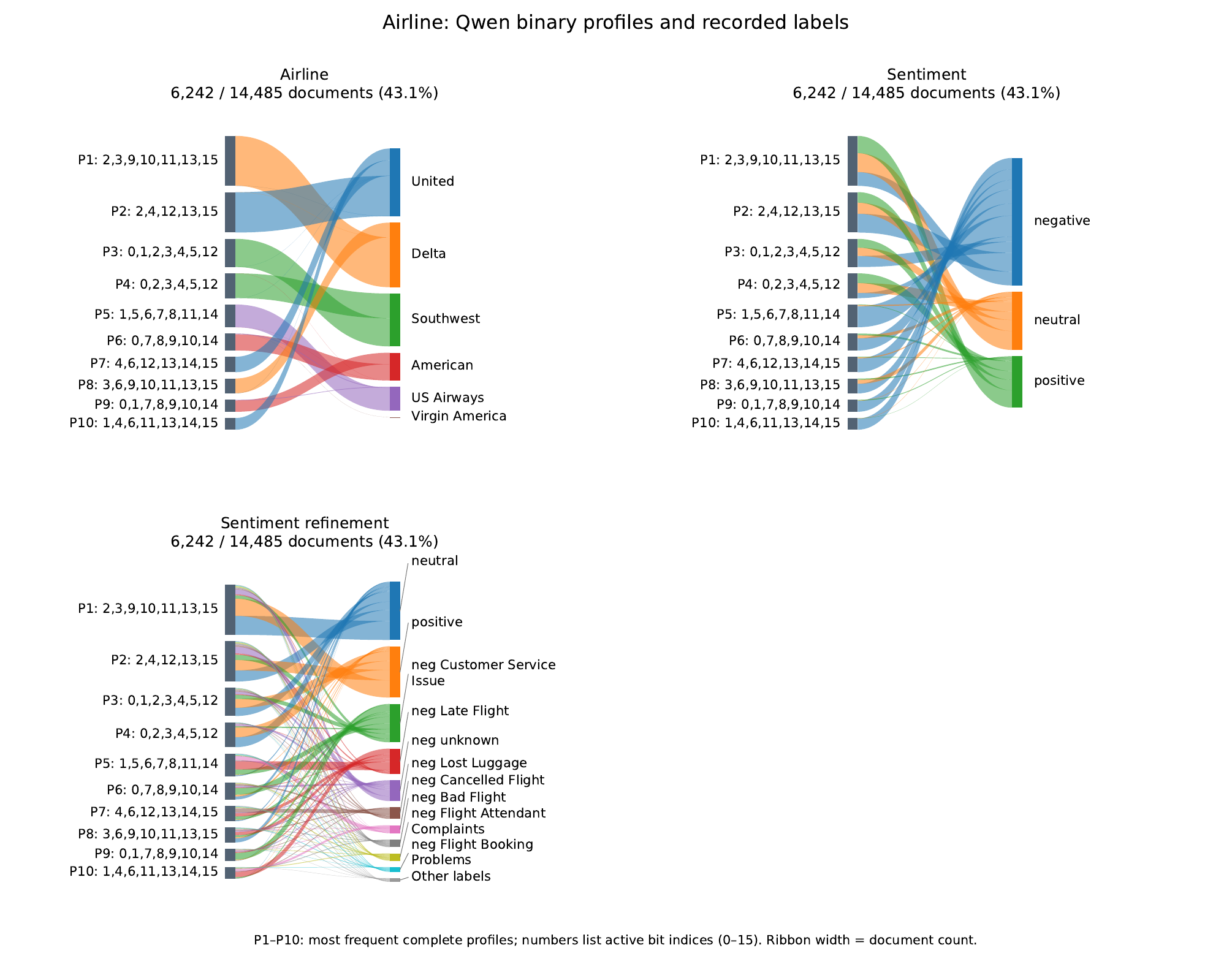}
\caption{Airline: label composition of document groups defined by the ten most frequent Qwen profiles. The same profiles are shown against carrier, sentiment, and its complaint-based refinement. Panel headings give the displayed sample and its corpus coverage; ribbon widths are document counts.}
\label{fig:corpus-profiles-airline}
\end{figure}

\subsection{GoEmotions}
\label{app:data-goemotions}

GoEmotions contains English Reddit comments annotated for 27
emotions or a neutral category \citep{DemszkyEtAl2020GoEmotions}. We use the
\texttt{simplified} configuration of
\href{https://huggingface.co/datasets/google-research-datasets/go_emotions}{\nolinkurl{google-research-datasets/go_emotions}},
pooling train, validation, and test. The present analysis retains comments with
exactly one supplied label and at least five characters after trimming, yielding
45,432 documents. The 28-category label is also aggregated into positive,
negative, ambiguous, and neutral affect. Positive comprises admiration,
amusement, approval, caring, desire, excitement, gratitude, joy, love, optimism,
pride, and relief; negative comprises anger, annoyance, disappointment,
disapproval, disgust, embarrassment, fear, grief, nervousness, remorse, and
sadness; ambiguous comprises confusion, curiosity, realization, and surprise.
Neutral remains separate. The  all-labels field coincides with the
primary label in this selected single-label corpus.

\begin{figure}[tbph]
\centering
\includegraphics[width=\textwidth]{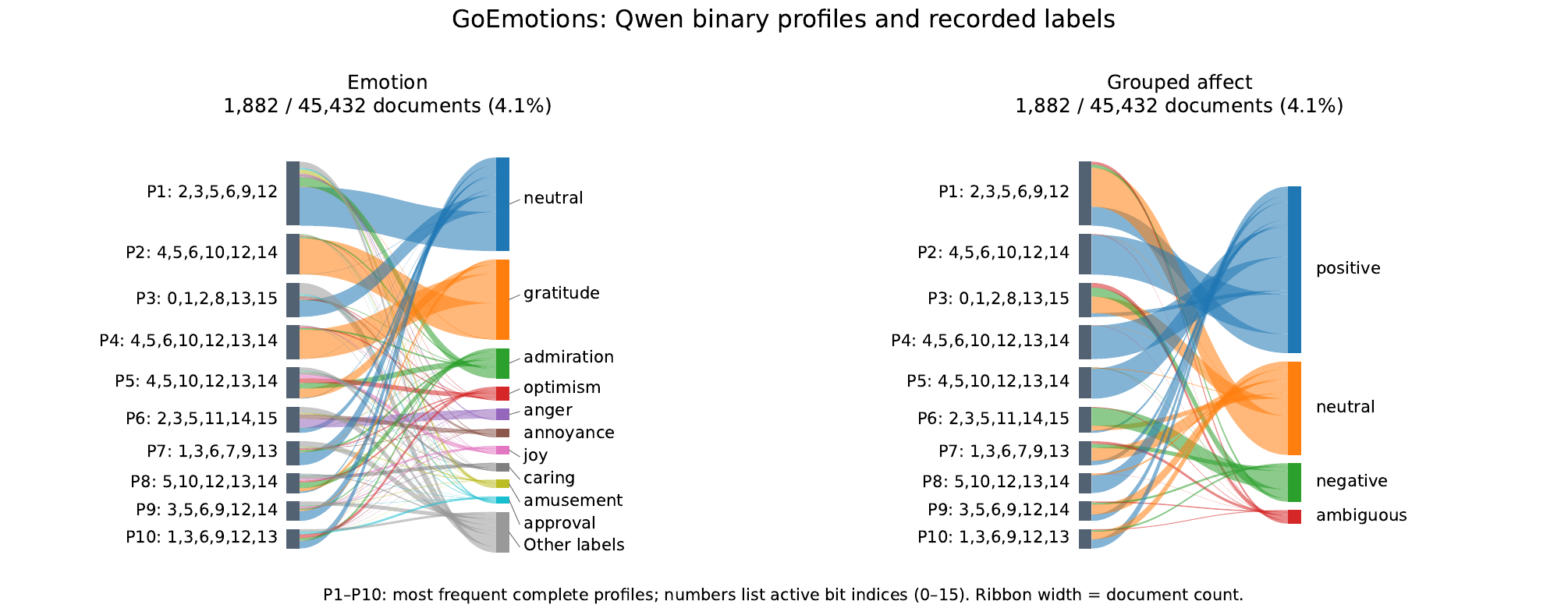}
\caption{GoEmotions: label composition of document groups defined by the ten most frequent Qwen profiles. The same profiles are shown against individual emotions and grouped affect. Panel headings give the displayed sample and its corpus coverage; ribbon widths are document counts.}
  \label{fig:corpus-profiles-goemotions}
\end{figure}

\subsection{TweetEval}
\label{app:data-tweeteval}

We use four English tweet-classification tasks from TweetEval
\citep{BarbieriEtAl2020TweetEval}, downloaded from
\href{https://huggingface.co/datasets/cardiffnlp/tweet_eval}{\nolinkurl{cardiffnlp/tweet_eval}}.
Sentiment distinguishes negative, neutral, and positive tweets and derives from
SemEval-2017 Task 4 \citep{RosenthalEtAl2017Sentiment}. Emotion distinguishes
anger, joy, optimism, and sadness, drawing on SemEval-2018 Task 1
\citep{MohammadEtAl2018Affect}. Hate distinguishes hateful from non-hateful tweets
in the task concerning immigrants and women \citep{BasileEtAl2019HateEval}.
Offensive distinguishes offensive from non-offensive language
\citep{ZampieriEtAl2019OffensEval}. Each task is treated as a separate corpus.
The prepared files pool the source training and test partitions, omit validation,
and require at least five characters after trimming. Their resulting sizes are
57,899, 4,678, 11,961, and 12,776, respectively. The dataset's category names are
retained alongside their numeric encodings.

\begin{figure}[tbph]
\centering
\includegraphics[width=\textwidth]{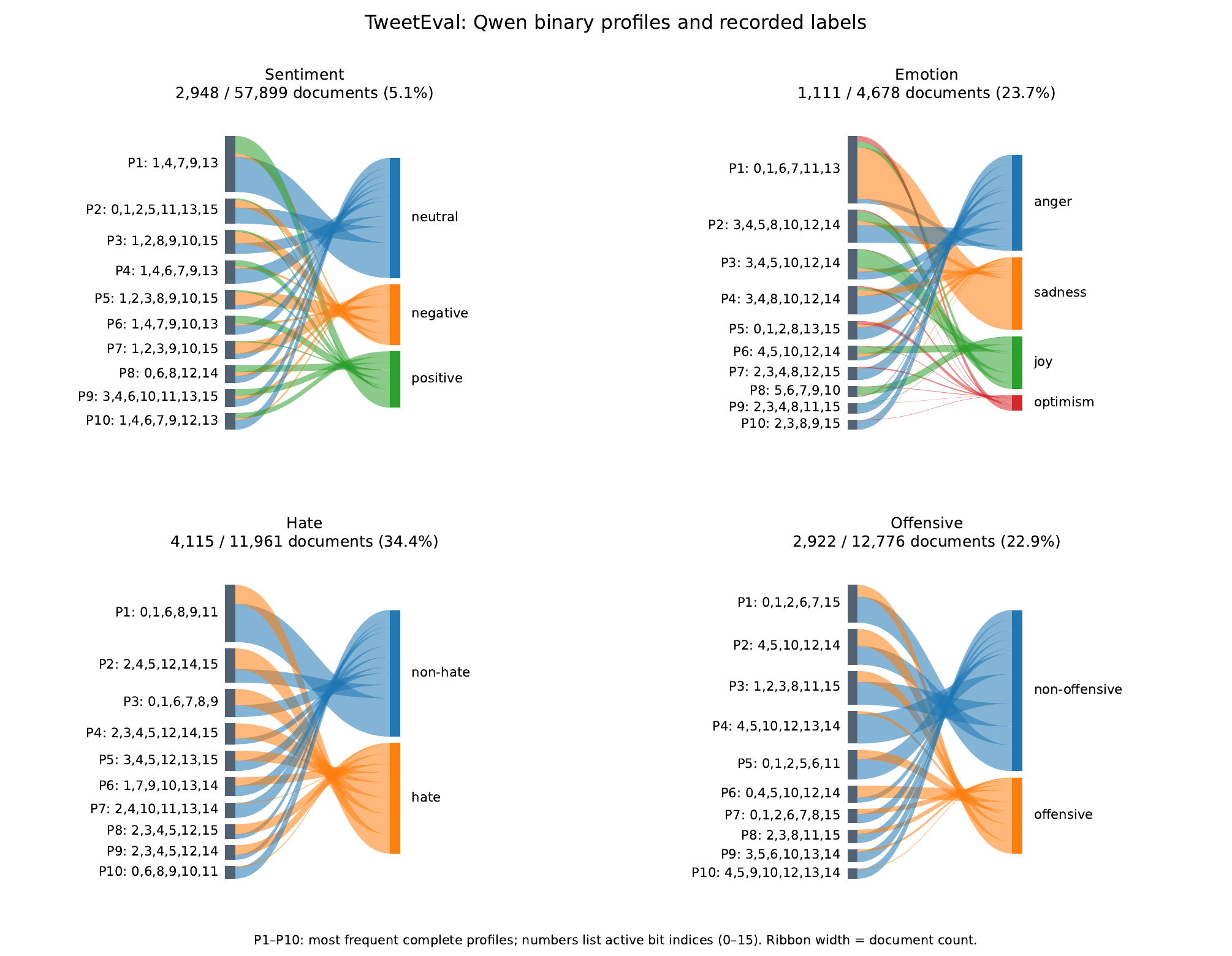}
\caption{TweetEval: label composition of document groups defined by the ten most frequent Qwen profiles. Each panel concerns a separate task-specific corpus and encoder; profile identifiers are local to that panel. Panel headings give the displayed sample and its corpus coverage; ribbon widths are document counts.}
  \label{fig:corpus-profiles-tweeteval}
\end{figure}

\subsection{FakeNewsNet}
\label{app:data-fakenewsnet}

FakeNewsNet combines collections associated with PolitiFact and
GossipCop \citep{ShuEtAl2020FakeNewsNet}. We download the four public CSV files
for real and fake items in the
\href{https://github.com/KaiDMML/FakeNewsNet}{authors' repository}. The text input
is the news title available in these files; article bodies and social-network
information are not collected for this experiment. After requiring at least 30
characters and removing duplicate source rows, the corpus contains 21,786 items:
20,907 from GossipCop and 879 from PolitiFact. We retain the supplied real/fake
classification (16,372 real and 5,414 fake), the source collection, and their
four-way cross-classification. Source denotes the fact-checking collection,
not the publishing outlet.

\begin{figure}[tbph]
\centering
\includegraphics[width=\textwidth]{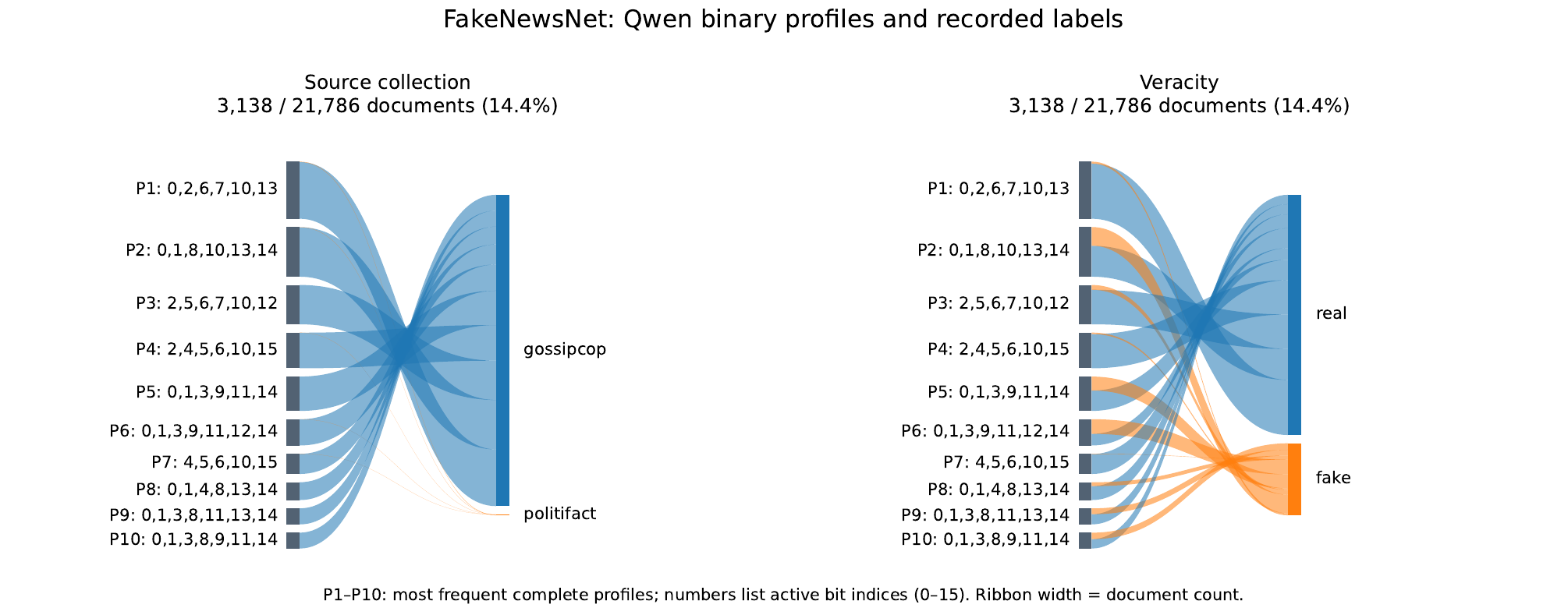}
\caption{FakeNewsNet: label composition of document groups defined by the ten most frequent Qwen profiles. The same profiles are shown against collection identity and veracity. Panel headings give the displayed sample and its corpus coverage; ribbon widths are document counts.}
\label{fig:corpus-profiles-fakenewsnet}
\end{figure}

\subsection{PHEME}
\label{app:data-pheme}

PHEME contains Twitter conversations around breaking-news events.
We use the official Figshare archive for rumour detection and veracity
classification \citep{KochkinaEtAl2018PHEME}, selecting source tweets from five
events: Charlie Hebdo, Ferguson, the Germanwings crash, the Ottawa shooting, and
the Sydney siege. Replies are not separate documents in our corpus. The prepared
sample contains 5,802 source tweets, comprising 3,830 non-rumours and 1,972 rumours.
For rumours, veracity is true (1,008), false (393), or unverified (571);
non-rumours receive a not-applicable code. We retain event identity, binary
rumour status, and veracity, and derive a four-category label comprising
non-rumour, true rumour, false rumour, and unverified rumour.  Retweet
and favorite counts provide the engagement-prediction application in
Section~\ref{sec:pheme-engagement}. These counts are snapshots: source tweets meet
retweet thresholds of 100 in four events and 25 in Germanwings, so this application
concerns engagement among already visible tweets.

\begin{figure}[tbph]
\centering
\includegraphics[width=\textwidth]{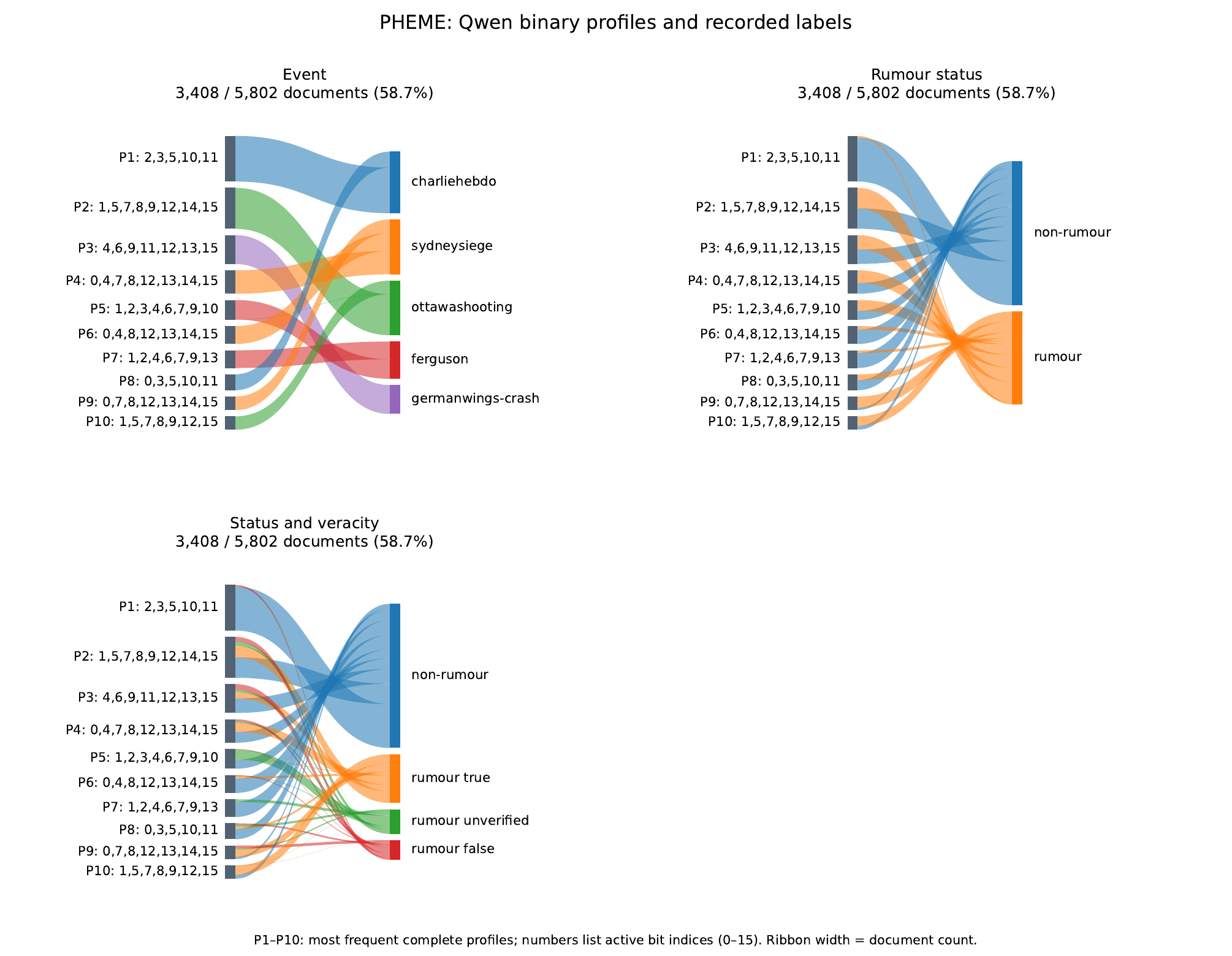}
\caption{PHEME: label composition of document groups defined by the ten most frequent Qwen profiles. The same profiles are shown against event, rumour status, and status--veracity labels. Panel headings give the displayed sample and its corpus coverage; ribbon widths are document counts.}
\label{fig:corpus-profiles-pheme}
\end{figure}

\subsection{Climate and migration literature}
\label{app:data-climmig}

The corpus called Climate Migration in our analyses contains
5,545 scholarly abstracts from the migration-literature resource associated with
LlaMig \citep{IacusQiHan2026Migration}. We use
\href{https://github.com/HaodongQi/deep_literature_review_migration/tree/main/analysis}{\texttt{allcoded.csv}
and \texttt{key-values.csv}} from the authors' repository, deduplicate document
identifiers, and retain abstracts of at least 20 characters. The model input is
the abstract; titles and DOIs are metadata. Six supplied coding fields describe
migration relevance (yes/no), migration drivers, sentiment toward migration,
climate or environmental hazards, discipline, and use of quantitative methods
(yes/no). The corpus includes papers outside migration: 4,175 abstracts are coded
as relevant and 1,370 as not relevant.

Drivers can include other, climate, environment, economic,
conflict, political, social, and family, with multiple codes per document.
The prepared hazard field contains 29 observed non-empty codes, including
pollution, infectious diseases, drought, flooding, and sea-level rise. Blank
source values in drivers, sentiment, hazards, and discipline are represented by
\texttt{none}; this denotes an absent code rather than an independently
established negative finding. Sentiment has positive, negative, neutral, and
\texttt{none}; the observed discipline values are social science, other, and
\texttt{none}. Drivers and hazards remain semicolon-separated fields. Semantic
profile summaries count their individual constituent labels. The existing
categorical predictive comparisons instead use the recorded combinations as
classes: drivers has 82 observed strings, reduced to 53 classes and 5,516 documents
after singleton exclusion; hazards has 51 strings, reduced to 27 classes and
5,521 documents. The other four targets use all 5,545 documents.

Table~\ref{tab:climmig-combination-evidence} reports the existing categorical driver and hazard evaluations. Their low macro-F1 scores and sparse class support motivate treating these combination tasks separately from the four simpler coding fields in the main text. BA averages recall over classes observed in the test set; the saved macro-F1 calculation averages over the union of true and predicted classes, assigning zero to undefined class scores.

\begin{table}[tbph]
\centering
{\small
\setlength{\tabcolsep}{3pt}
\begin{tabular}{llrrrll}
\toprule
Corpus & Attribute & $N$ & $C$ & AMI & BA [95\% CI] & Macro-F1 \\
\midrule
Climate Migration & Driver combinations & 5,516 & 53 (40) & 0.123 & 6.42 [5.41, 7.66] & 0.99 \\
 & Hazard combinations & 5,521 & 27 (21) & 0.015 & 13.12 [9.57, 16.81] & 2.93 \\
\bottomrule
\end{tabular}}
\caption{Additional Climate Migration tasks using the saved categorical encoding of driver and hazard combinations. BA and macro-F1 are percentages; columns follow Table~\ref{tab:cross-corpus-label-evidence}. These are exact-combination tasks, including the missing-code category, rather than separate predictions of each constituent annotation. Several retained rare classes do not occur in the test split.}
\label{tab:climmig-combination-evidence}
\end{table}

\begin{figure}[tbp]
\centering
\includegraphics[width=\textwidth]{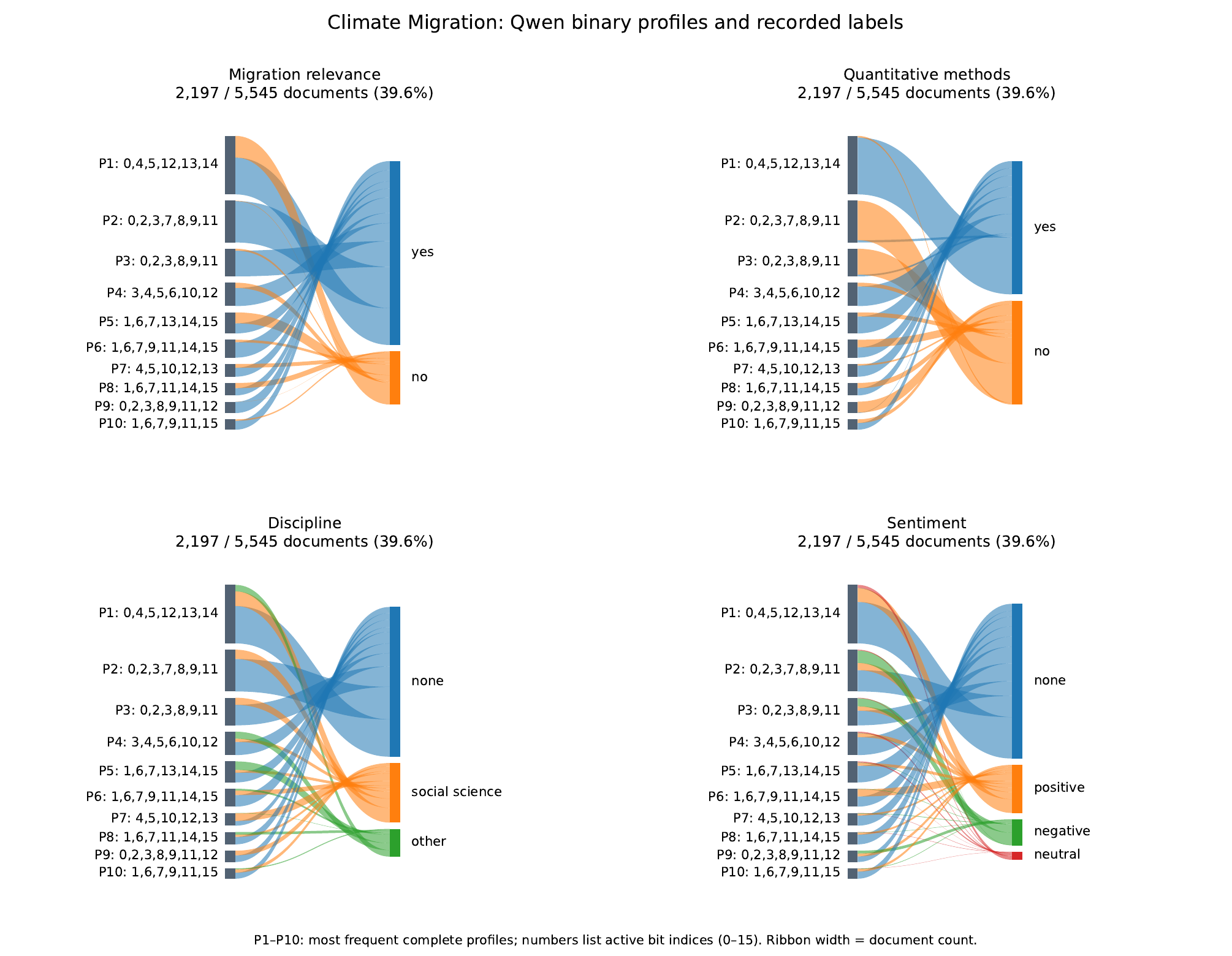}
\caption{Climate Migration: label composition of document groups defined by the ten most frequent Qwen profiles. The same profiles are shown against migration relevance, quantitative methods, discipline, and sentiment. Panel headings give the displayed sample and its corpus coverage; ribbon widths are document counts.}
\label{fig:corpus-profiles-climmig}
\end{figure}

\subsection{Mixed corpus}
\label{app:data-mixed}

The Mixed corpus combines 111,000 documents sampled without
replacement from the prepared sources, using master seed 1234 and an independent
deterministic random-number stream for each source. It includes 4,000 documents
from each of the four TweetEval tasks; 5,000 each from PHEME and Climate Migration;
10,000 each from AG News and Airline; 15,000 each from 20 Newsgroups, FakeNewsNet,
and GoEmotions; and 20,000 from IRA. The IRA component consists of
Internet Research Agency tweets collected by Linvill and Warren and distributed
through FiveThirtyEight's
\href{https://github.com/fivethirtyeight/russian-troll-tweets}{\nolinkurl{russian-troll-tweets}}
repository \citep{LinvillWarren2020}. It provides an additional source of tweets
within the heterogeneous mixture; IRA is not evaluated as a standalone corpus in this work.
A separate seeded permutation assigns new
identifiers and interleaves source rows, while the original identifiers are
retained as provenance. The AG News component was randomly sampled from an
earlier stratified extraction of 15,600 documents, prepared before the dedicated
AG News analysis was extended to all 127,600 documents. This sampling history
does not change the purpose of the Mixed experiment, which examines structure
across heterogeneous document sources.

The nine-valued source label merges the four TweetEval tasks
under one source. A second label records document type: tweet (51,000), news
(25,000), discussion post (15,000), comment (15,000), or abstract (5,000). The
23 source-specific label fields retain the four TweetEval targets; PHEME status,
veracity, and event; migration relevance, discipline, and sentiment; AG News
topic; fine and coarse newsgroups; FakeNewsNet veracity and source; IRA category,
type, and language; Airline sentiment and carrier; and GoEmotions primary,
grouped, and all-labels fields. Non-applicable entries are stored as
\texttt{unknown} and excluded when interpreting the corresponding source-specific
label. Cross-products and complaint-based refinements are omitted. Source and
document type provide corpus-wide reference partitions, while the remaining
labels support interpretation within their applicable subsets.

\begin{figure}[tbp]
\centering
\includegraphics[width=\textwidth]{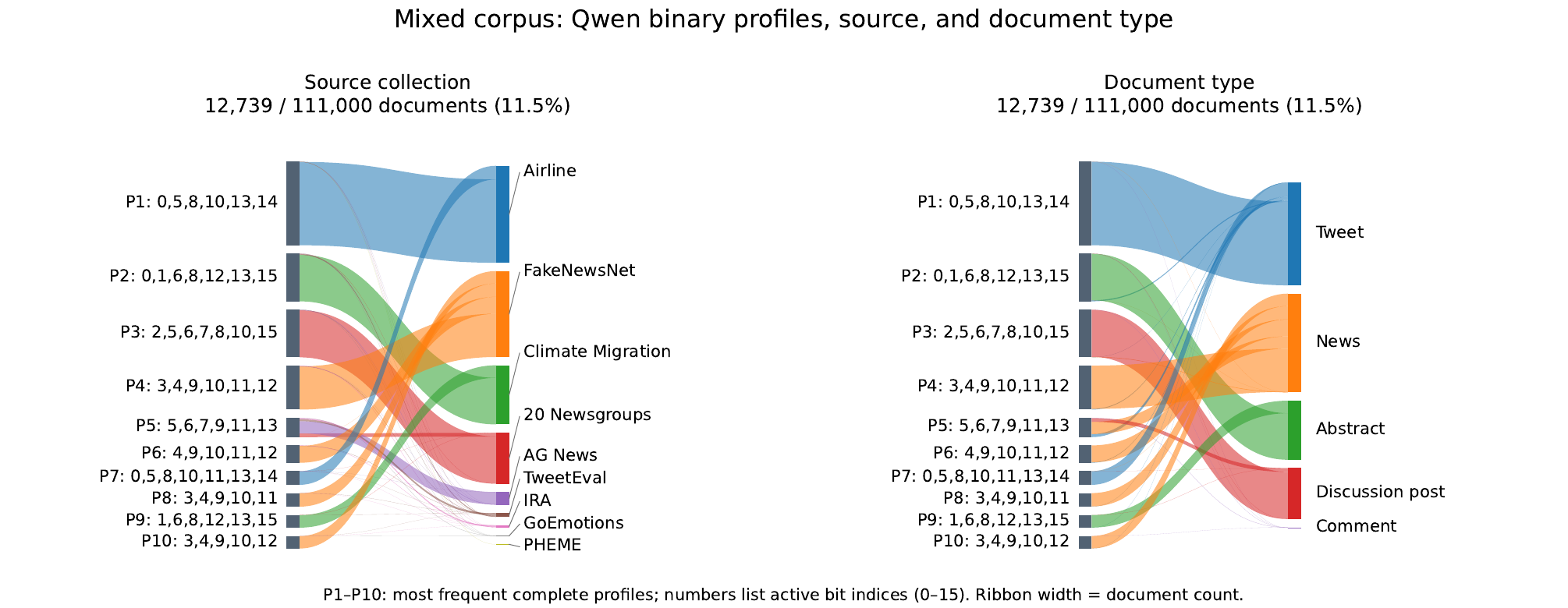}
\caption{Source and document-type composition of the ten frozen
Mixed Qwen profiles. Profile ranks P1--P10 correspond to P01--P10 in
Table~\ref{tab:mixed-profiles}; active bit indices are listed beside each profile.
The same 12,739 documents are shown in both panels, covering 11.48\% of the full
mixture. Ribbon widths and category-node heights refer to this displayed subset.}
\label{fig:mixed-profile-sources}
\end{figure}

\section{Embedding clusters and binary document groups}
\label{app:cluster-diagnostics}

\subsection{Residual entropy and partition agreement}
\label{app:cluster-partition-diagnostics}

These diagnostics supplement Section~\ref{sec:profiles-vs-clusters}.
Lloyd $k$-means is fitted to all 127,600 raw, unnormalised Qwen embeddings
at $M\in\{32,64,256\}$, without labels or binary codes. We use
scikit-learn's Lloyd implementation with k-means++ initialization,
ten starts, a maximum of 300 iterations per start, tolerance $10^{-4}$,
and random seed 1234, retaining the solution with the lowest within-cluster
sum of squared distances. The L2-normalized check at $M=32$ and $256$
uses the same settings. Complete profiles partition the same documents,
whereas conjunctions can overlap.

Let $C$ denote cluster membership and $B_k$ a binary coordinate.
The decomposition
\[
H(B_k)=I(B_k;C)+H(B_k\mid C)
\]
separates the bit entropy associated with cluster membership from
the variation remaining within clusters. We report the empirical
residual fraction $H(B_k\mid C)/H(B_k)$, which is zero when cluster
membership determines the bit and one when they are independent.
Averaged over the sixteen bits, this fraction is 0.578, 0.550, and
0.489 at $M=32$, 64, and 256. Independently permuting each bit column,
while preserving its prevalence, gives corresponding reference
values of 1.000, 1.000, and 0.998. Thus, clusters account for part
of the variation in bit activations, while substantial variation
remains within them.

For complete profiles, the residual entropy fraction decreases
from 0.656 at $M=32$ to 0.516 at $M=256$. The median effective number
of profiles within a cluster decreases from 124.3 to 43.8.
Conversely, the median three-bit conjunction requires 6, 11, and
30.5 clusters to cover 80\% of its documents.
For each conjunction, a support-preserving random reassignment
provides a reference for this coverage count. The median of these
conjunction-specific expected counts is 21.3, 41.3, and 149.8,
respectively (Supplementary Section~\ref{app:cluster-null-references}).
Clusters and conjunctions are therefore associated but not equivalent:
conjunctions concentrate within a subset of clusters, while extending
across more than one, and clusters contain multiple complete profiles.
The residual entropy of complete profiles also depends on the
granularity of the partitions being compared.

Chance-corrected agreement between complete profiles and $k$-means
ranges from 0.042 to 0.085 for ARI and from 0.402 to 0.412 for AMI
\citep{HubertArabie1985,VinhEppsBailey2010}. Across three ISBC fits
with different random seeds, the corresponding ranges are
0.147--0.181 and 0.349--0.363. These measures give different orderings
for partitions of different granularities; the complete profile
partition also varies across initializations.

Repeating clustering after L2 normalization changes the mean bit
residual entropy fraction from 0.578 to 0.583 at $M=32$ and from
0.489 to 0.488 at $M=256$. Profile--cluster AMI changes from
0.405 to 0.407 and from 0.402 to 0.404, respectively.
The distinction between the representations therefore persists when
clustering emphasizes angular geometry.

\subsection{Bit correspondence across initializations}
\label{app:seed-bit-correspondence}

We compare the individual coordinates of the three full-objective Qwen
fits used in Section~\ref{sec:loss-ablation}, with $K=16$, $A=6$, and seeds
1234, 2025, and 31415. All codes are evaluated on the same 127,600 AG News
documents, aligned by document identifier. For each pair of fits, the
Hungarian algorithm selects the one-to-one bit assignment maximizing
arithmetic-normalized mutual information,
$\operatorname{NMI}(B_k,B'_\ell)=2I(B_k;B'_\ell)/
[H(B_k)+H(B'_\ell)]$. This score is unaffected by complementing a bit,
so inverse activation patterns can also be matched.

\begin{table}[tbp]
\centering\small
\begin{tabular}{rrr}
\toprule
First seed & Second seed & Mean matched-bit NMI \\
\midrule
1234 & 2025 & 0.163 \\
1234 & 31415 & 0.163 \\
2025 & 31415 & 0.173 \\
\bottomrule
\end{tabular}
\caption{Bit correspondence across three AG News Qwen fits of the full
objective. Each value averages the sixteen matched-coordinate NMI scores
for one pair of fitted encoders. Scores describe activation correspondence
on the full corpus, not semantic agreement between assigned descriptions.}
\label{tab:qwen-seed-bit-correspondence}
\end{table}

The matched scores indicate limited one-to-one correspondence between
coordinates across these fits, even after accounting for permutations
and polarity. Similar aggregate predictive performance across seeds
therefore coexists with different bit memberships. The Mixed extension
addresses a separate question: whether selected associations persist for
additional documents when the encoder itself is held fixed.

\subsection{Conditional bit entropy for the ITQ comparison}
\label{app:itq-entropy}

Using the same raw-embedding $k$-means partitions and residual entropy
fraction as in Section~\ref{sec:profiles-vs-clusters}, we average over
the sixteen nonconstant coordinates. Table~\ref{tab:itq-bit-entropy} reports
these descriptive values for the unchanged ISBC code and both ITQ readouts.
Activity and bit prevalences differ across the representations, so the
comparison does not isolate an effect of encoder architecture.

\begin{table}[tbp]
\centering
\small
{
\begin{tabular}{lrrr}
\toprule
Representation & $M=32$ & $M=64$ & $M=256$ \\
\midrule
ISBC & 0.578 & 0.550 & 0.489 \\
ITQ top-6 & 0.656 & 0.625 & 0.569 \\
ITQ & 0.620 & 0.590 & 0.530 \\
\bottomrule
\end{tabular}}
\caption{Mean residual bit-entropy fraction conditional on
$M$ embedding clusters, evaluated on all 127,600 documents. Zero means
cluster membership determines the bit; one means it carries no information
about the bit. Values summarize association with the specified partitions.}
\label{tab:itq-bit-entropy}
\end{table}

\subsection{Category-based descriptions of conjunctions and clusters}
\label{app:cluster-category-composition}

We locate the two \texttt{Business} subsets examined in
Section~\ref{sec:agnews-taxonomy} within the $M=32$, 64, and 256
embedding-cluster partitions described in Section~\ref{sec:profiles-vs-clusters}. Each subset combines the \texttt{Business} label with a
conjunction of community bits; these targets are therefore defined
after fitting and are not an independent benchmark. The 147-document
technology-market subset spans 9, 20, and 39 clusters at the three
resolutions, while the 22-document \texttt{Business}--\texttt{World} subset spans
9, 11, and 17. Neither subset coincides with a single cluster.
Uniformly reassigning each target across documents while preserving
its size and the cluster sizes gives expected occupancy counts of
30.8, 54.3, and 105.3 for the technology-market subset, and
15.3, 18.2, and 20.9 for the \texttt{Business}--\texttt{World} subset.
The observed subsets are thus concentrated relative to this reference,
although their boundaries do not coincide with individual clusters. Table~\ref{tab:cluster-random-references}
collects the occupancy counts and their random references.

As a secondary description, we examine category composition.
Restricting three-bit conjunctions to the observed cluster-size
range, their mean dominant-category shares are 62.7\%, 59.1\%,
and 54.7\%, compared with 87.5\%, 86.6\%, and 89.1\% for clusters.
This restriction does not exactly match the group-size distributions.
Within these comparisons, conjunction-defined groups span the
editorial categories more broadly than clusters do; lower category
purity alone does not establish semantic coherence.

\subsection{Random references for cluster occupancy and coverage}
\label{app:cluster-null-references}

The reference randomly assigns a target of size $m$ to the
$N=127{,}600$ documents, keeping the fitted cluster sizes $n_c$ fixed.
The expected number of occupied clusters is
\[
\mathbb E[O_m]=\sum_{c=1}^{M}
\left[1-\frac{\binom{N-n_c}{m}}{\binom{N}{m}}\right].
\]
We evaluate this expression exactly for the two \texttt{Business} subsets.
For each of the $\binom{16}{3}=560$ three-bit conjunctions, we draw
2,000 cluster-count vectors from the corresponding multivariate
hypergeometric distribution, preserving its observed support.
In each draw, clusters are ordered by target count and we count how
many are needed to cover at least 80\% of the target documents.
Table~\ref{tab:cluster-random-references} reports the median of the
560 conjunction-specific Monte Carlo means, alongside the median
observed count. This is a collection of marginal reference values,
not a joint permutation test of the median across overlapping
conjunctions. The seed is 1234, with a separate reproducible stream
for each cluster resolution and conjunction.

\begin{table}[tbp]
\centering
\small
{
\begin{tabular}{r rr rr rr}
\toprule
& \multicolumn{2}{c}{Technology-market} & \multicolumn{2}{c}{\texttt{Business}--\texttt{World}} & \multicolumn{2}{c}{Three-bit coverage} \\
$M$ & Observed & Random & Observed & Random & Observed & Random \\
\midrule
32 & 9 & 30.8 & 9 & 15.3 & 6 & 21.3 \\
64 & 20 & 54.3 & 11 & 18.2 & 11 & 41.3 \\
256 & 39 & 105.3 & 17 & 20.9 & 30.5 & 149.8 \\
\bottomrule
\end{tabular}}
\caption{Observed cluster counts and random references.
The first two comparisons count occupied clusters for the 147- and
22-document subsets. The last compares the median number of clusters
covering 80\% of each three-bit conjunction with the median of its
conjunction-specific null means. Smaller observed counts indicate
concentration relative to random target assignment.}
\label{tab:cluster-random-references}
\end{table}

\section{AG News: capacity, reconstruction, geometry, and prediction}
\label{app:representation-diagnostics}

\subsection{Qwen code capacity and reconstruction diagnostics}
\label{app:qwen-capacity-reconstruction}

The following definitions and diagnostics support
Section~\ref{sec:qwen-compression}, using the same corpus and fitted codes.

Profile probabilities, empirical entropy $\widehat H(B)$, and effective
profile counts are defined in Section~\ref{sec:qwen-compression}.
Finite-sample entropy estimates tend to underestimate population entropy,
particularly when many profiles are rare or unobserved \citep{Paninski2003}.
We assess this sensitivity with the Miller--Madow correction,
\[
\widehat H_{\mathrm{MM}}(B)=\widehat H(B)+\frac{m-1}{2N\ln 2},
\]
where $m$ is the observed number of profiles. It adds 0.069 and 0.273 bits
per document for the 16/6 and 32/6 codes, and 0.597 for 32/16. These
increasing adjustments reflect the difficulty of estimating a more
varied profile distribution from the same corpus. The first-order
correction is a sensitivity diagnostic and may itself be inaccurate
for sparsely sampled distributions.

For the PCA reference justified in Section~\ref{sec:qwen-compression}
\citep{EckartYoung1936},
let $\lambda_1\ge\cdots\ge\lambda_D\ge0$ be the eigenvalues
of the empirical embedding covariance matrix
$\widehat\Sigma_x=(N-1)^{-1}\sum_{i=1}^{N}(x_i-\bar x)(x_i-\bar x)^\top$,
where $\bar x=N^{-1}\sum_{i=1}^{N}x_i$ is the corpus mean.
The cumulative proportion of variance explained by the first $K$ principal
components provides the optimal affine reconstruction reference at dimension $K$:
\[
R^2_{\mathrm{PCA},K}
=\frac{\sum_{j=1}^{K}\lambda_j}{\sum_{j=1}^{D}\lambda_j}.
\]
We approximate this quantity with randomized singular value decomposition
of the centered embedding matrix \citep{HalkoMartinssonTropp2011}.
The reference matches code dimension, with continuous rather than binary
coefficients.

The reconstruction score $R^2$ and trained/OLS decoder comparison are
defined in Section~\ref{sec:qwen-compression}. For every configuration,
ordinary least squares refits an intercept and $K$ coefficient vectors
to all corpus embeddings while keeping the learned hard codes fixed.
Table~\ref{tab:qwen-width-retention-proposal} reports the resulting
improvements over the saved decoder. Supplementary Section~\ref{app:activity-normalization}
details the role of internal layer normalization in the full pipeline.

\subsection{Qwen geometric preservation diagnostics}
\label{app:qwen-geometry-details}

These comparisons expand Section~\ref{sec:qwen-geometry}, using the same
sampled documents and pairs across all six configurations.

We assess geometric preservation by comparing distances between
the same document pairs in embedding and binary-code space. For each pair
$(i,j)$, angular separation is measured by cosine distance,
$d_{\cos}(u,v)=1-u^\top v/(\lVert u\rVert_2\lVert v\rVert_2)$,
applied separately to $(x_i,x_j)$ and $(B_i,B_j)$. We also compare
Euclidean embedding distances $\lVert x_i-x_j\rVert_2$, using the original
unnormalized embeddings, with Hamming and Jaccard distances between codes.
Writing $m_{ij}=\sum_{k=1}^K|B_{ik}-B_{jk}|$ for the number of coordinates
that differ, normalized Hamming distance is $m_{ij}/K$, whereas Jaccard
distance is $m_{ij}/\sum_k\max(B_{ik},B_{jk})$. Both depend on which bits
differ between documents. Jaccard normalizes these differences by the
number of coordinates active in at least one document.
Since $\lVert B_i-B_j\rVert_2^2=m_{ij}$, Hamming and binary Euclidean
distances induce the same ordering and therefore yield identical Spearman
correlations with embedding distances.

We use the paired-distance sample and average-rank treatment of ties
specified in Section~\ref{sec:qwen-geometry}; no sampled embedding or code
has zero norm. The 100,000 unordered pairs are sampled without replacement.
Table~\ref{tab:qwen-geometry-distances} reports all six
configurations, including the Jaccard comparisons.

\begin{table}[tbph]
  \centering
  \begingroup
  \small
  \setlength{\tabcolsep}{4pt}
  \begin{tabular}{rrrrrr}
    \toprule
    $K$ & $A$ & Cosine--Cosine & Cosine--Jaccard & Euclidean--Hamming & Euclidean--Jaccard \\
    \midrule
    8 & 4 & 0.449 & 0.449 & 0.291 & 0.290 \\
    8 & 6 & 0.453 & 0.452 & 0.335 & 0.336 \\
    16 & 6 & 0.494 & 0.493 & 0.330 & 0.311 \\
    32 & 6 & 0.495 & 0.495 & 0.356 & 0.320 \\
    32 & 8 & 0.496 & 0.495 & 0.374 & 0.312 \\
    32 & 16 & 0.501 & 0.500 & 0.344 & 0.333 \\
    \bottomrule
  \end{tabular}
  \endgroup
  \caption{Distance rank correlations for Qwen on AG News.
  Each entry is Spearman's $\rho$ on the same 100,000 document pairs.
  Column headings give the embedding distance followed by the binary-code
  distance. Euclidean--Hamming is rank-equivalent to Euclidean--Euclidean.
  $K$ is code width and $A$ is target activity.}
  \label{tab:qwen-geometry-distances}
\end{table}

For the local comparison in Table~\ref{tab:qwen-local-profile-agreement},
let $n_{B_i}$ be the number of corpus documents with the sampled document's
profile. Its probability of matching a uniformly selected other document
is $(n_{B_i}-1)/(N-1)$; averaging over the 2,000 sampled documents gives
the random reference. Exact-profile agreement becomes a stricter criterion
as more profiles are distinguished, so the higher agreement of eight-bit
codes should be read alongside their higher random agreement.

\subsection{Qwen predictive evaluation across widths and activity targets}
\label{app:qwen-prediction-details}

The following results supplement Section~\ref{sec:qwen-predictive-utility}.

Table~\ref{tab:qwen-predictive-performance} expands the logistic-probe
evaluation defined in Section~\ref{sec:qwen-predictive-utility}, using
the same fitted codes and label split.
Let $X$ denote the embedding of a randomly selected document,
$Y$ its assigned AG News category, and $e$ the fitted deterministic binary
encoder. Its code is $B=e(X)$. The data-processing inequality
gives $I(Y;B)\le I(Y;X)$. Predictive performance assesses how effectively
the chosen probe uses the representation for this particular taxonomy
\citep{XuEtAl2020}, while the profile analysis examines semantic structure
beyond that taxonomy. We also retain
cross-entropy contrasts: at target six, the excess over raw embeddings falls
from 0.242 to 0.163 bits/document when width increases from 16 to 32.

At target activity six, increasing width from 16 to 32 bits
raises reconstruction $R^2$ from 0.160 to 0.214 and BA from 85.62\% to
87.63\%. The same logistic probe applied to the original Qwen embeddings
achieves 91.09\% BA on the same split. The binary codes therefore achieve
93.99\% and 96.20\% of the embedding-based BA, respectively
(Table~\ref{tab:qwen-predictive-performance}).
The eight-bit code with the same activity target reaches 86.79\% BA,
also exceeding the sixteen-bit reference.
Meanwhile, cosine--Jaccard rank correlation changes little
(0.493 to 0.495; Table~\ref{tab:qwen-geometry-distances}).
At 32 bits, increasing target activity from six to sixteen further improves
reconstruction to 0.232, while BA decreases to 86.42\%, corresponding
to 94.87\% of the embedding-based BA.
The configurations therefore differ in reconstruction fidelity and
predictive utility for the evaluated taxonomy.

\begin{table}[tbp]
  \centering
  \small
  \setlength{\tabcolsep}{4pt}
  \begin{tabular}{lrrlrr}
    \toprule
    Representation & $K$ & $A$ & BA [95\% CI] & Relative BA & Macro-F1 \\
    \midrule
    Original embeddings & -- & -- & 91.09 [90.76, 91.44] & 100.00 & 91.08 \\
    \midrule
    Binary code & 8  & 4  & 85.17 [84.74, 85.59] & 93.50 & 85.10 \\
    Binary code & 8  & 6  & 86.79 [86.38, 87.19] & 95.28 & 86.73 \\
    Binary code & 16 & 6  & 85.62 [85.19, 86.04] & 93.99 & 85.57 \\
    Binary code & 32 & 6  & 87.63 [87.23, 88.02] & 96.20 & 87.60 \\
    Binary code & 32 & 8  & 86.94 [86.54, 87.35] & 95.44 & 86.91 \\
    Binary code & 32 & 16 & 86.42 [86.01, 86.83] & 94.87 & 86.38 \\
    \bottomrule
  \end{tabular}
  \caption{Predictive performance of Qwen representations on AG News.
  Logistic classifiers use the same 80/20 label split, with 102,080
  training documents and 25,520 test documents.
  The original embeddings have 1,024 dimensions; binary codes have
  width $K$ and target activity $A$.
  Balanced accuracy (BA) and macro-F1 are expressed as percentages.
  Brackets give 95\% bootstrap confidence intervals for BA.
  Relative BA is $100\,\mathrm{BA}/\mathrm{BA}_{\mathrm{embeddings}}$,
  calculated from unrounded scores.
  The raw-embedding reference is the saved probe used in the
  width/activity comparison; Supplementary Section~\ref{app:external-prediction} reports
  a separately refitted probe on the same split.
  The binary encoder is fitted without labels on the full corpus;
  labels are used only to train and evaluate the predictive classifiers.}
  \label{tab:qwen-predictive-performance}
\end{table}

\subsection{Loss ablation: representation diagnostics and predictive evaluation}
\label{app:loss-ablation}

This section supports Section~\ref{sec:loss-ablation}. Unless specified
otherwise, values are means and sample standard deviations across the
three paired encoder seeds. They describe variation across fitted
representations on the same corpus. The full objective yields
$6.01\pm0.05$ active bits and $2501.2\pm51.0$ effective profiles.
The latter averages across seeds; the seed-1234 fit has 2,560.2 effective
profiles and supplies the reference for the other AG News comparisons.

Reconstruction errors in Tables~\ref{tab:ablation-representation}
and~\ref{tab:ablation-diagnostics} are divided by the training calibration
variance $s_x$ in \eqref{eq:recon}. This differs from normalization by
the full-corpus mean-baseline error in Table~\ref{tab:external-reconstruction}.
For each fixed hard code, the OLS decoder refits an intercept and
coefficient vectors to all corpus embeddings. The trained decoder's
normalized MSE increases from $0.854\pm0.002$ to $1.035\pm0.000$ when
reconstruction is removed; refitting gives $0.852\pm0.002$ and
$0.918\pm0.003$, respectively. The latter comparison evaluates code
recoverability independently of the original decoder's optimization.

Profile entropy and effective profiles follow
Section~\ref{sec:qwen-compression}. For marginal bit entropy, let
$\widehat\pi_k=N^{-1}\sum_i B_{ik}$ denote the observed prevalence of
bit $k$. We average
$-\widehat\pi_k\log_2\widehat\pi_k-(1-\widehat\pi_k)\log_2(1-\widehat\pi_k)$
over the $K$ coordinates, with $0\log 0=0$. This measures binary
variation across documents. Soft-activation entropy instead averages
$h(p)=-p\ln p-(1-p)\ln(1-p)$ over documents and coordinates, as in
\eqref{eq:entropy}. It is measured in nats (natural units of information),
is zero at $p=0$ or $p=1$, and reaches $\ln2\simeq0.693$ at $p=0.5$.

The geometric correlations reuse the document sample and pair indices
of Section~\ref{sec:qwen-geometry}. Without activity control, the three
fits contain 104, 3, and 167 all-zero codes. Cosine distance is undefined
for these vectors, so pairs involving them are omitted from the angular
comparison, leaving 99,888, 100,000, and 99,899 pairs; all 100,000 pairs
remain in each Euclidean--Hamming comparison. Constant distances in the
collapsed no-balance fits yield undefined rank correlations. Mean absolute
bit correlation is computed over nonconstant coordinates and is likewise
undefined when every bit is constant.

\begin{table}[tbp]
\centering\small
\setlength{\tabcolsep}{3pt}
\resizebox{\textwidth}{!}{%
\begin{tabular}{lrrrrr}
\toprule
Objective & OLS MSE/$s_x$ & $\rho_{\mathrm{cos,cos}}$ & $\rho_{\mathrm{Euc,Ham}}$ & $\widehat H(B)$ & Mean bit entropy \\
\midrule
Full & $0.852\pm0.002$ & $0.484\pm0.009$ & $0.331\pm0.005$ & $11.288\pm0.029$ & $0.954\pm0.003$ \\
Without activity & $0.849\pm0.001$ & $0.450\pm0.022$ & $0.354\pm0.003$ & $11.431\pm0.063$ & $0.949\pm0.003$ \\
Without balance & $1.015\pm0.000$ & -- & -- & $0.000\pm0.000$ & $0.000\pm0.000$ \\
Without decorrelation & $0.852\pm0.002$ & $0.486\pm0.015$ & $0.335\pm0.003$ & $11.240\pm0.075$ & $0.953\pm0.002$ \\
Without entropy & $0.848\pm0.000$ & $0.471\pm0.002$ & $0.345\pm0.006$ & $12.129\pm0.053$ & $0.953\pm0.003$ \\
Without reconstruction & $0.918\pm0.003$ & $0.408\pm0.004$ & $0.267\pm0.006$ & $5.518\pm0.361$ & $0.943\pm0.001$ \\
\bottomrule
\end{tabular}%
}
\caption{Representation diagnostics in the loss ablation, as mean $\pm$
sample SD across three encoder seeds. Entropies are in bits. Correlations
compare cosine distances in both spaces or embedding Euclidean and binary
Hamming distances. Dashes denote undefined correlations for constant codes.}
\label{tab:ablation-representation}
\end{table}

\begin{table}[tbph]
\centering
\begingroup
\footnotesize
\setlength{\tabcolsep}{4pt}
\begin{tabular}{llrrr}
\toprule
Removed term & Diagnostic & Full & Ablated & Paired change \\
\midrule
Activity control & Absolute activity error
& $0.753\pm0.011$ & $1.319\pm0.025$ & $+0.566\pm0.036$ \\
Balance & Prevalence SD
& $0.014\pm0.002$ & $0.484\pm0.000$ & $+0.470\pm0.002$ \\
Decorrelation & Absolute correlation
& $0.120\pm0.006$ & $0.128\pm0.002$ & $+0.008\pm0.008$ \\
Entropy & Conditional entropy (nats)
& $0.0076\pm0.0002$ & $0.3026\pm0.0002$ & $+0.2950\pm0.0004$ \\
Reconstruction & Decoder MSE/$s_x$
& $0.854\pm0.002$ & $1.035\pm0.000$ & $+0.181\pm0.002$ \\
\bottomrule
\end{tabular}
\endgroup
\caption{Effect of each omitted term on its associated diagnostic.
Values are mean $\pm$ sample SD across three seeds, with paired changes.
Activity error is the corpus mean of $|\sum_k B_{ik}-6|$;
prevalence SD is calculated across the 16 bit prevalences.
Absolute correlation is averaged over pairs of nonconstant binary factors.
Conditional Bernoulli entropy is averaged over documents and bits and
expressed in nats (natural units of information), using natural logarithms.
Decoder MSE is normalized by the training calibration variance $s_x$.}
\label{tab:ablation-diagnostics}
\end{table}

The community analysis applies the positive-PMI graph protocol of
Supplementary Section~\ref{app:qwen-graph-protocol} independently to each
fit, retaining all 16 bit nodes and using weighted Leiden modularity at
resolution $\gamma=1$. Twenty optimizer seeds (1234--1253) assess partition
stability on each fixed graph. All full-objective fits have three
communities, as do fits without decorrelation, entropy, or reconstruction;
without activity control the counts are four, three, and three. Without
balance there are no positive-PMI edges and all nodes are isolated.

\begin{figure}[tbp]
\centering
\includegraphics[width=\textwidth]{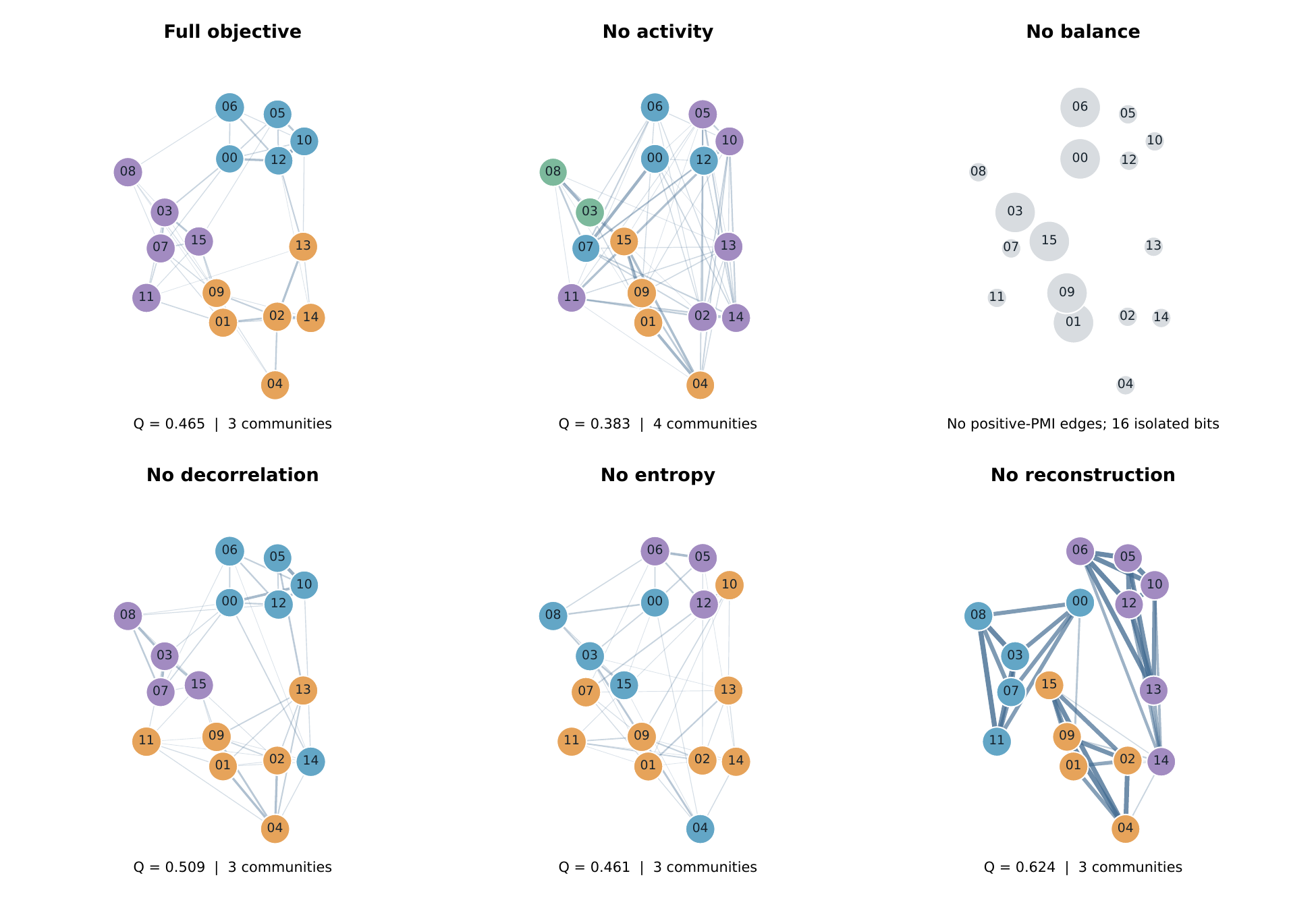}
\caption{Positive-PMI bit graphs for the loss ablation, using encoder
seed 1234. Colours identify communities fitted separately in each panel
by Leiden at resolution $\gamma=1$ and optimizer seed 1234; $Q$ is weighted
modularity. All panels reuse node positions from the full-objective graph.
Node area increases with activation prevalence; edge width and opacity
increase with PMI on a common scale. The no-balance code has 16 isolated
nodes, excluded from the community count. Shared node positions identify
encoder coordinates rather than an independently established semantic
alignment between ablated models.}
\label{fig:ablation-community-networks}
\end{figure}

Figure~\ref{fig:ablation-community-networks} shows the seed-1234 graphs.
Mean modularity across encoder fits is 0.486 for the full objective,
0.398 without activity, 0.480 without decorrelation, 0.439 without entropy,
and 0.610 without reconstruction. The last result accompanies a sharp
reduction in profile diversity and stronger bit dependence: more separated
co-activation groups can occur within a more repetitive code. Mean pairwise
adjusted Rand agreement across Leiden seeds is 1.000 for all nonempty
graphs except two no-entropy fits, where it is 0.958 and 0.979. This measures
optimizer stability on fixed graphs, separately from variation across
encoder fits.

As a secondary evaluation, class-balanced logistic classifiers
use the same 80/20 label split as the other AG News analyses, with either
deterministic binary codes or activation probabilities $p_i$ as inputs.
Probabilities are entered directly, rather than transformed into log-odds
as in Supplementary Section~\ref{app:external-prediction}. The representation
is fitted without labels on the full corpus. Table~\ref{tab:ablation-utility}
reports balanced accuracy and paired changes across encoder seeds.

Removing balance gives the constant-code result of 25\% BA. Removing
reconstruction reduces binary BA by $4.59\pm0.93$ percentage points,
although category information remains accessible at $81.02\pm0.81\%$ BA.
Without entropy regularization, continuous BA rises from 85.77\% to
87.90\%, while binary BA falls from 85.61\% to 84.99\%, widening the
continuous--binary difference. Removing activity or decorrelation has
little average effect on binary BA in these fits.

\begin{table}[tbph]
\centering
\begingroup
\footnotesize
\setlength{\tabcolsep}{4pt}
\begin{tabular}{lrrrr}
\toprule
Objective & Binary BA & $\Delta$ binary BA & Continuous BA & Effective profiles \\
\midrule
Full & $85.61\pm0.32$ & $0.00$ & $85.77\pm0.35$ & $2501.2\pm51.0$ \\
Without activity control & $85.63\pm0.24$ & $+0.02\pm0.27$ & $85.73\pm0.23$ & $2763.2\pm122.0$ \\
Without balance & $25.00\pm0.00$ & $-60.61\pm0.32$ & $25.00\pm0.00$ & $1.0\pm0.0$ \\
Without decorrelation & $85.80\pm0.68$ & $+0.18\pm0.64$ & $85.90\pm0.66$ & $2420.2\pm124.1$ \\
Without entropy & $84.99\pm0.08$ & $-0.63\pm0.24$ & $87.90\pm0.10$ & $4479.7\pm167.2$ \\
Without reconstruction & $81.02\pm0.81$ & $-4.59\pm0.93$ & $81.14\pm0.86$ & $46.8\pm11.2$ \\
\bottomrule
\end{tabular}
\endgroup
\caption{Predictive utility and profile diversity in the loss ablation. Values are mean $\pm$ sample SD across three
seeds. BA is expressed as a percentage; $\Delta$ is the paired change in
binary BA, in percentage points, relative to the full objective. The continuous
probe uses activation probabilities. Effective profiles are $2^{\widehat H(B)}$.}
\label{tab:ablation-utility}
\end{table}

\subsection{Organization across code capacities}
\label{app:qwen-capacity-organization}

We next examine how the organization of the code changes
with its capacity. Here, capacity refers to the representational budget provided
by $K$ binary coordinates, with at most $2^K$ possible profiles. The activity
target controls their use. We compare $K=8,16,32$ at a common target of six
active bits, corresponding to target activation fractions of 75\%, 37.5\%,
and 18.75\%. Activity is encouraged by a penalty rather than fixed exactly.
The three fitted codes contain 78, 12,174, and 48,371 observed profiles,
respectively. Thus, increasing width permits a finer empirical differentiation
of documents while keeping the target number of active coordinates constant.

\begin{figure}[ph]
  \centering
\includegraphics[width=0.98\textwidth,height=0.34\textheight,keepaspectratio]{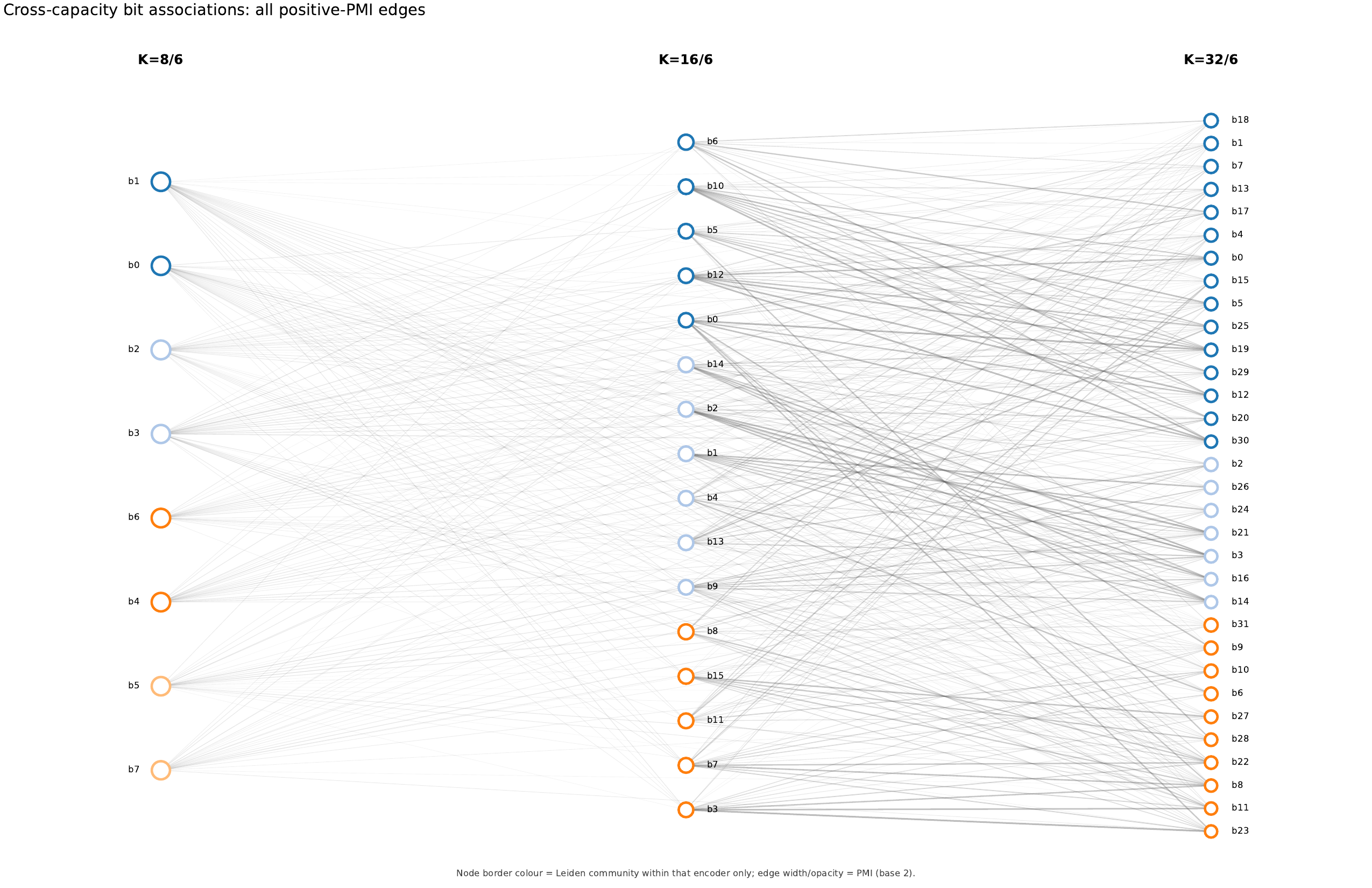}
  \par\medskip
\includegraphics[width=0.98\textwidth,height=0.30\textheight,keepaspectratio]{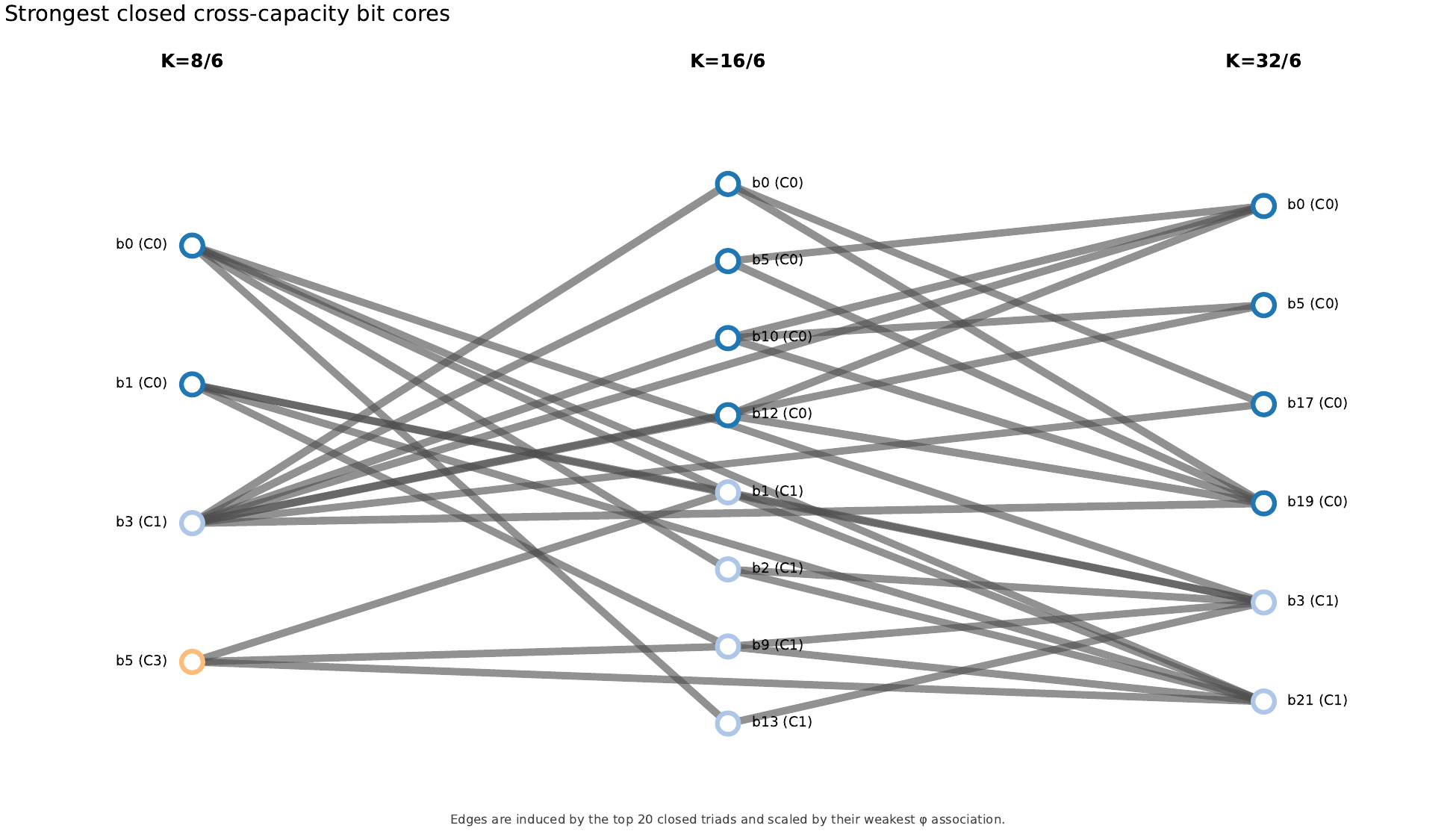}
  \caption{Tripartite bit-association graphs for Qwen on AG News.
  Columns represent the 8/6, 16/6, and 32/6 configurations, where the second
  number is target activity. Top: all positive-PMI edges; width and opacity
  increase with PMI. Bottom: the union of the 20 closed triads ranked highest
  by their minimum $\phi$. Edge widths increase with the largest minimum-$\phi$
  score among selected triads containing that edge. Only participating nodes
  are shown in the lower panel. Node-border colors and community IDs refer
  to within-encoder Leiden partitions; shared colors across columns do not
  encode a fitted community matching.}
  \label{fig:qwen-tripartite-capacity}
\end{figure}

To trace associations across these representations, we construct
a tripartite graph: its three node sets contain the bits of the 8/6, 16/6,
and 32/6 encoders, and edges connect bits from different encoders. Pairwise
associations are computed from their activations on the same 127,600 documents,
without category labels. The full graph contains 56 nodes and 456 edges with
finite positive PMI (Figure~\ref{fig:qwen-tripartite-capacity}, top).
Node colors record the Leiden communities fitted separately within each
encoder; community identifiers and bit indices are local to each code.

For a compact view, we identify closed triads containing one
bit from each encoder, with positive PMI and $\phi$ on all three edges.
We rank the 750 qualifying triads by their weakest $\phi$ association and
show the union of the 20 highest-ranked triads in the lower panel.
This criterion favors associations supported across all three pairs.
The strongest triad links bit 3 of 8/6, bit 0 of 16/6, and bit 19 of 32/6,
with minimum $\phi=0.270$ and mean $\phi=0.400$.

The selected triads reveal recurring groups of associations.
Bit 3 of 8/6 connects to bits 0, 5, 10, and 12 in a single 16/6 community,
which in turn connect to bits 0, 5, 17, and 19 in a single 32/6 community.
A second group links bit 5 of 8/6 to bits 1 and 9 of 16/6 and bit 21 of
32/6. These correspondences suggest that activation patterns represented by
individual coordinates in a narrower code can be distributed over several
coordinates in wider codes. At the document level, the conjunction of bits
2 and 3 in 8/6 selects 93,871 documents and is most strongly associated with
bit 10 of 16/6 ($\phi=0.436$) and bit 19 of 32/6 ($\phi=0.270$).
Together, the graph and conjunction analysis describe recurring co-activation
structure alongside the increasing diversity of complete profiles.

Cross-capacity associations also include inverse activation
patterns: there are 62 negative pairs between 8/6 and 16/6, 94 between 8/6
and 32/6, and 284 between 16/6 and 32/6. For example, bit~3 of 8/6 and
bit~21 of 32/6 have $\phi=-0.789$. The latter activates in 70.7\% of
documents where the former is inactive, compared with 0.47\% where it is
active. Thus, correspondence across capacities can involve the inactive
state of a coordinate as well as its active state. The positive-edge graphs
in Figure~\ref{fig:qwen-tripartite-capacity} summarize the co-activation
component of this relationship.

\section{Bit associations, activity constraints, and encoder normalization}
\label{app:activity-normalization}

\subsection{Qwen bit-association graph and sensitivity analysis}
\label{app:qwen-graph-protocol}

This subsection supplies the graph protocol for
Section~\ref{sec:agnews-bit-structure}. Each node represents one of the
sixteen Qwen bits. All 43 pairs with finite, strictly positive PMI enter
the undirected graph, with PMI as the edge weight and no additional
threshold. Leiden \citep{TraagEtAl2019} optimizes weighted modularity
with a configuration-model null \citep{ReichardtBornholdt2006}, comparing
within-community weights with the expectation based on weighted degrees.
The resolution parameter $\gamma=1$ gives standard modularity; larger
values favor finer partitions.

We run seeds 1234--1253 at each
$\gamma\in\{0.5,0.75,1,1.25,1.5,2\}$ and retain the partition with the
highest objective value within each resolution. Repeated seeds assess
sensitivity to the stochastic optimizer; the grid assesses coarser and
finer groupings. All seeds return the same partition at each tested
resolution, with pairwise adjusted Rand index (ARI) equal to one
\citep{HubertArabie1985}. The reference three-community partition is
unchanged at $\gamma=0.75$ and $1.25$; the graph has two communities at
$0.5$ and four at $1.5$ and $2$.

For non-constant binary variables, PMI and $\phi$ have the same sign.
We therefore overlay all 77 negative-$\phi$ pairs without changing the
43 edges used for community detection. Community fitting and spring layout
use positive PMI weights; displayed widths and opacity encode $|\phi|$
for both signs. The eight negative associations internal to communities
have $\phi$ between $-0.080$ and $-0.016$. The overlay is descriptive,
not an evaluation against an activity-preserving random reference.

In Figure~\ref{fig:qwen_PMI}, bit~2 co-activates with bits~1, 9, 13,
and 14 more often than marginal independence predicts, as does bit~0
with bit~12. Bits~0 and 2, and bit~1 paired with bits~10 or 12,
co-activate less than this reference predicts. Complete mutual exclusion
is the special case of zero joint activation, giving PMI $=-\infty$
when both marginal prevalences are positive. A \emph{dead} bit never
activates and a \emph{universal} bit is always active; neither occurs
here. Assessing individual pair associations beyond the activity constraint
would require a reference preserving both document-level activity and
marginal bit frequencies.

\subsection{Activity constraints and aggregate dependence}

Let $S=\sum_{k=1}^{K}B_k$ be document-level activity and
$q_k=\Pr(B_k=1)$. The variance identity
\[
\operatorname{Var}(S)=\sum_k q_k(1-q_k)
 +2\sum_{k<\ell}\operatorname{Cov}(B_k,B_\ell)
\]
relates activity variation to pairwise dependence. In the Qwen 16/6
codes, the empirical activity variance is 1.067, whereas the sum of
marginal Bernoulli variances is 3.732. The sum of pairwise covariances
is therefore $-1.333$, and the mean pairwise $\phi$ correlation is
$-0.0476$. For illustration, exchangeable bits with exactly fixed
activity would have pairwise correlation $-1/(K-1)$, or $-0.0667$
at $K=16$, even without different semantic roles. Our activity is
approximately, rather than exactly, fixed, and its bits have different
prevalences. The exchangeable fixed-activity value is therefore illustrative,
not a calibrated null for the observed graph. The identity accounts for
the aggregate covariance constraint; it does not determine which
specific pairs are negative or how much of each association is semantic.

\subsection{Encoder normalization and reconstruction references}

Layer normalization is applied within the encoder as
\[
\operatorname{LN}(x_i)_j=\gamma_j
 \frac{x_{ij}-\mu_i}{\sqrt{v_i+\epsilon}}+\beta_j,
\qquad
\mu_i=D^{-1}\sum_jx_{ij},\quad
v_i=D^{-1}\sum_j(x_{ij}-\mu_i)^2,
\]
where $\gamma_j$ and $\beta_j$ are learned parameters and $\epsilon$
is the numerical stabilizer. This removes the coordinate mean of
each vector and rescales its centered coordinates before encoding,
while reconstruction still targets the raw vector. Thus, the comparison
with raw embeddings or PCA concerns the full learned representation,
including normalization, nonlinear encoding, binary activations,
and the learning objective. It does not isolate the contribution of
binary discretization to reconstruction error or distance distortion.

\section{Cross-backbone correspondence and diagnostics}
\label{app:backbone-details}

\subsection{Backbone characteristics and relative performance}
\label{app:backbone-characteristics}

This supplementary section expands Section~\ref{sec:backbone-robustness}, using the
same five fitted encoders and all 127,600 AG News documents.

Qwen retains causal attention and uses the final token representation.
MiniLM uses a six-layer bidirectional BERT-style encoder with mean
pooling. The NVIDIA model adapts Llama 3.2 to bidirectional attention
and also uses mean pooling. Harrier uses a Gemma 3 decoder architecture
with causal attention and last-token pooling; its training includes
knowledge distillation from larger embedding models.
EmbeddingGemma 2 uses a Gemma 4-based bidirectional encoder with
mean pooling and a projection from 512 hidden dimensions to
768 output dimensions. We use its text representations.
In our experiments, the pooled embedding outputs are retained
without L2 normalization before fitting the binary encoder.
Each encoder uses $K=16$ bits with target activity $A=6$ and is trained
without labels. 

Predictive evaluation uses the common AG News 80/20 split and
logistic-classification settings in
Supplementary Section~\ref{app:classification-protocol}.
We first assess the relative impact of compression within each
backbone, then examine whether profiles learned from different
embedding spaces identify corresponding document groups.
Table~\ref{tab:backbone-relative-retention} reports predictive
performance relative to the original embeddings, the relative BA = 
$100\,\mathrm{BA}_{\mathrm{binary}}/
\mathrm{BA}_{\mathrm{embeddings}}$.
We also report the relative reconstruction = 
$100\,R^2_{\mathrm{trained}}/R^2_{\mathrm{PCA},16}$,
which compares the trained binary decoder with the continuous
16-dimensional PCA reference for the same embedding space.
The first ratio measures relative predictive performance;
the second measures the fraction of PCA-explained variance recovered
by the binary decoder.

Mean activity remains close to six for every backbone, while the
effective number of profiles ranges from 521.3 to 6488.6.
The same code budget therefore produces different distributions
of document groups. Binary representations achieve between
92.38\% and 95.24\% of their respective embedding-based BA.
Relative reconstruction ranges from 52.87\% to 70.73\% of the
PCA reference. The backbone ordering differs between these measures,
indicating that preservation of embedding variation and accessibility
of category information respond differently to compression.
Table~\ref{tab:backbone-size-correlations} reports descriptive
associations between backbone size and performance.
We log-transform parameter counts because they span more than
two orders of magnitude, while retaining embedding dimensions
on their original scale. Predictive performance is more strongly
associated with log parameter count than with embedding dimension,
for both original embeddings and binary codes.
For binary BA, the respective Pearson correlations are 0.849
and 0.688. Binary reconstruction shows the opposite pattern:
its Pearson correlation is 0.787 with embedding dimension and
0.447 with log parameter count. Spearman correlations support
the same ordering. 
 These associations
are descriptive, as the comparison includes only the five backbones considered here. 
\begin{table}[tbph]
  \centering
  \small
  \setlength{\tabcolsep}{4pt}
  \resizebox{\textwidth}{!}{%
  \begin{tabular}{lrrrrrrrr}
    \toprule
    & \multicolumn{2}{c}{Embedding BA}
    & \multicolumn{2}{c}{Binary BA}
    & \multicolumn{2}{c}{PCA reconstruction}
    & \multicolumn{2}{c}{Binary reconstruction} \\
    \cmidrule(lr){2-3}\cmidrule(lr){4-5}
    \cmidrule(lr){6-7}\cmidrule(lr){8-9}
    & Pearson $r$ & Spearman $\rho$
    & Pearson $r$ & Spearman $\rho$
    & Pearson $r$ & Spearman $\rho$
    & Pearson $r$ & Spearman $\rho$ \\
    \midrule
    $\log_{10}(\text{number of parameters})$
    & 0.793 & 0.667 & 0.849 & 0.821
    & 0.280 & 0.103 & 0.447 & 0.359 \\
    Embedding dimension
    & 0.616 & 0.600 & 0.688 & 0.700
    & 0.386 & 0.300 & 0.787 & 0.500 \\
    \bottomrule
  \end{tabular}%
  }
  \caption{Descriptive correlations between backbone size and
  predictive or reconstruction performance across five models.
  Parameter counts are log-transformed; embedding dimensions
  remain on their original scale.
  Embedding BA uses the original embeddings; binary BA uses
  16-bit codes with target activity six.
  PCA reconstruction is $R^2_{\mathrm{PCA},16}$;
  binary reconstruction is $R^2_{\mathrm{trained}}$.
  For EmbeddingGemma 2, parameter count refers to the text component.}
  \label{tab:backbone-size-correlations}
\end{table}

\subsection{Complete-profile partitions and document-based matching}
\label{app:backbone-profiles}

We first compare the complete document partitions induced by the
five binary encoders. Each distinct 16-bit profile defines one group,
and every document belongs to exactly one group within each backbone.
Table~\ref{tab:backbone-partition-sizes} describes their granularity.
The number of observed profiles ranges from 4,862 for NVIDIA to
20,470 for MiniLM. Documents with singleton profiles account for
1.26\% to 6.11\% of the corpus.

\begin{table}[tbph]
  \centering
  \small
  \begin{tabular}{lrrr}
    \toprule
    Backbone & Observed profiles &
    \shortstack{Mean documents\\per profile} &
    \shortstack{Documents in\\singleton profiles (\%)} \\
    \midrule
    Qwen             & 12,174 & 10.48 & 3.48 \\
    MiniLM           & 20,470 &  6.23 & 6.11 \\
    NVIDIA           &  4,862 & 26.24 & 1.26 \\
    Harrier          &  5,462 & 23.36 & 1.31 \\
    EmbeddingGemma 2  &  6,490 & 19.66 & 1.54 \\
    \bottomrule
  \end{tabular}
  \caption{Complete profile partitions on 127,600 AG News documents,
  with $K=16$ and target activity six. Observed profiles count distinct
  binary codes, whereas the effective profile counts reported earlier
  also account for their relative frequencies. Singleton percentages
  use all documents as the denominator.}
  \label{tab:backbone-partition-sizes}
\end{table}

For each backbone pair, we calculate the adjusted Rand index
(ARI; \citealp{HubertArabie1985}) and adjusted mutual information
(AMI; \citealp{VinhEppsBailey2010}). ARI measures agreement on whether
document pairs share the same profile. AMI measures the
information shared between profile assignments, correcting for its
expectation under random assignments with the observed group sizes
held fixed. Let $Z_m$ denote a document's profile identifier under
backbone $m$. We use the arithmetic normalization
\[
\operatorname{AMI}(Z_m,Z_\ell)
=
\frac{I(Z_m;Z_\ell)-\mathbb{E}_0[I(Z_m;Z_\ell)]}
{\tfrac{1}{2}\{H(Z_m)+H(Z_\ell)\}
-\mathbb{E}_0[I(Z_m;Z_\ell)]}.
\]
Both scores equal one for identical partitions and have a
chance-reference expectation of zero. They use all documents,
require neither category labels nor bit alignment, and involve
no nearest-neighbour selection or tie-breaking.
They assess exact grouping: distinct profiles remain separate
regardless of their Hamming distance.

\begin{table}[tbph]
  \centering
  \small
  \begin{tabular}{llrr}
    \toprule
    Backbone 1 & Backbone 2 & ARI & AMI \\
    \midrule
    Qwen    & MiniLM           & 0.095 & 0.233 \\
    Qwen    & NVIDIA           & 0.106 & 0.376 \\
    Qwen    & Harrier          & 0.087 & 0.284 \\
    Qwen    & EmbeddingGemma 2 & 0.059 & 0.221 \\
    MiniLM  & NVIDIA           & 0.077 & 0.295 \\
    MiniLM  & Harrier          & 0.065 & 0.213 \\
    MiniLM  & EmbeddingGemma 2 & 0.045 & 0.165 \\
    NVIDIA  & Harrier          & 0.169 & 0.369 \\
    NVIDIA  & EmbeddingGemma 2 & 0.099 & 0.304 \\
    Harrier & EmbeddingGemma 2 & 0.077 & 0.263 \\
    \bottomrule
  \end{tabular}
  \caption{Chance-adjusted agreement between complete profile
  partitions across backbones. Each comparison includes the same
  127,600 documents and all observed profiles, including singletons.
  ARI is the adjusted Rand index; AMI uses the arithmetic mean
  of partition entropies for normalization.
  Values describe the saved fitted encoders.}
  \label{tab:backbone-partition-agreement}
\end{table}

Table~\ref{tab:backbone-partition-agreement} shows positive
chance-adjusted agreement for every backbone pair, with ARI ranging
from 0.045 to 0.169 and AMI from 0.165 to 0.376.
NVIDIA and Harrier have the highest ARI, while Qwen and NVIDIA
have the highest AMI. MiniLM and EmbeddingGemma 2 have the lowest
values on both measures. Exact document-grouping agreement is
therefore limited, although the partitions share information.
The scores reflect differences in both group membership and
granularity: dividing a document group among several distinct profiles
changes the partition even when the resulting groups contain related
semantic content.

To examine individual correspondences within these global results,
we next match complete profiles through membership of their document groups.
Bit indices are local to each encoder, so matching operates on the
documents assigned to a profile rather than on the positions of its
active bits. Starting from a reference profile (Qwen in our case), we identify the most
frequent profile among documents in its group for each remaining backbone.
We select the backbone-profile pair with the largest conditional
share, restrict the document set to that intersection, and repeat
until all five backbones are represented. We call the resulting set of
matched profiles a matched profile family. At each step, we record
the retained share and its lift relative to the selected profile's
overall corpus frequency. This procedure allows several reference
profiles to correspond to the same profile in another backbone.
Neither category labels nor a correspondence between bit indices
enters the matching.

\begin{figure}[p]
  \centering
  \includegraphics[width=0.49\textwidth]
    {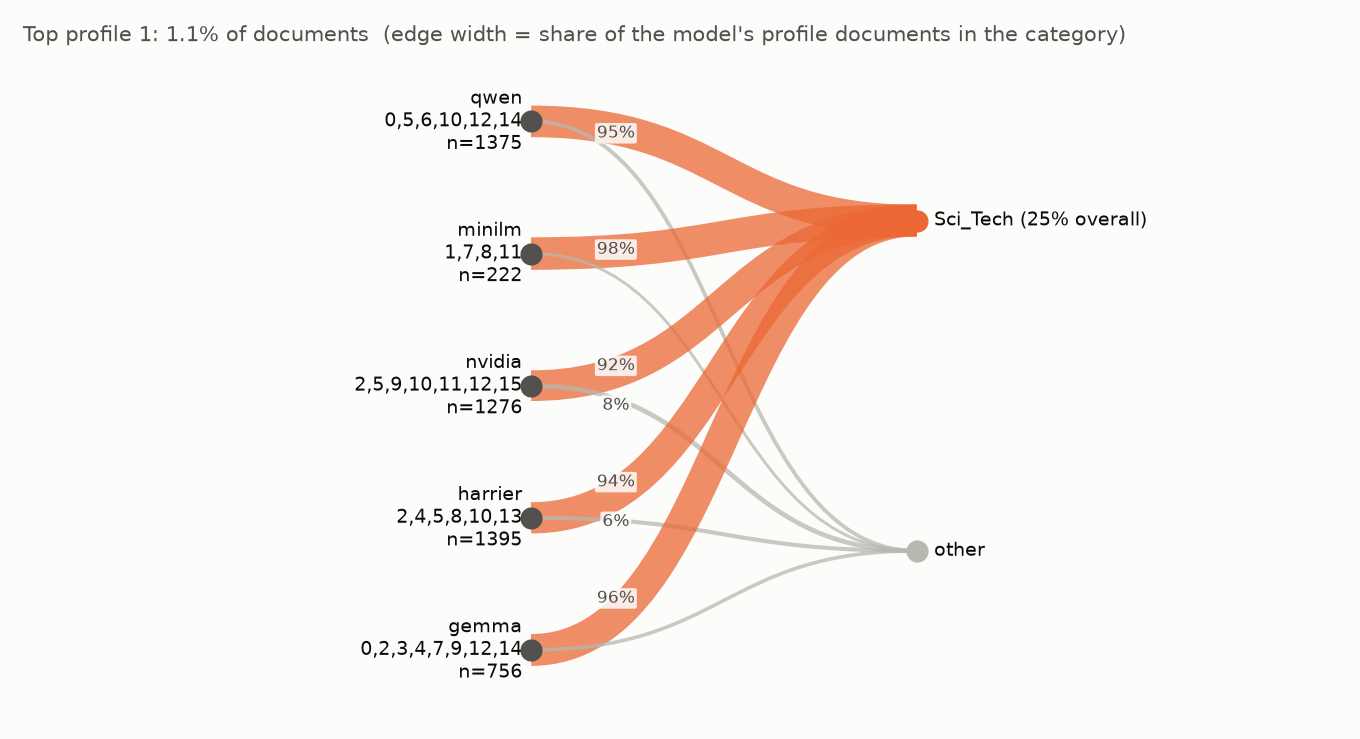}
  \hfill
  \includegraphics[width=0.49\textwidth]
    {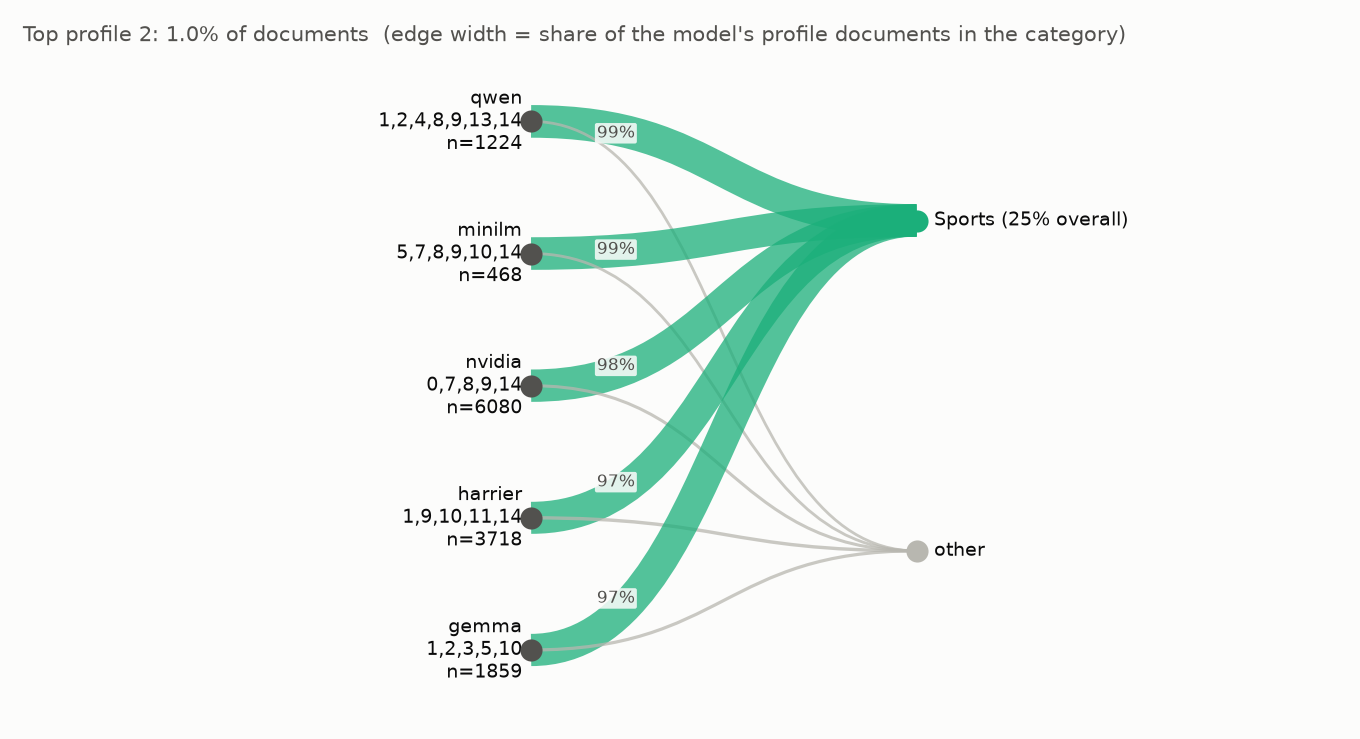}

  \medskip

  \includegraphics[width=0.49\textwidth]
    {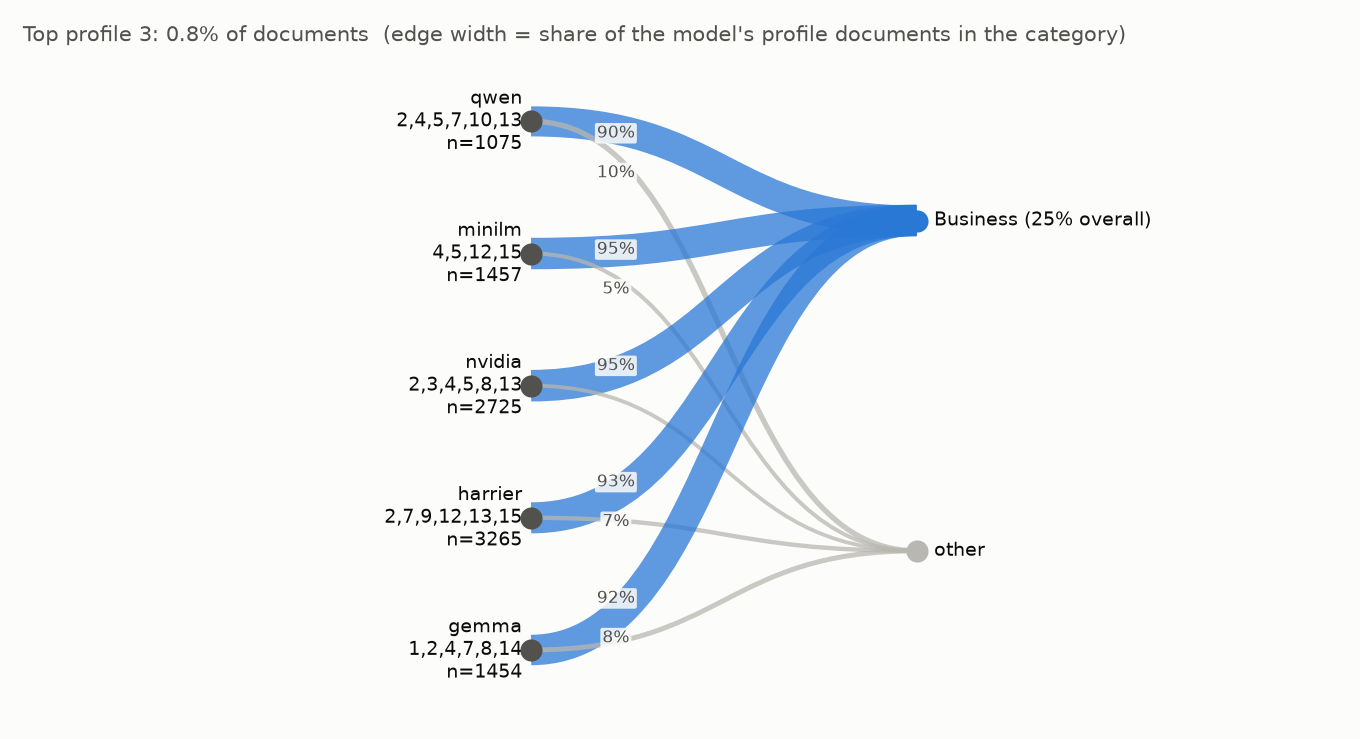}
  \hfill
  \includegraphics[width=0.49\textwidth]
    {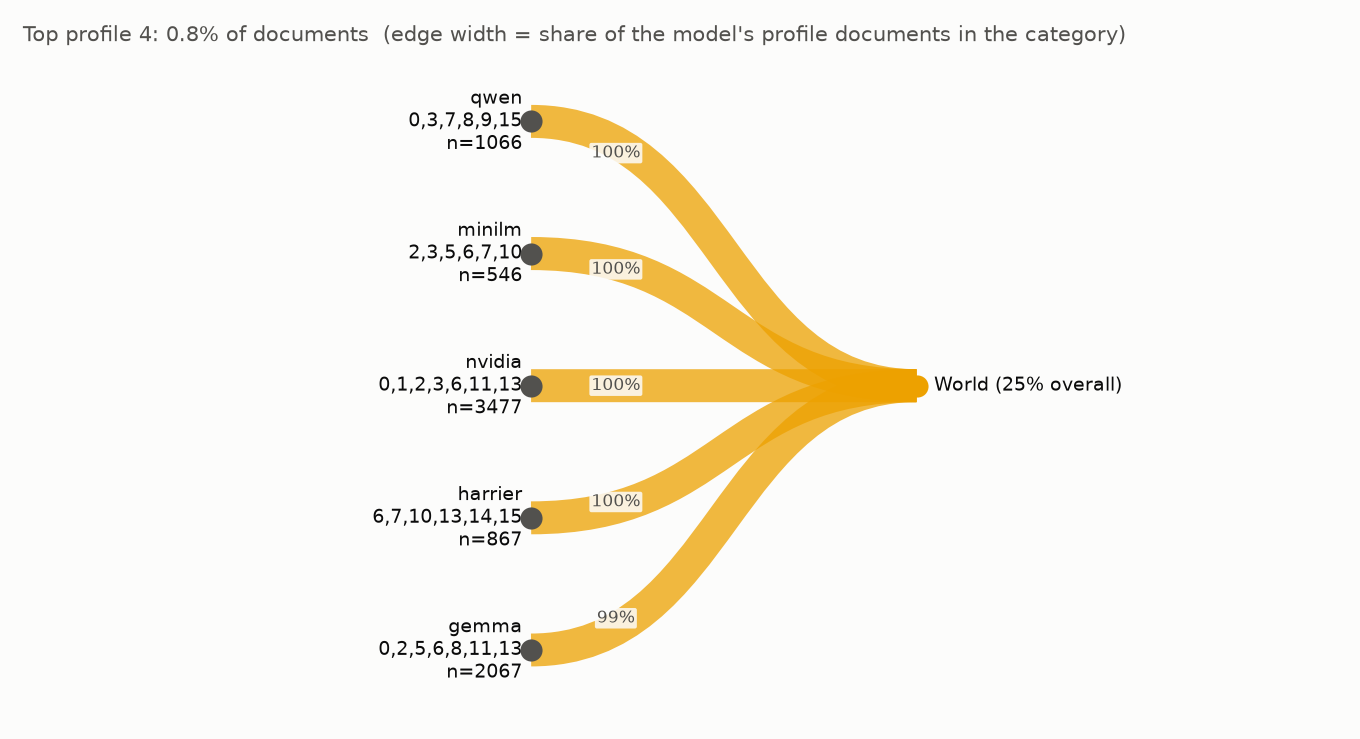}

  \caption{Matched profile families across five backbones, starting
  from the four most frequent Qwen profiles in descending frequency
  order: top left, top right, bottom left, and bottom right.
  Each model node lists the active bits and the full number of
  documents assigned to its matched profile.
  Ribbons show category proportions in the document group defined by each model's profile; small category contributions are grouped as Other.
  Matching uses document membership, while categories provide
  a subsequent interpretation.}
  \label{fig:backbone-profile-families}
\end{figure}

The matching artifacts use each backbone in turn as reference.
Here we focus on the 20 most frequent Qwen profiles, whose document groups
together contain 16,838 documents, or 13.20\% of the corpus.
Figure~\ref{fig:backbone-profile-families} presents the four most
frequent profiles and the category compositions of their matched
counterparts. Category distributions are calculated using the full
document membership of each matched profile, allowing us to examine
whether membership-based matches also have similar external
interpretations.

The examples show recurring category composition alongside differences
in the size and membership of their document groups. Approximately 90\% of the documents
in the second Qwen profile map to one NVIDIA profile; the matched
profiles are predominantly associated with \texttt{Sports}.
For the third Qwen profile, approximately 70\% map to one Harrier
profile, with \texttt{Business} predominant across the matched profile family.
For the fourth, approximately 60\% map to one NVIDIA profile,
and the matched profiles are predominantly associated with \texttt{World}.
The first Qwen profile, predominantly associated with \texttt{Sci\_Tech},
has a smaller initial overlap of approximately 19\% with its selected
Harrier counterpart. Its matched profile family nevertheless retains
similar category composition.

\begin{table}[tbph]
  \centering
  \begingroup
  \small
  \setlength{\tabcolsep}{5pt}
  \begin{tabular}{rrlrr}
    \toprule
    Rank & Reference documents & First matched backbone &
    \shortstack{Initial\\overlap (\%)} &
    \shortstack{Five-profile\\intersection} \\
    \midrule
    1 & 1375 & Harrier & 19 &  2 \\
    2 & 1224 & NVIDIA  & 90 & 68 \\
    3 & 1075 & Harrier & 70 & 69 \\
    4 & 1066 & NVIDIA  & 60 & 56 \\
    \bottomrule
  \end{tabular}
  \endgroup
  \caption{Document overlap for the four profile families in
  Figure~\ref{fig:backbone-profile-families}.
  Rank refers to Qwen profile frequency.
  Initial overlap is the percentage of reference documents retained
  after the first matching step, rounded to the nearest percent.
  The final column counts documents in the intersection of the five
  document groups defined by the matched profiles.}
  \label{tab:backbone-profile-overlap}
\end{table}

To summarize category-composition agreement across the 20 matched profile
families, let $\pi_m$ denote the category distribution of the document group
defined by the selected profile from backbone $m$. We define category coherence as
\[
\mathcal C
=1-\frac{1}{\binom{5}{2}}
  \sum_{m<\ell}\operatorname{JS}_2(\pi_m,\pi_\ell),
\]
where $\operatorname{JS}_2$ is Jensen--Shannon divergence computed
with base-two logarithms:
\[
\operatorname{JS}_2(p,q)
=H\!\left(\frac{p+q}{2}\right)
-\frac{H(p)+H(q)}{2}.
\]
Category coherence equals one when all five document groups in a matched
profile family have identical category distributions. Across the 20 matched
profile families, observed category coherence ranges
from 0.985 to approximately 1.000.

We compare these values with two reference procedures, each using
1,000 random draws. The first replaces each non-reference profile
with another profile from the same backbone whose document group has
approximately similar size. The second independently permutes document alignment
for the non-reference backbones and reruns the complete matching
procedure; category composition is then evaluated on the original
document memberships of the selected profiles.
Weighted by Qwen reference-profile frequency, observed mean category coherence
is 0.996, compared with 0.349 and 0.346 under the two respective
reference procedures. Thus, the matched families have substantially
more similar category composition than these reference pairings.
Document overlap provides a complementary view of this agreement.
Eight of the 20 Qwen-based families retain at least 30 documents
in the intersection of the document groups defined by all five matched profiles.
The four examples in Table~\ref{tab:backbone-profile-overlap} retain
between two and 69 documents. Similar category composition can
therefore accompany substantial differences in document-group boundaries,
as illustrated particularly by the first profile family.

\subsection{Matched bit communities and conditional patterns}
\label{app:backbone-bits}

We examine correspondence across backbones by matching individual
bits on the same 127,600 documents. For each of the ten backbone pairs,
the Hungarian algorithm selects a one-to-one assignment between their
16 bits that maximizes total normalized mutual information (NMI).
For binary coordinates $U$ and $V$, we use
$\operatorname{NMI}(U,V)=2I(U;V)/(H(U)+H(V))$.
The resulting network contains 80 nodes and 160 selected edges.
NMI measures dependence, including complementary activation patterns;
we therefore record the sign of the binary Pearson correlation $\phi$
separately from the NMI magnitude. Of the selected edges, 109 are positive
and 51 negative. Category labels enter neither matching nor clustering.

To group concordant activations, we apply Leiden
\citep{TraagEtAl2019} to the positive-edge subgraph, using NMI weights
and configuration-model modularity with resolution one. We retain the
highest-quality partition across 20 fixed seeds. Negative edges are
shown as associations between nodes but do not enter the clustering
objective. Figure~\ref{fig:backbone-bit-matching-activation} contains
11 connected communities and two singleton nodes. Ten communities
include all five backbones and one includes four. The communities
contain 94.1\% of positive-edge weight; all 51 negative selected edges
connect different communities or singleton nodes, although this was not imposed. Mean pairwise
ARI across optimization seeds is 0.942. Resolutions 0.5 and 2 yield
10 and 15 multi-node communities, respectively, with the same two
singletons and no negative selected edges within communities. Thus, the
separation of activation directions persists across these settings,
while community boundaries depend on resolution.

For each multi-node community, we select documents with an active
member bit in at least two distinct backbones. This defines overlapping
document groups and permits different members of a backbone to contribute.
For each singleton, we instead report all documents activating that bit.
Table~\ref{tab:backbone-community-composition} orders all four category
shares within each group; the corpus reference is 25\% per category.
These category descriptions are subsequent to the label-free graph analysis.

\begin{table}[tbp]
\centering\footnotesize
\setlength{\tabcolsep}{3pt}
\resizebox{\textwidth}{!}{%
\begin{tabular}{lrrllll}
\toprule
Community/node & Documents & Coverage (\%) & First & Second & Third & Fourth \\
\midrule
C1 & 104,797 & 82.1 & \texttt{World} (30.0) & \texttt{Sci\_Tech} (27.6) & \texttt{Business} (26.8) & \texttt{Sports} (15.6) \\
C2 & 80,745 & 63.3 & \texttt{Sports} (39.5) & \texttt{Business} (25.2) & \texttt{Sci\_Tech} (24.5) & \texttt{World} (10.8) \\
C3 & 82,792 & 64.9 & \texttt{Business} (36.9) & \texttt{Sci\_Tech} (31.9) & \texttt{World} (20.0) & \texttt{Sports} (11.2) \\
C4 & 100,368 & 78.7 & \texttt{World} (28.1) & \texttt{Business} (25.5) & \texttt{Sports} (25.1) & \texttt{Sci\_Tech} (21.4) \\
C5 & 97,065 & 76.1 & \texttt{Sports} (28.4) & \texttt{Business} (28.3) & \texttt{Sci\_Tech} (27.2) & \texttt{World} (16.1) \\
C6 & 64,187 & 50.3 & \texttt{Sports} (47.4) & \texttt{World} (39.9) & \texttt{Business} (7.0) & \texttt{Sci\_Tech} (5.6) \\
C7 & 71,821 & 56.3 & \texttt{Business} (37.8) & \texttt{Sci\_Tech} (28.3) & \texttt{World} (22.8) & \texttt{Sports} (11.1) \\
C8 & 59,550 & 46.7 & \texttt{World} (45.9) & \texttt{Sci\_Tech} (27.4) & \texttt{Business} (16.0) & \texttt{Sports} (10.7) \\
C9 & 60,580 & 47.5 & \texttt{Sci\_Tech} (49.2) & \texttt{World} (22.7) & \texttt{Business} (16.8) & \texttt{Sports} (11.3) \\
C10 & 57,055 & 44.7 & \texttt{Business} (47.2) & \texttt{Sci\_Tech} (46.1) & \texttt{World} (5.9) & \texttt{Sports} (0.9) \\
C11 & 57,879 & 45.4 & \texttt{Sports} (33.1) & \texttt{Business} (28.1) & \texttt{World} (20.7) & \texttt{Sci\_Tech} (18.1) \\
C12 (G15) & 45,023 & 35.3 & \texttt{Sports} (36.6) & \texttt{Sci\_Tech} (28.2) & \texttt{World} (17.8) & \texttt{Business} (17.4) \\
C13 (M1) & 47,326 & 37.1 & \texttt{Sports} (34.2) & \texttt{Sci\_Tech} (28.1) & \texttt{World} (21.6) & \texttt{Business} (16.1) \\
\bottomrule
\end{tabular}}
\caption{Category composition under the positive-edge community
partition. C1--C11 require an active member bit in at least two backbones;
C12--C13 report activation of their single bit. The two selection rules
are distinct. Category percentages use the selected group as denominator;
coverage uses all 127,600 documents. Groups overlap.}
\label{tab:backbone-community-composition}
\end{table}

The selected groups retain several category combinations.
\texttt{Business} and \texttt{Sci\_Tech} jointly account for 93.3\% of C10 and are also the
two largest categories in C3 and C7. \texttt{Sports} and \texttt{World} account for 87.3\%
of C6, while C8 combines \texttt{World} and \texttt{Sci\_Tech}. C11 has elevated \texttt{Sports}
and \texttt{Business} shares. C1, C4 and C5 cover broad portions of the corpus
and have more mixed compositions. These associations locate document
sets for interpretation. Their permissive membership rule differs from
the conjunction of all community bits used in the earlier Qwen case study,
so neither their coverage nor their thematic specificity is directly
comparable. In particular, a category mixture does not imply that each
selected document combines those topics.

Increasing resolution alone does not separate the paired
categories. Across resolutions one to eight, the number of communities and
singletons increases from 13 to 24. The members of C6 first separate across
communities at resolution four and those of C10 at eight, but the resulting
document groups still mix their principal categories. Greater specificity
appears in within-community activation patterns
(Table~\ref{tab:backbone-internal-patterns}). For example, C10 contains
patterns concentrated in \texttt{Business} and others concentrated in \texttt{Sci\_Tech},
despite the broad \texttt{Business}--\texttt{Sci\_Tech} composition of its pooled document group.
These patterns specify both active and inactive coordinates within a community
across backbones; they are not complete profiles of any individual encoder.

The pooled community composition consequently conceals
information carried by differences among its member activations. A community
can identify related coordinates across encoders while their combinations
distinguish document subsets within that broader association.

The singleton nodes describe substantial document sets.
EmbeddingGemma bit~15 (G15) activates in 45,023 documents and MiniLM
bit~01 (M1) in 47,326. Both have elevated \texttt{Sports} and \texttt{Sci\_Tech} shares,
with documents from all four categories. Their isolation follows from
the matching constraint: each has four selected partners, all negatively
associated. Positive associations exist outside the one-to-one assignments,
including G15--H11 ($\phi=0.281$) and M1--N14 ($\phi=0.265$).
Their singleton status therefore concerns the selected-edge graph,
rather than the absence of related coordinates in other backbones.

A seeded random sample of 30 active documents per singleton
provides an initial reading of their content. G15 includes commentary
and evaluative prose across topics, such as a metaphorical account of
the Red Sox (document 22481), an informal discussion of computer fears
(40571), and commentary on pressure for Kofi Annan's resignation (109984).
Full-corpus document-frequency contrasts also show enrichment of words
such as ``maybe'', ``really'', and ``thing'' among its active documents.
M1 includes concrete reports of sporting events and scientific or technical
developments, including Roddick's Bangkok semifinal qualification (46434),
astronaut candidates experiencing zero gravity (69096), and a Samsung
phone incorporating a hard drive (20889). Its enriched terms include
``Roger Clemens'', ``space station'', and ``Ferrari''.
These exploratory examples suggest a possible stylistic contribution for
G15 and heterogeneous subject matter for M1; they do not establish a
single semantic attribute for either coordinate.

\section{Alternative encoders: protocols and supplementary comparisons}
\label{app:alternative-encoder-diagnostics}

\subsection{Training, binary readouts, and reconstruction}
\label{app:external-encoder-details}

This supplementary section expands the comparison in Section~\ref{sec:external-encoders}.

All methods use the same 127,600 AG News documents and cached,
unnormalized 1,024-dimensional Qwen3-Embedding-0.6B vectors. Encoder
fitting uses the full corpus without labels. The ISBC
encoder and its saved codes are reused from the full-corpus experiment;
the other two autoencoders are fitted on this same corpus.

The ISBC model has target activity six; its hard code thresholds
probabilities at 0.5.
HypotheSAEs has 16 units and top-$k$ activity six. Its continuous features
are nonnegative sparse activations; its binary features indicate their
positive support. Training uses 100 epochs, batch size 512, learning
rate $5\times10^{-4}$, and auxiliary-loss coefficient $1/32$, with the
multi-top-$k$ loss disabled. BinarySentEmb uses the released linear
autoencoder with input and output dimensions adapted to 1,024 and a
16-bit representation. Training uses 20 epochs, effective document batch
size 128, learning rate $10^{-5}$, weight decay $10^{-5}$, and
similarity-preservation coefficient 0.8, following the learning-rate
and coefficient settings in Section~4.2 of \citet{ShenEtAl2019}.
Its similarity objective receives seeded random AG News document
pairs.
No activity target is imposed on this method. These settings adapt
the published compressor to our corpus and representation target.
PCA and ITQ are also fitted without labels on the complete corpus.
PCA retains sixteen centered scores without whitening; ITQ rotates these
scores before thresholding and has no fixed activity target.
We use sixteen PCA components for ITQ so that both ITQ and ISBC produce
sixteen-bit codes. The ISBC target of six active coordinates controls
sparsity within that code; using six PCA components for ITQ would instead
produce a six-bit code. The comparison therefore matches code width,
while observed activity is reported separately.

For each top-6 adaptation, we activate the six coordinates with the
largest signed scores and set the others to zero: encoder logits for
BinarySentEmb and rotated PCA scores for ITQ. Exact ties are resolved by
coordinate index, with the smaller index selected first. This changes only
the binary readout; the fitted encoder or rotation remains fixed. The
ISBC code is unchanged and retains its learned, variable activity around
the target of six.

We first compare reconstruction on the common fitting corpus
(Table~\ref{tab:external-reconstruction}). Refitting an affine decoder to
each fixed hard code provides a common measure of recoverable embedding
variation, alongside reconstruction without decoder refitting. With
refitted affine decoders, native ITQ reaches $R^2=0.172$, compared with
0.161 for ISBC, with mean activity
7.98 and 5.94, respectively. Applying top-6 to the same ITQ rotation
reduces $R^2$ to 0.152. Thus, ISBC reconstructs more variation at similar
mean activity in this comparison, while unconstrained ITQ reconstructs
more with a higher average activity. ISBC uses a soft activity target;
the top-6 readout fixes the count for every document.

For ITQ, let $P\in\mathbb{R}^{16\times D}$ contain the fitted PCA
axes as rows, $\mu$ be the PCA mean, and $R\in\mathbb{R}^{16\times16}$
be the learned orthogonal rotation. For a column-vector code $B_i$,
we also evaluate the fixed inverse reconstruction
\[
\widehat x_i=\mu+P^\top R(2B_i-\mathbf{1}).
\]
This reverses the rotation and PCA projection using signed unit-valued
codes, with no fitted amplitude or intercept correction. The same map
is used for the top-6 code. Its $R^2$ is 0.034 for native ITQ and 0.030
for ITQ top-6, compared with 0.172 and 0.152 after affine refitting.
The inverse fixes amplitudes at $-1/+1$ in the unwhitened PCA score
units, whereas OLS estimates reconstruction coefficients from the corpus.
The refitted comparison therefore measures the variation recoverable
by an affine decoder from each hard code.

The native HypotheSAEs decoder reaches $R^2=0.216$ using continuous
activation magnitudes. With affine decoders refitted to the hard
codes, the ISBC representation reaches $R^2=0.161$, compared
with 0.141 for HypotheSAEs and zero for BinarySentEmb
(Table~\ref{tab:external-reconstruction}). The latter code permits
only a constant reconstruction. These results distinguish the
reconstruction fidelity attainable from continuous activations from that
attainable from their binary support.

For BinarySentEmb top-6, refitting an affine decoder gives
$R^2=0.046$, confirming that the adapted code carries some embedding
variation. Passing that code to the original frozen decoder instead
gives negative $R^2$: the decoder fitted to zero-threshold codes does
not reconstruct effectively from the altered readout.

Table~\ref{tab:external-code-usage} summarizes how the methods use
the binary code space. HypotheSAEs has fewer observed profiles than
the ISBC encoder but a larger effective profile count, reflecting
a more even distribution across its profiles. BinarySentEmb has eight permanently active and eight
permanently inactive bits, giving one observed profile.

For BinarySentEmb, top-6 selection produces 20 profiles, but their effective count is only
6.1 and the largest contains 30.92\% of documents. The eight coordinates
with negative logits remain inactive throughout the corpus; the six
active coordinates are selected among the other eight. Thus, fixed
activity removes the single-profile collapse while leaving a
concentrated representation.

Original ITQ produces more observed and effective profiles than ISBC,
whereas ITQ top-6 produces fewer. Fixing activity at six limits the code
space to $\binom{16}{6}=8{,}008$ possible profiles; a soft activity target
also allows other activity counts. Profile entropy therefore reflects
both the admissible code space and its empirical use. Conditional bit
entropy relative to the embedding clusters is reported separately in
Supplementary Section~\ref{app:itq-entropy}.

\subsection{Document-based matching across autoencoder readouts}
\label{app:external-profile-details}

We apply the document-based matching procedure in
Supplementary Section~\ref{app:backbone-profiles} to the four binary readouts of
three autoencoders, using each as reference and starting from up to its
ten most frequent profiles. Matching uses no category labels. Category
coherence uses the same Jensen--Shannon definition with pairwise
averaging over the four readouts.
Table~\ref{tab:external-profile-matching} reports initial coverage,
final intersections, and retained fractions. The original BinarySentEmb
code places no restriction on membership, whereas its top-6 variant
adds a third nonconstant partition.

\begin{table}[tbph]
  \centering
  \footnotesize
  \setlength{\tabcolsep}{4pt}
\begin{tabular}{lrrrrrr}
\toprule
Reference & Profiles & Coverage (\%) & Final documents & Retained (\%) & Category coherence & Chain null \\
\midrule
ISBC & 10 & 8.06 & 295 & 2.87 & 0.732 & 0.608 \\
HypotheSAEs & 10 & 2.49 & 301 & 9.49 & 0.758 & 0.603 \\
BinarySentEmb top-6 & 10 & 99.54 & 271 & 0.21 & 0.708 & 0.582 \\
BinarySentEmb & 1 & 100.00 & 45 & 0.04 & 0.685 & 0.650 \\
\bottomrule
\end{tabular}
\caption{Directional matching of complete hard profiles across the four
binary readouts of the three autoencoders. Coverage is the initial percentage of the corpus;
retained percentage divides final intersection counts by initial
reference-document counts. Category coherence compares the full category
distributions of the document groups defined by the selected profiles, weighted by reference-profile
frequency. The chain null averages 1,000 seeded document-alignment
permutations with profile selection repeated in every draw.}
\label{tab:external-profile-matching}
\end{table}

The ten most frequent BinarySentEmb top-6 profiles cover 99.54\% of the corpus,
compared with 8.06\% for ISBC and 2.49\% for HypotheSAEs. Matching
retains only 0.21\% of the top-6 reference documents in the final
intersections. The corresponding fractions from ISBC and HypotheSAEs
are 2.87\% and 9.49\%. These differences reflect the coarse top-6
partition and the different boundaries of the three nonconstant codes.
Category coherence exceeds the chain-null mean in each direction,
indicating shared category composition despite limited exact overlap.
The coherence calculation includes the original BinarySentEmb profile,
whose composition is the corpus-wide category distribution.

\subsection{Supplementary predictive evaluation of alternative encoders}
\label{app:external-prediction}

We evaluate category prediction after unsupervised fitting, using the
common stratified split of 102,080 training and 25,520 test documents.
Test documents participate in representation fitting, but their labels
are held out from classifier training. The complete raw embeddings and
sixteen continuous PCA scores provide predictive references. Continuous
compressed probes use ISBC log-odds, HypotheSAEs activation magnitudes,
and BinarySentEmb pre-threshold logits; both BinarySentEmb binary
readouts share the same continuous features.

All downstream classifiers, including the raw-embedding reference,
are refitted using common settings without additional feature scaling.
Logistic regression follows Supplementary Section~\ref{app:classification-protocol}.
Additional probes among the autoencoder readouts use class-balanced
elastic-net logistic regression (SAGA, $C=1$, L1 ratio 0.5,
tolerance $10^{-4}$, and at most 5,000 iterations) and a class-balanced
classification forest (200 trees, Gini splitting, square-root predictor
subsampling, unrestricted depth, and minimum leaf size one).
Both use seed 1234.
Balanced-accuracy intervals use 2,000 shared class-stratified bootstrap
resamples of test documents. They quantify test-sample uncertainty
conditional on the fitted encoders and classifiers.

ISBC has the highest observed binary BA in this comparison and
retains approximately 94\% of the raw-embedding BA. Original ITQ is close,
while imposing exactly six active coordinates on its fitted rotation
reduces BA from 84.82\% to 82.58\%. These estimates describe the evaluated
fits; the bootstrap intervals do not measure variation across encoder
initializations. Continuous PCA and HypotheSAEs exceed ISBC's continuous
features, but ISBC loses less predictive performance when reduced to
hard activations. Equal coordinate count provides a dimensional reference;
sixteen continuous values and sixteen bits have different storage capacities.

\begin{table}[tbph]
  \centering
  \small
  \setlength{\tabcolsep}{4pt}
  \begin{tabular}{llrr}
    \toprule
    Representation & BA [95\% CI] & Macro-F1 & Relative BA \\
    \midrule
    ISBC, binary & {\bf 85.62} [85.20, 86.06] & 85.57 & {\bf 93.97 }\\
    ITQ, top-6 & 82.58 [82.14, 83.03] & 82.53 & 90.64 \\
    ITQ, binary & 84.82 [84.36, 85.26] & 84.79 & 93.10 \\
    HypotheSAEs, support & 75.37 [74.89, 75.89] & 75.10 & 82.72 \\
    BinarySentEmb, top-6 & 50.40 [49.81, 50.94] & 49.17 & 55.32 \\
    BinarySentEmb, binary & 25.00 [25.00, 25.00] & 10.00 & 27.44 \\
    \midrule
    PCA, continuous & \textbf{88.24} [87.83, 88.62] & 88.22 & \textbf{96.85} \\
    ISBC, log-odds & 86.17 [85.76, 86.59] & 86.12 & 94.58 \\
    HypotheSAEs, continuous & 87.99 [87.60, 88.37] & 87.97 & 96.58 \\
    BinarySentEmb, logits & 80.58 [80.12, 81.06] & 80.46 & 88.44 \\
    \midrule
    Raw Qwen & 91.11 [90.76, 91.45] & 91.10 & 100.00 \\
    \bottomrule
  \end{tabular}
  \caption{Logistic-probe performance on the common 25,520-document test set.
  BA and macro-F1 are percentages. Relative BA is
$100\,\mathrm{BA}/\mathrm{BA}_{\mathrm{embeddings}}$,
  using raw Qwen embeddings as the reference.
  Continuous compressed representations have 16 coordinates; raw Qwen
  embeddings have 1,024. All probes are refitted for this comparison.
  The refitted raw-embedding probe achieves 91.11\% BA,
  compared with 91.09\% for the saved probe in
  Table~\ref{tab:qwen-predictive-performance}, on the same document split.
  Each table uses its own raw-embedding reference for relative BA.
  PCA is a continuous dimensional reference. ITQ has mean binary
  activity 7.98; the ISBC mean is 5.94.
  The top-6 variants change only the binary readout of the fitted encoders; both have exactly six active coordinates.}
  \label{tab:external-prediction}
\end{table}

BinarySentEmb's logits support label prediction, but zero-thresholding
maps all documents to one code. Selecting the six largest logits restores
some predictive utility, although its binary BA remains below both
HypotheSAEs and ISBC. The ordering ISBC, HypotheSAEs support,
BinarySentEmb top-6, and BinarySentEmb zero-threshold code is unchanged
with elastic-net logistic regression and random forests. These results
concern the present adaptation and training configuration.

Figure~\ref{fig:binarysentemb-top6-gap} in Supplementary Section~\ref{app:external-rank-distributions} describes the separation at the
selection boundary through the document-level gap
$\Delta_i=z_{i,(6)}-z_{i,(7)}$. Its median is 3.027 and its mean is
3.716 logit units; the observed range is approximately 0.00003--22.657.
No document has an exact tie at this boundary. The gap therefore
characterizes how clearly the sixth coordinate is selected within each
document. Supplementary Section~\ref{app:external-rank-distributions} shows the
distributions of the continuous values both by within-document rank
and by fixed coordinate for each encoder.

\subsection{Linear references at sixteen coordinates}
\label{app:linear-reference-protocol}

All references use the same 127,600 raw Qwen embeddings.
PCA is fitted to centered embeddings without labels, using randomized SVD
with sixteen components, 32 oversampling directions, seven power iterations,
and seed 1234 \citep{HalkoMartinssonTropp2011}. The continuous probe receives
the sixteen component scores without whitening or additional scaling.
The classifier uses the same stratified 102,080/25,520 document split and
class-balanced logistic regression as Supplementary Section~\ref{app:external-prediction}.
All added probes converged. The confidence intervals
use the same 2,000 class-stratified bootstrap draws as the supplementary
comparison.

PCA-sign thresholds component scores at zero.
ITQ alternates binary assignments and an orthogonal Procrustes rotation
for fifty iterations, initialized with seed 1251, before thresholding
the rotated scores \citep{GongLazebnik2011}.
The random-projection control uses sixteen Gaussian directions
orthonormalized by QR decomposition (seed 1234), applied to centered
embeddings and thresholded at zero. PCA top-6, ITQ top-6, and random top-6
activate the six largest signed scores for each document; ITQ top-6 uses
the same fitted rotation as the original ITQ readout.
These controls match the activity count exactly, whereas PCA-sign,
ITQ, and random sign codes activate approximately half their coordinates.
The original ITQ codes are reproduced exactly for all 127,600 documents
when recovering its rotated scores. The other saved binary reference
codes are reused, and all logistic probes are refitted.

\begin{table}[tbp]
\centering
\small
\setlength{\tabcolsep}{4pt}
{
\begin{tabular}{lrrrr}
\toprule
Representation & Mean activity & BA [95\% CI] & Macro-F1 & $R^2_{\mathrm{OLS}}$ \\
\midrule
ISBC, binary & 5.94 & 85.62 [85.20, 86.06] & 85.57 & 0.161 \\
ITQ, top-6 & 6.00 & 82.58 [82.14, 83.03] & 82.53 & 0.152 \\
ITQ, binary & 7.98 & 84.82 [84.36, 85.26] & 84.79 & 0.172 \\
PCA-sign & 7.98 & 82.14 [81.66, 82.59] & 81.99 & 0.160 \\
PCA top-6 & 6.00 & 79.39 [78.89, 79.86] & 79.22 & 0.144 \\
Random sign & 7.99 & 54.77 [54.15, 55.36] & 54.49 & 0.069 \\
Random top-6 & 6.00 & 53.39 [52.80, 54.00] & 53.14 & 0.062 \\
PCA, continuous & -- & 88.24 [87.83, 88.62] & 88.22 & 0.245 \\
\bottomrule
\end{tabular}}
\caption{Linear and binary references on Qwen AG News.
All representations have sixteen coordinates. BA and macro-F1 are
test-set percentages; mean activity applies to binary codes only.
Affine reconstruction is evaluated on the full unsupervised fitting
corpus. For continuous PCA, this is the cumulative explained variance.
Intervals condition on the fitted representation and classifier.}
\label{tab:linear-reference-controls}
\end{table}

\section{Continuous activation distributions of the alternative encoders}
\label{app:external-rank-distributions}

We examine continuous values in two complementary ways:
by their rank within each document, and by following each fixed coordinate
across documents. The first view describes the values surrounding a
rank-based selection boundary; the second describes where the values
of each individual bit concentrate.

\subsection{Within-document ranks}

For each document, we order the sixteen continuous values
from largest to smallest. Each curve describes one rank across all
127,600 documents; its coordinate identity can differ between documents.
For ISBC and BinarySentEmb, the upper panels show ranks one through
six separately, and the lower panels show ranks seven through sixteen.
For HypotheSAEs, a single panel shows ranks one through six; the
remaining ranks are identically zero and are omitted. Shading has opacity
0.20 to make overlapping curves visible. Paired panels share a horizontal scale. ISBC uses log-odds of the stored probabilities,
clipped to $[10^{-7},1-10^{-7}]$ in single precision; HypotheSAEs uses
nonnegative sparse activations, and BinarySentEmb uses encoder logits.

Gaussian density estimates use Scott bandwidths and
reflection at finite support boundaries. ISBC clipping endpoints are represented separately as probability masses;
HypotheSAEs densities describe the positive continuous activations.
Each continuous curve integrates to the remaining probability, so a
rank concentrated at an endpoint has little or no continuous density.
The rank distributions describe the values available before binary
readout: coordinate identities, rather than ranks alone, determine the
resulting binary profiles.

\begin{figure}[tbp]
  \centering
  \includegraphics[width=\textwidth]{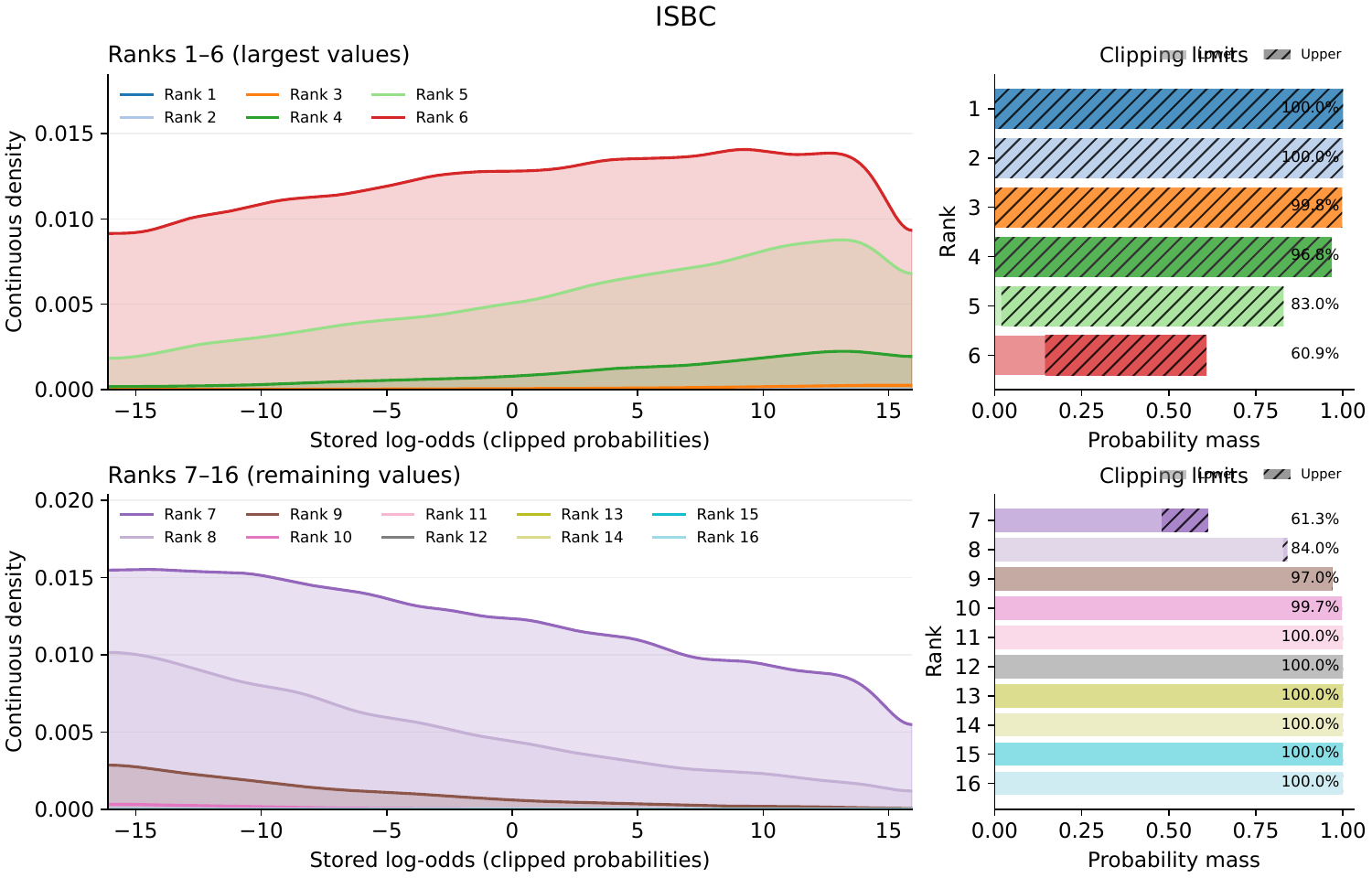}
  \caption{ISBC stored log-odds by within-document rank.
  Left: separate continuous densities for ranks 1--6 (top) and 7--16
  (bottom). Right: probability at the lower and upper clipping limits,
  with hatching indicating the upper limit; labels give their combined
  percentage. The endpoint masses reflect probabilities close to zero
  or one and the clipping used to calculate the stored log-odds.}
  \label{fig:external-rank-isbc}
\end{figure}

\begin{figure}[tbp]
  \centering
  \includegraphics[width=\textwidth]{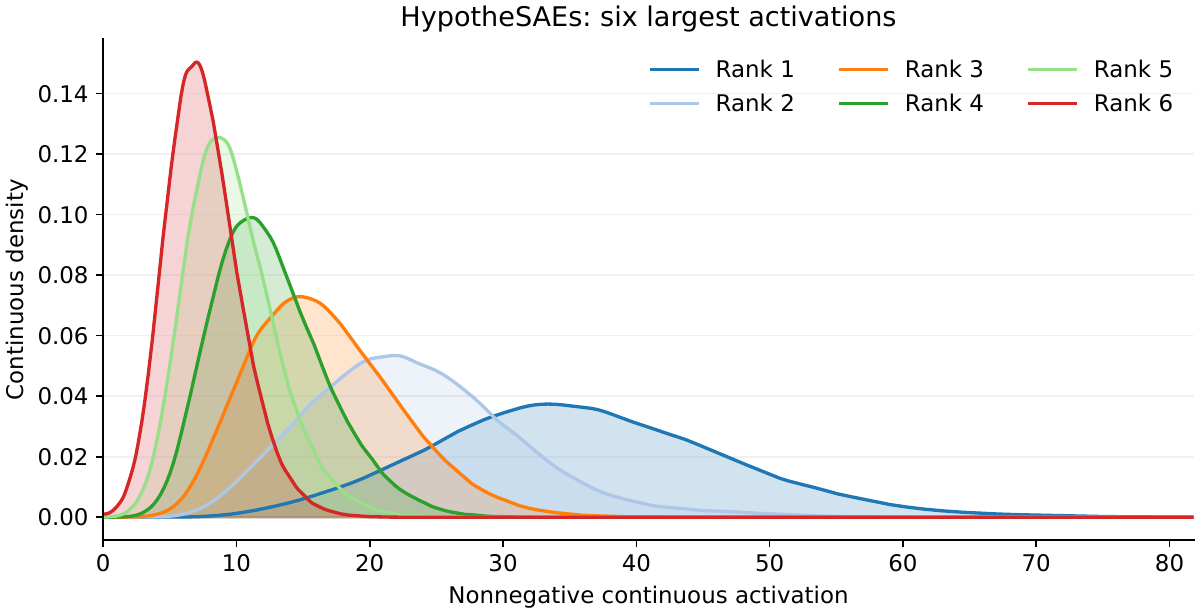}
  \caption{HypotheSAEs continuous activations by
  within-document rank. Each curve shows the continuous distribution of
  one of the six largest activations across all 127,600 documents.
  Ranks 7--16 are identically zero under the top-$k$ sparsity constraint
  and are omitted because their distributions are point masses at zero.}
  \label{fig:external-rank-hypothesaes}
\end{figure}

\begin{figure}[tbp]
  \centering
  \includegraphics[width=\textwidth]{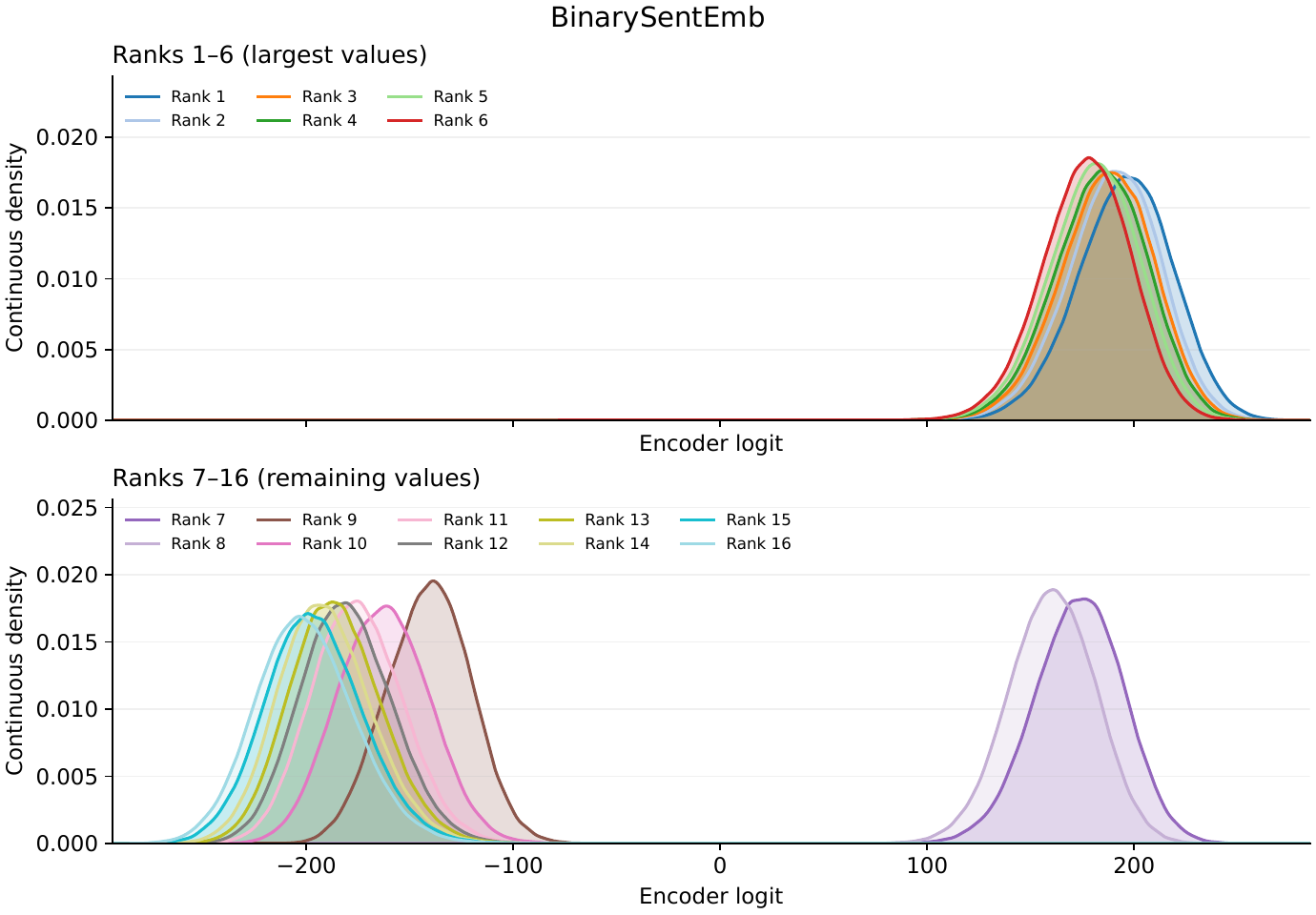}
  \caption{BinarySentEmb encoder logits by within-document
  rank, shared by the zero-threshold and top-6 readouts. Each rank has its
  own density. Ranks 7 and 8 remain positive although top-6 excludes them;
  ranks 9--16 are negative. The density overlap across documents is distinct
  from the within-document sixth--seventh gap in
  Figure~\ref{fig:binarysentemb-top6-gap}.}
  \label{fig:external-rank-binarysentemb}
\end{figure}

\begin{figure}[tbp]
  \centering
  \includegraphics[width=0.82\textwidth]{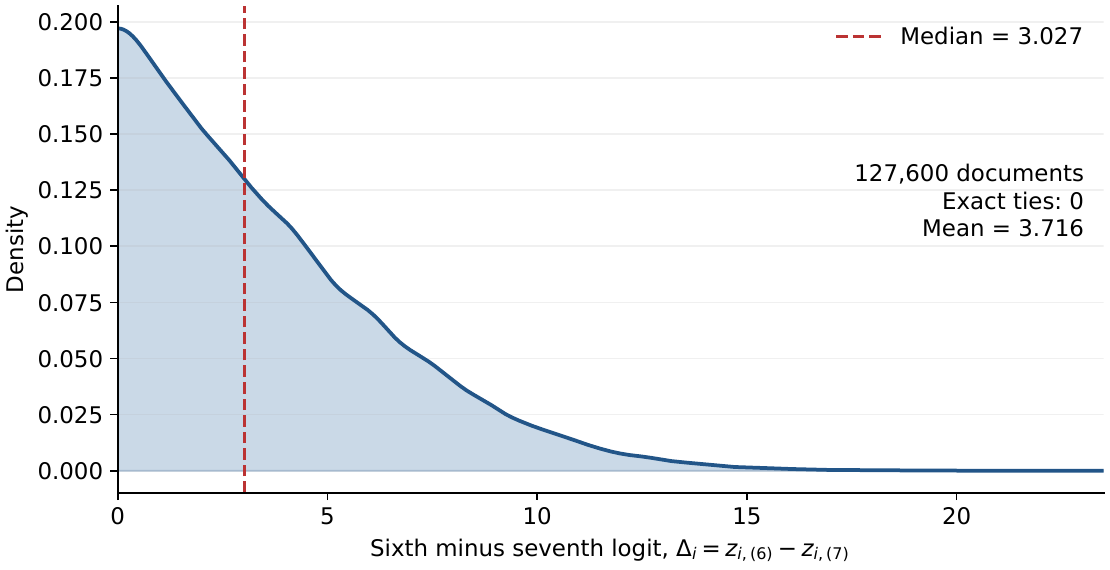}
  \caption{Document-level separation between the sixth and
  seventh largest BinarySentEmb logits across all 127,600 documents.
  The shaded curve is a Gaussian kernel density estimate reflected at
  zero; the dashed line marks the median. Small gaps indicate a narrow
  separation at the top-6 selection boundary. The horizontal scale is
  that of the fitted encoder's logits.}
  \label{fig:binarysentemb-top6-gap}
\end{figure}

\subsection{Fixed bit coordinates}
\label{app:external-bit-distributions}

\begin{figure}[tbp]
  \centering
  \includegraphics[width=\textwidth]{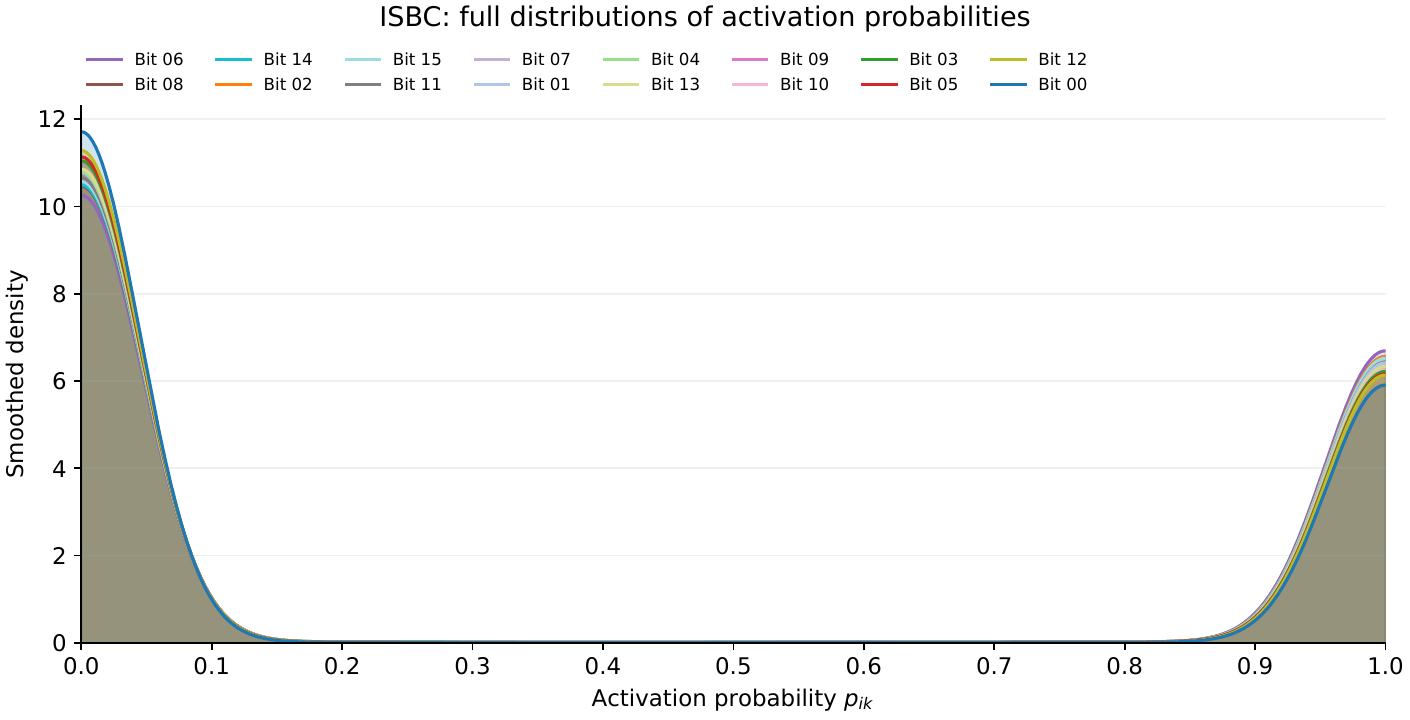}
  \caption{Full distributions of ISBC activation
  probabilities for all sixteen fixed bits across 127,600 documents.
  Each curve uses all original saved probabilities, including values
  at the endpoints, with Gaussian smoothing and reflection at zero and
  one. Each density integrates to one. The two peaks show concentration
  near inactive and active states; their widths and heights depend on
  smoothing, while their probability masses reflect the prevalence of
  these states. The legend retains the original bit indices,
  ordered by decreasing mean stored log-odds.}
  \label{fig:external-bit-isbc-probabilities}
\end{figure}

\begin{figure}[tbp]
  \centering
  \includegraphics[width=\textwidth]{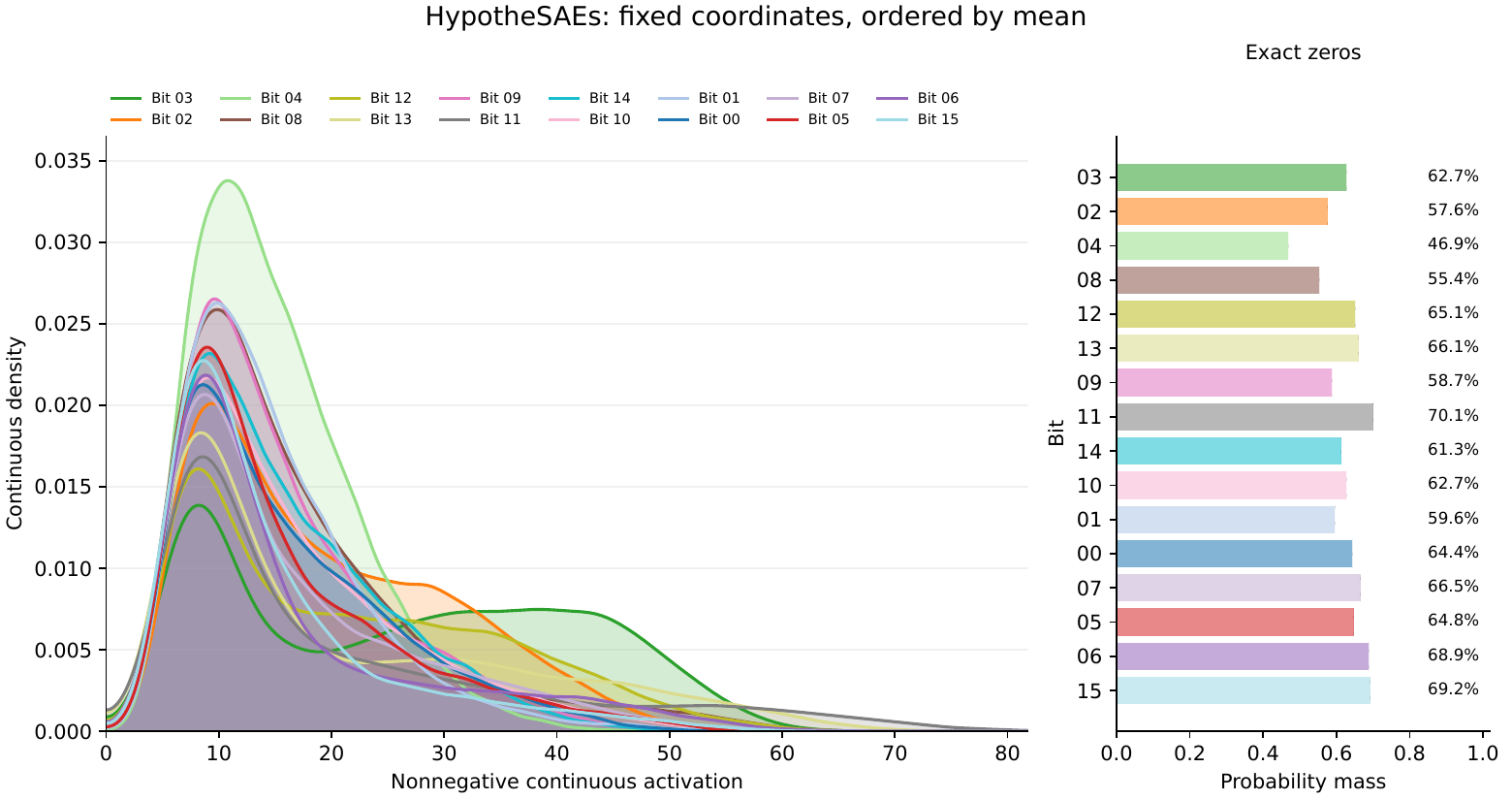}
  \caption{HypotheSAEs continuous values for each fixed
  coordinate, with all sixteen densities shown together and the legend
  ordered by mean activation. Left: distributions of positive magnitudes, weighted by each
  coordinate's probability of being positive. Right: probability of an
  exact zero. Unlike the within-document rank distributions, these curves
  track the same coordinate as it becomes active or inactive across
  documents.}
  \label{fig:external-bit-hypothesaes}
\end{figure}

\begin{figure}[tbp]
  \centering
  \includegraphics[width=\textwidth]{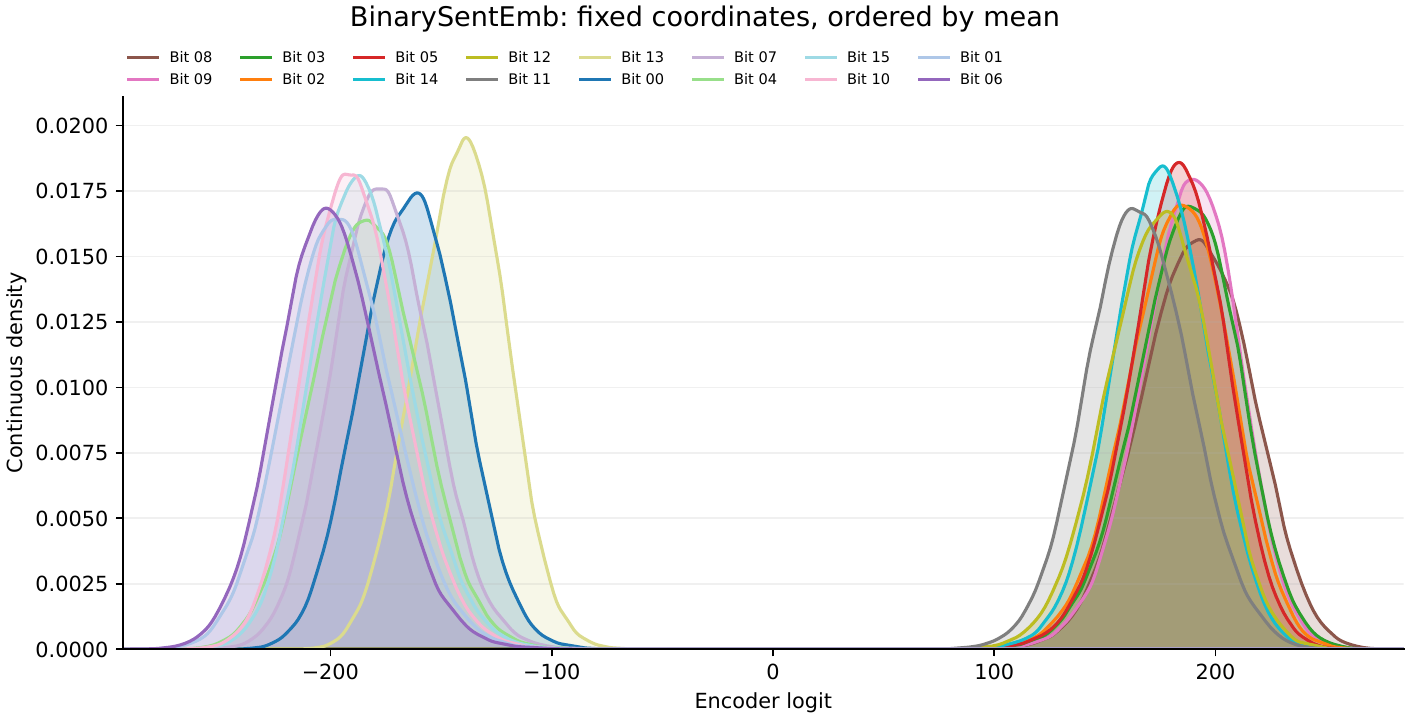}
  \caption{BinarySentEmb logits for each fixed coordinate,
  with all sixteen densities on common axes and the legend ordered by
  mean. The distributions separate into eight always-positive and eight
  always-negative coordinates. The original threshold assigns
  a fixed state to each, while the top-6 readout uses the relative
  ordering of positive coordinates within each document. The same saved
  logits support both readouts.}
  \label{fig:external-bit-binarysentemb}
\end{figure}

Figures~\ref{fig:external-bit-isbc-probabilities},
\ref{fig:external-bit-hypothesaes}, and~\ref{fig:external-bit-binarysentemb}
retain coordinate identity across documents. Each curve follows the
same bit throughout the corpus. All sixteen coordinate densities are
shown together on common axes for each encoder. Legends retain the
original bit indices, ordered by decreasing mean continuous value
(mean stored log-odds for ISBC). ISBC is shown on the original probability
scale with all values included. HypotheSAEs retains a separate display
of its zero masses, and BinarySentEmb is shown on the logit scale.

HypotheSAEs combines a mass at
zero with a distribution of positive magnitudes for each coordinate.
BinarySentEmb has eight coordinates that remain positive and eight that
remain negative; its top-6 readout selects among the positive coordinates
according to their document-specific ordering.

Figure~\ref{fig:external-bit-isbc-probabilities} shows the
complete distributions on the original probability scale, using all
saved activation probabilities, including values at or near zero and
one. Each bit has two concentrations, corresponding to predominantly
inactive and active states.

\section{Cross-corpus comparisons and the Mixed extension}
\label{app:corpora-mixed-diagnostics}

\subsection{Code use and conditional comparisons across corpora}
\label{app:cross-corpus-details}

These diagnostics expand Section~\ref{sec:cross-dataset}, using the separately
fitted Qwen encoders with $K=16$ and $A=6$.

Table~\ref{tab:cross-corpus-code-use} describes the complete binary
profiles. Mean activity ranges from 5.94 to 6.75 bits, while profile concentration
varies substantially. The document groups defined by the ten most frequent profiles contain
58.74\% of PHEME documents but only 4.14\% of GoEmotions comments. The effective profile count $2^{\widehat H}$ follows the definition in
Section~\ref{sec:qwen-compression}. Both observed and effective counts depend
on corpus size and composition; they summarize code use within each corpus.

Table~\ref{tab:cross-corpus-label-evidence} uses arithmetic-normalized
AMI \citep{VinhEppsBailey2010}, as defined in
Supplementary Section~\ref{app:backbone-profiles}, on all eligible documents for each
target. Class-balanced logistic probes use the saved 80/20 label splits
and class-stratified bootstrap intervals for BA. These intervals condition
on the fitted encoder and classifier. BA has random-prediction reference
$1/C$ when all $C$ classes occur in the test set; the encoder's access to
the full unlabelled corpus makes this evaluation transductive.

Agreement between partitions and prediction from individual bits
also describe different properties. For 20 Newsgroups, AMI is higher for the
20 original categories than for their five-group aggregation (0.357 versus
0.285), while BA is higher for the coarser target (80.70\% versus 49.08\%).
In FakeNewsNet, collection identity is predictable despite low full-partition
AMI (0.020), because prediction can combine many different profiles into the
same category. The remaining scores describe the specific annotation tasks
in each corpus; differences between tasks are not estimates of an intrinsic
capacity of binary coordinates to encode particular semantic attributes.

The profile diagrams in Supplementary Section~\ref{app:datasets} make these
differences inspectable. Each shows the ten most frequent complete profiles,
selected without labels, and reports their corpus coverage. In Airline, the
largest document group defined by a profile contains 1,315 tweets, of which 99.77\% concern Delta, but their
sentiment is distributed across negative (27.53\%), neutral (37.95\%), and
positive (34.52\%) labels (Figure~\ref{fig:corpus-profiles-airline}). This profile
groups tweets by carrier while spanning sentiment categories.

Climate Migration offers a complementary example. The same code
predicts migration relevance with 71.39\% BA and use of quantitative methods with
81.48\% BA; discipline and sentiment reach 61.60\% and 42.76\%, respectively.
Its most frequent profile is shared by 542 abstracts: 97.23\% are coded as using
quantitative methods, while 62.92\% concern migration. The second-most frequent profile
is shared by 387 abstracts, 98.97\% about migration but 94.57\% coded as not using
quantitative methods (Figure~\ref{fig:corpus-profiles-climmig}). These compositions
show that the profiles distinguish methodological as well as thematic
attributes. Driver and hazard combination tasks are reported separately in
Supplementary Section~\ref{app:data-climmig}, where their sparse categorical encoding is
explicit. The evidence across corpora supports evaluating a binary representation
against several available classifications: each reveals a different aspect of
the learned structure.

Single-bit profile comparisons show how these attributes vary within a fixed
configuration of the other coordinates. Applying the same minimum of 100
documents per profile as in the AG News comparison gives 23 eligible pairs in
Airline, 20 in 20 Newsgroups, three in Climate Migration, and seven in PHEME.
Table~\ref{tab:cross-corpus-single-bit} reports selected contrasts. The Airline
bit~01, Newsgroups, and PHEME examples maximize the smaller group size within
each corpus, without using labels. The additional Airline bit~14 and Climate
Migration examples were selected after examining their annotation contrasts.
All are exploratory descriptions of the fitted codes.

In Airline, the bit~01 pair remains almost exclusively about Southwest
(99.54\% and 99.87\%), while negative sentiment is more frequent when the bit
is active. The bit~14 pair is similarly concentrated on United (99.81\% and
100\%) but exhibits a larger sentiment contrast. Restricting each comparison
to its dominant carrier preserves the contrast. Moreover, negative sentiment
is higher on the active side in all six eligible bit~14 pairs, with differences
of 10.30--34.89 percentage points, and in all four eligible bit~01 pairs,
with differences of 17.59--22.75 points. Thus, the association recurs across
several configurations of the same fitted code.

In 20 Newsgroups, computing remains the dominant broad topic in both profiles
(98.41\% and 99.53\%), while the share of \texttt{comp.windows.x} increases
and that of \texttt{comp.graphics} decreases from 29.97\% to 11.32\%.
In Climate Migration, the higher migration-relevance share accompanies an
almost unchanged quantitative-methods share (58.26\% and 57.83\%).
The PHEME pair consists entirely of Sydney siege tweets, but differs in
rumour prevalence. These comparisons show that a coordinate can distinguish
content or annotations within a broader topic, entity, or event. Moderate
corpus-wide prediction of an attribute can therefore coexist with informative
contrasts in particular configurations. Profiles provide the conditioning
contexts for these comparisons; interpretation concerns the varying bit in
relation to the other coordinates and the documents they select.

\subsection{Mixed reference-corpus profiles and annotations}
\label{app:mixed-reference-details}

The following diagnostics supplement Section~\ref{sec:mixed}, using its
fixed Qwen representation and all 111,000 documents.

The resulting codes have 6.07 active bits on average and form
15,495 observed profiles, with empirical entropy 11.689 bits and an effective
profile count of 3,301.4. Table~\ref{tab:mixed-label-evidence} evaluates source
identity and document type using the hard codes. As in
Section~\ref{sec:cross-dataset}, AMI describes agreement between complete-profile
and label partitions, while a class-balanced logistic probe predicts labels
from individual bits. Each target uses a seeded 80\%/20\% document split;
representation learning has access to all documents without labels. Source and
type are related by construction: for example, the Climate Migration documents
are abstracts and the 20 Newsgroups documents are discussion posts. Their
association with the codes therefore describes overlapping aspects of the
mixture.

\begin{table}[tbph]
\centering
{\small
\setlength{\tabcolsep}{4pt}
\begin{tabular}{lrrlr}
\toprule
Target & Classes & AMI & BA [95\% CI] & Macro-F1 \\
\midrule
Source collection & 9 & 0.236 & 71.35 [70.79, 71.90] & 67.43 \\
Document type & 5 & 0.160 & 78.18 [77.62, 78.76] & 73.45 \\
\bottomrule
\end{tabular}}
\caption{Qwen hard-code evidence for the full Mixed corpus. BA and macro-F1 are percentages on held-out labels; AMI uses all 111,000 documents. BA intervals use 10,000 class-stratified bootstrap draws, conditional on the fitted probe and target-specific split.}
\label{tab:mixed-label-evidence}
\end{table}

For inspection, we retain the ten most frequent complete
profiles, each covering at least 0.45\% of the corpus. These profiles were
selected without labels. We use their previously frozen definitions and document
membership. Their document groups together contain 12,739 documents
(11.48\%). Table~\ref{tab:mixed-profiles} reports their source composition and
selected source-specific annotations. The source and document-type Sankey
diagrams appear in Figure~\ref{fig:mixed-profile-sources} in the supplementary dataset descriptions.
All percentages below use complete profile membership, rather than the
500-document inspection samples saved for each profile. These samples
are drawn without replacement with Python's \texttt{random.Random.sample}
from members in the saved corpus-row order. Each profile receives a seed
derived from the first 64 bits of a SHA-256 digest of the global seed 1234,
profile rank, and active-bit indices. The manifest records the exact seed,
input checksums, and sampled document IDs for each profile.

For nine of these ten frequent profiles, a single source accounts for
97.32--99.68\% of the corresponding document group. Several also distinguish
attributes within a source. The document group defined by P03 mainly contains computing discussions: 1,577 of its 1,921
20 Newsgroups posts belong to the five computing groups. The four document groups dominated by FakeNewsNet contain almost
exclusively GossipCop items, with
supplied real-label shares ranging from 54.13\% to 86.50\%. Thus, the frequent
profiles reflect both collection membership and variation within collections.

Two pairs illustrate the conditional interpretation of individual coordinates.
P01 and P07 differ only in bit~11, which is inactive in P01 and active in P07.
Among their Airline documents, negative sentiment accounts for 2,611 of 3,401
(76.77\%) and 257 of 556 (46.22\%), respectively. Carrier composition also
differs, so we examine the contrast separately within each airline
(Table~\ref{tab:mixed-within-airline}). The negative-sentiment share is lower
in P07 for all six carriers. This consistent direction shows that the aggregate
contrast is not explained solely by different carrier proportions, although
some carrier-specific groups are small. P02 and P09 differ only in bit~00,
active in P02: among their Climate Migration documents, migration relevance
accounts for 1,664 of 1,900 (87.58\%) and 304 of 512 (59.38\%). These comparisons
hold the other fifteen coordinates fixed and describe different attributes
within source-specific contexts. They use the previously selected frequent
profiles, with annotations examined subsequently.

\begin{table}[tbp]
\centering
\small
\setlength{\tabcolsep}{5pt}
\begin{tabular}{lrrrr}
\toprule
& \multicolumn{2}{c}{P01: bit 11 inactive} & \multicolumn{2}{c}{P07: bit 11 active} \\
\cmidrule(lr){2-3}\cmidrule(lr){4-5}
Carrier & Documents & Negative (\%) & Documents & Negative (\%) \\
\midrule
American & 929 & 76.32 & 117 & 54.70 \\
Delta & 353 & 63.17 & 32 & 28.12 \\
Southwest & 581 & 71.94 & 232 & 42.24 \\
US Airways & 1,018 & 81.63 & 87 & 60.92 \\
United & 456 & 84.87 & 26 & 73.08 \\
Virgin America & 64 & 67.19 & 62 & 22.58 \\
\bottomrule
\end{tabular}
\caption{Sentiment within airlines for two Mixed-corpus profiles differing
only in bit~11. Each percentage uses documents from the indicated carrier
and profile as its denominator. Counts include all Airline documents in
P01 and P07, not inspection samples. The bit belongs to the Mixed encoder
and is distinct from equally numbered bits in the standalone Airline encoder.}
\label{tab:mixed-within-airline}
\end{table}

P05 provides a case of thematic organization across sources.
Its 785 documents comprise 500 AG News items (63.69\%), 151 Newsgroups posts
(19.24\%), 77 TweetEval tweets (9.81\%), 48 IRA tweets (6.11\%), eight GoEmotions
comments (1.02\%), and one FakeNewsNet title (0.13\%). Among the AG News items,
494 of 500 are labelled \texttt{Sports}; all 151 Newsgroups posts belong to baseball
(89) or hockey (62). These independently supplied topic annotations identify
sports content within the same binary profile across news and discussion posts.
The document group defined by this profile also includes a TweetEval tweet about a cricketer missing a match
because of injury (Mixed document 37042) and a GoEmotions comment contrasting
playoff and regular-season performance (89685). These illustrative documents
extend the reading of the profile to sources whose labels concern affect or
evaluative language rather than sports.

\subsection{Additional documents encoded with the frozen Mixed model}
\label{app:mixed-new-details}

We apply the frozen Mixed encoder to 6,000 additional documents
from the same source collections. We sample 500 documents from each of the
eight non-TweetEval sources and 500 from each of the four TweetEval components,
without replacement, using global seed 20261030. Each component uses a
separate SHA-256-derived seed to permute its candidate rows; we retain the
first 500 eligible documents after the exclusions below. We exclude texts present in the
fitting corpus after normalized-text matching, duplicate texts across the new
sample, and identical model inputs after tokenization and truncation to
256 tokens. The current AG News source pool supplies the new news sample;
the original mixture retains its earlier extraction. The encoder, threshold,
and preprocessing are unchanged, and no classifier is fitted for this analysis.
Re-encoding 64 reference documents reproduces all their saved binary coordinates.
For exploratory reading, 20 of the 500 new documents per component were
sampled without replacement using separate component-specific seeds derived
from the same global seed. Their 240 IDs were saved before encoding the new
documents. The quoted texts illustrate selected content contrasts; they are
not used to estimate their prevalence. Sampling scripts, seed derivations,
input order, and selected IDs are retained with the analysis artifacts.

Previously observed complete profiles cover 5,675 of the 6,000
new documents (94.6\%). The new sample contains 2,915 distinct profiles,
including 309 absent from the fitting corpus. Mean activity is 6.16 bits.
Table~\ref{tab:mixed-new-coverage} reports coverage separately for each
sampling component. Across the 192 component--bit combinations, the mean
absolute difference in activation frequency between the reference and new
samples is 1.34 percentage points. This comparison conditions on the source
component, since the mixture proportions differ between samples. Profile
coverage describes access to reference documents with the same code;
interpretation additionally requires examining their content and annotations.

\begin{table}[tbp]
\centering\small
\begin{tabular}{lrrr}
\toprule
Sampling component & New documents & Mean active bits & Observed reference profile (\%) \\
\midrule
20 Newsgroups & 500 & 6.26 & 94.4 \\
AG News & 500 & 5.82 & 95.6 \\
Airline & 500 & 6.32 & 96.6 \\
Climate Migration & 500 & 6.81 & 99.0 \\
FakeNewsNet & 500 & 5.74 & 97.0 \\
GoEmotions & 500 & 5.78 & 94.4 \\
IRA & 500 & 6.04 & 90.8 \\
PHEME & 500 & 6.21 & 96.0 \\
TweetEval: emotion & 500 & 5.99 & 92.8 \\
TweetEval: hate & 500 & 6.40 & 95.0 \\
TweetEval: offensive & 500 & 6.59 & 92.8 \\
TweetEval: sentiment & 500 & 5.93 & 90.6 \\
\bottomrule
\end{tabular}
\caption{Frozen-encoder application to documents excluded from
ISBC fitting. The final column gives the percentage of new documents whose
complete binary profile occurs at least once in the original Mixed corpus.
The sampling allocation is fixed by source component, not by category.}
\label{tab:mixed-new-coverage}
\end{table}

The previously described conditional bit interpretations
retain their direction. Among new Airline documents, the negative-sentiment
share is lower in P07 than P01, which differ only in bit~11. For Climate
Migration, the migration-relevance share remains higher in P02 than P09,
which differ only in bit~00. To examine these associations beyond complete
profiles, we also construct three-bit contexts from the coordinates active
in both members of each pair. Using only reference-corpus annotations, we
choose the triplet with the highest source share among those covering at
least 25\% of that source. This post-hoc description uses annotations;
representation learning does not. It selects bits $\{0,10,13\}$ for Airline
and $\{1,12,15\}$ for Climate Migration.

The conditional contrasts persist when the remaining
coordinates are unrestricted (Table~\ref{tab:mixed-new-bit-contexts}).
Among all new documents activating the Airline context, 398 of 434 come
from Airline; the Climate Migration context contains 457 such abstracts
among 550 documents. For the new documents, the active-minus-inactive
difference is $-28.4$ percentage points for negative sentiment
(95\% interval $[-39.6,-16.1]$) and $18.4$ points for migration relevance
($[7.6,30.0]$). We combine the Wilson limits for each proportion using
Newcombe's score-based interval without continuity correction
\citep{Newcombe1998IndependentProportions}. These intervals use an
independent-document binomial approximation, conditional on the frozen
encoder, selected contexts, and observed group sizes, without a finite-population
correction. They describe sampling uncertainty within the specified groups.
These associations provide a reference for reading
a new text. For example, an Airline message praising a flight crew
(\texttt{new\_airline\_0477}) receives P07 and activates bit~11, whereas a
complaint about delays and poor communication
(\texttt{new\_airline\_0340}) receives P01 with that bit inactive.
P07 also contains a complaint about a credit
(\texttt{new\_airline\_0169}), illustrating the variability within the
same code. The interpretation concerns a conditional association with
sentiment rather than a deterministic sentiment label.

The sports reading of P05 also extends to new documents.
Its 38 new members comprise 28 AG News items labelled \texttt{Sports}, two
Newsgroups hockey posts, seven TweetEval tweets, and one PHEME tweet.
Inspection of the eight tweets identifies sports content, including a
TweetEval message about David Wright returning to the Mets lineup
(\texttt{new\_tweeteval\_sentiment\_0443}) and PHEME commentary on Kobe
Bryant's recent performance (\texttt{new\_pheme\_0161}). Both activate
bits $\{5,6,7,9,11,13\}$, with the remaining coordinates inactive,
linking them to the sports documents examined in the fitting corpus.

\section{PHEME: evaluation and interpretation diagnostics}
\label{app:pheme-tree-diagnostics}

\subsection{Predictive models and full evaluation}
\label{app:pheme-evaluation}

These results expand Section~\ref{sec:pheme-engagement}, using the same
frozen Qwen representation and engagement split. The ten specifications
per outcome use bits alone; bits with raw follower count and text length;
each of the three annotation sets with controls, with and without bits;
and all annotation sets jointly with controls, again with and without bits.
Every specification except bits only and the constant reference includes
the two controls. Categorical annotations enter as indicator variables,
and columns constant in the training set are excluded.

OLS and RF predict the standardized log outcome $z_i$ defined in
Section~\ref{sec:pheme-engagement}. Predictors retain their original scale.
The constant reference predicts the training mean of $z_i$, which is zero.
RF uses 1,000 trees, variance splitting, a minimum node size of ten, and
all available predictors considered at each split. We retain these settings
across specifications without tuning on test performance. Seeds are
$1234+100o+j$, where $o$ indexes retweets then favorites and $j$ indexes
bits with controls, bits only, veracity, veracity with bits, rumour status,
rumour status with bits, event, event with bits, all labels, and all labels
with bits, starting at one.

Both estimators are evaluated by test MSE. We use 2,000 shared
document-bootstrap resamples of the 1,161 test tweets for individual score
intervals and paired intervals for loss differences: each model versus
the constant reference, RF versus OLS, adding bits to each annotation set,
and adding controls to bits. These are pointwise 95\% intervals conditional
on the fitted models and split, without adjustment for multiple comparisons.

\begin{table}[tbph]
\centering\small
\setlength{\tabcolsep}{5pt}
\begin{tabular}{lrrrr}
\toprule
& \multicolumn{2}{c}{Retweets} & \multicolumn{2}{c}{Favorites} \\
Predictor specification & OLS & RF & OLS & RF \\
\midrule
Constant reference & 1.078 & 1.078 & 1.059 & 1.059 \\
Bits only & 1.020 & 1.022 & 0.963 & 0.962 \\
Bits + controls & 0.978 & 0.949 & 0.927 & 0.805 \\
Veracity & 1.033 & 1.090 & 0.960 & 0.939 \\
Veracity + bits & 0.978 & 0.936 & 0.896 & 0.792 \\
Rumour status & 1.038 & 1.099 & 0.966 & 0.949 \\
Rumour status + bits & 0.977 & 0.941 & 0.897 & 0.792 \\
Event & 0.968 & 0.953 & 0.966 & 0.869 \\
Event + bits & 0.965 & 0.946 & 0.908 & 0.804 \\
All labels & 0.967 & 0.959 & 0.906 & 0.840 \\
All labels + bits & 0.966 & 0.927 & 0.880 & 0.791 \\
\bottomrule
\end{tabular}
\caption{PHEME engagement prediction using Qwen hard bits. Entries are
MSE on the training-standardized $\log(1+y)$ outcome; lower is better.
Bits only uses the sixteen binary indicators without controls. The constant
reference uses no predictors and predicts zero. Every other row includes raw
follower count and text length. OLS and variance-split RF use the same training
and test tweets. Figure~\ref{fig:pheme-labelwise-ols} shows individual score
intervals.}
\label{tab:pheme-labelwise-scores}
\end{table}

Bits-only RF improves over the constant reference for both outcomes:
the paired MSE differences are $-0.056$ with interval $[-0.105,-0.005]$
for retweets and $-0.096$ with interval $[-0.149,-0.044]$ for favorites.
Its differences from bits-only OLS are small, with intervals
$[-0.017,0.021]$ and $[-0.027,0.025]$, respectively.
Adding controls reduces error for both estimators and outcomes, with
all four paired intervals below zero.

For retweets, adding bits to veracity or rumour status improves both
estimators, whereas adding them to event identity gives small differences
whose intervals include zero. With all annotations jointly, adding bits
improves RF, while the OLS difference remains small. For favorites,
adding bits improves every annotation-and-control specification for both
estimators, with all eight paired intervals below zero. RF also improves
on OLS in every favorite specification containing bits and controls;
for the corresponding retweet comparisons, all paired intervals include zero.
These results locate the clearer nonlinear predictive gain in models with
additional covariates, rather than in the bits-only specification.

\begin{figure}[tbph]
\centering
\includegraphics[width=\textwidth]{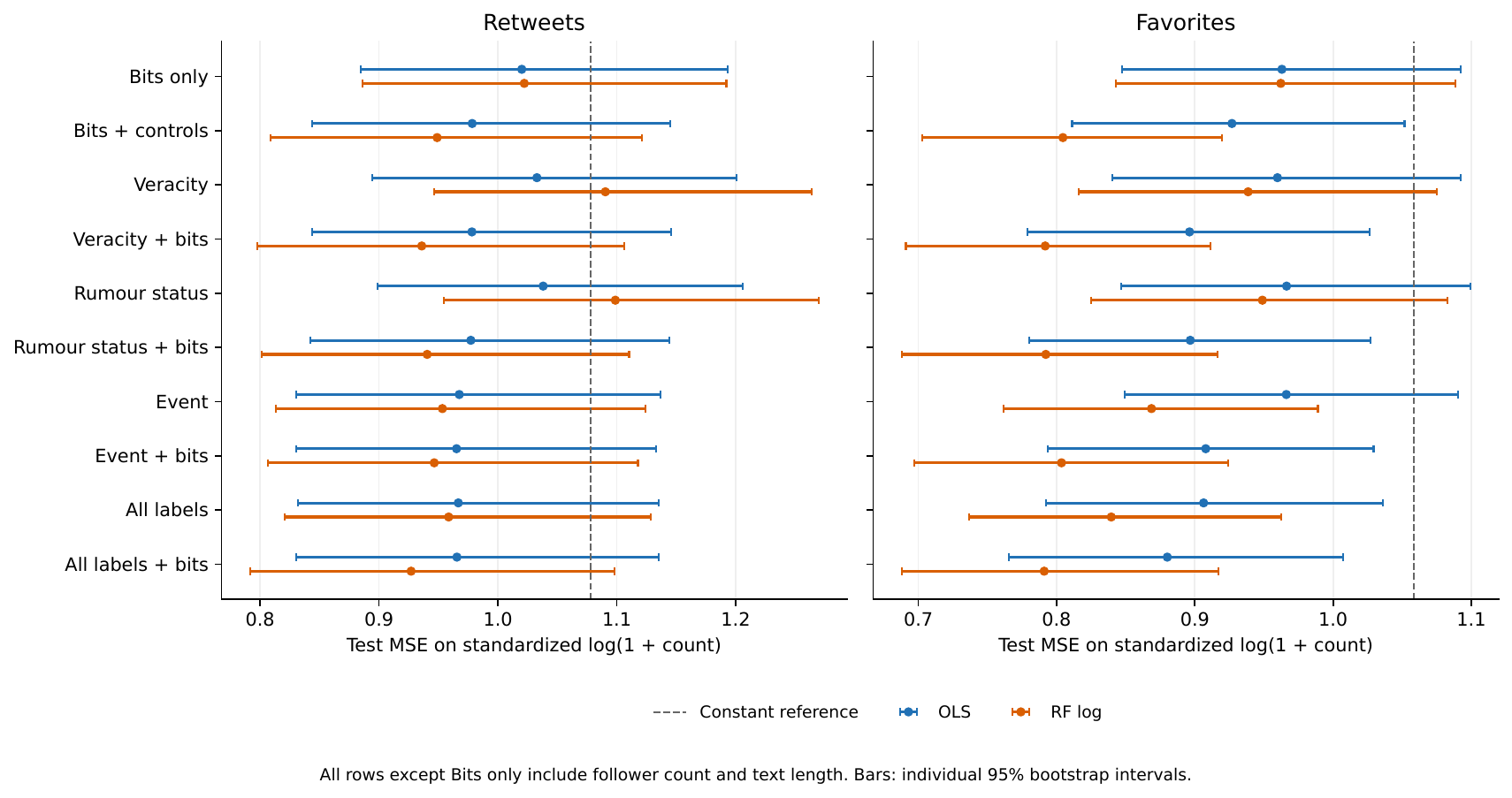}
\caption{OLS and RF prediction error on the training-standardized
$\log(1+y)$ outcome for retweets (left) and favorites (right). All
specifications except Bits only include follower count and text length.
Points are test MSE and bars are individual 95\% document-bootstrap
intervals conditional on the fitted models. Dashed lines mark the constant
reference. These bars are not intervals for differences between models;
the text reports paired comparisons.}
\label{fig:pheme-labelwise-ols}
\end{figure}

\subsection{Content inspection and category compositions}
\label{app:pheme-content}

We examine the content associated with bit~07 within the
Sydney event, where both activation states are sufficiently represented.
Within each state, training tweet IDs are sorted in ascending order and
12 tweets are sampled without replacement using pandas
\texttt{sample(n=12, random\_state=20261030)}. The seed is reset for each
state, and selection does not use engagement. Active-bit texts repeatedly report the
hostage situation, police activity, and developments at the cafe. Inactive-bit
texts include prayers, expressions of solidarity, and debate about religion
and community relations, including hostile reactions. For example, an
active-bit tweet reports police surrounding the cafe
(ID~\texttt{544274879055675392}); inactive-bit examples describe an interfaith
show of solidarity (ID~\texttt{544476492281827328}) and call for prayer and
peace (ID~\texttt{544331094523662337}). These readings identify a conditional
content distinction within a single event.

To describe this contrast across the selected document sets,
we count two groups of expressions identified from the training-text reading:
incident terms (\texttt{hostage}, \texttt{hostages}, \texttt{gunman},
\texttt{police}, \texttt{cafe}, and \texttt{caf\'e}) and
religion or public-response terms (\texttt{Muslim}, \texttt{Muslims},
\texttt{Islam}, \texttt{Islamophobia}, \texttt{solidarity},
\texttt{illridewithyou}, and words beginning with
\texttt{racis} or \texttt{pray}). Matching is case-insensitive; a leading hashtag symbol does not
prevent a match. These are descriptive lexical indicators.
The contrast persists in the test documents
(Table~\ref{tab:pheme-sydney-content}), with the other fifteen coordinates
unrestricted.

Within Sydney, active bit~07 is associated with lower
training mean counts. The lower median favorite count persists in the test subset, whereas
the mean reverses direction. One active-bit test tweet expressing concern
for the hostages has 69,825 favorites, illustrating both overlap between
the content groups and the influence of the upper tail. The content
contrast therefore provides an empirical interpretation of a predictive
coordinate, while its engagement association depends on the outcome
summary and sample.

The single-bit comparison in Table~\ref{tab:cross-corpus-single-bit}
provides a conditional view of bit~07. Holding the other fifteen coordinates fixed selects two
groups entirely associated with the Sydney siege; rumour prevalence is higher
in the group with bit~07 active. This connects the coordinate to a distinction
within event identity. It does not establish that rumour status explains the
engagement association, since the profile contrast and the tree split select
different document groups.

\begin{figure}[tbph]
\centering
\includegraphics[width=\textwidth]{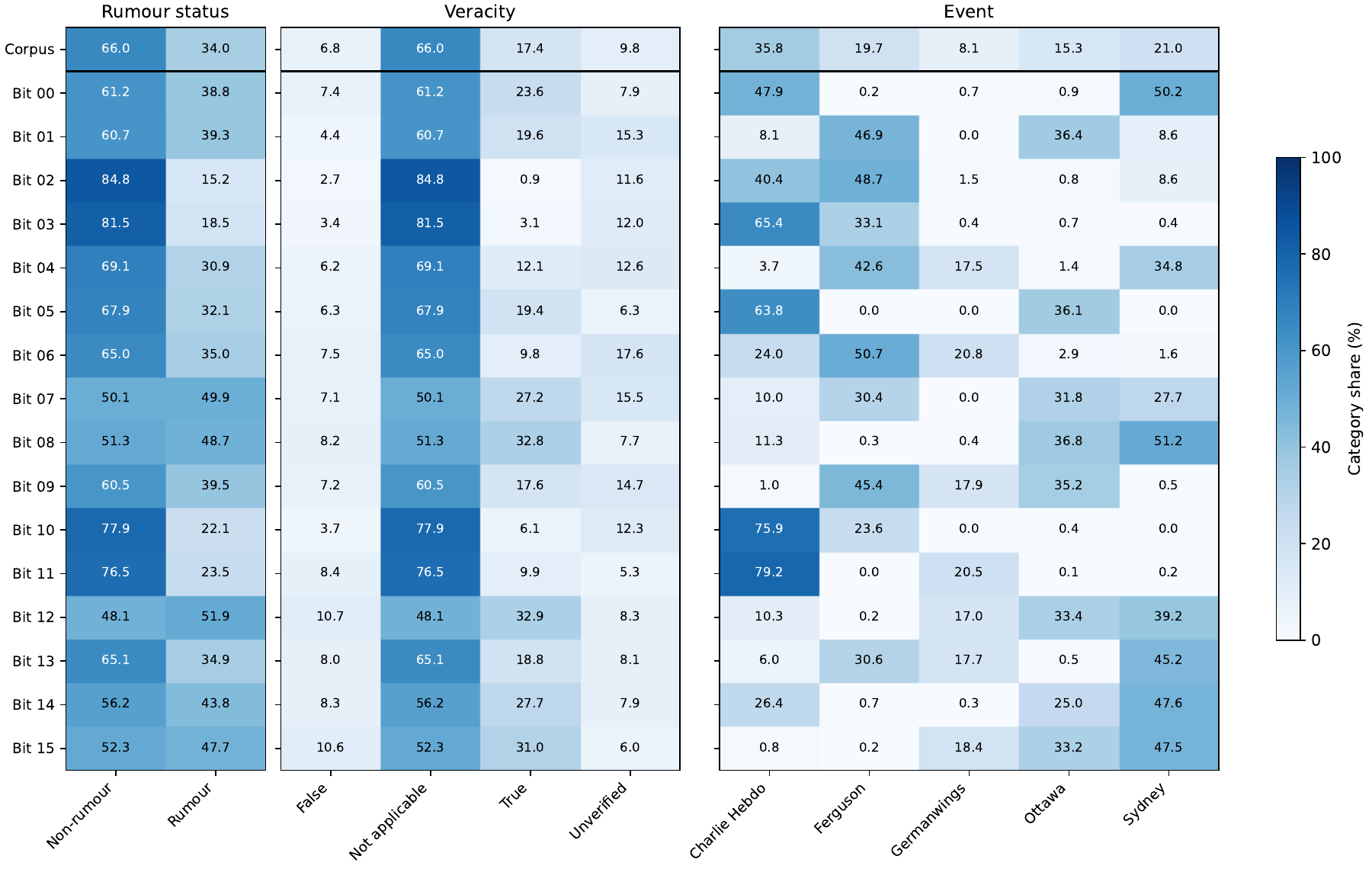}
\caption{PHEME category distributions conditional on each Qwen bit being active. Entries are percentages and sum to 100 within each taxonomy and row, up to rounding. The first row shows unconditional corpus frequencies. Bits overlap and select between 2,254 and 2,701 tweets each. These descriptive compositions use the full corpus and are separate from the held-out engagement evaluation.}
\label{fig:pheme-labelwise-composition}
\end{figure}

For each bit $k$, Figure~\ref{fig:pheme-labelwise-composition} reports the complete category distribution for each of the three taxonomies, using all 5,802 documents. Specifically, each cell estimates
$P(L^{(t)}=c\mid B_k=1)$,
where $L^{(t)}$ is event identity, rumour status, or veracity, and $c$ is one of its categories. The denominator is the number of documents activating bit $k$; probabilities sum to one separately within each taxonomy. Comparison with the unconditional corpus frequencies indicates which categories are more or less prevalent among the selected documents. Together, the three distributions describe each bit's relationship to the existing annotations without assigning it a single category.

For example, bit~06, which contributes the largest split-improvement share in the bits-only log-favorite tree, selects documents about Ferguson (50.71\%), Charlie Hebdo (23.96\%), Germanwings (20.81\%), Ottawa (2.93\%), and Sydney (1.60\%). Of these documents, 65.04\% are non-rumours, 9.80\% true rumours, 7.54\% false rumours, and 17.61\% unverified rumours. Bit~08 spans Sydney (51.25\%), Ottawa (36.78\%), and Charlie Hebdo (11.30\%), with smaller shares from the other events. Its rumour share is 48.71\%, compared with 33.99\% in the corpus; true rumours account for 32.80\% of all documents activating this bit. These distributions connect the predictive coordinates to several dimensions of the annotations simultaneously.

The full set of bits also reveals different combinations of the recorded categories. Active bit~05 selects Charlie Hebdo (63.83\%) and Ottawa (36.13\%), whereas active bit~11 selects Charlie Hebdo (79.19\%) and Germanwings (20.46\%). For bit~04, the largest event shares are Ferguson (42.59\%), Sydney (34.79\%), and Germanwings (17.51\%). The figure retains every category, including smaller shares, so that this interpretation reflects the composition of each selected document set. Predictive use and category composition together guide which documents to examine and which content differences to investigate.

\subsection{Engagement tails and conditional content}
\label{app:pheme-tails}

\subsubsection{Selection and reading}

This exploratory analysis complements Section~\ref{sec:pheme-tails} using
all 5,802 documents and the saved Qwen codes, without refitting the encoder
or a predictive model. For each outcome $Y$, event $e$, and
$q\in\{0.10,0.25\}$, the lower group contains documents with
$Y_i\leq Q_e(q)$ and the upper group those with $Y_i\geq Q_e(1-q)$,
where $Q_e$ is the empirical quantile with linear interpolation.
All boundary ties are retained, so group sizes need not be equal.
The thresholds are computed within the entire event, before selecting a bit
condition. Corpus-wide thresholds are used separately for the cross-outcome
overlap calculation below. No documents are sampled to balance group sizes.

The descriptive search covers all sixteen bits, their 120 pairs and 560
triples, and observed complete profiles. Conditional comparisons hold zero,
one, or two specified bits active while varying another bit, leaving the
remaining coordinates unrestricted. The examples were selected after examining
these contrasts and their content; the reported differences are not tests of
prespecified hypotheses or additional held-out predictive results.
The 10\% and 25\% analyses describe different degrees of outcome selectivity;
consistency across them is not an inclusion criterion.

Keyword counts use all documents in each selected group. Text is lowercased
and HTML-decoded; URLs, user mentions, English stopwords, and platform tokens
are removed. Hashtag words are retained. We count document presence of
unigrams and bigrams formed after stopword removal, using alphabetic tokens
of at least three characters, with internal hyphens and underscores allowed.
Terms are compared by differences in document prevalence. The displayed
contrasts select interpretable examples with at least five occurrences on
the enriched side; all observed terms remain in the analysis outputs.

\begin{figure}[tbp]
\centering
\includegraphics[width=\textwidth]{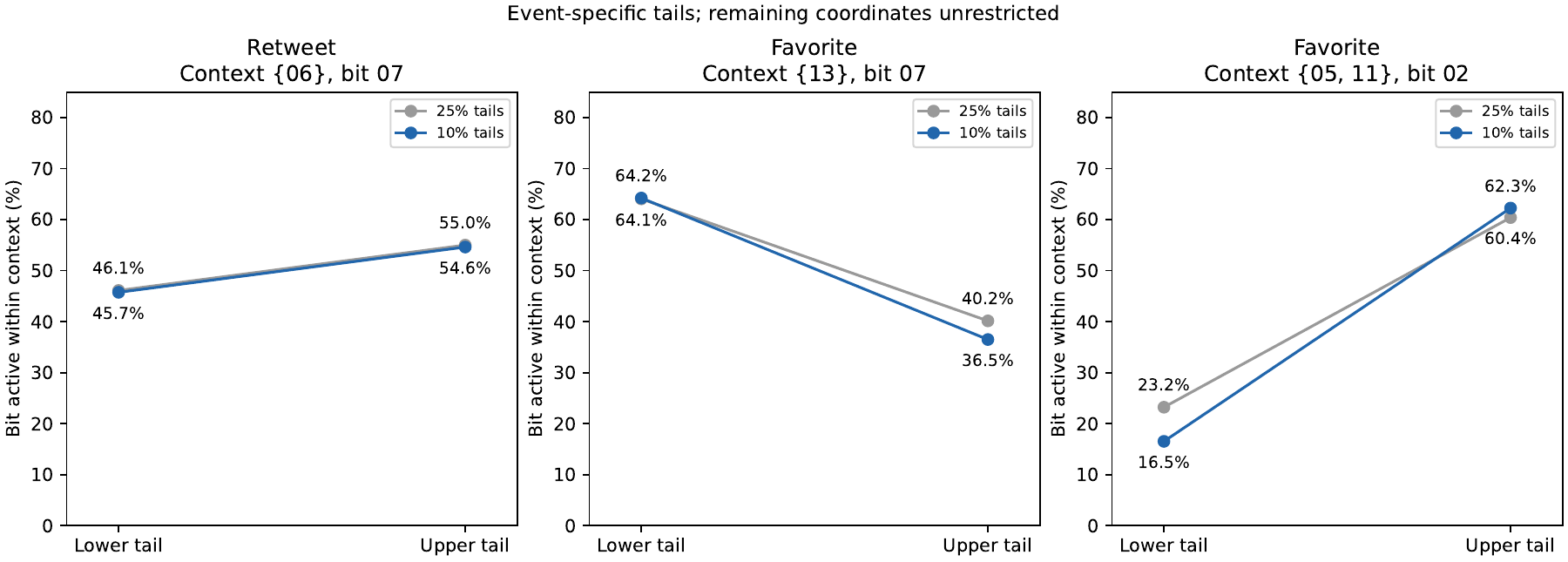}
\caption{Conditional bit prevalence in event-specific engagement tails.
Each context fixes the listed bits to one; all other coordinates except the
compared bit remain unrestricted. Points are the percentage with the compared
bit active among all documents satisfying that context in the indicated tail.
The first two panels pool documents after computing thresholds separately
within each event; the third context selects only Charlie Hebdo. Lines connect
descriptive group proportions, not repeated measurements or causal effects.}
\label{fig:pheme-tail-contexts}
\end{figure}

For the three contexts in Figure~\ref{fig:pheme-tail-contexts}, the initial
reading dossier samples up to twenty documents per decile-tail and bit-state
cell. IDs are sorted before sampling without replacement; the seed is the
integer encoded by the first four bytes of the SHA-256 digest of the saved
group name, in big-endian order. All 286 Charlie Hebdo documents in the two
decile tails of the $\{05,11\}$ context were subsequently read, including
both states of bit~02. Together with the 160 sampled reading rows for the
other two contexts, this gives 446 rows and 437 distinct documents.
The full-corpus Charlie Hebdo complement analysis separately samples up to
twelve documents from each of the eight $(B_5,B_{11},B_2)$ cells, again after
sorting IDs, with seed $20261100+2(2B_5+B_{11})+B_2$. Its small cells are
read in full, giving 83 reading rows. Additional follow-up reading covered
all fifty Ottawa favorite-decile documents with bit~14 inactive and the
seven Germanwings lower-retweet-decile documents with bit~02 active.
These reading sets can overlap. Quartile keyword statistics use every
selected document; the quartile groups were not read exhaustively.
Supplementary Section~\ref{app:reading-prompts} records the request types
and the retained dossier instruction. Qualitative descriptions are not
estimates of theme prevalence from these reading samples.

\subsubsection{The conjunction and its complements in the full corpus}

Table~\ref{tab:pheme-charlie-complements} exhaustively partitions the corpus
by bits~05 and~11. Their conjunction selects only Charlie Hebdo, but 603
additional Charlie Hebdo documents have other states of this pair. The pair
is therefore a selective event-associated condition, rather than an exact
recoding of the event annotation. To interpret the complements independently
of their differing event composition, we recompute keyword frequencies and
read documents within Charlie Hebdo alone.

\begin{table}[tbp]
\centering\small
\setlength{\tabcolsep}{4pt}
\begin{tabular}{lrrrrrr}
\toprule
$(B_5,B_{11})$ & All & Charlie Hebdo & Ferguson & Germanwings & Ottawa & Sydney \\
\midrule
$(0,0)$ & 2,673 & 269 & 1,142 & 7 & 40 & 1,215 \\
$(0,1)$ & 782 & 312 & 1 & 462 & 2 & 5 \\
$(1,0)$ & 871 & 22 & 0 & 0 & 848 & 1 \\
$(1,1)$ & 1,476 & 1,476 & 0 & 0 & 0 & 0 \\
\bottomrule
\end{tabular}
\caption{All PHEME documents partitioned by the states of bits~05 and~11.
Entries are document counts; the remaining fourteen bits are unrestricted.
Event labels are used after fitting to describe these disjoint groups.}
\label{tab:pheme-charlie-complements}
\end{table}

Within the $(1,1)$ group, Table~\ref{tab:pheme-charlie-bit-keywords}
shows a contrast between expressive content and incident reporting across
the full sample, not only its engagement tails. Documents with bit~02 active
include defenses of expression through words rather than bullets
(ID~553118384540643328) and appeals to stand together for freedom
(ID~552832738542514176). Inactive-bit examples report the death toll
(ID~552802128688078848) and the pursuit of suspects
(ID~553184390260338688). These themes overlap: an upper-tail document with
bit~02 inactive describes a tribute with pencils and silence
(ID~553205372719472640). The coordinate does not provide a categorical
annotation of emotion or solidarity.

\begin{table}[tbp]
\centering\small
\begin{tabular}{lrr}
\toprule
Term & $B_2=0$ ($n=737$) & $B_2=1$ ($n=739$) \\
\midrule
\texttt{jesuischarlie} & 79 (10.7\%) & 218 (29.5\%) \\
\texttt{cartoons} & 13 (1.8\%) & 68 (9.2\%) \\
\texttt{freedom} & 25 (3.4\%) & 56 (7.6\%) \\
\texttt{solidarity} & 15 (2.0\%) & 38 (5.1\%) \\
\texttt{attack} & 259 (35.1\%) & 84 (11.4\%) \\
\texttt{police} & 119 (16.1\%) & 3 (0.4\%) \\
\texttt{suspects} & 106 (14.4\%) & 2 (0.3\%) \\
\bottomrule
\end{tabular}
\caption{Selected term-presence counts and percentages within all 1,476
documents with $B_5=B_{11}=1$, separated by bit~02. All belong to Charlie
Hebdo. Percentages use the column denominators and are measured from the
complete groups; terms can co-occur.}
\label{tab:pheme-charlie-bit-keywords}
\end{table}

Within Charlie Hebdo, the $(0,1)$ group ($n=312$) includes discussion of
religion, condemnation, and collective blame. For example,
ID~552814410763603968 condemns the attack and expresses concern for its
victims, while ID~553136990787342336 questions unequal attribution of
collective responsibility. In its bit~02-active subset, \texttt{Islam} occurs
in 42/197 documents and \texttt{Muslims} in 45/197. The $(0,0)$ group
($n=269$) instead prominently describes the later hostage incidents:
\texttt{hostage} occurs in 94 documents, \texttt{kosher} in 70, and
\texttt{supermarket} in 45. Examples include reports
about the kosher supermarket (ID~553536887349325825) and a second
hostage situation in eastern Paris (ID~553540719714271233).
The $(1,0)$ group is small ($n=22$) and includes reports concerning police
officers and suspects, such as ID~553129559173115904; \texttt{police} occurs
in 15 of these 22 documents. Its only document with
bit~02 active asks readers not to circulate an execution video out of respect
for the victim (ID~552805956619628544). These examples characterize observed
subsets, not mutually exclusive semantic classes.

\subsubsection{Tail contrasts, other events, and outcome overlap}

\begin{figure}[tbp]
\centering
\includegraphics[width=\textwidth]{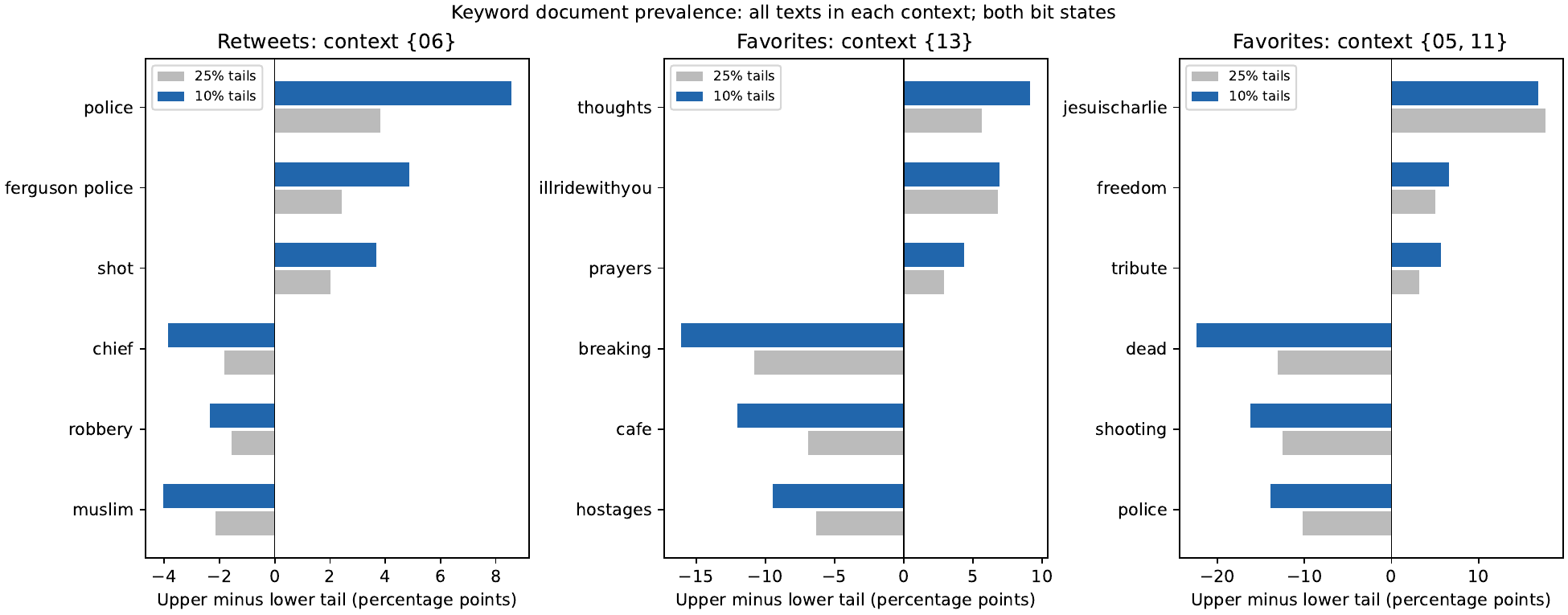}
\caption{Selected keyword contrasts in the same contexts as
Figure~\ref{fig:pheme-tail-contexts}, pooling both states of the compared bit.
Positive values indicate greater document prevalence in the upper tail;
negative values indicate greater prevalence in the lower tail. Counts use
all selected documents. The first two panels can reflect both within-event
content and changing event composition; the Charlie Hebdo context contains
only one event. Event-specific comparisons are reported separately in
Table~\ref{tab:pheme-event-tails} and the accompanying text.}
\label{fig:pheme-tail-keywords}
\end{figure}

Within the $\{05,11\}$ context, the lower and upper favorite deciles contain
127 and 159 documents, of which 21 and 99 activate bit~02. The corresponding
quartile counts are 72/310 and 241/399. The more expressive upper-tail reading
includes grief, identification with the victims, freedom of expression,
and anger or criticism, rather than a uniformly positive tone.
Figure~\ref{fig:pheme-tail-keywords} shows selected lexical contrasts.
For example, \texttt{jesuischarlie} occurs in 9/127 lower-decile documents and
38/159 upper-decile documents, whereas \texttt{dead} occurs in 30/127 and
2/159. Bit~02 also distinguishes favorite tails in the Charlie Hebdo
$(0,1)$ complement: it is active in 6/17 lower-decile and 23/36 upper-decile
documents. The small bit~02-active cells in the other complements do not
support an equally detailed comparison.

\begin{table}[tbp]
\centering\small
\setlength{\tabcolsep}{4pt}
\begin{tabular}{lrrrr}
\toprule
$(B_5,B_{11})$ & $B_2=1$, all & Median favorites & Lower 10\% & Upper 10\% \\
\midrule
$(0,0)$ & 10/269 & 42 & 3/63 (4.8\%) & 1/10 (10.0\%) \\
$(0,1)$ & 197/312 & 85 & 6/17 (35.3\%) & 23/36 (63.9\%) \\
$(1,0)$ & 1/22 & 43.5 & 0/4 (0.0\%) & 0/3 (0.0\%) \\
$(1,1)$ & 739/1476 & 88 & 21/127 (16.5\%) & 99/159 (62.3\%) \\
\bottomrule
\end{tabular}
\caption{Charlie Hebdo documents only: bit~02 prevalence and favorite
summaries in each state of bits~05 and~11. Fractions give bit~02-active
documents over all documents in the indicated group. The median uses every document
in the pair-state group. Decile boundaries are computed from all 2,079
Charlie Hebdo documents, before applying the bit conditions. Small fractions
are retained to show the limited support of some comparisons.}
\label{tab:pheme-charlie-complement-outcomes}
\end{table}

Table~\ref{tab:pheme-charlie-complement-outcomes} shows that the
$\{05,11\}$ conjunction is not the only group with substantial favorite
counts. In the $(0,1)$ complement, bit~02 also distinguishes the two tails,
within a different content context; the other complements contain very few
bit~02-active documents.

\begin{table}[tbp]
\centering\small
\setlength{\tabcolsep}{4pt}
\begin{tabular}{lllrrr}
\toprule
Event & Outcome & Bit & Lower 10\% & Upper 10\% & $\Delta_{25}$ (pp) \\
\midrule
Sydney & Favorites & 07 & 104/125 (83.2\%) & 43/123 (35.0\%) & -42.0 \\
Ottawa & Favorites & 14 & 75/93 (80.6\%) & 57/89 (64.0\%) & -13.8 \\
Ferguson & Retweets & 07 & 76/115 (66.1\%) & 91/116 (78.4\%) & +12.6 \\
Ferguson & Favorites & 07 & 95/116 (81.9\%) & 81/115 (70.4\%) & -9.8 \\
\bottomrule
\end{tabular}
\caption{Illustrative within-event bit contrasts. Fractions give active-bit
counts over all documents in each event-specific outcome decile.
$\Delta_{25}$ is upper minus lower bit prevalence for the corresponding
quartiles, in percentage points. Other bits are unrestricted.}
\label{tab:pheme-event-tails}
\end{table}

The Sydney favorite contrast agrees with the incident-reporting versus
public-response distinction examined in Section~\ref{sec:pheme-engagement}.
In its upper decile, \texttt{illridewithyou} occurs in 19/123 documents and
\texttt{thoughts} in 15/123, compared with zero occurrences of either term in
the 125 lower-decile documents. In Ottawa, upper-decile favorites similarly
include condolences and remembrance: \texttt{thoughts} appears in 22/89
documents versus 0/93 below, while \texttt{parliament} appears in 7/89 versus
30/93. Bit~14 is less prevalent in the upper favorite tail. Reading also
identifies unrelated entertainment and sports posts in the Ottawa collection,
which remain in the analysis; event metadata do not guarantee topical
homogeneity. Ferguson differs by outcome: bit~07 is more prevalent among
high-retweet documents and less prevalent among high-favorite documents.
Retweet contrasts include police force and demilitarization, while the
favorite reading also includes racial and moral evaluations. These are
contextual descriptions, not invariant meanings of bit~07.

The two outcomes select overlapping but distinct documents. Under corpus-wide
decile thresholds, each upper tail contains 582 tweets, with 361 shared
(62.0\% of either group). At quartile thresholds, 1,026 are shared between
upper groups of 1,454 retweet and 1,455 favorite documents, approximately
70.5\% of each. In the decile retweet-only group, bit~07 is active in 59.3\%,
compared with 38.0\% in the favorite-only group and 38.5\% in their intersection.
This overlap analysis follows, rather than replaces, the separate outcome
comparisons.

\subsubsection{Concentration of profiles and similarity of texts}

Profile entropy and embedding similarity describe different aspects of group
homogeneity. Table~\ref{tab:pheme-tail-homogeneity} reports empirical profile
entropy and mean cosine similarity across all distinct pairs of Qwen embeddings
within each tail of the three active-bit combinations. For the Charlie Hebdo
combination $\{05,11,02\}$, the upper decile has a more concentrated profile distribution,
but its mean embedding similarity is slightly lower. High engagement therefore
does not require a single, uniformly homogeneous text group. The entropy
values are descriptive plug-in estimates for unequal group sizes, not
sample-size-adjusted comparisons.

\begin{table}[tbp]
\centering\small
\setlength{\tabcolsep}{4pt}
\begin{tabular}{llrrrrrrr}
\toprule
Outcome & Active bits & Tail (\%) & $n_L$ & $n_U$ & $H_L$ & $H_U$ & Cosine$_L$ & Cosine$_U$ \\
\midrule
Retweets & $\{06,07\}$ & 10 & 113 & 124 & 3.668 & 3.291 & 0.455 & 0.474 \\
Retweets & $\{06,07\}$ & 25 & 273 & 302 & 3.631 & 3.484 & 0.452 & 0.470 \\
Favorites & $\{13,07\}$ & 10 & 176 & 100 & 3.211 & 3.006 & 0.471 & 0.461 \\
Favorites & $\{13,07\}$ & 25 & 435 & 265 & 3.277 & 3.120 & 0.464 & 0.451 \\
Favorites & $\{05,11,02\}$ & 10 & 21 & 99 & 0.792 & 0.244 & 0.556 & 0.535 \\
Favorites & $\{05,11,02\}$ & 25 & 72 & 241 & 0.884 & 0.490 & 0.558 & 0.536 \\
\bottomrule
\end{tabular}
\caption{Profile concentration and text similarity within selected
conjunctions. $L$ and $U$ denote lower and upper event-specific tails;
$H$ is empirical entropy of complete profiles in bits. Cosine is the mean
similarity of raw Qwen embeddings across all distinct document pairs within
the group. Counts, entropy, and similarity use every selected document.}
\label{tab:pheme-tail-homogeneity}
\end{table}

Content similarity alone does not determine engagement. Two tweets with the
same wording about thoughts for French cartoonists, their families, and
loved ones have 37,983 and one favorite, respectively
(IDs~552844593046097920 and~553149163181838336). Both activate bits~05, 11,
and~02; their recorded follower counts are 7,457 and 42. This example
illustrates why the tail descriptions cannot isolate a content effect from
account reach, timing, or other circumstances of circulation.

\subsection{Regression trees and importance diagnostics}
\label{app:pheme-trees}

Regression trees describe conditional groups through bit activations and
other covariates. For each outcome, we fit three trees on the standardized
log outcome: bits only, bits with controls, and bits with controls and all
labels. Each tree uses squared-error splitting, maximum depth eight,
minimum split and leaf sizes 30 and 15, complexity parameter 0.0005, and
no cross-validation pruning. For readability, the figures display the first
three split levels, collapsing deeper branches into their parent groups.
The reported importance uses all splits of the fitted trees.

\begin{figure}[ph]
\centering
\includegraphics[width=\textwidth]{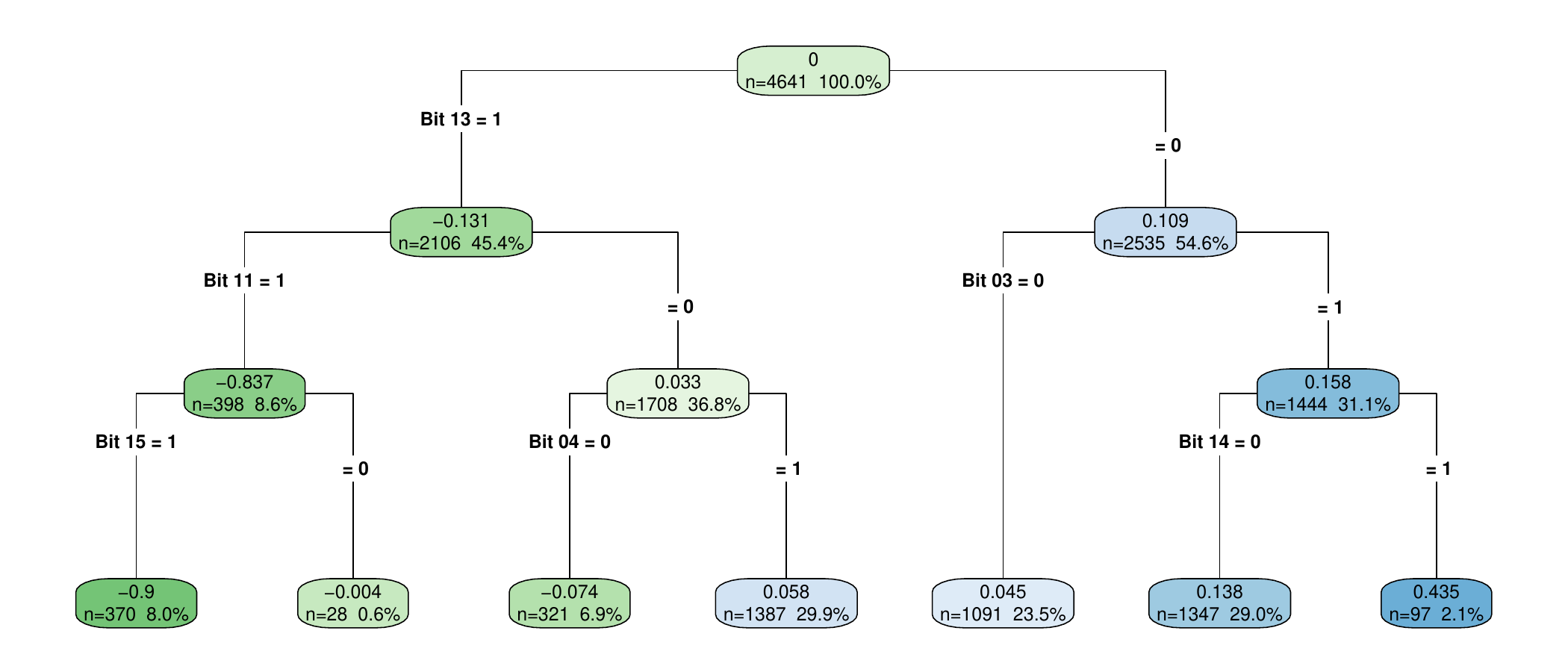}
\par\medskip
\includegraphics[width=\textwidth]{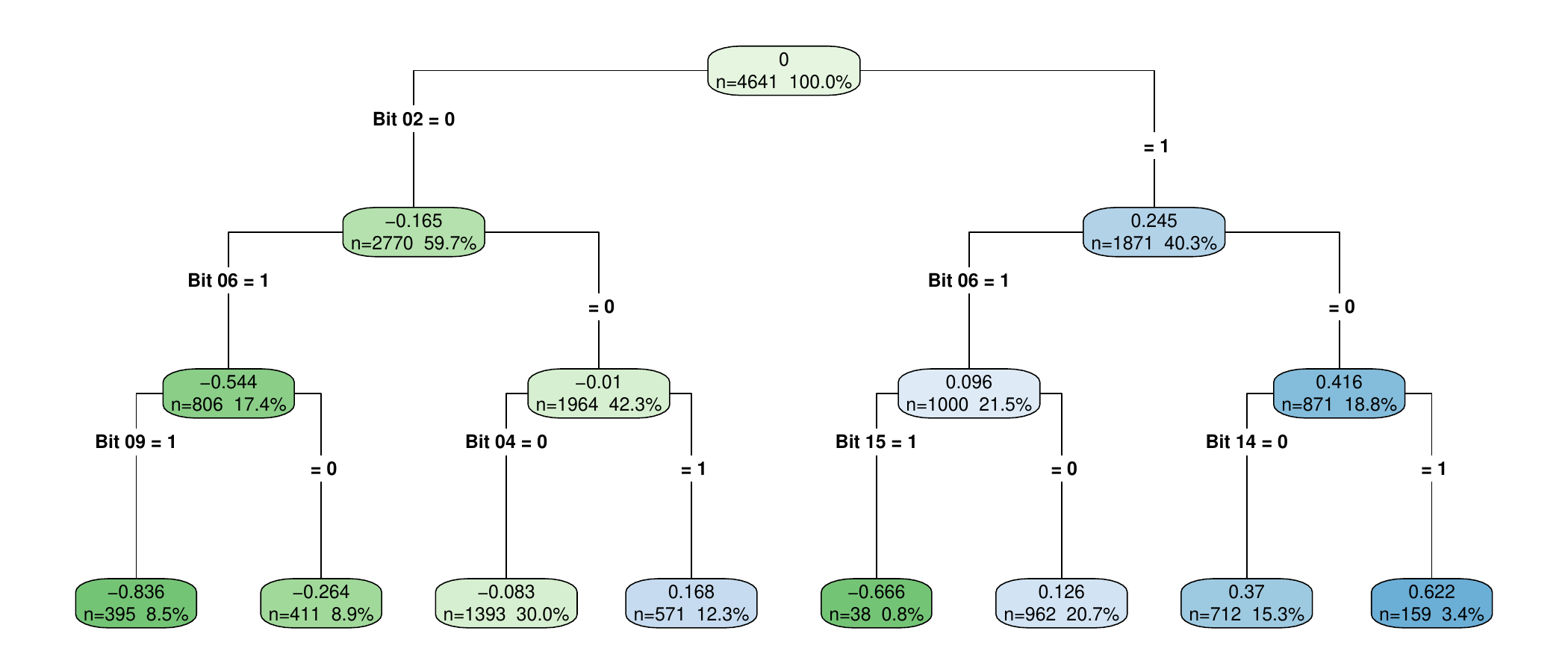}
\caption{First three split levels of the bits-only regression trees for
retweets (top) and favorites (bottom). Nodes show the training mean of
standardized $\log(1+y)$, number of documents, and percentage of the 4,641
training tweets. A negative node mean is below the training mean on this
scale. Branch conditions indicate bit activation or inactivity. Deeper
branches are collapsed for display; their parent means describe all
selected documents. Each path defines a subset for content inspection.}
\label{fig:pheme-bits-only-trees}
\end{figure}

In the bits-only retweet tree (Figure~\ref{fig:pheme-bits-only-trees}),
activating bit~13 is associated with a lower mean log outcome. Within its
active branch, bit~11 activation identifies a further decrease. For
favorites, bit~02 activation is associated with a higher mean log outcome;
within either activation state of bit~02, bit~06 activation identifies a
lower mean. Each comparison concerns the document group selected by the
preceding path. These descriptive training partitions are separate from
the held-out OLS and RF evaluation.

\begin{figure}[ph]
\centering
\includegraphics[width=\textwidth]{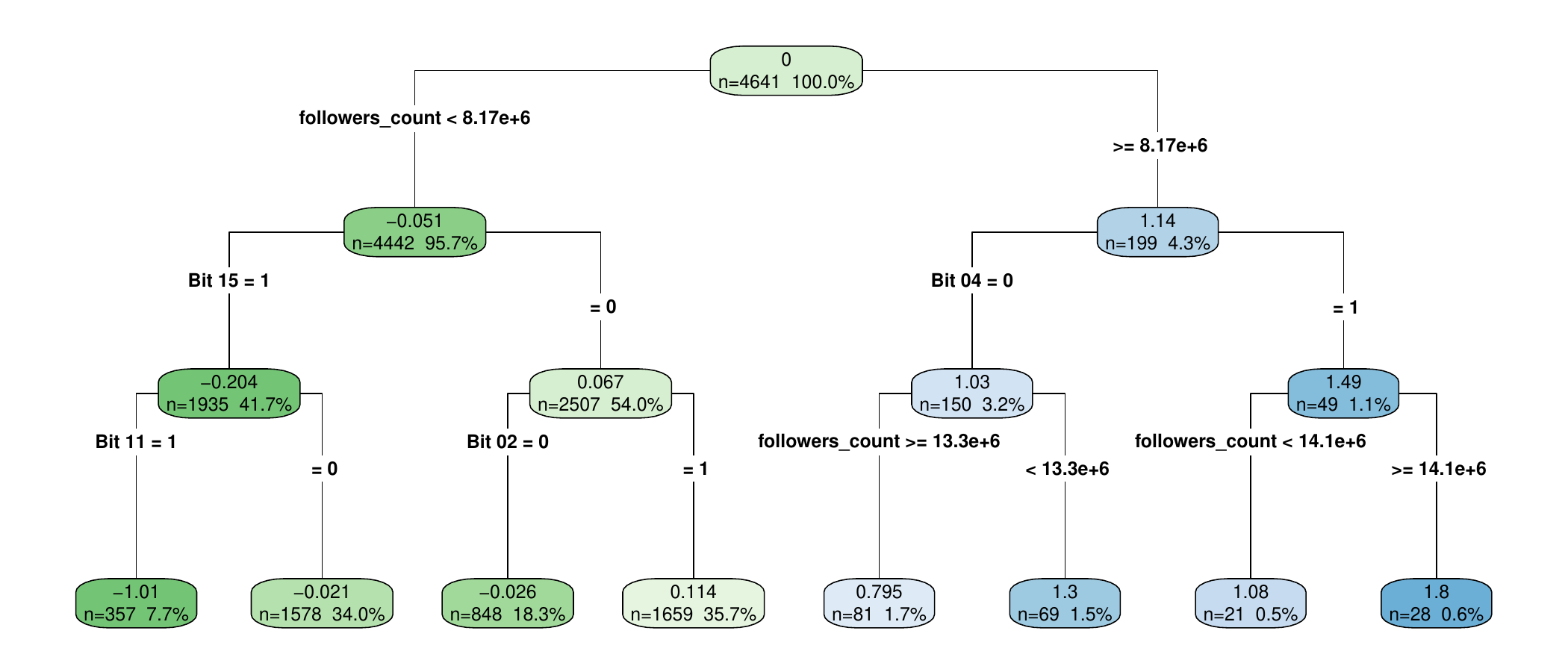}
\caption{First three split levels of the regression tree for retweets using bits, follower count, and text length. Nodes show training means of standardized $\log(1+y)$, document counts, and percentages of the 4,641 training tweets. Deeper branches are collapsed for display. Interpret each split conditional on its preceding path.}
\label{fig:pheme-tree-controls-retweets}
\end{figure}

\begin{figure}[ph]
\centering
\includegraphics[width=\textwidth]{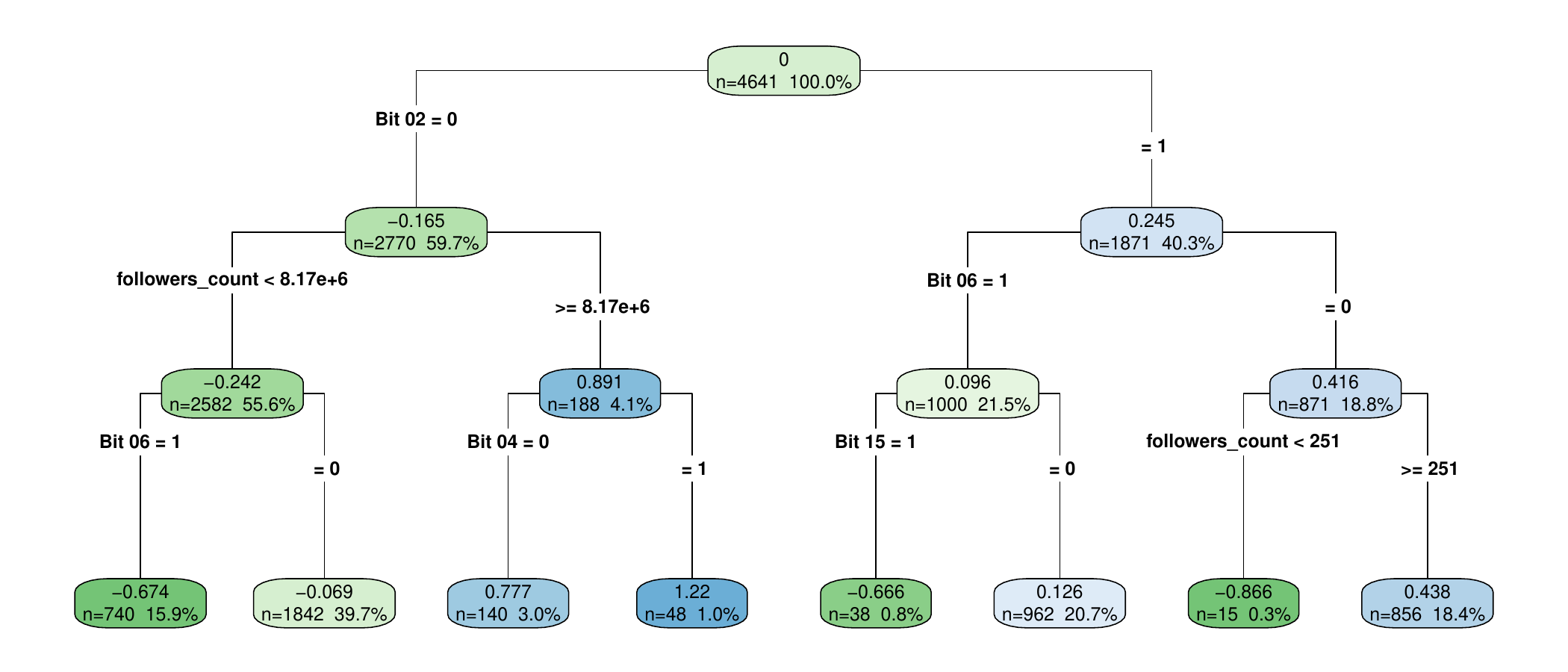}
\caption{First three split levels of the regression tree for favorites using bits, follower count, and text length. Nodes show training means of standardized $\log(1+y)$, document counts, and percentages of the 4,641 training tweets. Deeper branches are collapsed for display. Interpret each split conditional on its preceding path.}
\label{fig:pheme-tree-controls-favorites}
\end{figure}

\begin{figure}[ph]
\centering
\includegraphics[width=\textwidth]{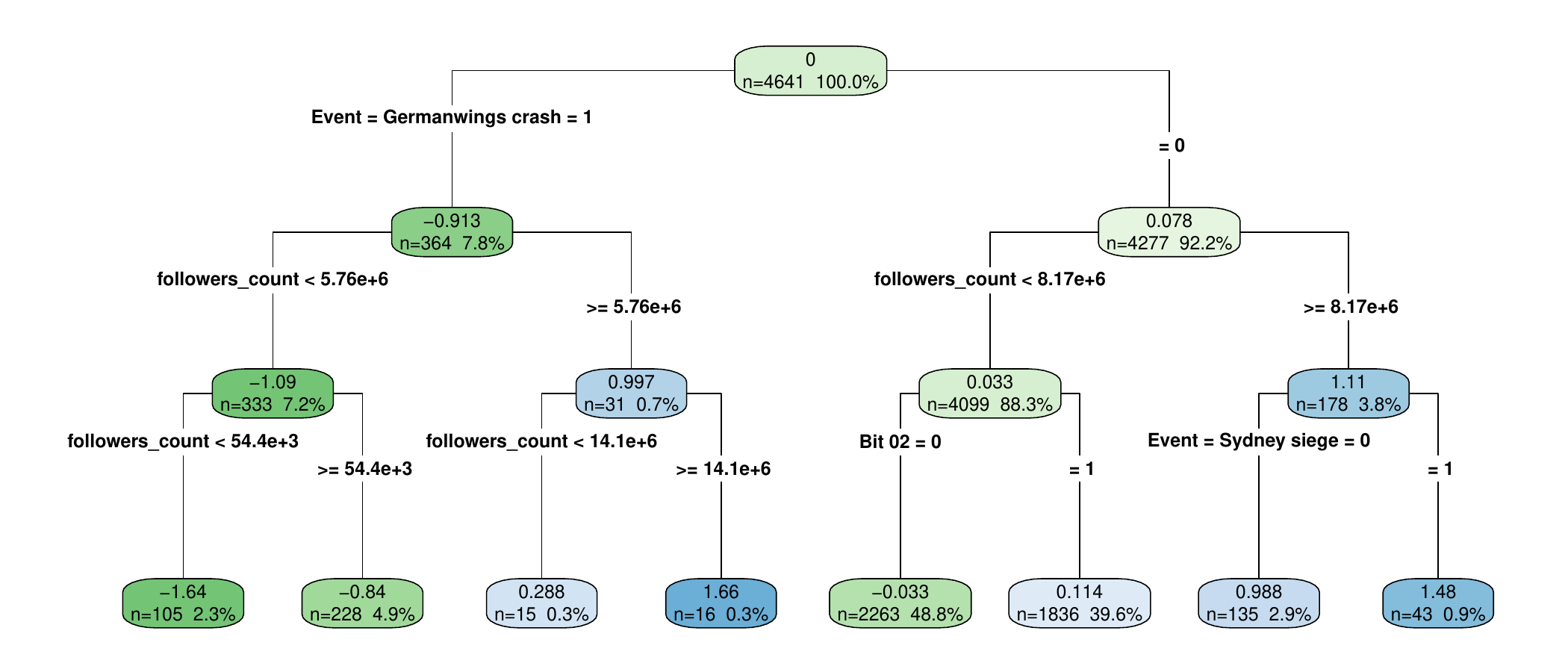}
\caption{First three split levels of the regression tree for retweets using bits, all three annotation sets, follower count, and text length. Nodes show training means of standardized $\log(1+y)$, document counts, and percentages of the 4,641 training tweets. Deeper branches are collapsed for display. Interpret each split conditional on its preceding path.}
\label{fig:pheme-tree-labels-retweets}
\end{figure}

\begin{figure}[ph]
\centering
\includegraphics[width=\textwidth]{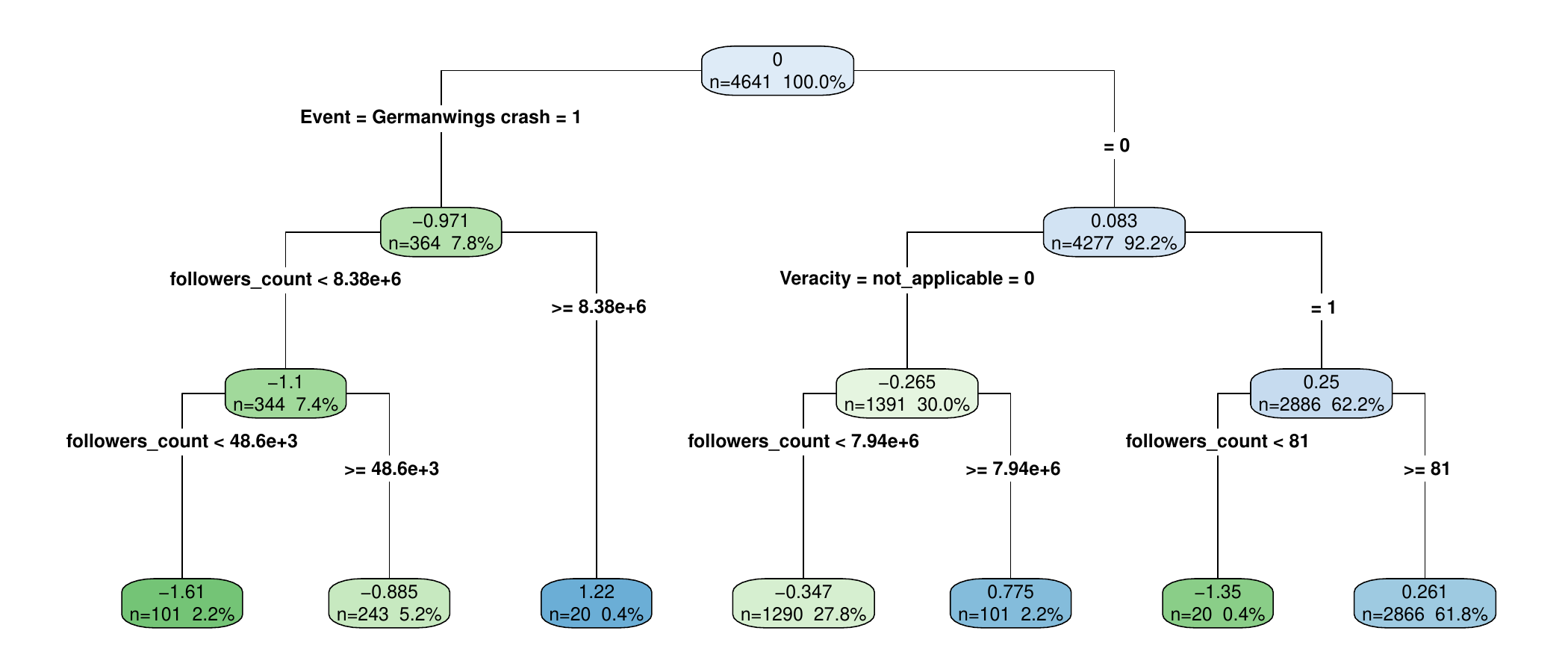}
\caption{First three split levels of the regression tree for favorites using bits, all three annotation sets, follower count, and text length. Nodes show training means of standardized $\log(1+y)$, document counts, and percentages of the 4,641 training tweets. Deeper branches are collapsed for display. Interpret each split conditional on its preceding path.}
\label{fig:pheme-tree-labels-favorites}
\end{figure}

Figures~\ref{fig:pheme-tree-controls-retweets}--\ref{fig:pheme-tree-labels-favorites}
show the corresponding trees with additional covariates. The retweet tree
with controls first splits on follower count; the favorite tree first splits
on bit~02. Both trees including labels first separate the Germanwings event.
Thus, the hierarchy of selected coordinates depends on which covariates
are available, as well as on the outcome.

\begin{figure}[ph]
\centering
\includegraphics[width=\textwidth,height=0.8\textheight,keepaspectratio]{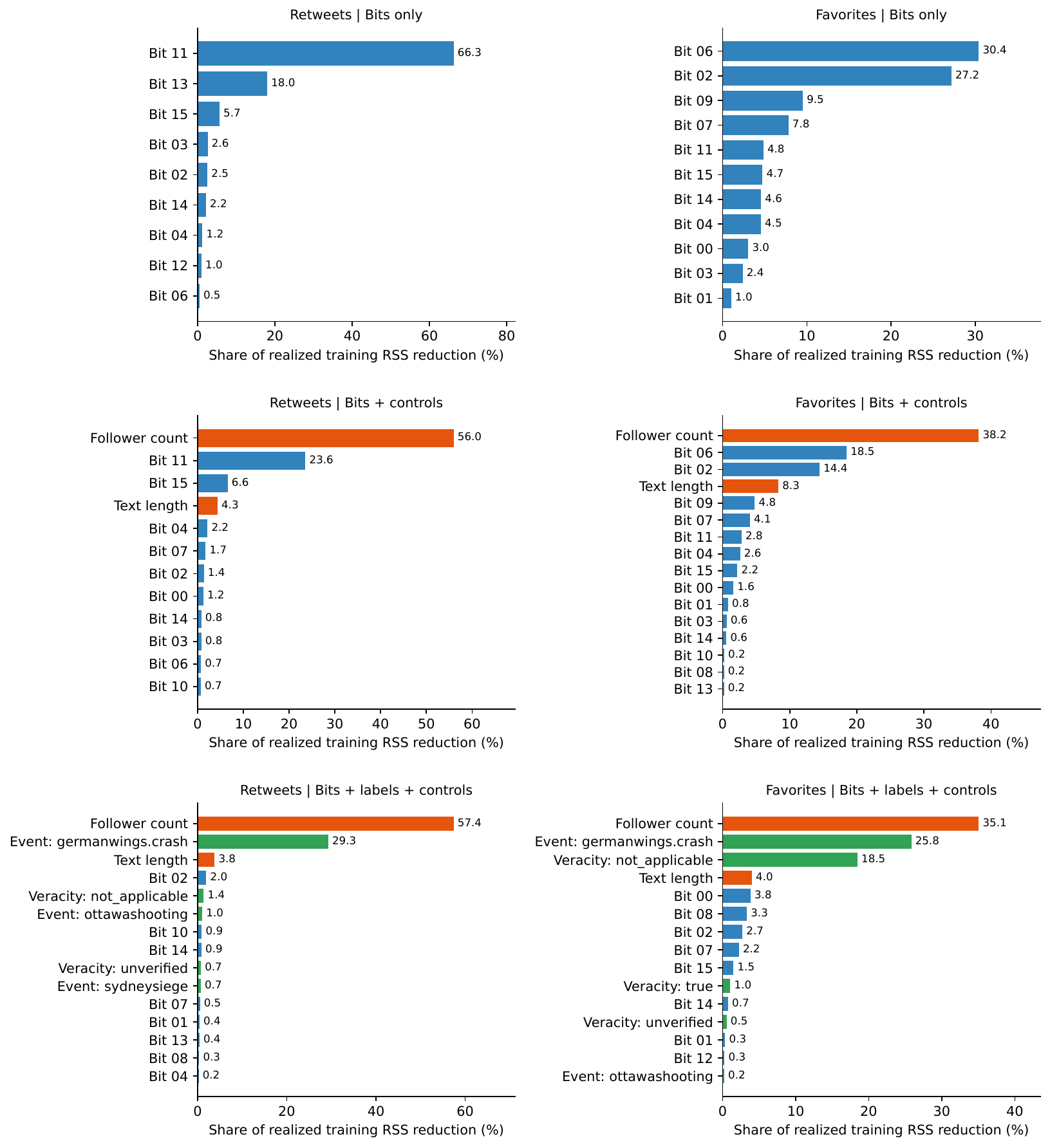}
\caption{Importance of variables selected by the six full regression trees
on standardized log outcomes. Bars and numbers give each variable's
percentage share of realized training residual-sum-of-squares reduction
from actual splits, excluding surrogate contributions. The denominator is
specific to each tree. Blue denotes bits, orange controls, and green
annotation indicators.}
\label{fig:pheme-tree-split-importance}
\end{figure}

Figure~\ref{fig:pheme-tree-split-importance} allocates each full tree's
realized reduction in training residual sum of squares to the variables
used in its actual splits. Shares sum to 100\% separately for each tree;
an omitted variable has no selected split. The calculation excludes
surrogate splits. These shares measure contributions to the fitted
partition's training error reduction, not the direction of an association
or the proportion of total outcome variance explained.

\begin{figure}[ph]
\centering
\includegraphics[width=\textwidth,height=0.8\textheight,keepaspectratio]{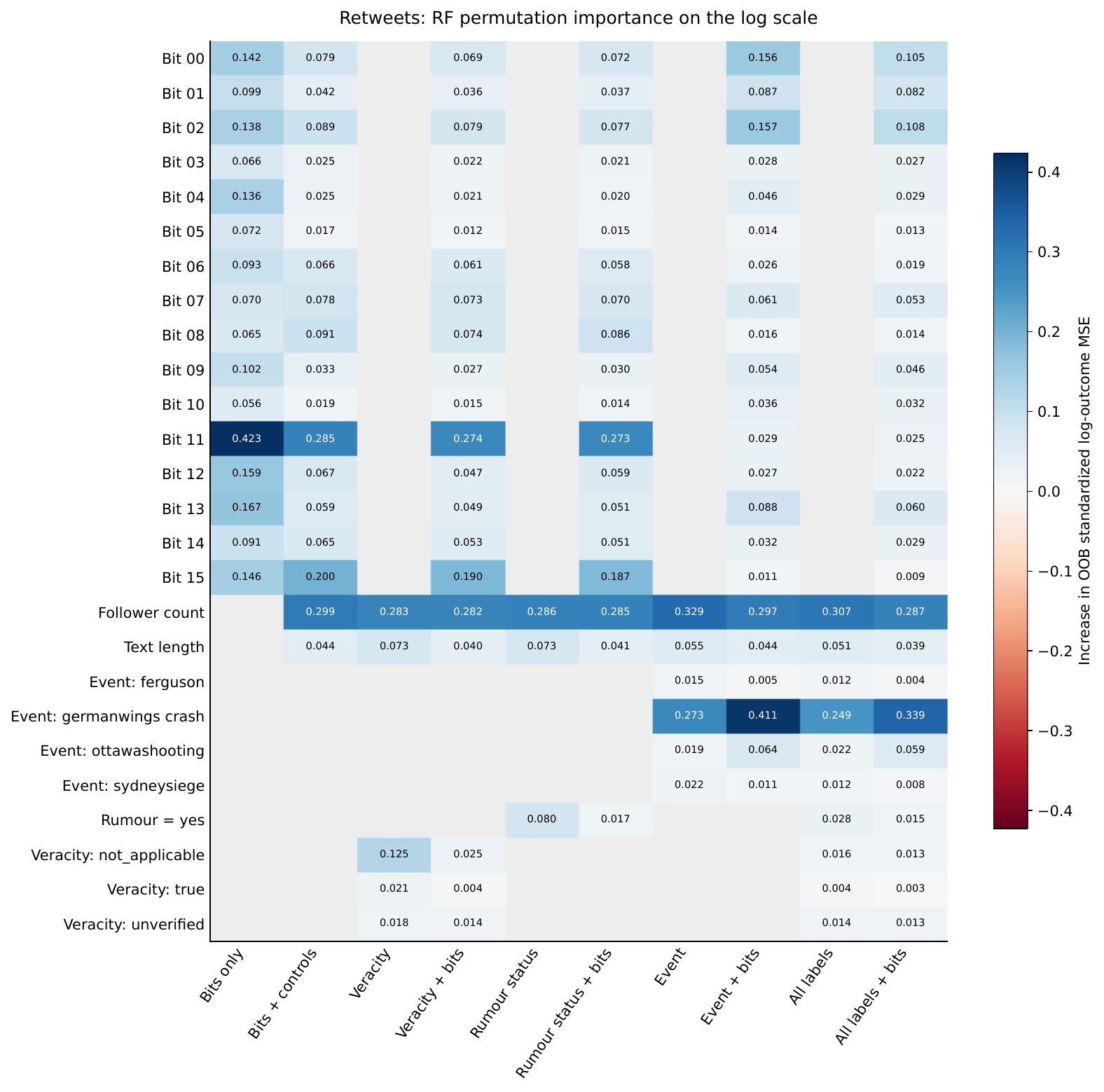}
\caption{RF permutation importance for retweets on the standardized log scale.
This perturbation-based ablation shuffles one predictor at a time among
each tree's out-of-bag observations, keeping the fitted forest fixed.
Rows identify predictors and columns specifications. Entries are the
increase in standardized log-outcome MSE after shuffling, averaged across
trees and rounded to three decimals. Blue positive cells indicate worse
prediction after shuffling; red negative cells indicate improved prediction.
The sign concerns error, not whether activation raises or lowers engagement.
Gray cells mark absent predictors; zero denotes rounding. All specifications
except Bits only include controls. Categorical predictors enter as indicators,
with reference categories omitted.}
\label{fig:pheme-rf-importance-retweets}
\end{figure}

\begin{figure}[ph]
\centering
\includegraphics[width=\textwidth,height=0.8\textheight,keepaspectratio]{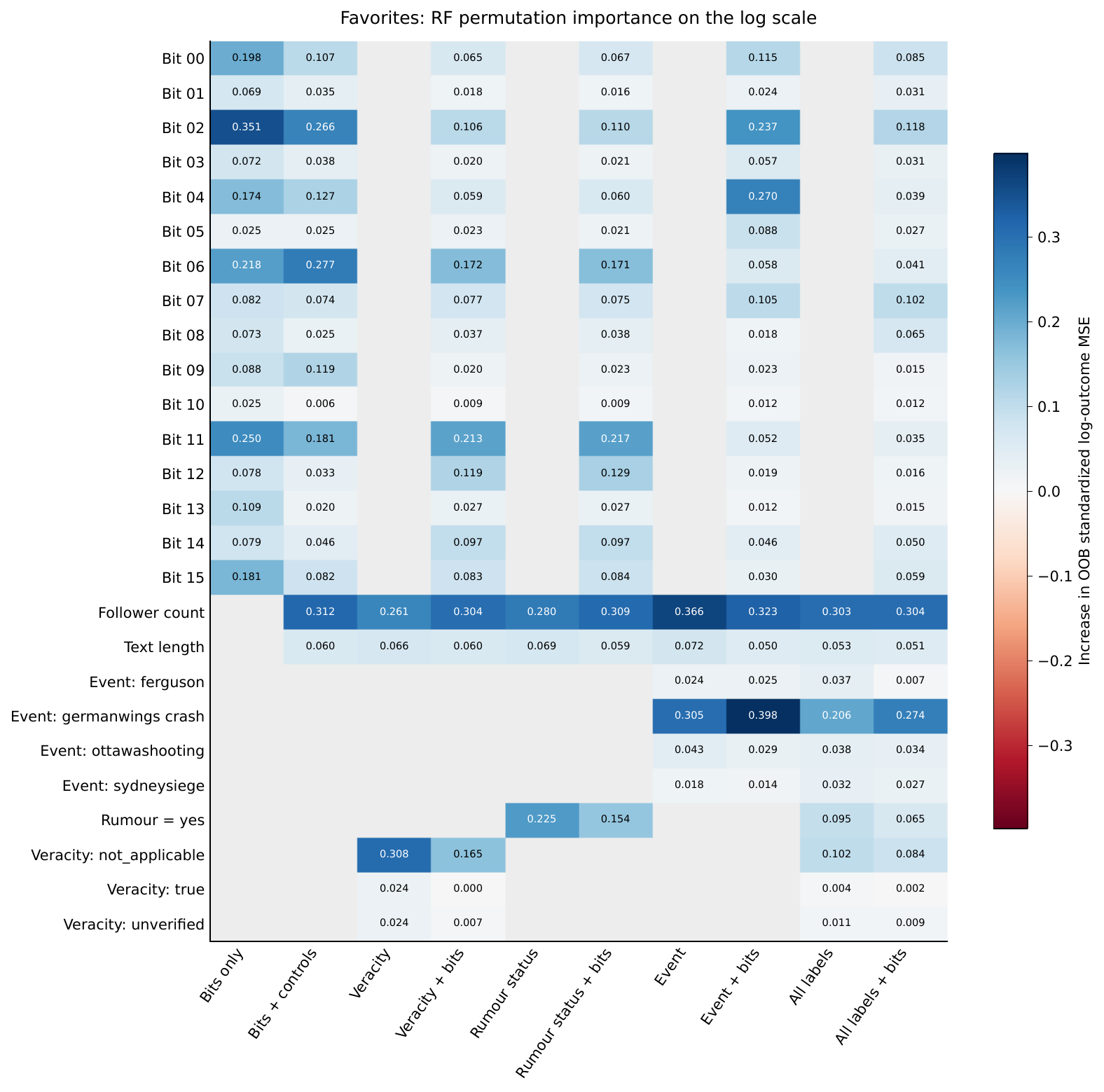}
\caption{RF permutation importance for favorites on the standardized log scale.
This perturbation-based ablation shuffles one predictor at a time among
each tree's out-of-bag observations, keeping the fitted forest fixed.
Rows identify predictors and columns specifications. Entries are the
increase in standardized log-outcome MSE after shuffling, averaged across
trees and rounded to three decimals. Blue positive cells indicate worse
prediction after shuffling; red negative cells indicate improved prediction.
The sign concerns error, not whether activation raises or lowers engagement.
Gray cells mark absent predictors; zero denotes rounding. All specifications
except Bits only include controls. Categorical predictors enter as indicators,
with reference categories omitted.}
\label{fig:pheme-rf-importance-favorites}
\end{figure}

Figures~\ref{fig:pheme-rf-importance-retweets} and~\ref{fig:pheme-rf-importance-favorites}
report RF permutation importance for all ten specifications. For each tree
in a forest, a predictor is shuffled among training observations omitted
from that tree's bootstrap sample, called out-of-bag observations. Importance
is the resulting increase in standardized log-outcome MSE, averaged across
trees. This diagnostic does not use the 1,161 test tweets. Positive values
indicate worse prediction after shuffling; negative values indicate improved
prediction after shuffling, not an association with lower engagement.

Bit~11 has the largest permutation importance in the bits-only retweet
forest, whereas bit~02 has the largest in the favorite forest. The
individual bits-only trees allocate their largest split-improvement shares
to bit~11 and bit~06, respectively. The two diagnostics therefore need not
rank coordinates identically. Correlated bits and labels can share
information, and shuffling breaks their dependence. The figures retain
all included predictors and negative importance values; they describe the
fitted models without confidence intervals or causal interpretation.

\end{document}